%% file: main.tex
\documentclass[letterpaper, 10 pt, journal, twoside]{IEEEtran}
\usepackage[utf8]{inputenc} 
\usepackage[T1]{fontenc}    
\usepackage{url}            
\usepackage{booktabs}       
\usepackage{amsfonts}       
\usepackage{nicefrac}       
\usepackage{microtype}      
\usepackage{xcolor}         

\usepackage{anyfontsize}
\usepackage{amsmath}
\usepackage{amssymb}
\usepackage{wrapfig}
\usepackage{subfig}
\usepackage{caption}
\usepackage{color}

\usepackage{physics}
\usepackage{bm}
\usepackage{algorithm}
\usepackage{algorithmic}
\usepackage{graphicx}

\usepackage{ragged2e}
\usepackage{fp}

\usepackage{ifthen}

\usepackage{pgfplots}
\usepackage{tikz}
\usepackage{tikz-3dplot}

\usepackage{ragged2e}
\usepackage{fp}

\usepackage{ifthen}

\usepackage{tikz}
\usetikzlibrary{
shapes.geometric, 
arrows,
 calc,
 automata,
 positioning,
 decorations,
 decorations.text,
 patterns}
 
 \usetikzlibrary{pgfplots.statistics, pgfplots.colorbrewer} 
\usepackage{pgfplotstable}

\usetikzlibrary{patterns.meta}

\usetikzlibrary{mindmap,snakes}

\usepackage{multirow}

\usepackage{algorithm}
\usepackage{algorithmic}
\usepackage{graphicx}
\usepackage[english]{babel}
\usepackage{array}
\usepackage[export]{adjustbox}

\usepackage{nicefrac}

\usepackage{xcolor}

\usepackage{colortbl}
\usepackage{makecell}
\usepackage{pifont}

\usepackage{mathptmx}

\usepackage[colorlinks=true,
    linkcolor=red,
    linkbordercolor=red,
    urlcolor=mylinkcolor,
    urlbordercolor=white
    ]{hyperref}

\pgfplotsset{
    /pgfplots/ybar legend/.style={
    /pgfplots/legend image code/.code={%
       \draw[##1,/tikz/.cd,yshift=-0.25em]
        (0cm,0cm) rectangle (3pt,0.8em);},
   },
}

\newcommand{\cmark}{\ding{51}}%
\newcommand{\xmark}{\ding{55}}

\definecolor{mylinkcolor}{rgb}{0.005, 0.3, 0.7}

\definecolor{mylinkcolor}{rgb}{0.005, 0.3, 0.7}
\definecolor{suppcolor}{rgb}{0.6, 0.0, 0.9}

\newcommand{\OursAcronym}{Jetson-SLAM}
\newcommand{\ORBBE}{Full-BA}

\usepackage[font=small]{caption}

\colorlet{baseclr}{gray}
\colorlet{rwclr}{white!90!baseclr}

\colorlet{dotaclr}{black}
\colorlet{dotbclr}{red}

\newcommand{\bdota}{\tikz{\node[circle, draw=dotaclr, fill=white!10!dotaclr,scale=0.4]{};}}
\newcommand{\bdotb}{\tikz{\node[circle, draw=dotbclr, fill=white!00!dotbclr,scale=0.4]{};}}

\title{High-Speed Stereo Visual SLAM for Low-Powered Computing Devices}

\author{Ashish~Kumar$^{\dagger}$, Jaesik Park$^{\ddagger}$, \IEEEmembership{Member, IEEE}, Laxmidhar~Behera$^{\dagger}$, \IEEEmembership{Senior Member, IEEE} \vspace{-1.75ex}  \\
\thanks{Manuscript received: August 18, 2023; Accepted October 16, 2023. This paper was recommended for publication by Editor Javier Civera upon evaluation of the Associate Editor and Reviewers' comments.}
\thanks{Jaesik Park was supported by IITP grant funded by the Korea government(MSIT) (NO.2021-0-01343 AI Graduate School Program (Seoul National University) \& (RS-2023-00227993: Detailed 3D reconstruction for urban areas from unstructured images)}
\thanks{$^{\dagger}$EE, Indian Institute of Technology (IIT), Kanpur, India.
{\tt\small \{krashish,lbehera\}@iitk.ac.in}~
$^{\ddagger}$CSE \& IPAI, Seoul National University (SNU), Republic of Korea.
{\tt\small jaesik.park@snu.ac.kr}}%
\thanks{Digital Object Identifier (DOI): see top of this page.}
}

\begin{document}

\maketitle

\begin{justify}

\begin{abstract}
\input{abstract}
\end{abstract}

\end{justify}

\vspace{-0.75ex}
\begin{IEEEkeywords}
Aerial Systems: Applications; SLAM; Embedded Systems for Robotic and Automation
\end{IEEEkeywords}

\vspace{-2.00ex}
\section*{\textcolor{suppcolor}{Supplementary Material}}
\noindent
\textbf{Code:} \textcolor{mylinkcolor} {\footnotesize \url{https://github.com/ashishkumar822/\OursAcronym}} \\
\textbf{Video:}~\textcolor{gray}{\footnotesize See attachment.}

\IEEEpeerreviewmaketitle

%


\vspace{-1.5ex}
\section{Introduction}
\label{sec:intro}
\input{introduction}
\vspace{-1.0ex}
\section{Methodology}
\label{sec:method}
\input{methodology}
\input{bounded_rectification}
\input{pyca_fc}
\input{pyca_pfa}
\section{System Integration}
\label{sec:sysint}
\input{system_integration}
%
%
\section{Experiments}
\label{sec:exp}
\input{experiments}
\section{Conclusion \& Future Work}
\label{sec:conc}
\input{conclusion}
\bibliographystyle{ieeetr}
\bibliography{bibfile}









\end{document}

%% file: abstract.tex
We present an accurate and GPU-accelerated Stereo Visual SLAM design called \OursAcronym{}. It exhibits frame-processing rates above $\mathbf{60}$FPS on NVIDIA's low-powered $\mathbf{10}$W Jetson-NX embedded computer and above $\mathbf{200}$FPS on desktop-grade $\mathbf{200}$W GPUs, even in stereo configuration and in the multiscale setting. Our contributions are threefold: (\textit{i}) a Bounded Rectification technique to prevent tagging many non-corner points as a corner in FAST detection, improving SLAM accuracy. (\textit{ii}) A novel Pyramidal Culling and Aggregation (\texttt{PyCA}) technique that yields robust features while suppressing redundant ones at high speeds by harnessing a GPU device. \texttt{PyCA} uses our new Multi-Location Per Thread culling strategy (\texttt{MLPT}) and Thread-Efficient Warp-Allocation (\texttt{TEWA}) scheme for GPU to enable \OursAcronym{} achieving high accuracy and speed on embedded devices. (\textit{iii}) \OursAcronym{} library achieves resource efficiency by having a data-sharing mechanism. Our experiments on three challenging datasets: KITTI, EuRoC, and KAIST-VIO, and two highly accurate SLAM backends: \ORBBE~and ICE-BA show that \OursAcronym{} is the fastest available accurate and GPU-accelerated SLAM system (Fig.~\ref{fig:intro}). 

%% file: introduction.tex
\IEEEPARstart{A} centimeter-accurate local positioning system is crucial for complex robotic and autonomous flight systems to execute navigation, control, and visual servoing tasks precisely \cite{towards}. Visual odometry (VO) can be employed for this purpose, but it discards older environmental observations and lacks global consistency \cite{orb2}. This causes pose-estimation drifts over time, although an agent navigates in the same area. 
\par
In contrast, visual SLAM offers drift-free localization and mapping, which enables the precise execution of autonomous tasks. In this context, \textit{Stereo visual SLAM} is particularly interesting due to its high metric accuracy and low-cost sensor demands. However, its compute-intensive frontend (feature detection-extraction-matching, stereo-matching) and backend (graph optimization, loop-closure, localization and mapping) quickly exhaust low-powered devices. Also, the limited computing power of these devices forces a SLAM system to drop intermediate frames \cite{kaistvio}, resulting in reduced frame rate and tracking failure \cite{orb2}.
\par
This situation becomes difficult due to shared computing resources in the presence of co-existing modules such as data acquisition, control, grasping, and compute-intensive deep networks. It can easily cause a catastrophic system failure, e.g., the divergence of the control system from the desired trajectory due to delays in position feedback. For these reasons, UAV-based autonomous manipulation \cite{towards} executed SLAM on a remote computer. \cite{edgeslam} explored network computing for SLAM but is unsuitable for isolated robotic systems. 
\input {plots/intro_figure}
\par
Although there have been efforts to speed up VO systems, SLAM still remains untouched. For instance, recent \cite{kaistvio} only benchmarks runtime of existing VO systems \cite{forster2016svo, dmvio, rebecq2021simultaneous} onto Jetson devices. In VO, restricting feature count by dividing an image into grids and picking one feature per grid is commonly used. \cite{fastergpu} is such an approach for GPU. However, its design is limited only to monocular VO, preventing its use in SLAM. Its unconfigurable scale factor of two in a multiscale setting reduces the image resolution aggressively, resulting in fewer features which are further trimmed to only one feature per-grid irrespective of the number of scales. It leads to inadequate feature points in SLAM, causing tracking failures (see video). Moreover, it is inefficient for smaller grids when deployed onto embedded computing devices.
\par
ORB-SLAM$2$~\cite{orb2}, ICE-BA~\cite{iceba} are highly accurate SLAM systems. However, their complex CPU-only workload turns them slower and exhausts the low-powered devices. SLAMCore~\cite{slamcore} is CPU-efficient SLAM but is not open source and is not benchmarked alongside deep networks. Despite these limitations, a high-speed SLAM system is the need of the current time with a huge scope in modern autonomous systems. 
\par
In visual SLAM, the speed is mainly governed by the frontend load, which varies with image resolution and doubles in stereo mode (of our interest). Also, the number of features yielded by the frontend affects the backend load. Thus, we develop a GPU-accelerated frontend from scratch which produces a sufficiently smaller number of impactful features at high-speed by harnessing the on-chip GPU of the embedded computers. However, in this endeavor, the scarcity of GPU cores appears as a bottleneck that limits the maximum achievable speed, and an inefficient use of the cores degrades the GPU throughput. We tackle this issue via our algorithmic and system development contributions, resulting in a high-speed, accurate, and resource-efficient GPU-accelerated SLAM system available to date, called \emph{\OursAcronym}. Our contributions are:
%
%
%
%
%
\paragraph{Bounded Rectification}
prevents misclassification of non-corners as corners in FAST features \cite{fast}, and improves SLAM accuracy by producing impactful corners (Sec.~\ref{sec:fast}). 
\paragraph{Pyramidal Culling and Aggregation (\texttt{PyCA})}
It yields high-quality multiscale features via our Multi-Location Per-Thread (\texttt{MLPT}) culling, and Thread Efficient Warp-Allocation (\texttt{TEWA}) to deliver high speeds ($2000$ FPS) and high computing efficiency even in the scarcity of GPU cores (Sec.~\ref{sec:pyca}).
\paragraph{Frontend--Middle-end--Backend Design of \OursAcronym}
We develop a new SLAM component called \emph{Middle-end} that houses stereo-matching, feature matching, feature-tracking, and performs data-sharing to avoid CPU-GPU memory-transfer overhead of duplicating-and-accessing intermediate results needed across SLAM components (Sec.~\ref{sec:sysint}).
\par
Despite we contribute in the frontend and system design, the middle-end and backend performance also gets boosted. It turns \OursAcronym{} efficient and accurate while reaching above $60$FPS $@432\times240$, even at eight scales in stereo mode on Jetson-NX alongside VGG \cite{vgg} deep neural network. The high speed minimizes tracking failures during camera rotations (video), and facilitates developing autonomous UAVs which still rely on external positioning systems \cite{towards}. 
\par
Next, we discuss our algorithmic contributions (Sec.~\ref{sec:method}), and system development contributions (Sec.~\ref{sec:sysint}). Experiments are presented in Sec.~\ref{sec:exp}, with conclusions in Sec.~\ref{sec:conc}.

%% file: plots/intro_figure.tex
\begin{figure}[t]
    \centering
\begin{tikzpicture}

\colorlet{frontendclr}{white!60!gray}
\colorlet{frontenddclr}{white!10!gray}

\colorlet{cpuclr}{white!60!magenta}
\colorlet{cpudclr}{white!20!magenta}

\colorlet{gpuclr}{white!80!blue}
\colorlet{gpudclr}{white!20!blue}

\colorlet{boundaryclr}{white!95!black}

\FPeval{\nodescale}{1.0}

\node (a) [xshift=-17.5ex, yshift=-9.25ex, scale=0.8]{(a)};
\node (b) [xshift=20.9ex, yshift=-9.25ex, scale=0.8]{(b)};
\node (c) [xshift=-17.9ex, yshift=-23.5ex, scale=0.8]{(c)};
\node (d) [xshift=11.0ex, yshift=-23.5ex,scale=0.8]{(d)};

\node (im) [xshift=-0.0ex, scale=\nodescale]{
\tikz{
%
%
%
%
\node (s1) [xshift = 0.0ex,yshift=0.0ex]{\includegraphics[width=35.2ex, height=10ex]{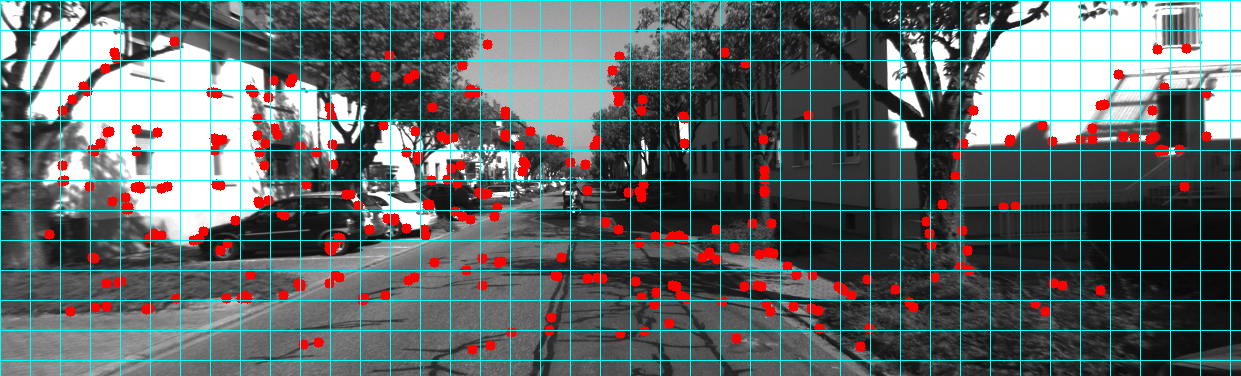}};
\node (s2) [xshift = -10.0ex,yshift=-8.0ex]{\includegraphics[scale=0.055]{images/slam_kitti_00.png}};
\node (s3) [xshift = 3.9ex,yshift=-8.0ex]{\includegraphics[scale=0.042]{images/slam_kitti_00.png}};
\node (s4) [xshift = 13.9ex,yshift=-8.0ex]{\includegraphics[scale=0.027]{images/slam_kitti_00.png}};
\draw [<->,white!20!black] ($(s1.west)-(-0.40ex,-5.0ex)$) |- ($(s1.east)-(0.5ex, 11.5ex)$) node [fill=white, xshift=-17.9ex,yshift=-0.0ex, scale=0.53]{\textbf{\texttt{Pyramidal Culling and Aggregation (\texttt{PyCA})}}};
}};

\FPeval{\xshifta}{0}
\FPeval{\xshiftb}{0}
\FPeval{\xshiftc}{27}

\FPeval{\yshifta}{0}
\FPeval{\yshiftb}{0+0}
\FPeval{\yshiftc}{0-0.3}

\FPeval{\ticklabelscale}{1.0}
\FPeval{\xlabelrotate}{0}

\FPeval{\labelscale}{1.5}
\FPeval{\labelscaleb}{1.3}
\FPeval{\labelscalec}{1.3}
\FPeval{\ticklabelscaleb}{1.1}
\FPeval{\ticklabelscalec}{1.1}

\FPeval{\gpuscale}{1.3}
\FPeval{\gpuxshifta}{23.8}
\FPeval{\gpuxshiftb}{23.8}

\FPeval{\gpuyshifta}{18.8}
\FPeval{\gpuyshiftb}{18.8}

\FPeval{\barw}{7}
\FPeval{\barsep}{0.8}
\FPeval{\linew}{1.0}
\FPeval{\barcorner}{0.2}

\FPeval{\linewb}{0.3}
\FPeval{\linewc}{1.0}

\FPeval{\mrksize}{0.5}
\FPeval{\legendscale}{1.0}
\FPeval{\legendimscale}{0.5}

\FPeval{\legendxshifta}{11.0}
\FPeval{\legendxshiftb}{35.5}
\FPeval{\legendxshiftc}{35.5}

\FPeval{\legendyshifta}{26.5}
\FPeval{\legendyshiftb}{15.5}
\FPeval{\legendyshiftc}{15.5}

\FPeval{\ylabelxshifta}{3.5}
\FPeval{\ylabelxshiftb}{2.5}
\FPeval{\ylabelxshiftc}{0.0}

\FPeval{\ylabelyshifta}{12.5}
\FPeval{\ylabelyshiftb}{10}
\FPeval{\ylabelyshiftc}{10}

\FPeval{\xlabelxshifta}{13}
\FPeval{\xlabelxshiftb}{23}
\FPeval{\xlabelxshiftc}{23}

\FPeval{\xlabelyshifta}{0+1.5}
\FPeval{\xlabelyshiftb}{0+0.3}
\FPeval{\xlabelyshiftc}{0-0.8}

\colorlet{gridclr}{white!90!black}
\colorlet{dlegendclr}{white!80!black}
\colorlet{axisclr}{white!70!black}
\colorlet{axisbgclr}{white!100!black}

\colorlet{dlegendclr}{white!80!black}
\colorlet{legendclr}{white!100!black}

\colorlet{clr1}{white!0!magenta}
\colorlet{clr2}{black!50!green}
\colorlet{clr3}{white!0!cyan}

\colorlet{clr1fill}{black!30!clr1}
\colorlet{clr2fill}{black!30!clr2}
\colorlet{clr3fill}{black!30!clr3}

\colorlet{gpudclr}{white!80!black}
\colorlet{gpuclr}{white!90!black}

\colorlet{deltaclr}{white!0!black}
\colorlet{gputxtclr}{white!0!black}

\colorlet{gtclr}{white!0!magenta}
\colorlet{predclr}{blue!10!brown}
\colorlet{orbclr}{blue!50!black}

\node (im) [xshift=27.0ex, yshift=0.2ex, scale=\nodescale]{
\tikz{
%
\FPeval{\pltw}{47}
\FPeval{\plth}{47}

\FPeval{\bscale}{0.39}%
\node (na) [draw=none, xshift =\xshifta ex, yshift= \yshifta ex]{
\scalebox{\bscale}
{\tikz{
\pgfplotsset{width=\pltw ex, height=\plth ex}
\begin{axis}[
   axis background style={fill=axisbgclr},
    title={},
    xlabel={x(m)},
    ylabel={y(m)},
    xmin=-300, xmax=300,
    ymin=-40, ymax=570,
    ytick={-40, 265,  570},
     axis line style={axisclr},
    legend image post style={scale =\legendimscale},
    legend style={at={(\legendxshifta ex,\legendyshifta ex)},anchor=south, legend columns = 1, draw = {dlegendclr}, fill={legendclr}, nodes={scale=\legendscale}},
    ymajorgrids=true, 
    xmajorgrids=true,
    grid style={dashed, gridclr},
    major tick length=1ex,
    y label style={at={(\ylabelxshifta ex, \ylabelyshifta ex)}, scale=\labelscale},
    x label style={at={(\xlabelxshifta ex, \xlabelyshifta ex)}, scale=\labelscale},
    yticklabel style={scale=\ticklabelscale},
    xticklabel style={scale=\ticklabelscale},
    legend cell align={left}
]
\input{plots/traj_coords_kitti_00}
%
%
 \legend{Ground-truth, ORB-SLAM$2$ \cite{orb2}, \OursAcronym}
\end{axis}
%
%
}
}};
}};

\node (im) [xshift=9.0ex, yshift=-18.0ex, scale=\nodescale]{
\tikz{
\FPeval{\pltw}{60}
\FPeval{\plth}{31}
\FPeval{\bscale}{0.43}
\node (na) [draw=none, xshift =\xshiftb ex, yshift= \yshiftb ex]{
\scalebox{\bscale}
{\tikz{
\pgfplotsset{width=\pltw ex, height=\plth ex}
\begin{axis}[
   axis background style={fill=axisbgclr},
    title={},
    xlabel={Frame Resolution},
    ylabel={FPS (Hz)},
    xmin=-0.2, xmax=2.2,
    ymin=0, ymax=80,
    xtick={0, 1, 2, 3, 4},
    ytick={0, 20, 40, 60, 80},
   xticklabels={$320$$\times$$240$, $752$$\times$$480$, $1242$$\times$$375$},
         yticklabels={$0$, $20$, $40$, $60$, $80$},
     axis line style={axisclr},
    legend image post style={scale =\legendimscale},
    legend style={at={(\legendxshiftb ex,\legendyshiftb ex)},anchor=north, legend columns = 1, draw = {dlegendclr}, fill={legendclr}, nodes={scale=\legendscale}},
    ymajorgrids=true, 
    xmajorgrids=true,
    grid style={dashed, gridclr},
    major tick length=1ex,
    y label style={at={(\ylabelxshiftb ex, \ylabelyshiftb ex)}, scale=\labelscaleb},
    x label style={at={(\xlabelxshiftb ex, \xlabelyshiftb ex)}, scale=\labelscaleb},
    xticklabel style={rotate=\xlabelrotate},
    yticklabel style={scale=\ticklabelscaleb},
    xticklabel style={scale=\ticklabelscaleb},
    yticklabel style={xshift=0.5ex},
    legend cell align={left}
]
\addplot[
    fill=none,draw = clr1,
    very thick,
    line width= \linewb ex,
    mark = square*,
    mark size = \mrksize ex
    ]
    coordinates {
(0, 22)
(1, 11)
(2, 8)
    };
\addplot[
    fill=none,draw = clr2,
    very thick,
    line width= \linewb ex,
    mark = square*,
    mark size = \mrksize ex
    ]
    coordinates {
(0, 60)
(1, 31)
(2, 22)
};
\legend{ORB-SLAM$2$ \cite{orb2}, Jetson-SLAM}
\node [draw=gpudclr,fill=gpuclr,rounded corners=0.2ex, minimum width=23.7ex, xshift=\gpuxshifta ex, yshift=\gpuyshifta ex, scale=\gpuscale]{\textcolor{gputxtclr}{\textbf{Jetson-NX}$\mathbf{~@}$~\textbf{Scales}~$\mathbf{=8}$}};
\end{axis}
}
}};
\node (nc) [draw=none, xshift =\xshiftc ex, yshift= \yshiftc ex]{
\scalebox{\bscale}
{\tikz{
\pgfplotsset{width=\pltw ex, height=\plth ex}
\begin{axis}[
   axis background style={fill=axisbgclr},
    title={},
    xlabel={SLAM Pipeline},
    ylabel={ATE Error (m)},
    xmin=-0.0, xmax=1.5,
    ymin=0, ymax=1.0,
    xtick={0.15, 0.7},
    ytick={0, 0.25, 0.50, 0.75, 1.00},
   xticklabels={ORB-SLAM$2$, \OursAcronym},
         yticklabels={$0$, $0.25$, $0.50$, $0.75$, $1.00$},
     axis line style={axisclr},
    legend image post style={scale =\legendimscale},
    legend style={at={(\legendxshiftc ex,\legendyshiftc ex)},anchor=north, legend columns = 1, draw = {dlegendclr}, fill={legendclr}, nodes={scale=\legendscale}},
    ymajorgrids=true, 
    xmajorgrids=true,
    grid style={dashed, gridclr},
    major tick length=1ex,
    major tick length=-0.5ex,
    ytick align=outside,
    y label style={at={(\ylabelxshiftc ex, \ylabelyshiftc ex)}, scale=\labelscalec},
    x label style={at={(\xlabelxshiftc ex, \xlabelyshiftc ex)}, scale=\labelscalec},
    xticklabel style={yshift=-1ex,rotate=\xlabelrotate},
    yticklabel style={xshift=-0.5ex},
    yticklabel style={scale=\ticklabelscalec},
    xticklabel style={scale=\ticklabelscalec},
    ybar=\barsep ex,
   bar width = \barw pt,
    legend cell align={left}
    ]
\addplot[
    fill=clr1fill,draw = clr1,
    very thick,
    line width= \linew pt,
    rounded corners=\barcorner ex
    ]
    coordinates {
(0.20, 0.96)
    };
\addplot[
    fill=clr2fill,draw = clr2,
    very thick,
    line width= \linew pt,
    rounded corners=\barcorner ex
    ]
    coordinates {
(0.65, 0.66)
};
%
 \legend{ORB-SLAM$2$ \cite{orb2}, Jetson-SLAM}
\node [draw=gpudclr,fill=gpuclr,rounded corners=0.2ex, minimum width=23.7ex, xshift=\gpuxshiftb ex, yshift=\gpuyshiftb ex, scale=\gpuscale]{\textcolor{gputxtclr}{\textbf{Jetson-NX}$\mathbf{~@}$~\textbf{Scales}~$\mathbf{=8}$}};
\end{axis}
}
}};

}};

\end{tikzpicture}
\vspace{-4.0ex}
\caption{(a) Output of \OursAcronym's GPU-accelerated and resource-efficient Frontend--Middle-end design, (b) the output trajectory, (c) Frames-Per-Second benchmarking on Jetson-NX embedded computer, and (d) SLAM performance on a KITTI sequence.}
\label{fig:intro}
\vspace{-3.0ex}
\end{figure}

%% file: bounded_rectification.tex
\subsection{Bounded Rectification for Corner Detection}
\label{sec:fast}

%
In FAST \cite{fast} detection, the number of consecutive dark ($N_d$) and bright pixels ($N_b$) are computed for a segment around each image pixel, known as Bresenhem circle \cite{fast} of radius $3$ and a length $N_{seg}=16$ pixels. If there exists $P_{min}$ brighter or darker pixels, the center pixel is labeled as a corner (Eq.~\ref{eq:label}).
\begin{equation}
\footnotesize
L_p =  
\begin{cases}
\texttt{bright},~~~~ I_c-I_p <-\epsilon \\
\texttt{similar},~~  -\epsilon < I_c-I_p <\epsilon \\
\texttt{dark},~~~~~~ I_c-I_p > \epsilon
\end{cases}
\label{eq:label}
\end{equation}
%
%
\begin{equation}
\footnotesize
L_c =  
\begin{cases}
\texttt{Corner},~~~~ N_b,~N_d \geq P_{min} \\
\texttt{Non-corner},~~  \text{otherwise}
\end{cases}
\label{eq:corner}
\end{equation}
where, $L_p$ is the pixel label, $I_c$ and $I_p$ are the intensities of the center pixel and any pixel on the segment respectively. $\epsilon$ is often set to $20$, and $L_c$ denotes the center pixel label.
\par
We note that the above process misclassifies many non-corner points as corners. Such points are essentially statistical outliers having a shot-noise-like or tiny blob-like appearance. Interestingly, they satisfy Eq.~\ref{eq:label},~\ref{eq:corner}, but can not be used as map points due to their highly similar appearance which confuses the stereo-matcher. For instance, all segment pixels in Fig.~\ref{fig:noncorner} are bright and $N_b > P_{min}$, yet it is not a corner. Fig.~\ref{fig:shotnoiseexample} shows a real-world case. 
This issue needs to be resolved since the error in the frontend propagates to the other SLAM components. Thus we aim to discard such points at the detection stage itself because it reduces the number of features reaching the stereo-matcher which lowers the matching time. To this end, we propose a key enhancement called \emph{Bounded Rectification} which not only improves the corner quality but also improves the metric accuracy of \OursAcronym~significantly. 
\paragraph*{Rectification} we propose to rewrite Eq.~\ref{eq:corner} which performs upper and lower bounded rectification (Eq.~\ref{eq:newcorner}).
%
%
%
%
\begin{equation}
\footnotesize
L_c =  
\begin{cases}
\texttt{Corner},~P_{min} \leq N_b,N_d \leq P_{max} \\
\texttt{Non-corner},~~  \text{otherwise}
\end{cases}
\label{eq:newcorner}
\end{equation}
\input{plots/br_circle__non_corners}
This formulation rejects the cases similar to Fig.~\ref{fig:noncorner} by upper and lower bounding $N_b$ or $N_d$ because for the center pixel to be a corner, all of its surrounding pixels can not be \texttt{bright} or \texttt{dark} simultaneously. This can be assured \textit{iff} $P_{max} < 16$, which is the drawback of Eq.~\ref{eq:corner}. $P_{min}$ is generally set to $9$ whereas $P_{max}$ can be set anywhere $\in (P_{min},16)$. We use $P_{max} = 13$ in our case which does not restrict corner diversity but can be adjusted based on the nature of the visual scenes.
\par
Bounded rectification-based FAST detection outputs a Corner Response Matrix (CRF-matrix, Sec.~\ref{sec:fastesetgpu}), which is utilized by our \texttt{PyCA} technique. Next, we brief GPU terminologies for better grasping of the upcoming text.
%
%

%% file: plots/br_circle__non_corners.tex
\begin{figure}[!t]
\centering

\colorlet{bdrwclr}{black!0!cyan}

\begin{tikzpicture}

\FPeval{\xshiftb}{28}
\FPeval{\xshiftc}{82}

\FPeval{\yshiftb}{0}
\FPeval{\yshiftc}{0-0.3}

\FPeval{\bh}{3}
\FPeval{\bw}{3}

\colorlet{bdrwclr}{white!60!black}
\colorlet{cbfillclr}{white!50!black}
\colorlet{bfillclr}{white!80!black}

\colorlet{bdrwclr}{black!0!cyan}
\colorlet{centerbdrwclr}{white!40!red}

\FPeval{\linew}{0.3}
\FPeval{\clinew}{0.5}

\FPeval{\pathradii}{14.5}
\FPeval{\pathlinew}{0.5}
\colorlet{pathclr}{white!40!black}

\FPeval{\scalea}{0.29}
\FPeval{\scaleb}{0.7}
\FPeval{\scalec}{0.7}

\node (a) [scale= \scalea]{
\tikz{
\node [yshift=0ex,scale=1.0]{
\tikz{
\draw[->,line width=\pathlinew ex,pathclr] (0,0) arc (-88:268:\pathradii ex);
}};
\draw[->,line width=\pathlinew ex,pathclr] ($(-7.9 * \bw ex, -7.9*\bh ex)$) -- ($(7.9 * \bw ex, -7.9 *\bh ex)$);
\node (c) [draw=centerbdrwclr, fill=cbfillclr, line width=\clinew ex, minimum height=\bh ex, minimum width=\bw ex]{};
\node (0) [draw=bdrwclr, fill=bfillclr, line width=\linew ex, minimum height=\bh ex, minimum width=\bw ex, xshift = 0*\bw ex, yshift=-3*\bh ex]{$1$};
\node (1) [draw=bdrwclr, fill=bfillclr, line width=\linew ex, minimum height=\bh ex, minimum width=\bw ex, xshift = 1*\bw ex, yshift=-3*\bh ex]{$1$};
\node (2) [draw=bdrwclr, fill=bfillclr, line width=\linew ex, minimum height=\bh ex, minimum width=\bw ex, xshift = 2*\bw ex, yshift=-2*\bh ex]{$1$};
\node (3) [draw=bdrwclr, fill=bfillclr, line width=\linew ex, minimum height=\bh ex, minimum width=\bw ex, xshift = 3*\bw ex, yshift=-1*\bh ex]{$1$};
\node (4) [draw=bdrwclr, fill=bfillclr, line width=\linew ex, minimum height=\bh ex, minimum width=\bw ex, xshift = 3*\bw ex, yshift=0*\bh ex]{$1$};
\node (5) [draw=bdrwclr, fill=bfillclr, line width=\linew ex, minimum height=\bh ex, minimum width=\bw ex, xshift = 3*\bw ex, yshift=1*\bh ex]{$1$};
\node (6) [draw=bdrwclr, fill=bfillclr, line width=\linew ex, minimum height=\bh ex, minimum width=\bw ex, xshift = 2*\bw ex, yshift=2*\bh ex]{$1$};
\node (7) [draw=bdrwclr, fill=bfillclr, line width=\linew ex, minimum height=\bh ex, minimum width=\bw ex, xshift = 1*\bw ex, yshift=3*\bh ex]{$1$};
\node (8) [draw=bdrwclr, fill=bfillclr, line width=\linew ex, minimum height=\bh ex, minimum width=\bw ex, xshift = 0*\bw ex, yshift=3*\bh ex]{$1$};
\node (9) [draw=bdrwclr, fill=bfillclr, line width=\linew ex, minimum height=\bh ex, minimum width=\bw ex, xshift = -1*\bw ex, yshift=3*\bh ex]{$1$};
\node (10) [draw=bdrwclr, fill=bfillclr, line width=\linew ex, minimum height=\bh ex, minimum width=\bw ex, xshift = -2*\bw ex, yshift=2*\bh ex]{$1$};
\node (11) [draw=bdrwclr, fill=bfillclr, line width=\linew ex, minimum height=\bh ex, minimum width=\bw ex, xshift = -3*\bw ex, yshift=1*\bh ex]{$1$};
\node (12) [draw=bdrwclr, fill=bfillclr, line width=\linew ex, minimum height=\bh ex, minimum width=\bw ex, xshift = -3*\bw ex, yshift=0*\bh ex]{$1$};
\node (13) [draw=bdrwclr, fill=bfillclr, line width=\linew ex, minimum height=\bh ex, minimum width=\bw ex, xshift = -3*\bw ex, yshift=-1*\bh ex]{$1$};
\node (14) [draw=bdrwclr, fill=bfillclr, line width=\linew ex, minimum height=\bh ex, minimum width=\bw ex, xshift = -2*\bw ex, yshift=-2*\bh ex]{$1$};
\node (15) [draw=bdrwclr, fill=bfillclr, line width=\linew ex, minimum height=\bh ex, minimum width=\bw ex, xshift = -1*\bw ex, yshift=-3*\bh ex]{$1$};
\node (0) [draw=bdrwclr, fill=bfillclr, line width=\linew ex, minimum height=\bh ex, minimum width=\bw ex, xshift = -7.5*\bw ex, yshift=-6*\bh ex]{$1$};
\node (0) [draw=bdrwclr, fill=bfillclr, line width=\linew ex, minimum height=\bh ex, minimum width=\bw ex, xshift = -6.5*\bw ex, yshift=-6*\bh ex]{$1$};
\node (0) [draw=bdrwclr, fill=bfillclr, line width=\linew ex, minimum height=\bh ex, minimum width=\bw ex, xshift = -5.5*\bw ex, yshift=-6*\bh ex]{$1$};
\node (0) [draw=bdrwclr, fill=bfillclr, line width=\linew ex, minimum height=\bh ex, minimum width=\bw ex, xshift = -4.5*\bw ex, yshift=-6*\bh ex]{$1$};
\node (0) [draw=bdrwclr, fill=bfillclr, line width=\linew ex, minimum height=\bh ex, minimum width=\bw ex, xshift = -3.5*\bw ex, yshift=-6*\bh ex]{$1$};
\node (0) [draw=bdrwclr, fill=bfillclr, line width=\linew ex, minimum height=\bh ex, minimum width=\bw ex, xshift = -2.5*\bw ex, yshift=-6*\bh ex]{$1$};
\node (0) [draw=bdrwclr, fill=bfillclr, line width=\linew ex, minimum height=\bh ex, minimum width=\bw ex, xshift = -1.5*\bw ex, yshift=-6*\bh ex]{$1$};
\node (0) [draw=bdrwclr, fill=bfillclr, line width=\linew ex, minimum height=\bh ex, minimum width=\bw ex, xshift = -0.5*\bw ex, yshift=-6*\bh ex]{$1$};
\node (0) [draw=bdrwclr, fill=bfillclr, line width=\linew ex, minimum height=\bh ex, minimum width=\bw ex, xshift = 0.5*\bw ex, yshift=-6*\bh ex]{$1$};
\node (0) [draw=bdrwclr, fill=bfillclr, line width=\linew ex, minimum height=\bh ex, minimum width=\bw ex, xshift = 1.5*\bw ex, yshift=-6*\bh ex]{$1$};
\node (0) [draw=bdrwclr, fill=bfillclr, line width=\linew ex, minimum height=\bh ex, minimum width=\bw ex, xshift = 2.5*\bw ex, yshift=-6*\bh ex]{$1$};
\node (0) [draw=bdrwclr, fill=bfillclr, line width=\linew ex, minimum height=\bh ex, minimum width=\bw ex, xshift = 3.5*\bw ex, yshift=-6*\bh ex]{$1$};
\node (0) [draw=bdrwclr, fill=bfillclr, line width=\linew ex, minimum height=\bh ex, minimum width=\bw ex, xshift = 4.5*\bw ex, yshift=-6*\bh ex]{$1$};
\node (0) [draw=bdrwclr, fill=bfillclr, line width=\linew ex, minimum height=\bh ex, minimum width=\bw ex, xshift = 5.5*\bw ex, yshift=-6*\bh ex]{$1$};
\node (0) [draw=bdrwclr, fill=bfillclr, line width=\linew ex, minimum height=\bh ex, minimum width=\bw ex, xshift = 6.5*\bw ex, yshift=-6*\bh ex]{$1$};
\node (0) [draw=bdrwclr, fill=bfillclr, line width=\linew ex, minimum height=\bh ex, minimum width=\bw ex, xshift = 7.5*\bw ex, yshift=-6*\bh ex]{$1$};
\node (0) [draw=none, fill=none, minimum height=\bh ex, minimum width=\bw ex, xshift = 0*\bw ex, yshift=-4*\bh ex]{$0$};
\node (1) [draw=none, fill=none, minimum height=\bh ex, minimum width=\bw ex, xshift = 1*\bw ex, yshift=-4*\bh ex]{$1$};
\node (2) [draw=none, fill=none, minimum height=\bh ex, minimum width=\bw ex, xshift = 3*\bw ex, yshift=-3*\bh ex]{$2$};
\node (3) [draw=none, fill=none, minimum height=\bh ex, minimum width=\bw ex, xshift = 4*\bw ex, yshift=-1*\bh ex]{$3$};
\node (4) [draw=none, fill=none, minimum height=\bh ex, minimum width=\bw ex, xshift = 4*\bw ex, yshift=0*\bh ex]{$4$};
\node (5) [draw=none, fill=none, minimum height=\bh ex, minimum width=\bw ex, xshift = 4*\bw ex, yshift=1*\bh ex]{$5$};
\node (6) [draw=none, fill=none, minimum height=\bh ex, minimum width=\bw ex, xshift = 3*\bw ex, yshift=3*\bh ex]{$6$};
\node (7) [draw=none, fill=none, minimum height=\bh ex, minimum width=\bw ex, xshift = 1*\bw ex, yshift=4*\bh ex]{$7$};
\node (8) [draw=none, fill=none, minimum height=\bh ex, minimum width=\bw ex, xshift = 0*\bw ex, yshift=4*\bh ex]{$8$};
\node (9) [draw=none, fill=none, minimum height=\bh ex, minimum width=\bw ex, xshift = -1*\bw ex, yshift=4*\bh ex]{$9$};
\node (10) [draw=none, fill=none, minimum height=\bh ex, minimum width=\bw ex, xshift = -3*\bw ex, yshift=3*\bh ex]{$10$};
\node (11) [draw=none, fill=none, minimum height=\bh ex, minimum width=\bw ex, xshift = -4*\bw ex, yshift=1*\bh ex]{$11$};
\node (12) [draw=none, fill=none, minimum height=\bh ex, minimum width=\bw ex, xshift = -4*\bw ex, yshift=0*\bh ex]{$12$};
\node (13) [draw=none, fill=none, minimum height=\bh ex, minimum width=\bw ex, xshift = -4*\bw ex, yshift=-1*\bh ex]{$13$};
\node (14) [draw=none, fill=none, minimum height=\bh ex, minimum width=\bw ex, xshift = -3*\bw ex, yshift=-3*\bh ex]{$14$};
\node (15) [draw=none, fill=none, minimum height=\bh ex, minimum width=\bw ex, xshift = -1*\bw ex, yshift=-4*\bh ex]{$15$};
\node (0) [draw=none, fill=none, minimum height=\bh ex, minimum width=\bw ex, xshift = -7.5*\bw ex, yshift=-7*\bh ex]{$0$};
\node (0) [draw=none, fill=none, minimum height=\bh ex, minimum width=\bw ex, xshift = -6.5*\bw ex, yshift=-7*\bh ex]{$1$};
\node (0) [draw=none, fill=none, minimum height=\bh ex, minimum width=\bw ex, xshift = -5.5*\bw ex, yshift=-7*\bh ex]{$2$};
\node (0) [draw=none, fill=none, minimum height=\bh ex, minimum width=\bw ex, xshift = -4.5*\bw ex, yshift=-7*\bh ex]{$3$};
\node (0) [draw=none, fill=none, minimum height=\bh ex, minimum width=\bw ex, xshift = -3.5*\bw ex, yshift=-7*\bh ex]{$4$};
\node (0) [draw=none, fill=none, minimum height=\bh ex, minimum width=\bw ex, xshift = -2.5*\bw ex, yshift=-7*\bh ex]{$5$};
\node (0) [draw=none, fill=none, minimum height=\bh ex, minimum width=\bw ex, xshift = -1.5*\bw ex, yshift=-7*\bh ex]{$6$};
\node (0) [draw=none, fill=none, minimum height=\bh ex, minimum width=\bw ex, xshift = -0.5*\bw ex, yshift=-7*\bh ex]{$7$};
\node (0) [draw=none, fill=none, minimum height=\bh ex, minimum width=\bw ex, xshift = 0.5*\bw ex, yshift=-7*\bh ex]{$8$};
\node (0) [draw=none, fill=none, minimum height=\bh ex, minimum width=\bw ex, xshift = 1.5*\bw ex, yshift=-7*\bh ex]{$9$};
\node (0) [draw=none, fill=none, minimum height=\bh ex, minimum width=\bw ex, xshift = 2.5*\bw ex, yshift=-7*\bh ex]{$10$};
\node (0) [draw=none, fill=none, minimum height=\bh ex, minimum width=\bw ex, xshift = 3.5*\bw ex, yshift=-7*\bh ex]{$11$};
\node (0) [draw=none, fill=none, minimum height=\bh ex, minimum width=\bw ex, xshift = 4.5*\bw ex, yshift=-7*\bh ex]{$12$};
\node (0) [draw=none, fill=none, minimum height=\bh ex, minimum width=\bw ex, xshift = 5.5*\bw ex, yshift=-7*\bh ex]{$13$};
\node (0) [draw=none, fill=none, minimum height=\bh ex, minimum width=\bw ex, xshift = 6.5*\bw ex, yshift=-7*\bh ex]{$14$};
\node (0) [draw=none, fill=none, minimum height=\bh ex, minimum width=\bw ex, xshift = 7.5*\bw ex, yshift=-7*\bh ex]{$15$};
}};

%
%
%
%
%

%
%
\FPeval{\bh}{3.20}
\FPeval{\bw}{3.20}

\FPeval{\imw}{7.5}
\FPeval{\imh}{7.5}

\FPeval{\linew}{0.3}
\FPeval{\clinew}{0.5}

\FPeval{\circscale}{0.335}

\colorlet{bdrwclr}{black!0!cyan}
\colorlet{centerbdrwclr}{white!40!red}

\colorlet{cbfillclr}{white!65!black}
\colorlet{bfillclr}{white!90!black}
\colorlet{imnoiseboundaryclr}{white!90!red}
\colorlet{imboundaryclr}{white!100!white}

\node (b) [xshift=\xshiftb ex, yshift=\yshiftb ex, scale=\scaleb]
{
\tikz{
\node () [xshift=0ex + 19.25ex, yshift=0ex]
{
\tikz{
\node (n1) [xshift=0ex + 0ex, yshift=0ex]{\includegraphics[width=\imw ex, height=\imh ex]{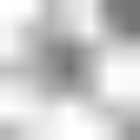}};
\node ()[draw=imnoiseboundaryclr,minimum width=\imw ex, minimum height=\imh ex]{};
\node (1) [scale= \circscale]{
\tikz{
\node (c) [draw=centerbdrwclr, fill=none,line width=\clinew ex, minimum height=\bh ex, minimum width=\bw ex]{};
\node (0) [draw=bdrwclr, fill=none,line width=\linew ex, minimum height=\bh ex, minimum width=\bw ex, xshift = 0*\bw ex, yshift=-3*\bh ex]{};
\node (1) [draw=bdrwclr, fill=none,line width=\linew ex, minimum height=\bh ex, minimum width=\bw ex, xshift = 1*\bw ex, yshift=-3*\bh ex]{};
\node (2) [draw=bdrwclr, fill=none,line width=\linew ex, minimum height=\bh ex, minimum width=\bw ex, xshift = 2*\bw ex, yshift=-2*\bh ex]{};
\node (3) [draw=bdrwclr, fill=none,line width=\linew ex, minimum height=\bh ex, minimum width=\bw ex, xshift = 3*\bw ex, yshift=-1*\bh ex]{};
\node (4) [draw=bdrwclr, fill=none,line width=\linew ex, minimum height=\bh ex, minimum width=\bw ex, xshift = 3*\bw ex, yshift=0*\bh ex]{};
\node (5) [draw=bdrwclr, fill=none,line width=\linew ex, minimum height=\bh ex, minimum width=\bw ex, xshift = 3*\bw ex, yshift=1*\bh ex]{};
\node (6) [draw=bdrwclr, fill=none,line width=\linew ex, minimum height=\bh ex, minimum width=\bw ex, xshift = 2*\bw ex, yshift=2*\bh ex]{};
\node (7) [draw=bdrwclr, fill=none,line width=\linew ex, minimum height=\bh ex, minimum width=\bw ex, xshift = 1*\bw ex, yshift=3*\bh ex]{};
\node (8) [draw=bdrwclr, fill=none,line width=\linew ex, minimum height=\bh ex, minimum width=\bw ex, xshift = 0*\bw ex, yshift=3*\bh ex]{};
\node (9) [draw=bdrwclr, fill=none,line width=\linew ex, minimum height=\bh ex, minimum width=\bw ex, xshift = -1*\bw ex, yshift=3*\bh ex]{};
\node (10) [draw=bdrwclr, fill=none,line width=\linew ex, minimum height=\bh ex, minimum width=\bw ex, xshift = -2*\bw ex, yshift=2*\bh ex]{};
\node (11) [draw=bdrwclr, fill=none,line width=\linew ex, minimum height=\bh ex, minimum width=\bw ex, xshift = -3*\bw ex, yshift=1*\bh ex]{};
\node (12) [draw=bdrwclr, fill=none,line width=\linew ex, minimum height=\bh ex, minimum width=\bw ex, xshift = -3*\bw ex, yshift=0*\bh ex]{};
\node (13) [draw=bdrwclr, fill=none,line width=\linew ex, minimum height=\bh ex, minimum width=\bw ex, xshift = -3*\bw ex, yshift=-1*\bh ex]{};
\node (14) [draw=bdrwclr, fill=none,line width=\linew ex, minimum height=\bh ex, minimum width=\bw ex, xshift = -2*\bw ex, yshift=-2*\bh ex]{};
\node (15) [draw=bdrwclr, fill=none,line width=\linew ex, minimum height=\bh ex, minimum width=\bw ex, xshift = -1*\bw ex, yshift=-3*\bh ex]{};
}};
}};
\node () [xshift=\imw ex + 0.25ex + 19.25ex, yshift=0ex]
{
\tikz{
\node (n1) []{\includegraphics[width=\imw ex, height=\imh ex]{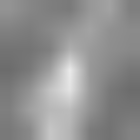}};
\node ()[draw=imnoiseboundaryclr,minimum width=\imw ex, minimum height=\imh ex]{};
\node (1) [scale= \circscale]{
\tikz{
\node (c) [draw=centerbdrwclr, fill=none,line width=\clinew ex, minimum height=\bh ex, minimum width=\bw ex]{};
\node (0) [draw=bdrwclr, fill=none,line width=\linew ex, minimum height=\bh ex, minimum width=\bw ex, xshift = 0*\bw ex, yshift=-3*\bh ex]{};
\node (1) [draw=bdrwclr, fill=none,line width=\linew ex, minimum height=\bh ex, minimum width=\bw ex, xshift = 1*\bw ex, yshift=-3*\bh ex]{};
\node (2) [draw=bdrwclr, fill=none,line width=\linew ex, minimum height=\bh ex, minimum width=\bw ex, xshift = 2*\bw ex, yshift=-2*\bh ex]{};
\node (3) [draw=bdrwclr, fill=none,line width=\linew ex, minimum height=\bh ex, minimum width=\bw ex, xshift = 3*\bw ex, yshift=-1*\bh ex]{};
\node (4) [draw=bdrwclr, fill=none,line width=\linew ex, minimum height=\bh ex, minimum width=\bw ex, xshift = 3*\bw ex, yshift=0*\bh ex]{};
\node (5) [draw=bdrwclr, fill=none,line width=\linew ex, minimum height=\bh ex, minimum width=\bw ex, xshift = 3*\bw ex, yshift=1*\bh ex]{};
\node (6) [draw=bdrwclr, fill=none,line width=\linew ex, minimum height=\bh ex, minimum width=\bw ex, xshift = 2*\bw ex, yshift=2*\bh ex]{};
\node (7) [draw=bdrwclr, fill=none,line width=\linew ex, minimum height=\bh ex, minimum width=\bw ex, xshift = 1*\bw ex, yshift=3*\bh ex]{};
\node (8) [draw=bdrwclr, fill=none,line width=\linew ex, minimum height=\bh ex, minimum width=\bw ex, xshift = 0*\bw ex, yshift=3*\bh ex]{};
\node (9) [draw=bdrwclr, fill=none,line width=\linew ex, minimum height=\bh ex, minimum width=\bw ex, xshift = -1*\bw ex, yshift=3*\bh ex]{};
\node (10) [draw=bdrwclr, fill=none,line width=\linew ex, minimum height=\bh ex, minimum width=\bw ex, xshift = -2*\bw ex, yshift=2*\bh ex]{};
\node (11) [draw=bdrwclr, fill=none,line width=\linew ex, minimum height=\bh ex, minimum width=\bw ex, xshift = -3*\bw ex, yshift=1*\bh ex]{};
\node (12) [draw=bdrwclr, fill=none,line width=\linew ex, minimum height=\bh ex, minimum width=\bw ex, xshift = -3*\bw ex, yshift=0*\bh ex]{};
\node (13) [draw=bdrwclr, fill=none,line width=\linew ex, minimum height=\bh ex, minimum width=\bw ex, xshift = -3*\bw ex, yshift=-1*\bh ex]{};
\node (14) [draw=bdrwclr, fill=none,line width=\linew ex, minimum height=\bh ex, minimum width=\bw ex, xshift = -2*\bw ex, yshift=-2*\bh ex]{};
\node (15) [draw=bdrwclr, fill=none,line width=\linew ex, minimum height=\bh ex, minimum width=\bw ex, xshift = -1*\bw ex, yshift=-3*\bh ex]{};
}};
}};
\node () [xshift=2*\imw ex + 0.5ex + 19.25ex, yshift=0ex]
{
\tikz{
\node (n1) []{\includegraphics[width=\imw ex, height=\imh ex]{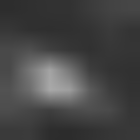}};
\node ()[draw=imnoiseboundaryclr,minimum width=\imw ex, minimum height=\imh ex]{};
\node (1) [scale= \circscale]{
\tikz{
\node (c) [draw=centerbdrwclr, fill=none,line width=\clinew ex, minimum height=\bh ex, minimum width=\bw ex]{};
\node (0) [draw=bdrwclr, fill=none,line width=\linew ex, minimum height=\bh ex, minimum width=\bw ex, xshift = 0*\bw ex, yshift=-3*\bh ex]{};
\node (1) [draw=bdrwclr, fill=none,line width=\linew ex, minimum height=\bh ex, minimum width=\bw ex, xshift = 1*\bw ex, yshift=-3*\bh ex]{};
\node (2) [draw=bdrwclr, fill=none,line width=\linew ex, minimum height=\bh ex, minimum width=\bw ex, xshift = 2*\bw ex, yshift=-2*\bh ex]{};
\node (3) [draw=bdrwclr, fill=none,line width=\linew ex, minimum height=\bh ex, minimum width=\bw ex, xshift = 3*\bw ex, yshift=-1*\bh ex]{};
\node (4) [draw=bdrwclr, fill=none,line width=\linew ex, minimum height=\bh ex, minimum width=\bw ex, xshift = 3*\bw ex, yshift=0*\bh ex]{};
\node (5) [draw=bdrwclr, fill=none,line width=\linew ex, minimum height=\bh ex, minimum width=\bw ex, xshift = 3*\bw ex, yshift=1*\bh ex]{};
\node (6) [draw=bdrwclr, fill=none,line width=\linew ex, minimum height=\bh ex, minimum width=\bw ex, xshift = 2*\bw ex, yshift=2*\bh ex]{};
\node (7) [draw=bdrwclr, fill=none,line width=\linew ex, minimum height=\bh ex, minimum width=\bw ex, xshift = 1*\bw ex, yshift=3*\bh ex]{};
\node (8) [draw=bdrwclr, fill=none,line width=\linew ex, minimum height=\bh ex, minimum width=\bw ex, xshift = 0*\bw ex, yshift=3*\bh ex]{};
\node (9) [draw=bdrwclr, fill=none,line width=\linew ex, minimum height=\bh ex, minimum width=\bw ex, xshift = -1*\bw ex, yshift=3*\bh ex]{};
\node (10) [draw=bdrwclr, fill=none,line width=\linew ex, minimum height=\bh ex, minimum width=\bw ex, xshift = -2*\bw ex, yshift=2*\bh ex]{};
\node (11) [draw=bdrwclr, fill=none,line width=\linew ex, minimum height=\bh ex, minimum width=\bw ex, xshift = -3*\bw ex, yshift=1*\bh ex]{};
\node (12) [draw=bdrwclr, fill=none,line width=\linew ex, minimum height=\bh ex, minimum width=\bw ex, xshift = -3*\bw ex, yshift=0*\bh ex]{};
\node (13) [draw=bdrwclr, fill=none,line width=\linew ex, minimum height=\bh ex, minimum width=\bw ex, xshift = -3*\bw ex, yshift=-1*\bh ex]{};
\node (14) [draw=bdrwclr, fill=none,line width=\linew ex, minimum height=\bh ex, minimum width=\bw ex, xshift = -2*\bw ex, yshift=-2*\bh ex]{};
\node (15) [draw=bdrwclr, fill=none,line width=\linew ex, minimum height=\bh ex, minimum width=\bw ex, xshift = -1*\bw ex, yshift=-3*\bh ex]{};
}};
}};
\node () [xshift=0ex + 19.25ex, yshift=-\imh ex - 0.35ex]
{
\tikz{
\node (n1) []{\includegraphics[width=\imw ex, height=\imh ex]{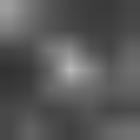}};
\node ()[draw=imnoiseboundaryclr,minimum width=\imw ex, minimum height=\imh ex]{};
\node (1) [scale= \circscale]{
\tikz{
\node (c) [draw=centerbdrwclr, fill=none,line width=\clinew ex, minimum height=\bh ex, minimum width=\bw ex]{};
\node (0) [draw=bdrwclr, fill=none,line width=\linew ex, minimum height=\bh ex, minimum width=\bw ex, xshift = 0*\bw ex, yshift=-3*\bh ex]{};
\node (1) [draw=bdrwclr, fill=none,line width=\linew ex, minimum height=\bh ex, minimum width=\bw ex, xshift = 1*\bw ex, yshift=-3*\bh ex]{};
\node (2) [draw=bdrwclr, fill=none,line width=\linew ex, minimum height=\bh ex, minimum width=\bw ex, xshift = 2*\bw ex, yshift=-2*\bh ex]{};
\node (3) [draw=bdrwclr, fill=none,line width=\linew ex, minimum height=\bh ex, minimum width=\bw ex, xshift = 3*\bw ex, yshift=-1*\bh ex]{};
\node (4) [draw=bdrwclr, fill=none,line width=\linew ex, minimum height=\bh ex, minimum width=\bw ex, xshift = 3*\bw ex, yshift=0*\bh ex]{};
\node (5) [draw=bdrwclr, fill=none,line width=\linew ex, minimum height=\bh ex, minimum width=\bw ex, xshift = 3*\bw ex, yshift=1*\bh ex]{};
\node (6) [draw=bdrwclr, fill=none,line width=\linew ex, minimum height=\bh ex, minimum width=\bw ex, xshift = 2*\bw ex, yshift=2*\bh ex]{};
\node (7) [draw=bdrwclr, fill=none,line width=\linew ex, minimum height=\bh ex, minimum width=\bw ex, xshift = 1*\bw ex, yshift=3*\bh ex]{};
\node (8) [draw=bdrwclr, fill=none,line width=\linew ex, minimum height=\bh ex, minimum width=\bw ex, xshift = 0*\bw ex, yshift=3*\bh ex]{};
\node (9) [draw=bdrwclr, fill=none,line width=\linew ex, minimum height=\bh ex, minimum width=\bw ex, xshift = -1*\bw ex, yshift=3*\bh ex]{};
\node (10) [draw=bdrwclr, fill=none,line width=\linew ex, minimum height=\bh ex, minimum width=\bw ex, xshift = -2*\bw ex, yshift=2*\bh ex]{};
\node (11) [draw=bdrwclr, fill=none,line width=\linew ex, minimum height=\bh ex, minimum width=\bw ex, xshift = -3*\bw ex, yshift=1*\bh ex]{};
\node (12) [draw=bdrwclr, fill=none,line width=\linew ex, minimum height=\bh ex, minimum width=\bw ex, xshift = -3*\bw ex, yshift=0*\bh ex]{};
\node (13) [draw=bdrwclr, fill=none,line width=\linew ex, minimum height=\bh ex, minimum width=\bw ex, xshift = -3*\bw ex, yshift=-1*\bh ex]{};
\node (14) [draw=bdrwclr, fill=none,line width=\linew ex, minimum height=\bh ex, minimum width=\bw ex, xshift = -2*\bw ex, yshift=-2*\bh ex]{};
\node (15) [draw=bdrwclr, fill=none,line width=\linew ex, minimum height=\bh ex, minimum width=\bw ex, xshift = -1*\bw ex, yshift=-3*\bh ex]{};
}};
}};
\node () [xshift=\imw ex + 0.25ex + 19.25ex, yshift=-\imh ex - 0.35ex]
{
\tikz{
\node (n1) []{\includegraphics[width=\imw ex, height=\imh ex]{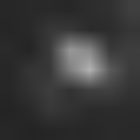}};
\node ()[draw=imnoiseboundaryclr,minimum width=\imw ex, minimum height=\imh ex]{};
\node (1) [scale= \circscale]{
\tikz{
\node (c) [draw=centerbdrwclr, fill=none,line width=\clinew ex, minimum height=\bh ex, minimum width=\bw ex]{};
\node (0) [draw=bdrwclr, fill=none,line width=\linew ex, minimum height=\bh ex, minimum width=\bw ex, xshift = 0*\bw ex, yshift=-3*\bh ex]{};
\node (1) [draw=bdrwclr, fill=none,line width=\linew ex, minimum height=\bh ex, minimum width=\bw ex, xshift = 1*\bw ex, yshift=-3*\bh ex]{};
\node (2) [draw=bdrwclr, fill=none,line width=\linew ex, minimum height=\bh ex, minimum width=\bw ex, xshift = 2*\bw ex, yshift=-2*\bh ex]{};
\node (3) [draw=bdrwclr, fill=none,line width=\linew ex, minimum height=\bh ex, minimum width=\bw ex, xshift = 3*\bw ex, yshift=-1*\bh ex]{};
\node (4) [draw=bdrwclr, fill=none,line width=\linew ex, minimum height=\bh ex, minimum width=\bw ex, xshift = 3*\bw ex, yshift=0*\bh ex]{};
\node (5) [draw=bdrwclr, fill=none,line width=\linew ex, minimum height=\bh ex, minimum width=\bw ex, xshift = 3*\bw ex, yshift=1*\bh ex]{};
\node (6) [draw=bdrwclr, fill=none,line width=\linew ex, minimum height=\bh ex, minimum width=\bw ex, xshift = 2*\bw ex, yshift=2*\bh ex]{};
\node (7) [draw=bdrwclr, fill=none,line width=\linew ex, minimum height=\bh ex, minimum width=\bw ex, xshift = 1*\bw ex, yshift=3*\bh ex]{};
\node (8) [draw=bdrwclr, fill=none,line width=\linew ex, minimum height=\bh ex, minimum width=\bw ex, xshift = 0*\bw ex, yshift=3*\bh ex]{};
\node (9) [draw=bdrwclr, fill=none,line width=\linew ex, minimum height=\bh ex, minimum width=\bw ex, xshift = -1*\bw ex, yshift=3*\bh ex]{};
\node (10) [draw=bdrwclr, fill=none,line width=\linew ex, minimum height=\bh ex, minimum width=\bw ex, xshift = -2*\bw ex, yshift=2*\bh ex]{};
\node (11) [draw=bdrwclr, fill=none,line width=\linew ex, minimum height=\bh ex, minimum width=\bw ex, xshift = -3*\bw ex, yshift=1*\bh ex]{};
\node (12) [draw=bdrwclr, fill=none,line width=\linew ex, minimum height=\bh ex, minimum width=\bw ex, xshift = -3*\bw ex, yshift=0*\bh ex]{};
\node (13) [draw=bdrwclr, fill=none,line width=\linew ex, minimum height=\bh ex, minimum width=\bw ex, xshift = -3*\bw ex, yshift=-1*\bh ex]{};
\node (14) [draw=bdrwclr, fill=none,line width=\linew ex, minimum height=\bh ex, minimum width=\bw ex, xshift = -2*\bw ex, yshift=-2*\bh ex]{};
\node (15) [draw=bdrwclr, fill=none,line width=\linew ex, minimum height=\bh ex, minimum width=\bw ex, xshift = -1*\bw ex, yshift=-3*\bh ex]{};
}};
}};
\node () [xshift=2*\imw ex + 0.5ex + 19.25ex, yshift=-\imh ex - 0.35ex]
{
\tikz{
\node (n1) []{\includegraphics[width=\imw ex, height=\imh ex]{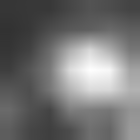}};
\node ()[draw=imnoiseboundaryclr,minimum width=\imw ex, minimum height=\imh ex]{};
\node (1) [scale= \circscale]{
\tikz{
\node (c) [draw=centerbdrwclr, fill=none,line width=\clinew ex, minimum height=\bh ex, minimum width=\bw ex]{};
\node (0) [draw=bdrwclr, fill=none,line width=\linew ex, minimum height=\bh ex, minimum width=\bw ex, xshift = 0*\bw ex, yshift=-3*\bh ex]{};
\node (1) [draw=bdrwclr, fill=none,line width=\linew ex, minimum height=\bh ex, minimum width=\bw ex, xshift = 1*\bw ex, yshift=-3*\bh ex]{};
\node (2) [draw=bdrwclr, fill=none,line width=\linew ex, minimum height=\bh ex, minimum width=\bw ex, xshift = 2*\bw ex, yshift=-2*\bh ex]{};
\node (3) [draw=bdrwclr, fill=none,line width=\linew ex, minimum height=\bh ex, minimum width=\bw ex, xshift = 3*\bw ex, yshift=-1*\bh ex]{};
\node (4) [draw=bdrwclr, fill=none,line width=\linew ex, minimum height=\bh ex, minimum width=\bw ex, xshift = 3*\bw ex, yshift=0*\bh ex]{};
\node (5) [draw=bdrwclr, fill=none,line width=\linew ex, minimum height=\bh ex, minimum width=\bw ex, xshift = 3*\bw ex, yshift=1*\bh ex]{};
\node (6) [draw=bdrwclr, fill=none,line width=\linew ex, minimum height=\bh ex, minimum width=\bw ex, xshift = 2*\bw ex, yshift=2*\bh ex]{};
\node (7) [draw=bdrwclr, fill=none,line width=\linew ex, minimum height=\bh ex, minimum width=\bw ex, xshift = 1*\bw ex, yshift=3*\bh ex]{};
\node (8) [draw=bdrwclr, fill=none,line width=\linew ex, minimum height=\bh ex, minimum width=\bw ex, xshift = 0*\bw ex, yshift=3*\bh ex]{};
\node (9) [draw=bdrwclr, fill=none,line width=\linew ex, minimum height=\bh ex, minimum width=\bw ex, xshift = -1*\bw ex, yshift=3*\bh ex]{};
\node (10) [draw=bdrwclr, fill=none,line width=\linew ex, minimum height=\bh ex, minimum width=\bw ex, xshift = -2*\bw ex, yshift=2*\bh ex]{};
\node (11) [draw=bdrwclr, fill=none,line width=\linew ex, minimum height=\bh ex, minimum width=\bw ex, xshift = -3*\bw ex, yshift=1*\bh ex]{};
\node (12) [draw=bdrwclr, fill=none,line width=\linew ex, minimum height=\bh ex, minimum width=\bw ex, xshift = -3*\bw ex, yshift=0*\bh ex]{};
\node (13) [draw=bdrwclr, fill=none,line width=\linew ex, minimum height=\bh ex, minimum width=\bw ex, xshift = -3*\bw ex, yshift=-1*\bh ex]{};
\node (14) [draw=bdrwclr, fill=none,line width=\linew ex, minimum height=\bh ex, minimum width=\bw ex, xshift = -2*\bw ex, yshift=-2*\bh ex]{};
\node (15) [draw=bdrwclr, fill=none,line width=\linew ex, minimum height=\bh ex, minimum width=\bw ex, xshift = -1*\bw ex, yshift=-3*\bh ex]{};
}};
}};
\node (n1) [xshift=-0.25ex, yshift=-3.95ex]{
\tikz{
\FPeval{\imw}{31}
\FPeval{\imh}{15.5}
%
\node ()[]{\includegraphics[width= \imw ex, height=\imh ex]{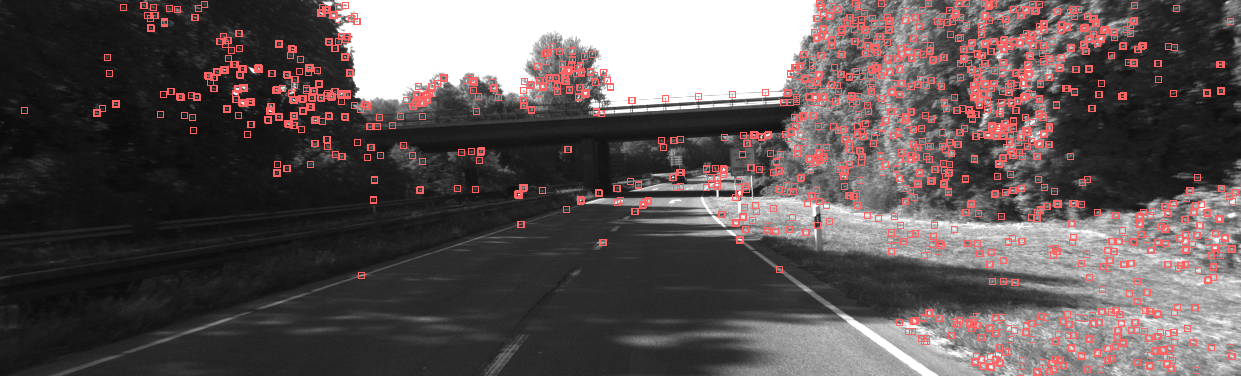}};
\node ()[draw=imboundaryclr,minimum width=\imw ex, minimum height=\imh ex]{};
}};
\node (a) [draw=none, xshift = -0 ex, yshift=-13.5 ex]{\footnotesize Image from seq. KITTI-$01$, Frame-$76$};
\node (b) [draw=none, xshift = 27.0 ex, yshift=-13.5 ex]{\footnotesize Non-corner examples};
}};
%
%
%
%
\node (a) [draw=none, xshift = -6.0 ex, yshift=-2 ex]{\footnotesize (a)};
\node (b) [draw=none, xshift = 7.8 ex, yshift=-2 ex]{\footnotesize (b)};

\end{tikzpicture}
\subfloat{\label{fig:noncorner}}
\subfloat{\label{fig:shotnoiseexample}}
\vspace{-3.0ex}
\newcommand{\rectbox}{\raisebox{0.3ex}{\tikz{\node (c0) [draw=bdrwclr, fill=none, minimum width = 1 ex, minimum height=1 ex,scale=0.6]{};}}}
\caption{(a) A non-corner but all bright pixels on its Bresenhem circle \cite{fast}, (b) Real-image examples of such points tagged as a corner. Each rectangle (\protect\rectbox) denotes a pixel in a $7\times 7$ patch of the image.}
\vspace{-3.5ex}
\end{figure}
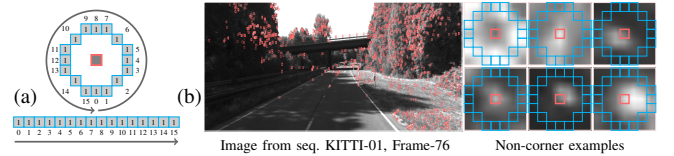

%% file: pyca_fc.tex
%
\subsection{GPU Fundamentals}
NVIDIA GPUs comprise \emph{streaming multiprocessors} (SM), each having multiple GPU \emph{cores}, and an on-chip \textit{shared memory} with low memory access cost (MAC). Off-chip \textit{global memory} is also present but has a higher MAC. A GPU performs computing in \emph{warp} consisting of $32$ \emph{threads} that are concurrently executed on an SM. A \emph{block} comprises many such warps which all are executed on the same SM even if the other SMs are sitting idle. It is so because the threads of a block residing onto an SM may need to communicate with each other, and if warps are executed on different SMs, communication can only be achieved via global memory which has higher MAC, in contrast to the shared memory which has lower MAC but is inaccessible to the other SMs.
\subsection{Pyramidal Culling and Aggregation (\texttt{PyCA})}
\label{sec:pyca}
\texttt{PyCA} detects robust features at high speeds (in $\mu$S) on GPU via its Feature Culling (\texttt{FC}) and  Pyramidal Feature Aggregation (\texttt{PFA}) steps. \texttt{FC} harvests the strongest corners by suppressing the weaker ones, mimicking a culling behaviour, while \texttt{PFA} harvests robust features from the output of \texttt{FC} applied at multiple scales. \texttt{PyCA} remains aware of GPU core scarcity, which, if left unaddressed, reduces the throughput. 
%
%
%
\input{plots/fc_figure}
\subsubsection{Feature Culling (\texttt{FC})} It divides the CRF-matrix into non-overlapping cells of a size ($c_h, c_w$) pixels, and then performs vertical, and horizontal culling in each cell (Fig.~\ref{fig:featureculling}).
\paragraph{Vertical Feature Culling}
In this step, a cell is traversed vertically, and the maximum response is recorded for each column. To achieve that, we propose Multi-Location Per Thread (\texttt{MLPT}) culling in which we divide the vertical cell dimension $c_h$ into chunks (Multi-Location), and then process each of them with a single thread (Per-Thread). The total number of threads required in this process is given by:
\vspace{-0.25ex}
\begin{equation}
\footnotesize
N_t = \min(1,~\lceil c_h/N_{max} \rceil),
\vspace{-0.25ex}
\end{equation}
where $N_{max}$ is the maximum location that a single thread processes. It is adjustable per the needs. Based on our analysis, it can be set in the range $[1,10]$ for common cell sizes. 
\par
Now each of the $N_t$ threads stores its result in shared memory which is utilized by the first thread (T$_{0j}$) of each column. The thread T$_{0j}$ finds the strongest response among $N_t$ values and stores the result for horizontal feature culling.
\par
For the above operation, $\log_2$-reduction \cite{cudareduction} can also be used, however, it processes only two locations per thread, thus requiring a huge number of threads per cell which incurs GPU kernel-launch overhead. Moreover, it is repeated $\lceil \log_2(c_h) \rceil$ times which increases the number of warps required and thus increased runtime to process multiple cells. Also, culling an entire column using one thread becomes slower and inefficient for larger cells in a scenario of core scarcity. 
\par
In contrast, \texttt{MLPT} avoids these issues by using fewer threads regardless of the cell size and also conforms to memory coalescing (Fig.~\ref{fig:memorycoalescing}). This can be verified via Table~\ref{tab:mlpt}, that $\log_2$-reduction requires more warps which induce overhead, even for one cell on Jetson-NX-like embedded devices having core scarcity (Table~\ref{tab:specs}). Since multiple cells exist in an image, \texttt{MLPT} can run faster by the order of milliseconds.
\input{plots/mem_coalescing}
\input{tables/mlpt_vs_log2}
\input{plots/tewa_scheme}
\paragraph{Horizontal Feature Culling}
Now feature culling is performed in the horizontal direction over $c_w$ responses, stored in the shared memory by the previous step. In this case, memory is consecutive, therefore memory coalescing can not be achieved, but memory transactions can be minimized. To achieve that, a thread in \texttt{MLPT} always accesses an element at a stride of $N_{max}$ instead of a consecutive location, unlike vertical culling. This combines the memory transactions of $N_t$ threads of a warp into one. The remaining process is the same as the previous i.e. \texttt{MLPT} performed over $c_w$ locations which produce the strongest corner in the cell if present. 
\input{plots/pfa_figure}
\paragraph{Thread Efficient Warp-Allocation (\texttt{TEWA})}
\label{sec:tewa}
A na\"ive way to allocate GPU for \texttt{FC} is to assign a block size equal to the cell size. However, a job requiring less than $32$ warp threads ($W$) leads to a wastage of leftover threads since they do not involve in the computations but are still part of a warp. This leads to poor throughput on Jetson devices due to core scarcity. The multiscale scenario becomes more challenging due to smaller cells at lower scales. For instance, a block consists of $9$ threads for a $3 \times 3$ cell which if executed in a warp, will lead to a wastage of $23$ threads, indicating severely low \emph{warp-efficiency}, defined as:
\vspace{-1ex}
\begin{equation}
\footnotesize
\eta_w = \frac{N_{ta}}{WN_w}
\label{eq:warpefficiency}
\end{equation}
where, $N_{ta}, N_w, W$ are the number of active threads, number of warps required, and threads per warp respectively. 
\par
Hence, we propose a thread-efficient warp allocation scheme (\texttt{TEWA}) that offers high warp efficiency regardless of the cell size. In this scheme, we assign multiple cells to a single thread block by virtually partitioning the warps into chunks, each chunk handling one cell. We fix $x$ dimension (\texttt{block.x}) of the thread block to $128$, and find the maximum number of the cells that can be fit into \texttt{block.x}, while $y$ dimension (\texttt{block.y}) of the block is set to $N_t$. Together both of these numbers are utilized to launch the GPU kernels. It might be the case that a few threads in the block might still sit idle because the warp-size $W$ is not an integer multiple of cell-size, however, the wastage is minimal in \texttt{TEWA}.
\par
Fig.~\ref{fig:warpalloc} depicts the \texttt{TEWA} design, where, \texttt{FC} for the entire CRF-matrix occurs in one CUDA kernel-launch call that avoids additional launch overhead. Moreover, multiple cells are processed in a single warp which reduces the number of warps and cores required. This leads to high throughput in \texttt{FC} contrary to the na\"ive allocation (Table~\ref{tab:warpefficiency}), eventually reducing the runtime and latency, even in GPU core scarcity.

\input{tables/tewa}

%% file: plots/fc_figure.tex
\begin{figure}[!t]
\centering
\colorlet{thrdclr}{black!20!cyan}
\colorlet{idlethrdclr}{black!20!magenta}
\colorlet{warpclr}{white!80!white}
\colorlet{warpdclr}{white!20!gray}
\colorlet{gridclr}{white!40!cyan}
\colorlet{minigriddclr}{white!80!black}
\colorlet{minigridclr}{white!100!black}
\colorlet{pixeldotclr}{green!50!black}
\colorlet{pixeldotscale}{green!50!black}
\colorlet{threadclr}{white!90!black}
\FPeval{\bw}{3.0}
\FPeval{\bh}{3.0}
\FPeval{\w}{4}
\FPeval{\h}{5}
\FPeval{\pixelcornerradii}{0.4}
\FPeval{\pixeldotscale}{0.15}
\FPeval{\thtop}{1.9}
\FPeval{\scal}{0.71}
\begin{tikzpicture}

\node (im) [scale=\scal]
{
\tikz{
{
\FPeval{\i}{0}
\FPeval{\j}{0}
\node (c\i\j) [draw=none, fill=white!50!magenta, minimum width = \bw ex, minimum height=\bh ex, xshift=\j*\bw ex, yshift=-\i*\bh ex]{};
\FPeval{\i}{1}
\FPeval{\j}{0}
\node (c\i\j) [draw=none, fill=white!50!magenta, minimum width = \bw ex, minimum height=\bh ex, xshift=\j*\bw ex, yshift=-\i*\bh ex]{};
\FPeval{\i}{2}
\FPeval{\j}{0}
\node (c\i\j) [draw=none, fill=white!50!magenta, minimum width = \bw ex, minimum height=\bh ex, xshift=\j*\bw ex, yshift=-\i*\bh ex]{};
\FPeval{\i}{3}
\FPeval{\j}{0}
\node (c\i\j) [draw=none, fill=white!50!blue, minimum width = \bw ex, minimum height=\bh ex, xshift=\j*\bw ex, yshift=-\i*\bh ex]{};
\FPeval{\i}{4}
\FPeval{\j}{0}
\node (c\i\j) [draw=none, fill=white!50!blue, minimum width = \bw ex, minimum height=\bh ex, xshift=\j*\bw ex, yshift=-\i*\bh ex]{};
\FPeval{\i}{5}
\FPeval{\j}{0}
\node (c\i\j) [draw=none, fill=white!50!blue, minimum width = \bw ex, minimum height=\bh ex, xshift=\j*\bw ex, yshift=-\i*\bh ex]{};
}
\foreach \i in {0,1,...,\h}
{
\foreach \j in {0,1,...,\w}
{
\FPeval{\a}{clip(\i*(\h+1)+\j)}
\node (c\i\j) [draw=minigriddclr, fill=none, minimum width = \bw ex, minimum height=\bh ex, xshift=\j*\bw ex, yshift=-\i*\bh ex]{};
}
}
\foreach \i in {0,1,...,\h}
{
\foreach \j in {0,1,...,\w}
{
\node (dc\i\j) [draw=none, circle,fill=pixeldotclr,minimum width = \bw ex, minimum height=\bh ex, xshift=\j*\bw ex+0.0*\bw ex, yshift=-\i*\bh ex+0.0*\bh ex,scale=\pixeldotscale]{};
}
}
\draw [very thick, gridclr] ($(0ex,0ex)+(-0.5*\bw ex, 0.5*\bh ex)$) -- ($(0ex,0ex)+(\w * \bw ex + 0.5*\bw ex, 0.5*\bh ex)$);
\draw [very thick, gridclr] ($(0ex,0ex)+(-0.5*\bw ex, 0.5*\bh ex- 6*\bh ex)$) -- ($(0ex,0ex)+(\w * \bw ex + 0.5*\bw ex, 0.5*\bh ex- 6*\bh ex)$);
\draw [very thick, gridclr] ($(0ex,0ex)+(-0.5*\bw ex, 0.5*\bh ex)$) -- ($(0ex,0ex)+(-0.5*\bw ex, -0.5*\bh ex - \h * \bh ex)$);
\draw [very thick, gridclr] ($(0ex,0ex)+(-0.5*\bw ex + 5*\bw ex, 0.5*\bh ex)$) -- ($(0ex,0ex)+(-0.5*\bw ex + 5*\bw ex, -0.5*\bh ex - \h * \bh ex)$);
\draw [<->] ($(0ex,0ex)+(-0.75*\bw ex , 0.5*\bh ex)$) -- ($(0ex,0ex)+(-0.75*\bw ex, 0.5*\bh ex - 3*\bw ex )$) node [xshift=-1.0ex, yshift=4ex, scale=0.7,rotate=90]{$N_{max}$};
\draw [<->] ($(0ex,0ex)+(-0.75*\bw ex, 0.5*\bh ex-3*\bw ex)$) -- ($(0ex,0ex)+(-0.75*\bw ex, 0.5*\bh ex - 6*\bw ex )$) node [xshift=-1ex, yshift=4ex, scale=0.7,rotate=90]{$N_{max}$};
%
\foreach \i in {0,1,2,3,4}
{
\node (t0\i) [fill=threadclr, xshift=\i*\bw ex, yshift=1.2*\bh ex, scale=0.5]{T$_{0\i}$};
\FPeval{\lshift}{0.3*\bw}
%
%
%
\draw [gray, ->, rounded corners=\pixelcornerradii ex] ($(t0\i.south)+(0 ex, 0.0 ex)$) -- ($(dc0\i.north)+(0 ex, 0ex)$);
\draw [gray, ->, rounded corners=\pixelcornerradii ex] ($(t0\i.south)+(-\lshift ex, 0.0 ex)$) |- ($(dc1\i.west)+(0 ex, 0ex)$);
\draw [gray, ->, rounded corners=\pixelcornerradii ex] ($(t0\i.south)+(\lshift ex, 0.0 ex)$) |- ($(dc2\i.east)+(0 ex, 0ex)$);
\node (t1\i) [fill=threadclr, xshift=\i*\bw ex, yshift=-\h* \bh ex - 1.2*\bh ex, scale=0.5]{T$_{1\i}$};
\FPeval{\lshift}{0.3*\bw}
%
%
%
%
\draw [gray, ->, rounded corners=\pixelcornerradii ex] ($(t1\i.north)+(0 ex, 0.0 ex)$) -- ($(dc5\i.south)+(0 ex, 0ex)$);
\draw [gray, ->, rounded corners=\pixelcornerradii ex] ($(t1\i.north)+(-\lshift ex, 0.0 ex)$) |- ($(dc4\i.west)+(0 ex, 0ex)$);
\draw [gray, ->, rounded corners=\pixelcornerradii ex] ($(t1\i.north)+(\lshift ex, 0.0 ex)$) |- ($(dc3\i.east)+(0 ex, 0ex)$);
}
}};

\node (im1) [xshift=14.8ex,yshift=1ex, scale=\scal]
{
\tikz{
\FPeval{\h}{2-1}
{
\FPeval{\i}{0}
\FPeval{\j}{0}
\node (c\i\j) [draw=none, fill=white!50!magenta, minimum width = \bw ex, minimum height=\bh ex, xshift=\j*\bw ex, yshift=-\i*\bh ex]{};
\FPeval{\i}{1}
\FPeval{\j}{0}
\node (c\i\j) [draw=none, fill=white!50!blue, minimum width = \bw ex, minimum height=\bh ex, xshift=\j*\bw ex, yshift=-\i*\bh ex]{};
}
\foreach \i in {0,1,...,\h}
{
\foreach \j in {0,1,...,\w}
{
\FPeval{\a}{clip(\i*(\h+1)+\j)}
\node (c\i\j) [draw=minigriddclr, fill=none, minimum width = \bw ex, minimum height=\bh ex, xshift=\j*\bw ex, yshift=-\i*\bh ex]{};
}
}
\foreach \i in {0,1,...,\h}
{
\foreach \j in {0,1,...,\w}
{
\node (dc\i\j) [draw=none, circle,fill=pixeldotclr,minimum width = \bw ex, minimum height=\bh ex, xshift=\j*\bw ex+0.0*\bw ex, yshift=-\i*\bh ex+0.0*\bh ex,scale=\pixeldotscale]{};
}
}
\draw [very thick, gridclr] ($(0ex,0ex)+(-0.5*\bw ex, 0.5*\bh ex)$) -- ($(0ex,0ex)+(\w * \bw ex + 0.5*\bw ex, 0.5*\bh ex)$);
\draw [very thick, gridclr] ($(0ex,0ex)+(-0.5*\bw ex, 0.5*\bh ex- 2*\bh ex)$) -- ($(0ex,0ex)+(\w * \bw ex + 0.5*\bw ex, 0.5*\bh ex- 2*\bh ex)$);
\draw [very thick, gridclr] ($(0ex,0ex)+(-0.5*\bw ex, 0.5*\bh ex)$) -- ($(0ex,0ex)+(-0.5*\bw ex, -0.5*\bh ex - \h * \bh ex)$);
\draw [very thick, gridclr] ($(0ex,0ex)+(-0.5*\bw ex + 5*\bw ex, 0.5*\bh ex)$) -- ($(0ex,0ex)+(-0.5*\bw ex + 5*\bw ex, 0.5*\bh ex - 2* \bh ex)$);
\foreach \i in {0,1,2,3,4}
{
\node (t0\i) [fill=threadclr, xshift=\i*\bw ex, yshift=-2.2*\bh ex, scale=0.5]{T$_{0\i}$};
\FPeval{\lshift}{0.3*\bw}
%
%
%
%
\draw [gray, ->, rounded corners=\pixelcornerradii ex] ($(t0\i.north)+(-\lshift ex, 0.0 ex)$) |- ($(dc0\i.west)+(0ex, 0ex)$);
\draw [gray, ->, rounded corners=\pixelcornerradii ex] ($(t0\i.north)+(\lshift ex, 0.0 ex)$)  |- ($(dc1\i.east)+(0 ex, 0ex)$);
}
\draw [<->] ($(0ex,0ex)+(-0.5*\bw ex, 0.7*\bh ex)$) -- ($(0ex,0ex)+(2.5*\bw ex, 0.7*\bh ex)$) node [xshift=-5ex, yshift=0.8ex, scale=0.7]{$N_{max}$};
}};

\node (im3) [xshift=28.8ex,scale=\scal]
{
\tikz{
\FPeval{\h}{1-1}
\FPeval{\w}{5-1}
{
\FPeval{\i}{0}
\FPeval{\j}{0}
\node (c\i\j) [draw=none, fill=white!50!magenta, minimum width = \bw ex, minimum height=\bh ex, xshift=\j*\bw ex, yshift=-\i*\bh ex]{};
\FPeval{\i}{0}
\FPeval{\j}{2}
\node (c\i\j) [draw=none, fill=white!50!magenta, minimum width = \bw ex, minimum height=\bh ex, xshift=\j*\bw ex, yshift=-\i*\bh ex]{};
\FPeval{\i}{0}
\FPeval{\j}{4}
\node (c\i\j) [draw=none, fill=white!50!magenta, minimum width = \bw ex, minimum height=\bh ex, xshift=\j*\bw ex, yshift=-\i*\bh ex]{};
\FPeval{\i}{0}
\FPeval{\j}{1}
\node (c\i\j) [draw=none, fill=white!50!blue, minimum width = \bw ex, minimum height=\bh ex, xshift=\j*\bw ex, yshift=-\i*\bh ex]{};
\FPeval{\i}{0}
\FPeval{\j}{3}
\node (c\i\j) [draw=none, fill=white!50!blue, minimum width = \bw ex, minimum height=\bh ex, xshift=\j*\bw ex, yshift=-\i*\bh ex]{};
}
\foreach \i in {0,...,\h}
{
\foreach \j in {0,...,\w}
{
\FPeval{\a}{clip(\i*(\h+1)+\j)}
\node (c\i\j) [draw=minigriddclr, fill=none, minimum width = \bw ex, minimum height=\bh ex, xshift=\j*\bw ex, yshift=-\i*\bh ex]{};
}
}
\foreach \i in {0,...,\h}
{
\foreach \j in {0,...,\w}
{
\node (dc\i\j) [draw=none, circle,fill=pixeldotclr,minimum width = \bw ex, minimum height=\bh ex, xshift=\j*\bw ex+0.0*\bw ex, yshift=-\i*\bh ex+0.0*\bh ex,scale=\pixeldotscale]{};
}
}
\draw [very thick, gridclr] ($(0ex,0ex)+(-0.5*\bw ex, 0.5*\bh ex)$) -- ($(0ex,0ex)+(\w * \bw ex + 0.5*\bw ex, 0.5*\bh ex)$);
\draw [very thick, gridclr] ($(0ex,0ex)+(-0.5*\bw ex, 0.5*\bh ex- 1*\bh ex)$) -- ($(0ex,0ex)+(\w * \bw ex + 0.5*\bw ex, 0.5*\bh ex- 1*\bh ex)$);
\draw [very thick, gridclr] ($(0ex,0ex)+(-0.5*\bw ex, 0.5*\bh ex)$) -- ($(0ex,0ex)+(-0.5*\bw ex, -0.5*\bh ex - \h * \bh ex)$);
\draw [very thick, gridclr] ($(0ex,0ex)+(-0.5*\bw ex + 5*\bw ex, 0.5*\bh ex)$) -- ($(0ex,0ex)+(-0.5*\bw ex + 5*\bw ex, 0.5*\bh ex - 1* \bh ex)$);
{
\FPeval{\i}{0}
\node (t00) [fill=threadclr, xshift=2.0*\bw ex, yshift=-1.25*\bh ex, scale=0.5]{ T$_{0\i}$};
\FPeval{\i}{1}
\node (t01) [fill=threadclr, xshift=1.5*\i*\bw ex+0.5*\bw ex, yshift=1.25*\bh ex, scale=0.5]{ T$_{0\i}$};
\draw [->, gray, rounded corners=\pixelcornerradii ex] (t00.west) -| (dc00.north);
\draw [->, gray, rounded corners=\pixelcornerradii ex] (t00.north) -- (dc02.south);
\draw [->, gray, rounded corners=\pixelcornerradii ex] (t00.east) -| (dc04.south);
\draw [->, gray, rounded corners=\pixelcornerradii ex] (t01.west) -| (dc01.north);
\draw [->, gray, rounded corners=\pixelcornerradii ex] (t01.east) -| (dc03.north);
%
}
}};

\node (im5) [xshift=39.6ex,yshift=1.25ex, scale=\scal]
{
\tikz{
\FPeval{\h}{1-1}
\FPeval{\w}{2-1}
{
\FPeval{\i}{0}
\FPeval{\j}{0}
\node (c\i\j) [draw=none, fill=white!50!magenta, minimum width = \bw ex, minimum height=\bh ex, xshift=\j*\bw ex, yshift=-\i*\bh ex]{};
\FPeval{\i}{0}
\FPeval{\j}{1}
\node (c\i\j) [draw=none, fill=white!50!blue, minimum width = \bw ex, minimum height=\bh ex, xshift=\j*\bw ex, yshift=-\i*\bh ex]{};
}
\foreach \i in {0,...,\h}
{
\foreach \j in {0,...,\w}
{
\FPeval{\a}{clip(\i*(\h+1)+\j)}
\node (c\i\j) [draw=minigriddclr, fill=none, minimum width = \bw ex, minimum height=\bh ex, xshift=\j*\bw ex, yshift=-\i*\bh ex]{};
}
}
\foreach \i in {0,...,\h}
{
\foreach \j in {0,...,\w}
{
\node (dc\i\j) [draw=none, circle,fill=pixeldotclr,minimum width = \bw ex, minimum height=\bh ex, xshift=\j*\bw ex+0.0*\bw ex, yshift=-\i*\bh ex+0.0*\bh ex,scale=\pixeldotscale]{};
}
}
\draw [very thick, gridclr] ($(0ex,0ex)+(-0.5*\bw ex, 0.5*\bh ex)$) -- ($(0ex,0ex)+(\w * \bw ex + 0.5*\bw ex, 0.5*\bh ex)$);
\draw [very thick, gridclr] ($(0ex,0ex)+(-0.5*\bw ex, 0.5*\bh ex- 1*\bh ex)$) -- ($(0ex,0ex)+(\w * \bw ex + 0.5*\bw ex, 0.5*\bh ex- 1*\bh ex)$);
\draw [very thick, gridclr] ($(0ex,0ex)+(-0.5*\bw ex, 0.5*\bh ex)$) -- ($(0ex,0ex)+(-0.5*\bw ex, -0.5*\bh ex - \h * \bh ex)$);
\draw [very thick, gridclr] ($(0ex,0ex)+(-0.5*\bw ex + 2*\bw ex, 0.5*\bh ex)$) -- ($(0ex,0ex)+(-0.5*\bw ex + 2*\bw ex, 0.5*\bh ex - 1* \bh ex)$);
{
\node (t00) [fill=threadclr, xshift=2*0*\bw ex+0.5*\bw ex, yshift=1.25*\bh ex, scale=0.5]{ T$_{00}$};
\draw [->, gray, rounded corners=\pixelcornerradii ex] ($(t00.west)$) -| ($(dc00.north)-(0ex, 0 ex)$);
\draw [->, gray, rounded corners=\pixelcornerradii ex] ($(t00.east)$) -| ($(dc01.north)-(-0ex, 0 ex)$);
}
}};

\node (im6) [xshift=46ex,scale=\scal]
{
\tikz{
\FPeval{\h}{1-1}
\FPeval{\w}{1-1}
\foreach \i in {0,...,\h}
{
\foreach \j in {0,...,\w}
{
\FPeval{\a}{clip(\i*(\h+1)+\j)}
\node (c\i\j) [draw=minigriddclr, fill=minigridclr, minimum width = \bw ex, minimum height=\bh ex, xshift=\j*\bw ex, yshift=-\i*\bh ex]{};
}
}
\foreach \i in {0,...,\h}
{
\foreach \j in {0,...,\w}
{
\node (dc\i\j) [draw=none, circle,fill=pixeldotclr,minimum width = \bw ex, minimum height=\bh ex, xshift=\j*\bw ex+0.0*\bw ex, yshift=-\i*\bh ex+0.0*\bh ex,scale=\pixeldotscale]{};
}
}
\draw [very thick, gridclr] ($(0ex,0ex)+(-0.5*\bw ex, 0.5*\bh ex)$) -- ($(0ex,0ex)+(\w * \bw ex + 0.5*\bw ex, 0.5*\bh ex)$);
\draw [very thick, gridclr] ($(0ex,0ex)+(-0.5*\bw ex, 0.5*\bh ex- 1*\bh ex)$) -- ($(0ex,0ex)+(\w * \bw ex + 0.5*\bw ex, 0.5*\bh ex- 1*\bh ex)$);
\draw [very thick, gridclr] ($(0ex,0ex)+(-0.5*\bw ex, 0.5*\bh ex)$) -- ($(0ex,0ex)+(-0.5*\bw ex, -0.5*\bh ex - \h * \bh ex)$);
\draw [very thick, gridclr] ($(0ex,0ex)+(-0.5*\bw ex + 1*\bw ex, 0.5*\bh ex)$) -- ($(0ex,0ex)+(-0.5*\bw ex + 1*\bw ex, 0.5*\bh ex - 1* \bh ex)$);
}};

\draw [->, black,] (im.east) -- ($(im1.west)-(0ex,1ex)$);
\draw [->, black,] ($(im1.east)-(0ex,1ex)$) -- (im3.west);
\draw [->, black,] (im3.east) -- ($(im5.west)-(0ex, 1.25ex)$);
\draw [->, black,] ($(im5.east)-(0ex, 1.25ex)$) -- (im6.west);

\draw [<->, ] ($(im.west)+(0.5ex, -10.0 ex)$) -- ($(im6.east)+(0ex, -10.0 ex)$) node [fill=white,xshift=-33.0ex, yshift=-0.1ex, scale=0.6]{Proposed  Multi-location Per-thread Culling (\texttt{MLPT})};

\draw [<->, ] ($(im.west)+(0.5ex, 10.0 ex)$) -- ($(im.east)+(15.5ex, 10.0 ex)$) node [fill=white,xshift=-14.25ex, yshift=0.1ex, scale=0.6]{Vertical Culling (VC)};

\draw [<->, ] ($(im3.west)+(-0.3ex, 10.0 ex)$) -- ($(im6.east)+(0ex, 10.0 ex)$) node [fill=white,xshift=-12.25ex, yshift=0.1ex, scale=0.6]{Horizontal Culling (HC)};

\end{tikzpicture}

\newcommand{\cellelement}{\raisebox{0.2ex}{\tikz{\node (c0) [draw=none, circle,fill=pixeldotclr, minimum width = 1 ex, minimum height=1 ex,scale=0.25]{};}}}
\vspace{-0ex}
\caption{Feature Culling (\texttt{FC}) for a $6 \times 5$ cell. T$_{ij}$ is a CUDA-thread of CUDA kernel \cite{cudaguide}. A `\protect\cellelement' indicates the corner strength of a pixel.}
\label{fig:featureculling}
\vspace{-1.0ex}
\end{figure}
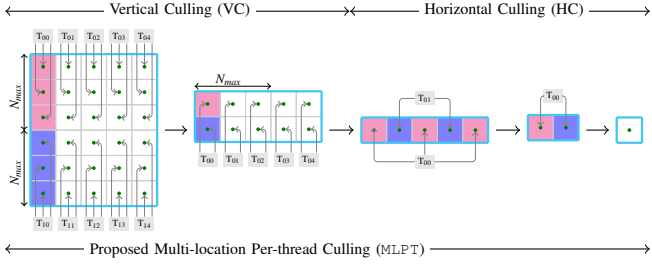

%% file: plots/mem_coalescing.tex
\begin{figure}[t]
\centering
\colorlet{thrdclr}{black!20!cyan}
\colorlet{idlethrdclr}{black!20!magenta}
\colorlet{warpclr}{white!80!white}
\colorlet{warpdclr}{white!20!gray}
\colorlet{gridclr}{white!40!magenta}
\colorlet{minigriddclr}{white!20!magenta}
\colorlet{minigridclr}{white!100!black}
\colorlet{threadclr}{white!90!black}
\FPeval{\bw}{3.5}
\FPeval{\bh}{3.5}
\FPeval{\w}{7}
\FPeval{\h}{5}
\FPeval{\thtop}{1.9}
\FPeval{\scal}{0.75}
\begin{tikzpicture}

\node (im4) [xshift=0ex,scale=\scal]
{
\tikz{
\node (c0) [draw=minigriddclr, fill=minigridclr, minimum width = \bw ex, minimum height=\bh ex, xshift=0*\bw ex, yshift=-0*\bh ex]{};
\node (c1) [draw=minigriddclr, fill=minigridclr, minimum width = \bw ex, minimum height=\bh ex, xshift=1*\bw ex, yshift=-0*\bh ex]{};
\node (c2) [draw=minigriddclr, fill=minigridclr, minimum width = \bw ex, minimum height=\bh ex, xshift=2*\bw ex, yshift=-0*\bh ex]{};
\node (c3) [draw=minigriddclr, fill=minigridclr, minimum width = \bw ex, minimum height=\bh ex, xshift=3*\bw ex, yshift=-0*\bh ex]{};
\node (c4) [draw=minigriddclr, fill=minigridclr, minimum width = \bw ex, minimum height=\bh ex, xshift=4*\bw ex, yshift=-0*\bh ex]{};
\node (c5) [draw=none, fill=none, minimum width = \bw ex, minimum height=\bh ex, xshift=5*\bw ex, yshift=-0*\bh ex]{\tikz{\draw [dotted] (c0.west) -- (c0.east);}};
\node (c6) [draw=minigriddclr, fill=minigridclr, minimum width = \bw ex, minimum height=\bh ex, xshift=6*\bw ex, yshift=-0*\bh ex]{};
\node (c7) [draw=minigriddclr, fill=minigridclr, minimum width = \bw ex, minimum height=\bh ex, xshift=7*\bw ex, yshift=-0*\bh ex]{};
\node (t0) [fill=threadclr, minimum width = \bw ex, minimum height=\bh ex, xshift=0*\bw ex, yshift=1.5*\bh ex, scale=0.6]{T$_0$};
\node (t1) [fill=threadclr, minimum width = \bw ex, minimum height=\bh ex, xshift=1*\bw ex, yshift=1.5*\bh ex, scale=0.6]{T$_1$};
\node (t2) [fill=threadclr, minimum width = \bw ex, minimum height=\bh ex, xshift=2*\bw ex, yshift=1.5*\bh ex, scale=0.6]{T$_2$};
\node (t3) [fill=threadclr, minimum width = \bw ex, minimum height=\bh ex, xshift=3*\bw ex, yshift=1.5*\bh ex, scale=0.6]{T$_3$};
\node (t4) [fill=threadclr, minimum width = \bw ex, minimum height=\bh ex, xshift=4*\bw ex, yshift=1.5*\bh ex, scale=0.6]{T$_4$};
\node (t5) [draw=none, fill=none, minimum width = \bw ex, minimum height=\bh ex, xshift=5*\bw ex, yshift=1.5*\bh ex]{\tikz{\draw [dotted] (c0.west) -- (c0.east);}};
\node (t6) [fill=threadclr, minimum width = \bw ex, minimum height=\bh ex, xshift=6*\bw ex, yshift=1.5*\bh ex, scale=0.6]{T$_{30}$};
\node (t7) [fill=threadclr, minimum width = \bw ex, minimum height=\bh ex, xshift=7*\bw ex, yshift=1.5*\bh ex, scale=0.6]{T$_{31}$};
\draw [->] (t0) -- (c0);
\draw [->] (t1) -- (c1);
\draw [->] (t2) -- (c2);
\draw [->] (t3) -- (c3);
\draw [->] (t4) -- (c4);
\draw [->] (t6) -- (c6);
\draw [->] (t7) -- (c7);
}};

\node (im4) [xshift=28.5ex,scale=\scal]
{
\tikz{
\node (c0) [draw=minigriddclr, fill=minigridclr, minimum width = \bw ex, minimum height=\bh ex, xshift=0*\bw ex, yshift=-0*\bh ex]{};
\node (c1) [draw=minigriddclr, fill=minigridclr, minimum width = \bw ex, minimum height=\bh ex, xshift=1*\bw ex, yshift=-0*\bh ex]{};
\node (c2) [draw=minigriddclr, fill=minigridclr, minimum width = \bw ex, minimum height=\bh ex, xshift=2*\bw ex, yshift=-0*\bh ex]{};
\node (c3) [draw=minigriddclr, fill=minigridclr, minimum width = \bw ex, minimum height=\bh ex, xshift=3*\bw ex, yshift=-0*\bh ex]{};
\node (c4) [draw=minigriddclr, fill=minigridclr, minimum width = \bw ex, minimum height=\bh ex, xshift=4*\bw ex, yshift=-0*\bh ex]{};
\node (c5) [draw=none, fill=none, minimum width = \bw ex, minimum height=\bh ex, xshift=5*\bw ex, yshift=-0*\bh ex]{\tikz{\draw [dotted] (c0.west) -- (c0.east);}};
\node (c6) [draw=minigriddclr, fill=minigridclr, minimum width = \bw ex, minimum height=\bh ex, xshift=6*\bw ex, yshift=-0*\bh ex]{};
\node (c7) [draw=minigriddclr, fill=minigridclr, minimum width = \bw ex, minimum height=\bh ex, xshift=7*\bw ex, yshift=-0*\bh ex]{};
\node (t0) [fill=threadclr, minimum width = \bw ex, minimum height=\bh ex, xshift=0*\bw ex, yshift=1.5*\bh ex, scale=0.6]{T$_0$};
\node (t1) [fill=threadclr, minimum width = \bw ex, minimum height=\bh ex, xshift=1*\bw ex, yshift=1.5*\bh ex, scale=0.6]{T$_1$};
\node (t2) [fill=threadclr, minimum width = \bw ex, minimum height=\bh ex, xshift=2*\bw ex, yshift=1.5*\bh ex, scale=0.6]{T$_2$};
\node (t3) [fill=threadclr, minimum width = \bw ex, minimum height=\bh ex, xshift=3*\bw ex, yshift=1.5*\bh ex, scale=0.6]{T$_3$};
\node (t4) [fill=threadclr, minimum width = \bw ex, minimum height=\bh ex, xshift=4*\bw ex, yshift=1.5*\bh ex, scale=0.6]{T$_4$};
\node (t5) [draw=none, fill=none, minimum width = \bw ex, minimum height=\bh ex, xshift=5*\bw ex, yshift=1.5*\bh ex]{\tikz{\draw [dotted] (c0.west) -- (c0.east);}};
\node (t6) [fill=threadclr, minimum width = \bw ex, minimum height=\bh ex, xshift=6*\bw ex, yshift=1.5*\bh ex, scale=0.6]{T$_{30}$};
\node (t7) [fill=threadclr, minimum width = \bw ex, minimum height=\bh ex, xshift=7*\bw ex, yshift=1.5*\bh ex, scale=0.6]{T$_{31}$};
\draw [->] (t0.south) -- (c4.north);
\draw [->] (t1.south) -- (c2.north);
\draw [->] (t2.south) -- (c7.north);
\draw [->] (t3.south) -- (c3.north);
\draw [->] (t4.south) -- (c0.north);
\draw [->] (t6.south) -- (c6.north);
\draw [->] (t7.south) -- (c1.north);
}};


\node (a) [minimum width = 1 ex, minimum height=1 ex, xshift=-13.0 ex, yshift=-2.8 ex, scale=0.7]{(a)};
\node (b) [minimum width = 1 ex, minimum height=1 ex, xshift=15.5 ex, yshift=-2.8 ex, scale=0.7]{(b)};

\end{tikzpicture}
\newcommand{\memory}{\raisebox{-0.0ex}{\tikz{\node (c0) [draw=minigriddclr, fill=minigridclr, minimum width = \bw ex, minimum height=\bh ex, scale=0.3]{};}}}
\newcommand{\thread}{\raisebox{-0.0ex}{\tikz{\node (c0) [fill=threadclr, minimum width = \bw ex, minimum height=\bh ex, scale=0.4]{T$_{i}$};}}}
\vspace{-2.0ex}
\caption{(a) coalesced, and (b) non-coalesced memory access. `\protect\memory' denotes contiguous memory block, and a `\protect\thread' denotes $i^{th}$ warp thread. In coalesced access, $32$ threads read in one machine-cycle, whereas in non-coalesced access, the memory transactions are serialized \cite{cudaguide}.}
\label{fig:memorycoalescing}
\vspace{-2.5ex}
\end{figure}
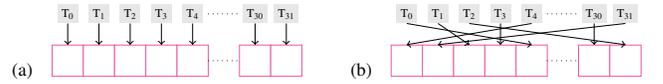

%% file: tables/mlpt_vs_log2.tex
\begin{table}[t]
\centering
\caption{\texttt{MLPT} vs $\log_2$-reduction. $N_w$: number of warps.}
\label{tab:mlpt}
\arrayrulecolor{white!70!black}
\tiny
\setlength{\tabcolsep}{7.5pt}
\vspace{-1ex}
\begin{tabular}{c c c c c c c c}
\hline
\multirow{2}{*}{Cell-size} & \multicolumn{1}{|c}{\multirow{2}{*}{\makecell{Culling \\Scheme}}} & \multicolumn{1}{|c}{\multirow{2}{*}{$N_{max}$}}  & \multicolumn{1}{|c}{\multirow{2}{*}{$N_{t}$}}  & \multicolumn{1}{|c}{\multirow{2}{*}{\makecell{Total \\ Threads}}} &  \multicolumn{1}{|c}{\multirow{2}{*}{$N_{w}$}}  & \multicolumn{2}{|c}{Time ($\mu$S} \\ \cline{7-8}

& \multicolumn{1}{|c}{} & \multicolumn{1}{|c}{} & \multicolumn{1}{|c}{} & \multicolumn{1}{|c}{} & \multicolumn{1}{|c}{} & \multicolumn{1}{|c}{RTX-$2070$}  & \multicolumn{1}{|c}{Jetson-NX} \\ \hline

\multirow{2}{*}{} &  \multicolumn{1}{|l}{\bdota{} $\log_2$}   & \multicolumn{1}{|c}{$-$}  &  \multicolumn{1}{|c}{$10$}   &  \multicolumn{1}{|c}{$320$}  &  \multicolumn{1}{|c}{$10$}   &  \multicolumn{1}{|c}{$4.9\mu$S}  &  \multicolumn{1}{|c}{$11.4\mu$S}  \\
\rowcolor{rwclr}
\multirow{-2}{*}{$20 \times 32$} &  \multicolumn{1}{|l}{\bdotb{} \texttt{MLPT}}   & \multicolumn{1}{|c}{$5$}  &  \multicolumn{1}{|c}{$4$}     &  \multicolumn{1}{|c}{$\mathbf{160}$}    &  \multicolumn{1}{|c}{$\mathbf{5}$}    &  \multicolumn{1}{|c}{$\mathbf{4.6}\mu$S}   &  \multicolumn{1}{|c}{$\mathbf{8.7}\mu$S} \\ \hline
\multirow{2}{*}{} &  \multicolumn{1}{|l}{\bdota{} $\log_2$}   & \multicolumn{1}{|c}{$-$}     &  \multicolumn{1}{|c}{$25$} &  \multicolumn{1}{|c}{$800$}    &  \multicolumn{1}{|c}{$25$}    &  \multicolumn{1}{|c}{$3.2\mu$S}  &  \multicolumn{1}{|c}{$20.4\mu$S} \\
\rowcolor{rwclr}
\multirow{-2}{*}{$50 \times 32$}  &  \multicolumn{1}{|l}{\bdotb{} \texttt{MLPT}}   & \multicolumn{1}{|c}{$5$}  &  \multicolumn{1}{|c}{$4$}     &  \multicolumn{1}{|c}{$\mathbf{320}$}     &  \multicolumn{1}{|c}{$\mathbf{10}$}   &  \multicolumn{1}{|c}{$\mathbf{2.9}\mu$S}   &  \multicolumn{1}{|c}{$\mathbf{16.3}\mu$S} \\ \hline
\multirow{2}{*}{} &  \multicolumn{1}{|l}{\bdota{} $\log_2$}   & \multicolumn{1}{|c}{$-$}     &  \multicolumn{1}{|c}{$50$} &  \multicolumn{1}{|c}{$1600$}   &  \multicolumn{1}{|c}{$50$}   &  \multicolumn{1}{|c}{$3.8\mu$S}  &  \multicolumn{1}{|c}{$14.0\mu$S}  \\
\rowcolor{rwclr}
\multirow{-2}{*}{$100 \times 32$}  &  \multicolumn{1}{|l}{\bdotb{} \texttt{MLPT}}   & \multicolumn{1}{|c}{$10$}  &  \multicolumn{1}{|c}{$10$}     &  \multicolumn{1}{|c}{$\mathbf{320}$}    &  \multicolumn{1}{|c}{$\mathbf{10}$}    &  \multicolumn{1}{|c}{$\mathbf{2.6}\mu$S} &  \multicolumn{1}{|c}{$\mathbf{10.6}\mu$S}   \\  

 \hline
\end{tabular}
\vspace{-2.5ex}
\end{table}

%% file: plots/tewa_scheme.tex
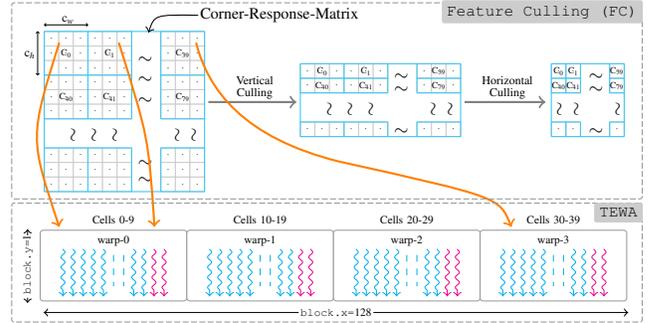
\begin{figure}[!t]
\centering
\colorlet{thrdclr}{black!20!cyan}
\colorlet{idlethrdclr}{black!20!magenta}
\colorlet{warpclr}{white!80!white}
\colorlet{warpdclr}{white!20!gray}
\colorlet{gridclr}{white!40!cyan}
\colorlet{minigriddclr}{white!80!black}
\colorlet{minigridclr}{white!100!black}
\FPeval{\bw}{3.5}
\FPeval{\bh}{3.5}
\FPeval{\w}{10}
\FPeval{\h}{10}
\FPeval{\thtop}{1.9}
\begin{tikzpicture}

\node (outer) [scale=1.0]
{
\tikz{
\node (im) [xshift=-1ex,scale=0.35]
{
\tikz{
\foreach \i in {0,1,...,\h}
{
\foreach \j in {0,1,...,\w}
{
\FPeval{\a}{clip(\i*(\h+1)+\j)}
\ifthenelse{\(\i < 6 \OR \i > 7\) \AND \(\j < 6 \OR \j > 7\)}
{
\node (c\i\j) [draw=minigriddclr, fill=minigridclr, minimum width = \bw ex, minimum height=\bh ex, xshift=\j*\bw ex, yshift=-\i*\bh ex]{$\cdot$};
}
}
}
\draw [very thick, gridclr] ($(0ex,0ex)+(-0.5*\bw ex, 0.5*\bh ex)$) -- ($(0ex,0ex)+(\w * \bw ex + 0.5*\bw ex, 0.5*\bh ex)$);
\draw [very thick, gridclr] ($(0ex,0ex)+(-0.5*\bw ex, 0.5*\bh ex- 11*\bh ex)$) -- ($(0ex,0ex)+(\w * \bw ex + 0.5*\bw ex, 0.5*\bh ex- 11*\bh ex)$);
\draw [very thick, gridclr] ($(0ex,0ex)+(-0.5*\bw ex, 0.5*\bh ex)$) -- ($(0ex,0ex)+(-0.5*\bw ex, -0.5*\bh ex - \h * \bh ex)$);
\draw [very thick, gridclr] ($(0ex,0ex)+(-0.5*\bw ex + 11*\bw ex, 0.5*\bh ex)$) -- ($(0ex,0ex)+(-0.5*\bw ex + 11*\bw ex, -0.5*\bh ex - \h * \bh ex)$);
\draw [very thick, gridclr] ($(0ex,0ex)+(-0.5*\bw ex, 0.5*\bh ex - 3*\bh ex)$) -- ($(0ex,0ex)+(5 * \bw ex + 0.5*\bw ex, 0.5*\bh ex- 3*\bh ex)$);
\draw [very thick, gridclr] ($(0ex,0ex)+(-0.5*\bw ex, 0.5*\bh ex- 6*\bh ex)$) -- ($(0ex,0ex)+(5 * \bw ex + 0.5*\bw ex, 0.5*\bh ex- 6*\bh ex)$);
\draw [very thick, gridclr] ($(0ex,0ex)+(-0.5*\bw ex+8*\bw ex, 0.5*\bh ex - 3*\bh ex)$) -- ($(0ex,0ex)+(\w * \bw ex + 0.5*\bw ex, 0.5*\bh ex- 3*\bh ex)$);
\draw [very thick, gridclr] ($(0ex,0ex)+(-0.5*\bw ex+8*\bw ex, 0.5*\bh ex- 6*\bh ex)$) -- ($(0ex,0ex)+(\w * \bw ex + 0.5*\bw ex, 0.5*\bh ex- 6*\bh ex)$);
\draw [very thick, gridclr] ($(0ex,0ex)+(-0.5*\bw ex, 0.5*\bh ex- 8*\bh ex)$) -- ($(0ex,0ex)+(5 * \bw ex + 0.5*\bw ex, 0.5*\bh ex- 8*\bh ex)$);
\draw [very thick, gridclr] ($(0ex,0ex)+(-0.5*\bw ex+8*\bw ex, 0.5*\bh ex- 8*\bh ex)$) -- ($(0ex,0ex)+(\w * \bw ex + 0.5*\bw ex, 0.5*\bh ex- 8*\bh ex)$);
\draw [very thick, gridclr] ($(0ex,0ex)+(-0.5*\bw ex + 3*\bw ex, 0.5*\bh ex)$) -- ($(0ex,0ex)+(-0.5*\bw ex + 3*\bw ex, -0.5*\bh ex - 5 * \bh ex)$);
\draw [very thick, gridclr] ($(0ex,0ex)+(-0.5*\bw ex + 6*\bw ex, 0.5*\bh ex)$) -- ($(0ex,0ex)+(-0.5*\bw ex + 6*\bw ex, -0.5*\bh ex - 5 * \bh ex)$);
\draw [very thick, gridclr] ($(0ex,0ex)+(-0.5*\bw ex + 3*\bw ex, 0.5*\bh ex - 8*\bh ex )$) -- ($(0ex,0ex)+(-0.5*\bw ex + 3*\bw ex, -0.5*\bh ex - \h * \bh ex)$);
\draw [very thick, gridclr] ($(0ex,0ex)+(-0.5*\bw ex + 6*\bw ex, 0.5*\bh ex- 8*\bh ex )$) -- ($(0ex,0ex)+(-0.5*\bw ex + 6*\bw ex, -0.5*\bh ex - \h * \bh ex)$);
\draw [very thick, gridclr] ($(0ex,0ex)+(-0.5*\bw ex + 8*\bw ex, 0.5*\bh ex)$) -- ($(0ex,0ex)+(-0.5*\bw ex + 8*\bw ex, -0.5*\bh ex - 5 * \bh ex)$);
\draw [very thick, gridclr] ($(0ex,0ex)+(-0.5*\bw ex + 8*\bw ex, 0.5*\bh ex- 8*\bh ex )$) -- ($(0ex,0ex)+(-0.5*\bw ex + 8*\bw ex, -0.5*\bh ex - \h * \bh ex)$);
\node (break) [xshift=22.5ex, yshift =-5ex, scale=2.0]{$\mathbf{\sim}$};
\node (break) [xshift=22.5ex, yshift =-10ex, scale=2.0]{$\mathbf{\sim}$};
\node (break) [xshift=22.5ex, yshift =-15ex, scale=2.0]{$\mathbf{\sim}$};
\node (break) [xshift=22.5ex, yshift =-30ex, scale=2.0]{$\mathbf{\sim}$};
\node (break) [xshift=22.5ex, yshift =-34ex, scale=2.0]{$\mathbf{\sim}$};
\node (break) [xshift=5.5ex, yshift =-23ex, scale=2.0, rotate=90]{$\mathbf{\sim}$};
\node (break) [xshift=10.5ex, yshift =-23ex, scale=2.0, rotate=90]{$\mathbf{\sim}$};
\node (break) [xshift=15.5ex, yshift =-23ex, scale=2.0, rotate=90]{$\mathbf{\sim}$};
\node (break) [xshift=30.5ex, yshift =-23ex, scale=2.0, rotate=90]{$\mathbf{\sim}$};
\node (break) [xshift=34.5ex, yshift =-23ex, scale=2.0, rotate=90]{$\mathbf{\sim}$};
\node [xshift=\bw ex, yshift=-\bh ex, scale=1.0]{C$_0$};
\node [xshift=4*\bw ex, yshift=-\bh ex, scale=1.0]{C$_1$};
\node [xshift=9*\bw ex, yshift=-\bh ex, scale=1.0]{C$_{39}$};
\node [xshift=\bw ex, yshift=-4*\bh ex, scale=1.0]{C$_{40}$};
\node [xshift=4*\bw ex, yshift=-4*\bh ex, scale=1.0]{C$_{41}$};
\node [xshift=9*\bw ex, yshift=-4*\bh ex, scale=1.0]{C$_{79}$};
\draw [<->] ($(0ex,0ex)+(-0.95*\bw ex, 0.5* \bh ex)$) -- ($(0ex,0ex)+(-0.95*\bw ex, 0.5* \bh ex - 3*\bh ex)$) node [xshift=-1.9 ex, yshift=4.8 ex, scale=1.3]{c$_{h}$};
\draw [<->] ($(0ex,0ex)+(-0.5*\bw ex, 0.95* \bh ex)$) -- ($(0ex,0ex)+(-0.5*\bw ex + 3*\bw ex, 0.95* \bh ex)$) node [xshift=-4.8 ex, yshift=1.25 ex, scale=1.3]{c$_{w}$};
}};
%
%
%
%
\node (im1) [xshift=21.5ex, scale=0.35]
{
\tikz{
\FPeval{\nh}{clip(5)-1}
\foreach \i in {0,1,...,\nh}
{
\foreach \j in {0,1,...,\w}
{
\FPeval{\a}{clip(\i*27+\j)}
\ifthenelse{\(\i <2 \OR \i > 3\) \AND \(\j < 6 \OR \j > 7\)}
{
\node (c\i\j) [draw=minigriddclr, fill=minigridclr, minimum width = \bw ex, minimum height=\bh ex, xshift=\j*\bw ex, yshift=-\i*\bh ex]{$\cdot$};
}
}
}
\draw [very thick, gridclr] ($(0ex,0ex)+(-0.5*\bw ex, 0.5*\bh ex)$) -- ($(0ex,0ex)+(\w * \bw ex + 0.5*\bw ex, 0.5*\bh ex)$);
\draw [very thick, gridclr] ($(0ex,0ex)+(-0.5*\bw ex, 0.5*\bh ex- 5*\bh ex)$) -- ($(0ex,0ex)+(\w * \bw ex + 0.5*\bw ex, 0.5*\bh ex- 5*\bh ex)$);
\draw [very thick, gridclr] ($(0ex,0ex)+(-0.5*\bw ex, 0.5*\bh ex)$) -- ($(0ex,0ex)+(-0.5*\bw ex, -0.5*\bh ex - \nh * \bh ex)$);
\draw [very thick, gridclr] ($(0ex,0ex)+(-0.5*\bw ex + 11*\bw ex, 0.5*\bh ex)$) -- ($(0ex,0ex)+(-0.5*\bw ex + 11*\bw ex, -0.5*\bh ex - \nh * \bh ex)$);
\draw [very thick, gridclr] ($(0ex,0ex)+(-0.5*\bw ex, 0.5*\bh ex - 1*\bh ex)$) -- ($(0ex,0ex)+(5 * \bw ex + 0.5*\bw ex, 0.5*\bh ex- 1*\bh ex)$);
\draw [very thick, gridclr] ($(0ex,0ex)+(-0.5*\bw ex, 0.5*\bh ex- 2*\bh ex)$) -- ($(0ex,0ex)+(5 * \bw ex + 0.5*\bw ex, 0.5*\bh ex- 2*\bh ex)$);
\draw [very thick, gridclr] ($(0ex,0ex)+(-0.5*\bw ex+8*\bw ex, 0.5*\bh ex - 1*\bh ex)$) -- ($(0ex,0ex)+(\w * \bw ex + 0.5*\bw ex, 0.5*\bh ex- 1*\bh ex)$);
\draw [very thick, gridclr] ($(0ex,0ex)+(-0.5*\bw ex+8*\bw ex, 0.5*\bh ex- 2*\bh ex)$) -- ($(0ex,0ex)+(\w * \bw ex + 0.5*\bw ex, 0.5*\bh ex- 2*\bh ex)$);
\draw [very thick, gridclr] ($(0ex,0ex)+(-0.5*\bw ex, 0.5*\bh ex- 4*\bh ex)$) -- ($(0ex,0ex)+(5 * \bw ex + 0.5*\bw ex, 0.5*\bh ex- 4*\bh ex)$);
\draw [very thick, gridclr] ($(0ex,0ex)+(-0.5*\bw ex+8*\bw ex, 0.5*\bh ex- 4*\bh ex)$) -- ($(0ex,0ex)+(\w * \bw ex + 0.5*\bw ex, 0.5*\bh ex- 4*\bh ex)$);
\draw [very thick, gridclr] ($(0ex,0ex)+(-0.5*\bw ex + 3*\bw ex, 0.5*\bh ex)$) -- ($(0ex,0ex)+(-0.5*\bw ex + 3*\bw ex, -0.5*\bh ex - 1 * \bh ex)$);
\draw [very thick, gridclr] ($(0ex,0ex)+(-0.5*\bw ex + 6*\bw ex, 0.5*\bh ex)$) -- ($(0ex,0ex)+(-0.5*\bw ex + 6*\bw ex, -0.5*\bh ex - 1 * \bh ex)$);
\draw [very thick, gridclr] ($(0ex,0ex)+(-0.5*\bw ex + 3*\bw ex, 0.5*\bh ex - 4*\bh ex )$) -- ($(0ex,0ex)+(-0.5*\bw ex + 3*\bw ex, -0.5*\bh ex - \nh * \bh ex)$);
\draw [very thick, gridclr] ($(0ex,0ex)+(-0.5*\bw ex + 6*\bw ex, 0.5*\bh ex- 4*\bh ex )$) -- ($(0ex,0ex)+(-0.5*\bw ex + 6*\bw ex, -0.5*\bh ex - \nh * \bh ex)$);
\draw [very thick, gridclr] ($(0ex,0ex)+(-0.5*\bw ex + 8*\bw ex, 0.5*\bh ex)$) -- ($(0ex,0ex)+(-0.5*\bw ex + 8*\bw ex, -0.5*\bh ex - 1 * \bh ex)$);
\draw [very thick, gridclr] ($(0ex,0ex)+(-0.5*\bw ex + 8*\bw ex, 0.5*\bh ex- 4*\bh ex )$) -- ($(0ex,0ex)+(-0.5*\bw ex + 8*\bw ex, -0.5*\bh ex - \nh * \bh ex)$);
\node (break) [xshift=22.5ex, yshift =-1.5ex, scale=2.0]{$\mathbf{\sim}$};
\node (break) [xshift=22.5ex, yshift =-4.5ex, scale=2.0]{$\mathbf{\sim}$};
\node (break) [xshift=22.5ex, yshift =-14.5ex, scale=2.0]{$\mathbf{\sim}$};
\node (break) [xshift=5.5ex, yshift =-9ex, scale=2.0, rotate=90]{$\mathbf{\sim}$};
\node (break) [xshift=10.5ex, yshift =-9ex, scale=2.0, rotate=90]{$\mathbf{\sim}$};
\node (break) [xshift=15.5ex, yshift =-9ex, scale=2.0, rotate=90]{$\mathbf{\sim}$};
\node (break) [xshift=30.5ex, yshift =-9ex, scale=2.0, rotate=90]{$\mathbf{\sim}$};
\node (break) [xshift=34.5ex, yshift =-9ex, scale=2.0, rotate=90]{$\mathbf{\sim}$};
\node [xshift=\bw ex, yshift=0*\bh ex, scale=1.0]{C$_0$};
\node [xshift=4*\bw ex, yshift=0*\bh ex, scale=1.0]{C$_1$};
\node [xshift=9*\bw ex, yshift=0*\bh ex, scale=1.0]{C$_{39}$};
\node [xshift=\bw ex, yshift=-1*\bh ex, scale=1.0]{C$_{40}$};
\node [xshift=4*\bw ex, yshift=-1*\bh ex, scale=1.0]{C$_{41}$};
\node [xshift=9*\bw ex, yshift=-1*\bh ex, scale=1.0]{C$_{79}$};
}};
%
%
%
%
\node (im2) [xshift=39ex, scale=0.35]
{
\tikz{
\FPeval{\nh}{clip(5)-1}
\FPeval{\nw}{clip(5)-1}
\foreach \i in {0,1,...,\nh}
{
\foreach \j in {0,1,...,\nw}
{
\FPeval{\a}{clip(\i*27+\j)}
\ifthenelse{\(\i <2 \OR \i > 3\) \AND \(\j < 2 \OR \j > 3\)}
{
\node (c\i\j) [draw=minigriddclr, fill=minigridclr, minimum width = \bw ex, minimum height=\bh ex, xshift=\j*\bw ex, yshift=-\i*\bh ex]{$\cdot$};
}
}
}
\draw [very thick, gridclr] ($(0ex,0ex)+(-0.5*\bw ex, 0.5*\bh ex)$) -- ($(0ex,0ex)+(\nw * \bw ex + 0.5*\bw ex, 0.5*\bh ex)$);
\draw [very thick, gridclr] ($(0ex,0ex)+(-0.5*\bw ex, 0.5*\bh ex- 5*\bh ex)$) -- ($(0ex,0ex)+(\nw * \bw ex + 0.5*\bw ex, 0.5*\bh ex- 5*\bh ex)$);
\draw [very thick, gridclr] ($(0ex,0ex)+(-0.5*\bw ex, 0.5*\bh ex)$) -- ($(0ex,0ex)+(-0.5*\bw ex, -0.5*\bh ex - \nh * \bh ex)$);
\draw [very thick, gridclr] ($(0ex,0ex)+(-0.5*\bw ex + 5*\bw ex, 0.5*\bh ex)$) -- ($(0ex,0ex)+(-0.5*\bw ex + 5*\bw ex, -0.5*\bh ex - \nh * \bh ex)$);
\draw [very thick, gridclr] ($(0ex,0ex)+(-0.5*\bw ex, 0.5*\bh ex - 1*\bh ex)$) -- ($(0ex,0ex)+(1 * \bw ex + 0.5*\bw ex, 0.5*\bh ex- 1*\bh ex)$);
\draw [very thick, gridclr] ($(0ex,0ex)+(-0.5*\bw ex, 0.5*\bh ex- 2*\bh ex)$) -- ($(0ex,0ex)+(1 * \bw ex + 0.5*\bw ex, 0.5*\bh ex- 2*\bh ex)$);
\draw [very thick, gridclr] ($(0ex,0ex)+(-0.5*\bw ex+4*\bw ex, 0.5*\bh ex - 1*\bh ex)$) -- ($(0ex,0ex)+(\nw * \bw ex + 0.5*\bw ex, 0.5*\bh ex- 1*\bh ex)$);
\draw [very thick, gridclr] ($(0ex,0ex)+(-0.5*\bw ex+4*\bw ex, 0.5*\bh ex- 2*\bh ex)$) -- ($(0ex,0ex)+(\nw * \bw ex + 0.5*\bw ex, 0.5*\bh ex- 2*\bh ex)$);
\draw [very thick, gridclr] ($(0ex,0ex)+(-0.5*\bw ex, 0.5*\bh ex- 4*\bh ex)$) -- ($(0ex,0ex)+(1 * \bw ex + 0.5*\bw ex, 0.5*\bh ex- 4*\bh ex)$);
\draw [very thick, gridclr] ($(0ex,0ex)+(-0.5*\bw ex+4*\bw ex, 0.5*\bh ex- 4*\bh ex)$) -- ($(0ex,0ex)+(\nw * \bw ex + 0.5*\bw ex, 0.5*\bh ex- 4*\bh ex)$);
\draw [very thick, gridclr] ($(0ex,0ex)+(-0.5*\bw ex + 1*\bw ex, 0.5*\bh ex)$) -- ($(0ex,0ex)+(-0.5*\bw ex + 1*\bw ex, -0.5*\bh ex - 1 * \bh ex)$);
\draw [very thick, gridclr] ($(0ex,0ex)+(-0.5*\bw ex + 2*\bw ex, 0.5*\bh ex)$) -- ($(0ex,0ex)+(-0.5*\bw ex + 2*\bw ex, -0.5*\bh ex - 1 * \bh ex)$);
\draw [very thick, gridclr] ($(0ex,0ex)+(-0.5*\bw ex + 1*\bw ex, 0.5*\bh ex - 4*\bh ex )$) -- ($(0ex,0ex)+(-0.5*\bw ex + 1*\bw ex, -0.5*\bh ex - \nh * \bh ex)$);
\draw [very thick, gridclr] ($(0ex,0ex)+(-0.5*\bw ex + 2*\bw ex, 0.5*\bh ex- 4*\bh ex )$) -- ($(0ex,0ex)+(-0.5*\bw ex + 2*\bw ex, -0.5*\bh ex - \nh * \bh ex)$);
\draw [very thick, gridclr] ($(0ex,0ex)+(-0.5*\bw ex + 4*\bw ex, 0.5*\bh ex)$) -- ($(0ex,0ex)+(-0.5*\bw ex + 4*\bw ex, -0.5*\bh ex - 1 * \bh ex)$);
\draw [very thick, gridclr] ($(0ex,0ex)+(-0.5*\bw ex + 4*\bw ex, 0.5*\bh ex- 4*\bh ex )$) -- ($(0ex,0ex)+(-0.5*\bw ex + 4*\bw ex, -0.5*\bh ex - \nh * \bh ex)$);
\node (break) [xshift=9.0ex, yshift =-1.5ex, scale=2.0]{$\mathbf{\sim}$};
\node (break) [xshift=9.0ex, yshift =-4.5ex, scale=2.0]{$\mathbf{\sim}$};
\node (break) [xshift=9.0ex, yshift =-14.5ex, scale=2.0]{$\mathbf{\sim}$};
\node (break) [xshift=0.5ex, yshift =-9ex, scale=2.0, rotate=90]{$\mathbf{\sim}$};
\node (break) [xshift=4.5ex, yshift =-9ex, scale=2.0, rotate=90]{$\mathbf{\sim}$};
\node (break) [xshift=14.5ex, yshift =-9ex, scale=2.0, rotate=90]{$\mathbf{\sim}$};
\node [xshift=0*\bw ex, yshift=0*\bh ex, scale=1.0]{C$_0$};
\node [xshift=1*\bw ex, yshift=0*\bh ex, scale=1.0]{C$_1$};
\node [xshift=4*\bw ex, yshift=0*\bh ex, scale=1.0]{C$_{39}$};
\node [xshift=0*\bw ex, yshift=-1*\bh ex, scale=1.0]{C$_{40}$};
\node [xshift=1*\bw ex, yshift=-1*\bh ex, scale=1.0]{C$_{41}$};
\node [xshift=4*\bw ex, yshift=-1*\bh ex, scale=1.0]{C$_{79}$};
}};
\draw [->, gray,thick] (im.east) -- (im1.west) node [black, xshift=-3.5ex,yshift=1.2ex, scale=0.45]{\shortstack{Vertical \\ Culling}};
\draw [->, gray,thick] (im1.east) -- (im2.west)node [black, xshift=-3.2ex,yshift=1.2ex, scale=0.45]{\shortstack{Horizontal \\ Culling}};
\colorlet{thrdclr}{black!0!cyan}
\colorlet{idlethrdclr}{black!0!magenta}
\colorlet{warpclr}{white!80!white}
\colorlet{warpdclr}{white!20!gray}
\FPeval{\bw}{14.5}
\FPeval{\bh}{7}
\FPeval{\lw}{0.1}
\FPeval{\thtop}{1.9}
%
%
%
%
%
\node (nmsg) [xshift=17ex, yshift=-13.5ex, scale=0.85]{
\tikz
{
%
%
%
\node (w1) []{
\tikz
{
\node (w1) [line width=0.1ex, draw = warpdclr,fill=warpclr, rounded corners=0.3ex,minimum width=\bw ex, minimum height=\bh ex]{};
\draw [line width=\lw ex, thrdclr,->, snake=snake, segment amplitude=0.2ex,segment length=1.0	ex,line after snake=0.5ex] ($(w1.north)-(5ex,\thtop ex)$) -- ($(w1.south)-(5ex,-0.5ex)$);
\draw [line width=\lw ex, thrdclr,->, snake=snake, segment amplitude=0.2ex,segment length=1.0	ex,line after snake=0.5ex] ($(w1.north)-(4ex,\thtop ex)$) -- ($(w1.south)-(4ex,-0.5ex)$);
\draw [line width=\lw ex, thrdclr,->, snake=snake, segment amplitude=0.2ex,segment length=1.0	ex,line after snake=0.5ex] ($(w1.north)-(3ex,\thtop ex)$) -- ($(w1.south)-(3ex,-0.5ex)$);
\draw [line width=\lw ex, thrdclr,->, snake=snake, segment amplitude=0.2ex,segment length=1.0	ex,line after snake=0.5ex] ($(w1.north)-(2ex,\thtop ex)$) -- ($(w1.south)-(2ex,-0.5ex)$);
\draw [line width=\lw ex, thrdclr,->, snake=snake, segment amplitude=0.2ex,segment length=1.0	ex,line after snake=0.5ex] ($(w1.north)-(1ex,\thtop ex)$) -- ($(w1.south)-(1ex,-0.5ex)$);
\draw [line width=\lw ex, thrdclr,-, dashed] ($(w1.north)-(0ex,2.5ex)$) -- ($(w1.south)-(0ex,-1.5ex)$);
\draw [line width=\lw ex, thrdclr,-, dashed] ($(w1.north)-(-1ex,2.5ex)$) -- ($(w1.south)-(-1ex,-1.5ex)$);
\draw [line width=\lw ex, thrdclr,->, snake=snake, segment amplitude=0.2ex,segment length=1.0	ex,line after snake=0.5ex] ($(w1.north)-(-2ex,\thtop ex)$) -- ($(w1.south)-(-2ex,-0.5ex)$);
\draw [line width=\lw ex, thrdclr,->, snake=snake, segment amplitude=0.2ex,segment length=1.0	ex,line after snake=0.5ex] ($(w1.north)-(-3ex,\thtop ex)$) -- ($(w1.south)-(-3ex,-0.5ex)$);
\draw [line width=\lw ex, idlethrdclr,->, snake=snake, segment amplitude=0.2ex,segment length=1.0	ex,line after snake=0.5ex] ($(w1.north)-(-4ex,\thtop ex)$) -- ($(w1.south)-(-4ex,-0.5ex)$);
\draw [line width=\lw ex, idlethrdclr,->, snake=snake, segment amplitude=0.2ex,segment length=1.0	ex,line after snake=0.5ex] ($(w1.north)-(-5ex,\thtop ex)$) -- ($(w1.south)-(-5ex,-0.5ex)$);
\node (warp) [yshift=2.5ex, scale=0.5]{warp-$0$};
\node (grid) [yshift=4.5ex, scale=0.5]{Cells $0$-$9$};
}
};
\node (w2) [xshift=\bw ex]{
\tikz
{
\node (w1) [line width=0.1ex, draw = warpdclr,fill=warpclr, rounded corners=0.3ex,minimum width=\bw ex, minimum height=\bh ex]{};
\draw [line width=\lw ex, thrdclr,->, snake=snake, segment amplitude=0.2ex,segment length=1.0	ex,line after snake=0.5ex] ($(w1.north)-(5ex,\thtop ex)$) -- ($(w1.south)-(5ex,-0.5ex)$);
\draw [line width=\lw ex, thrdclr,->, snake=snake, segment amplitude=0.2ex,segment length=1.0	ex,line after snake=0.5ex] ($(w1.north)-(4ex,\thtop ex)$) -- ($(w1.south)-(4ex,-0.5ex)$);
\draw [line width=\lw ex, thrdclr,->, snake=snake, segment amplitude=0.2ex,segment length=1.0	ex,line after snake=0.5ex] ($(w1.north)-(3ex,\thtop ex)$) -- ($(w1.south)-(3ex,-0.5ex)$);
\draw [line width=\lw ex, thrdclr,->, snake=snake, segment amplitude=0.2ex,segment length=1.0	ex,line after snake=0.5ex] ($(w1.north)-(2ex,\thtop ex)$) -- ($(w1.south)-(2ex,-0.5ex)$);
\draw [line width=\lw ex, thrdclr,->, snake=snake, segment amplitude=0.2ex,segment length=1.0	ex,line after snake=0.5ex] ($(w1.north)-(1ex,\thtop ex)$) -- ($(w1.south)-(1ex,-0.5ex)$);
\draw [line width=\lw ex, thrdclr,-, dashed] ($(w1.north)-(0ex,2.5ex)$) -- ($(w1.south)-(0ex,-1.5ex)$);
\draw [line width=\lw ex, thrdclr,-, dashed] ($(w1.north)-(-1ex,2.5ex)$) -- ($(w1.south)-(-1ex,-1.5ex)$);
\draw [line width=\lw ex, thrdclr,->, snake=snake, segment amplitude=0.2ex,segment length=1.0	ex,line after snake=0.5ex] ($(w1.north)-(-2ex,\thtop ex)$) -- ($(w1.south)-(-2ex,-0.5ex)$);
\draw [line width=\lw ex, thrdclr,->, snake=snake, segment amplitude=0.2ex,segment length=1.0	ex,line after snake=0.5ex] ($(w1.north)-(-3ex,\thtop ex)$) -- ($(w1.south)-(-3ex,-0.5ex)$);
\draw [line width=\lw ex, idlethrdclr,->, snake=snake, segment amplitude=0.2ex,segment length=1.0	ex,line after snake=0.5ex] ($(w1.north)-(-4ex,\thtop ex)$) -- ($(w1.south)-(-4ex,-0.5ex)$);
\draw [line width=\lw ex, idlethrdclr,->, snake=snake, segment amplitude=0.2ex,segment length=1.0	ex,line after snake=0.5ex] ($(w1.north)-(-5ex,\thtop ex)$) -- ($(w1.south)-(-5ex,-0.5ex)$);
\node (warp) [yshift=2.5ex, scale=0.5]{warp-$1$};
\node (grid) [yshift=4.5ex, scale=0.5]{Cells $10$-$19$};
}
};
\node (w3) [xshift=2*\bw ex]{
\tikz
{
\node (w1) [line width=0.1ex, draw = warpdclr,fill=warpclr, rounded corners=0.3ex,minimum width=\bw ex, minimum height=\bh ex]{};
\draw [line width=\lw ex, thrdclr,->, snake=snake, segment amplitude=0.2ex,segment length=1.0	ex,line after snake=0.5ex] ($(w1.north)-(5ex,\thtop ex)$) -- ($(w1.south)-(5ex,-0.5ex)$);
\draw [line width=\lw ex, thrdclr,->, snake=snake, segment amplitude=0.2ex,segment length=1.0	ex,line after snake=0.5ex] ($(w1.north)-(4ex,\thtop ex)$) -- ($(w1.south)-(4ex,-0.5ex)$);
\draw [line width=\lw ex, thrdclr,->, snake=snake, segment amplitude=0.2ex,segment length=1.0	ex,line after snake=0.5ex] ($(w1.north)-(3ex,\thtop ex)$) -- ($(w1.south)-(3ex,-0.5ex)$);
\draw [line width=\lw ex, thrdclr,->, snake=snake, segment amplitude=0.2ex,segment length=1.0	ex,line after snake=0.5ex] ($(w1.north)-(2ex,\thtop ex)$) -- ($(w1.south)-(2ex,-0.5ex)$);
\draw [line width=\lw ex, thrdclr,->, snake=snake, segment amplitude=0.2ex,segment length=1.0	ex,line after snake=0.5ex] ($(w1.north)-(1ex,\thtop ex)$) -- ($(w1.south)-(1ex,-0.5ex)$);
\draw [line width=\lw ex, thrdclr,-, dashed] ($(w1.north)-(0ex,2.5ex)$) -- ($(w1.south)-(0ex,-1.5ex)$);
\draw [line width=\lw ex, thrdclr,-, dashed] ($(w1.north)-(-1ex,2.5ex)$) -- ($(w1.south)-(-1ex,-1.5ex)$);
\draw [line width=\lw ex, thrdclr,->, snake=snake, segment amplitude=0.2ex,segment length=1.0	ex,line after snake=0.5ex] ($(w1.north)-(-2ex,\thtop ex)$) -- ($(w1.south)-(-2ex,-0.5ex)$);
\draw [line width=\lw ex, thrdclr,->, snake=snake, segment amplitude=0.2ex,segment length=1.0	ex,line after snake=0.5ex] ($(w1.north)-(-3ex,\thtop ex)$) -- ($(w1.south)-(-3ex,-0.5ex)$);
\draw [line width=\lw ex, idlethrdclr,->, snake=snake, segment amplitude=0.2ex,segment length=1.0	ex,line after snake=0.5ex] ($(w1.north)-(-4ex,\thtop ex)$) -- ($(w1.south)-(-4ex,-0.5ex)$);
\draw [line width=\lw ex, idlethrdclr,->, snake=snake, segment amplitude=0.2ex,segment length=1.0	ex,line after snake=0.5ex] ($(w1.north)-(-5ex,\thtop ex)$) -- ($(w1.south)-(-5ex,-0.5ex)$);
\node (warp) [yshift=2.5ex, scale=0.5]{warp-$2$};
\node (grid) [yshift=4.5ex, scale=0.5]{Cells $20$-$29$};
}
};
\node (w4) [xshift=3* \bw ex]{
\tikz
{
\node (w1) [line width=0.1ex, draw = warpdclr,fill=warpclr, rounded corners=0.3ex,minimum width=\bw ex, minimum height=\bh ex]{};
\draw [line width=\lw ex, thrdclr,->, snake=snake, segment amplitude=0.2ex,segment length=1.0	ex,line after snake=0.5ex] ($(w1.north)-(5ex,\thtop ex)$) -- ($(w1.south)-(5ex,-0.5ex)$);
\draw [line width=\lw ex, thrdclr,->, snake=snake, segment amplitude=0.2ex,segment length=1.0	ex,line after snake=0.5ex] ($(w1.north)-(4ex,\thtop ex)$) -- ($(w1.south)-(4ex,-0.5ex)$);
\draw [line width=\lw ex, thrdclr,->, snake=snake, segment amplitude=0.2ex,segment length=1.0	ex,line after snake=0.5ex] ($(w1.north)-(3ex,\thtop ex)$) -- ($(w1.south)-(3ex,-0.5ex)$);
\draw [line width=\lw ex, thrdclr,->, snake=snake, segment amplitude=0.2ex,segment length=1.0	ex,line after snake=0.5ex] ($(w1.north)-(2ex,\thtop ex)$) -- ($(w1.south)-(2ex,-0.5ex)$);
\draw [line width=\lw ex, thrdclr,->, snake=snake, segment amplitude=0.2ex,segment length=1.0	ex,line after snake=0.5ex] ($(w1.north)-(1ex,\thtop ex)$) -- ($(w1.south)-(1ex,-0.5ex)$);
\draw [line width=\lw ex, thrdclr,-, dashed] ($(w1.north)-(0ex,2.5ex)$) -- ($(w1.south)-(0ex,-1.5ex)$);
\draw [line width=\lw ex, thrdclr,-, dashed] ($(w1.north)-(-1ex,2.5ex)$) -- ($(w1.south)-(-1ex,-1.5ex)$);
\draw [line width=\lw ex, thrdclr,->, snake=snake, segment amplitude=0.2ex,segment length=1.0	ex,line after snake=0.5ex] ($(w1.north)-(-2ex,\thtop ex)$) -- ($(w1.south)-(-2ex,-0.5ex)$);
\draw [line width=\lw ex, thrdclr,->, snake=snake, segment amplitude=0.2ex,segment length=1.0	ex,line after snake=0.5ex] ($(w1.north)-(-3ex,\thtop ex)$) -- ($(w1.south)-(-3ex,-0.5ex)$);
\draw [line width=\lw ex, idlethrdclr,->, snake=snake, segment amplitude=0.2ex,segment length=1.0	ex,line after snake=0.5ex] ($(w1.north)-(-4ex,\thtop ex)$) -- ($(w1.south)-(-4ex,-0.5ex)$);
\draw [line width=\lw ex, idlethrdclr,->, snake=snake, segment amplitude=0.2ex,segment length=1.0	ex,line after snake=0.5ex] ($(w1.north)-(-5ex,\thtop ex)$) -- ($(w1.south)-(-5ex,-0.5ex)$);
\node (warp) [yshift=2.5ex, scale=0.5]{warp-$3$};
\node (grid) [yshift=4.5ex, scale=0.5]{Cells $30$-$39$};
}
};
\node (blockx) [xshift=22ex, yshift=-5.5ex, scale=0.5]{\texttt{block.x=$128$}};
\draw [->] ($(blockx.west)-(-0.2ex,0.0ex)$) -- ($(blockx.west)-(25.0ex,0.0ex)$);
\draw [->] ($(blockx.east)-(0.2ex,0.0ex)$) -- ($(blockx.east)-(-25.0ex,0.0ex)$);
\node (blocky) [xshift=-8.5ex, yshift=-0.7ex, rotate=90, scale=0.5]{\texttt{block.y=$1$}};
\draw [->] ($(blocky.west)-(0.0ex,0.0ex)$) -- ($(blocky.west)-(0.0ex,0.5ex)$);
\draw [->] ($(blocky.east)-(0.0ex,0.5ex)$) -- ($(blocky.east)-(0.0ex,-0.1ex)$);
}};
\draw [-> , orange,   thick] ($(-5.5ex,5.0ex)$) .. controls (-8ex, -3ex) .. ($(-5.5ex,-10.5ex)$);
\draw [-> , orange,   thick] ($(-0.5ex,5.0ex)$) .. controls (2ex, -3ex) .. ($(2.5ex,-10.5ex)$);
\draw [-> , orange,   thick] ($(6.0ex,5.0ex)$) .. controls (7ex, 1ex) and (10ex, -3ex) .. (18ex, -5ex) .. controls (31.5ex, -8ex) .. ($(32.5ex,-10.5ex)$);
\node (crm) [draw=none, xshift=13ex, yshift=7.3ex, scale=0.6]{Corner-Response-Matrix};
\draw [->] ($(crm.west)-(-0.3ex, 0ex)$) .. controls ($(im.north)-(-3.5ex,0.5ex)$) .. ($(im.north)-(-3ex,2.0ex)$);
\node (pyca) [draw=white!60!black, dash pattern= on 0.5ex off 0.2ex, rounded corners=0.2ex, minimum height = 16.5ex, minimum width =53 ex, xshift=17ex, yshift=0.1ex]{};
\node (pyca) [draw=none,fill=white!80!black, , rounded corners=0.2ex, xshift=35.1ex, yshift=7.6ex, scale=0.6]{\texttt{Feature Culling (\texttt{FC})}};
\node (tewa) [draw=white!60!black, dash pattern= on 0.5ex off 0.2ex, rounded corners=0.2ex, minimum height = 10.0ex, minimum width =53 ex, xshift=17ex, yshift=-13.5ex]{};
\node (tewa) [draw=none,fill=white!80!black, , rounded corners=0.2ex, xshift=41.5ex, yshift=-9.3ex, scale=0.6]{\texttt{TEWA}};
}};

\end{tikzpicture}

\vspace{-0.3ex}
\newcommand{\usethread}{\raisebox{-0.5ex}{\tikz{\node(a)[scale=0.75]{\tikz{\draw [thick, thrdclr,->, snake=snake, segment amplitude=0.2ex,segment length=0.8	ex,line after snake=0.5ex] ($(0ex,1ex)$) --($(0ex,-1ex)$);}};}}}
\newcommand{\wastethread}{\raisebox{-0.5ex}{\tikz{\node(a)[scale=0.75]{\tikz{\draw [thick, idlethrdclr,->, snake=snake, segment amplitude=0.2ex,segment length=0.8	ex,line after snake=0.5ex] ($(0ex,1ex)$) --($(0ex,-1ex)$);}};}}}

\vspace{0ex}

\caption{Illustration of \texttt{FC} + (\texttt{TEWA}) scheme. \texttt{FC} is applied over the CRF-Matrix which produces the strongest corner in a cell. A `$C_i$' is a cell, and a `\protect\usethread' and a `\protect\wastethread' denote a working and an idle/wasted thread in a warp respectively for a $3\times3$ cell-size.}
\label{fig:warpalloc}
\vspace{-2.25ex}
\end{figure}

%% file: plots/pfa_figure.tex
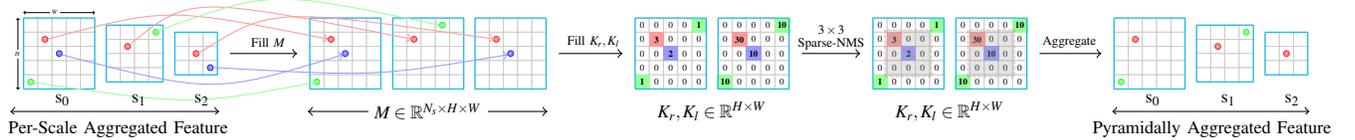
\begin{figure*}[!t]
\centering
\colorlet{thrdclr}{black!20!cyan}
\colorlet{idlethrdclr}{black!20!magenta}
\colorlet{warpclr}{white!80!white}
\colorlet{warpdclr}{white!20!gray}
\colorlet{gridclr}{white!20!cyan}
\colorlet{minigriddclr}{white!80!black}
\colorlet{minigridclr}{white!100!black}
\colorlet{threadclr}{white!90!black}
\FPeval{\bw}{3.5}
\FPeval{\bh}{3.5}
\FPeval{\w}{4}
\FPeval{\h}{4}
\colorlet{p1dclr}{white!10!red}
\colorlet{p1clr}{white!60!red}
\colorlet{p2dclr}{white!10!blue}
\colorlet{p2clr}{white!60!blue}
\colorlet{p3dclr}{white!10!green}
\colorlet{p3clr}{white!60!green}
\colorlet{nmsdclr}{white!0!red}
\colorlet{nmsclr}{white!30!gray}
\FPeval{\thtop}{1.9}
\begin{tikzpicture}

\node (outer)[scale=0.975]
{\tikz{
\node (step1) []
{
\tikz{
\node (im) [scale=0.35]
{
\tikz{
\foreach \i in {0,1,...,\h}
{
\foreach \j in {0,1,...,\w}
{
\FPeval{\a}{clip(\i*(\h+1)+\j)}
\node (c\i\j) [draw=minigriddclr, fill=minigridclr, minimum width = \bw ex, minimum height=\bh ex, xshift=\j*\bw ex, yshift=-\i*\bh ex]{};
}
}
{
\FPeval{\i}{1}
\FPeval{\j}{1}
\node (c0) [draw=p1dclr, circle,fill=p1clr, minimum width = \bw ex, minimum height=\bh ex, xshift=\j*\bw ex+0.0*\bw ex, yshift=-\i*\bh ex+0.0*\bh ex,scale=0.4]{};
\FPeval{\i}{2}
\FPeval{\j}{2}
\node (c0) [draw=p2dclr, circle,fill=p2clr, minimum width = \bw ex, minimum height=\bh ex, xshift=\j*\bw ex+0.0*\bw ex, yshift=-\i*\bh ex+0.0*\bh ex,scale=0.4]{};
\FPeval{\i}{4}
\FPeval{\j}{0}
\node (c0) [draw=p3dclr, circle,fill=p3clr, minimum width = \bw ex, minimum height=\bh ex, xshift=\j*\bw ex+0.0*\bw ex, yshift=-\i*\bh ex+0.0*\bh ex,scale=0.4]{};
}
\FPeval{\ow}{\w+1}
\FPeval{\oh}{\h+1}
\draw [very thick, gridclr] ($(0ex,0ex)+(-0.5*\bw ex, 0.5*\bh ex)$) -- ($(0ex,0ex)+(\ow * \bw ex - 0.5*\bw ex, 0.5*\bh ex)$);
\draw [very thick, gridclr] ($(0ex,0ex)+(-0.5*\bw ex, 0.5*\bh ex- \oh*\bh ex)$) -- ($(0ex,0ex)+(\ow * \bw ex - 0.5*\bw ex, 0.5*\bh ex- \oh*\bh ex)$);
\draw [very thick, gridclr] ($(0ex,0ex)+(-0.5*\bw ex, 0.5*\bh ex)$) -- ($(0ex,0ex)+(-0.5*\bw ex, -0.5*\bh ex - \h * \bh ex)$);
\draw [very thick, gridclr] ($(0ex,0ex)+(-0.5*\bw ex + \ow*\bw ex, 0.5*\bh ex)$) -- ($(0ex,0ex)+(-0.5*\bw ex + \ow*\bw ex, -0.5*\bh ex - \h * \bh ex)$);
}};
\FPeval{\w}{3}
\FPeval{\h}{3}
\node (im1) [xshift=6.5ex,scale=0.35]
{
\tikz{
\foreach \i in {0,1,...,\h}
{
\foreach \j in {0,1,...,\w}
{
\FPeval{\a}{clip(\i*(\h+1)+\j)}
\node (c\i\j) [draw=minigriddclr, fill=minigridclr, minimum width = \bw ex, minimum height=\bh ex, xshift=\j*\bw ex, yshift=-\i*\bh ex]{};
}
}
{
\FPeval{\i}{1}
\FPeval{\j}{1}
\node (c0) [draw=p1dclr, circle,fill=p1clr, minimum width = \bw ex, minimum height=\bh ex, xshift=\j*\bw ex+0.0*\bw ex, yshift=-\i*\bh ex+0.0*\bh ex,scale=0.4]{};
\FPeval{\i}{0}
\FPeval{\j}{3}
\node (c0) [draw=p3dclr, circle,fill=p3clr, minimum width = \bw ex, minimum height=\bh ex, xshift=\j*\bw ex+0.0*\bw ex, yshift=-\i*\bh ex+0.0*\bh ex,scale=0.4]{};
}
\FPeval{\ow}{\w+1}
\FPeval{\oh}{\h+1}
\draw [very thick, gridclr] ($(0ex,0ex)+(-0.5*\bw ex, 0.5*\bh ex)$) -- ($(0ex,0ex)+(\ow * \bw ex - 0.5*\bw ex, 0.5*\bh ex)$);
\draw [very thick, gridclr] ($(0ex,0ex)+(-0.5*\bw ex, 0.5*\bh ex- \oh*\bh ex)$) -- ($(0ex,0ex)+(\ow * \bw ex - 0.5*\bw ex, 0.5*\bh ex- \oh*\bh ex)$);
\draw [very thick, gridclr] ($(0ex,0ex)+(-0.5*\bw ex, 0.5*\bh ex)$) -- ($(0ex,0ex)+(-0.5*\bw ex, -0.5*\bh ex - \h * \bh ex)$);
\draw [very thick, gridclr] ($(0ex,0ex)+(-0.5*\bw ex + \ow*\bw ex, 0.5*\bh ex)$) -- ($(0ex,0ex)+(-0.5*\bw ex + \ow*\bw ex, -0.5*\bh ex - \h * \bh ex)$);
}};
\FPeval{\w}{2}
\FPeval{\h}{2}
\node (im2) [xshift=11.8ex,scale=0.35]
{
\tikz{
\foreach \i in {0,1,...,\h}
{
\foreach \j in {0,1,...,\w}
{
\FPeval{\a}{clip(\i*(\h+1)+\j)}
\node (c\i\j) [draw=minigriddclr, fill=minigridclr, minimum width = \bw ex, minimum height=\bh ex, xshift=\j*\bw ex, yshift=-\i*\bh ex]{};
}
}
{
\FPeval{\i}{1}
\FPeval{\j}{1}
\node (c0) [draw=p1dclr, circle,fill=p1clr, minimum width = \bw ex, minimum height=\bh ex, xshift=\j*\bw ex+0.0*\bw ex, yshift=-\i*\bh ex+0.0*\bh ex,scale=0.4]{};
\FPeval{\i}{2}
\FPeval{\j}{2}
\node (c0) [draw=p2dclr, circle,fill=p2clr, minimum width = \bw ex, minimum height=\bh ex, xshift=\j*\bw ex+0.0*\bw ex, yshift=-\i*\bh ex+0.0*\bh ex,scale=0.4]{};
}
\FPeval{\ow}{\w+1}
\FPeval{\oh}{\h+1}
\draw [very thick, gridclr] ($(0ex,0ex)+(-0.5*\bw ex, 0.5*\bh ex)$) -- ($(0ex,0ex)+(\ow * \bw ex - 0.5*\bw ex, 0.5*\bh ex)$);
\draw [very thick, gridclr] ($(0ex,0ex)+(-0.5*\bw ex, 0.5*\bh ex- \oh*\bh ex)$) -- ($(0ex,0ex)+(\ow * \bw ex - 0.5*\bw ex, 0.5*\bh ex- \oh*\bh ex)$);
\draw [very thick, gridclr] ($(0ex,0ex)+(-0.5*\bw ex, 0.5*\bh ex)$) -- ($(0ex,0ex)+(-0.5*\bw ex, -0.5*\bh ex - \h * \bh ex)$);
\draw [very thick, gridclr] ($(0ex,0ex)+(-0.5*\bw ex + \ow*\bw ex, 0.5*\bh ex)$) -- ($(0ex,0ex)+(-0.5*\bw ex + \ow*\bw ex, -0.5*\bh ex - \h * \bh ex)$);
}};
}};
\node [xshift=-5.75ex,yshift=0.42ex, scale=0.35]
{
\tikz{
\FPeval{\ow}{\w+1}
\FPeval{\oh}{\h+1}
\draw [<->] (-0.9*\bw ex, 0.5 * \bh ex) -- ( -0.9*\bw ex, 0.5 * \bh ex -\oh * \bh ex);
\node [fill=white, xshift=-3.3ex, yshift=-7ex, scale=0.7]{$H$};
\draw [<->] (-0.5*\bw ex, 0.9 * \bh ex) -- ( -0.5*\bw ex + \ow * \bw ex , 0.9 * \bh ex);
\node [fill=white, xshift=6.ex, yshift=3.3ex, scale=0.7]{$W$};
}};
%
%
%
%
%
%
%
%
\node (step2) [xshift=26.5ex]
{
\tikz{
\node (im) [scale=0.35]
{
\tikz{
\foreach \i in {0,1,...,\h}
{
\foreach \j in {0,1,...,\w}
{
\FPeval{\a}{clip(\i*(\h+1)+\j)}
\node (c\i\j) [draw=minigriddclr, fill=minigridclr, minimum width = \bw ex, minimum height=\bh ex, xshift=\j*\bw ex, yshift=-\i*\bh ex]{};
}
}
{
\FPeval{\i}{1}
\FPeval{\j}{1}
\node (c0) [draw=p1dclr, circle,fill=p1clr, minimum width = \bw ex, minimum height=\bh ex, xshift=\j*\bw ex+0.0*\bw ex, yshift=-\i*\bh ex+0.0*\bh ex,scale=0.4]{};
\FPeval{\i}{2}
\FPeval{\j}{2}
\node (c0) [draw=p2dclr, circle,fill=p2clr, minimum width = \bw ex, minimum height=\bh ex, xshift=\j*\bw ex+0.0*\bw ex, yshift=-\i*\bh ex+0.0*\bh ex,scale=0.4]{};
\FPeval{\i}{4}
\FPeval{\j}{0}
\node (c0) [draw=p3dclr, circle,fill=p3clr, minimum width = \bw ex, minimum height=\bh ex, xshift=\j*\bw ex+0.0*\bw ex, yshift=-\i*\bh ex+0.0*\bh ex,scale=0.4]{};
}
\FPeval{\ow}{\w+1}
\FPeval{\oh}{\h+1}
\draw [very thick, gridclr] ($(0ex,0ex)+(-0.5*\bw ex, 0.5*\bh ex)$) -- ($(0ex,0ex)+(\ow * \bw ex - 0.5*\bw ex, 0.5*\bh ex)$);
\draw [very thick, gridclr] ($(0ex,0ex)+(-0.5*\bw ex, 0.5*\bh ex- \oh*\bh ex)$) -- ($(0ex,0ex)+(\ow * \bw ex - 0.5*\bw ex, 0.5*\bh ex- \oh*\bh ex)$);
\draw [very thick, gridclr] ($(0ex,0ex)+(-0.5*\bw ex, 0.5*\bh ex)$) -- ($(0ex,0ex)+(-0.5*\bw ex, -0.5*\bh ex - \h * \bh ex)$);
\draw [very thick, gridclr] ($(0ex,0ex)+(-0.5*\bw ex + \ow*\bw ex, 0.5*\bh ex)$) -- ($(0ex,0ex)+(-0.5*\bw ex + \ow*\bw ex, -0.5*\bh ex - \h * \bh ex)$);
}};
\FPeval{\w}{4}
\FPeval{\h}{4}
\node (im1) [xshift=7.15ex,scale=0.35]
{
\tikz{
\foreach \i in {0,1,...,\h}
{
\foreach \j in {0,1,...,\w}
{
\FPeval{\a}{clip(\i*(\h+1)+\j)}
\node (c\i\j) [draw=minigriddclr, fill=minigridclr, minimum width = \bw ex, minimum height=\bh ex, xshift=\j*\bw ex, yshift=-\i*\bh ex]{};
}
}
{
\FPeval{\i}{1}
\FPeval{\j}{1}
\node (c0) [draw=p1dclr, circle,fill=p1clr, minimum width = \bw ex, minimum height=\bh ex, xshift=\j*\bw ex+0.0*\bw ex, yshift=-\i*\bh ex+0.0*\bh ex,scale=0.4]{};
\FPeval{\i}{0}
\FPeval{\j}{3}
\node (c0) [draw=p3dclr, circle,fill=p3clr, minimum width = \bw ex, minimum height=\bh ex, xshift=\j*\bw ex+0.0*\bw ex, yshift=-\i*\bh ex+0.0*\bh ex,scale=0.4]{};
}
\FPeval{\ow}{\w+1}
\FPeval{\oh}{\h+1}
\draw [very thick, gridclr] ($(0ex,0ex)+(-0.5*\bw ex, 0.5*\bh ex)$) -- ($(0ex,0ex)+(\ow * \bw ex - 0.5*\bw ex, 0.5*\bh ex)$);
\draw [very thick, gridclr] ($(0ex,0ex)+(-0.5*\bw ex, 0.5*\bh ex- \oh*\bh ex)$) -- ($(0ex,0ex)+(\ow * \bw ex - 0.5*\bw ex, 0.5*\bh ex- \oh*\bh ex)$);
\draw [very thick, gridclr] ($(0ex,0ex)+(-0.5*\bw ex, 0.5*\bh ex)$) -- ($(0ex,0ex)+(-0.5*\bw ex, -0.5*\bh ex - \h * \bh ex)$);
\draw [very thick, gridclr] ($(0ex,0ex)+(-0.5*\bw ex + \ow*\bw ex, 0.5*\bh ex)$) -- ($(0ex,0ex)+(-0.5*\bw ex + \ow*\bw ex, -0.5*\bh ex - \h * \bh ex)$);
}};
\FPeval{\w}{4}
\FPeval{\h}{4}
\node (im1) [xshift=14.25ex,scale=0.35]
{
\tikz{
\foreach \i in {0,1,...,\h}
{
\foreach \j in {0,1,...,\w}
{
\FPeval{\a}{clip(\i*(\h+1)+\j)}
\node (c\i\j) [draw=minigriddclr, fill=minigridclr, minimum width = \bw ex, minimum height=\bh ex, xshift=\j*\bw ex, yshift=-\i*\bh ex]{};
}
}
{
\FPeval{\i}{1}
\FPeval{\j}{1}
\node (c0) [draw=p1dclr, circle,fill=p1clr, minimum width = \bw ex, minimum height=\bh ex, xshift=\j*\bw ex+0.0*\bw ex, yshift=-\i*\bh ex+0.0*\bh ex,scale=0.4]{};
\FPeval{\i}{2}
\FPeval{\j}{2}
\node (c0) [draw=p2dclr, circle,fill=p2clr, minimum width = \bw ex, minimum height=\bh ex, xshift=\j*\bw ex+0.0*\bw ex, yshift=-\i*\bh ex+0.0*\bh ex,scale=0.4]{};
}
\FPeval{\ow}{\w+1}
\FPeval{\oh}{\h+1}
\draw [very thick, gridclr] ($(0ex,0ex)+(-0.5*\bw ex, 0.5*\bh ex)$) -- ($(0ex,0ex)+(\ow * \bw ex - 0.5*\bw ex, 0.5*\bh ex)$);
\draw [very thick, gridclr] ($(0ex,0ex)+(-0.5*\bw ex, 0.5*\bh ex- \oh*\bh ex)$) -- ($(0ex,0ex)+(\ow * \bw ex - 0.5*\bw ex, 0.5*\bh ex- \oh*\bh ex)$);
\draw [very thick, gridclr] ($(0ex,0ex)+(-0.5*\bw ex, 0.5*\bh ex)$) -- ($(0ex,0ex)+(-0.5*\bw ex, -0.5*\bh ex - \h * \bh ex)$);
\draw [very thick, gridclr] ($(0ex,0ex)+(-0.5*\bw ex + \ow*\bw ex, 0.5*\bh ex)$) -- ($(0ex,0ex)+(-0.5*\bw ex + \ow*\bw ex, -0.5*\bh ex - \h * \bh ex)$);
}};
}};
%
%
%
%
%
%
\node (step3) [xshift=51ex]
{
\tikz{
\node (im) [scale=0.35]
{
\tikz{
\foreach \i in {0,1,...,\h}
{
\foreach \j in {0,1,...,\w}
{
\FPeval{\a}{clip(\i*(\h+1)+\j)}
\ifthenelse{\( \NOT \(\i = 0 \AND \j =4 \) \AND \NOT \(\i = 1 \AND \j = 1 \) \AND \NOT \(\i = 2 \AND \j = 2 \) \AND \NOT \(\i = 4 \AND \j = 0 \) \)}
{\node (c\i\j) [draw=minigriddclr, fill=minigridclr, minimum width = \bw ex, minimum height=\bh ex, xshift=\j*\bw ex, yshift=-\i*\bh ex]{$0$};}
}
}
{
\FPeval{\i}{0}
\FPeval{\j}{4}
\node (c0) [draw=none, fill=p3clr, minimum width = \bw ex, minimum height=\bh ex, xshift=\j*\bw ex+0.0*\bw ex, yshift=-\i*\bh ex+0.0*\bh ex,scale=1.0]{$\mathbf{1}$};
\FPeval{\i}{1}
\FPeval{\j}{1}
\node (c0) [draw=none, ,fill=p1clr, minimum width = \bw ex, minimum height=\bh ex, xshift=\j*\bw ex+0.0*\bw ex, yshift=-\i*\bh ex+0.0*\bh ex,scale=1.0]{$\mathbf{3}$};
\FPeval{\i}{2}
\FPeval{\j}{2}
\node (c0) [draw=none, ,fill=p2clr, minimum width = \bw ex, minimum height=\bh ex, xshift=\j*\bw ex+0.0*\bw ex, yshift=-\i*\bh ex+0.0*\bh ex,scale=1.0]{$\mathbf{2}$};
\FPeval{\i}{4}
\FPeval{\j}{0}
\node (c0) [draw=none, ,fill=p3clr, minimum width = \bw ex, minimum height=\bh ex, xshift=\j*\bw ex+0.0*\bw ex, yshift=-\i*\bh ex+0.0*\bh ex,scale=1.0	]{$\mathbf{1}$};
}
\FPeval{\ow}{\w+1}
\FPeval{\oh}{\h+1}
\draw [very thick, gridclr] ($(0ex,0ex)+(-0.5*\bw ex, 0.5*\bh ex)$) -- ($(0ex,0ex)+(\ow * \bw ex - 0.5*\bw ex, 0.5*\bh ex)$);
\draw [very thick, gridclr] ($(0ex,0ex)+(-0.5*\bw ex, 0.5*\bh ex- \oh*\bh ex)$) -- ($(0ex,0ex)+(\ow * \bw ex - 0.5*\bw ex, 0.5*\bh ex- \oh*\bh ex)$);
\draw [very thick, gridclr] ($(0ex,0ex)+(-0.5*\bw ex, 0.5*\bh ex)$) -- ($(0ex,0ex)+(-0.5*\bw ex, -0.5*\bh ex - \h * \bh ex)$);
\draw [very thick, gridclr] ($(0ex,0ex)+(-0.5*\bw ex + \ow*\bw ex, 0.5*\bh ex)$) -- ($(0ex,0ex)+(-0.5*\bw ex + \ow*\bw ex, -0.5*\bh ex - \h * \bh ex)$);
}};
\FPeval{\w}{4}
\FPeval{\h}{4}
\node (im1) [xshift=7.15ex,scale=0.35]
{
\tikz{
\foreach \i in {0,1,...,\h}
{
\foreach \j in {0,1,...,\w}
{
\FPeval{\a}{clip(\i*(\h+1)+\j)}
\ifthenelse{\( \NOT \(\i = 0 \AND \j =4 \) \AND \NOT \(\i = 1 \AND \j = 1 \) \AND \NOT \(\i = 2 \AND \j = 2 \) \AND \NOT \(\i = 4 \AND \j = 0 \) \)}
{\node (c\i\j) [draw=minigriddclr, fill=minigridclr, minimum width = \bw ex, minimum height=\bh ex, xshift=\j*\bw ex, yshift=-\i*\bh ex]{$0$};}
}
}
{
\FPeval{\i}{0}
\FPeval{\j}{4}
\node (c0) [draw=none, fill=p3clr, minimum width = \bw ex, minimum height=\bh ex, xshift=\j*\bw ex+0.0*\bw ex, yshift=-\i*\bh ex+0.0*\bh ex,scale=1.0]{$\mathbf{10}$};
\FPeval{\i}{1}
\FPeval{\j}{1}
\node (c0) [draw=none, fill=p1clr, minimum width = \bw ex, minimum height=\bh ex, xshift=\j*\bw ex+0.0*\bw ex, yshift=-\i*\bh ex+0.0*\bh ex,scale=1.0]{$\mathbf{30}$};
\FPeval{\i}{2}
\FPeval{\j}{2}
\node (c0) [draw=none, fill=p2clr, minimum width = \bw ex, minimum height=\bh ex, xshift=\j*\bw ex+0.0*\bw ex, yshift=-\i*\bh ex+0.0*\bh ex,scale=1.0]{$\mathbf{10}$};
\FPeval{\i}{4}
\FPeval{\j}{0}
\node (c0) [draw=none, fill=p3clr, minimum width = \bw ex, minimum height=\bh ex, xshift=\j*\bw ex+0.0*\bw ex, yshift=-\i*\bh ex+0.0*\bh ex,scale=1.0	]{$\mathbf{10}$};
}
\FPeval{\ow}{\w+1}
\FPeval{\oh}{\h+1}
\draw [very thick, gridclr] ($(0ex,0ex)+(-0.5*\bw ex, 0.5*\bh ex)$) -- ($(0ex,0ex)+(\ow * \bw ex - 0.5*\bw ex, 0.5*\bh ex)$);
\draw [very thick, gridclr] ($(0ex,0ex)+(-0.5*\bw ex, 0.5*\bh ex- \oh*\bh ex)$) -- ($(0ex,0ex)+(\ow * \bw ex - 0.5*\bw ex, 0.5*\bh ex- \oh*\bh ex)$);
\draw [very thick, gridclr] ($(0ex,0ex)+(-0.5*\bw ex, 0.5*\bh ex)$) -- ($(0ex,0ex)+(-0.5*\bw ex, -0.5*\bh ex - \h * \bh ex)$);
\draw [very thick, gridclr] ($(0ex,0ex)+(-0.5*\bw ex + \ow*\bw ex, 0.5*\bh ex)$) -- ($(0ex,0ex)+(-0.5*\bw ex + \ow*\bw ex, -0.5*\bh ex - \h * \bh ex)$);
}};
}};
%
%
%
%
%
%
\node (step4) [xshift=71.5ex]
{
\tikz{
\node (im) [scale=0.35]
{
\tikz{
\foreach \i in {0,1,...,\h}
{
\foreach \j in {0,1,...,\w}
{
\FPeval{\a}{clip(\i*(\h+1)+\j)}
\ifthenelse{\( \NOT \(\i = 0 \AND \j =4 \) \AND \NOT \(\i = 1 \AND \j = 1 \) \AND \NOT \(\i = 2 \AND \j = 2 \) \AND \NOT \(\i = 4 \AND \j = 0 \) \)}
{\node (c\i\j) [draw=minigriddclr, fill=minigridclr, minimum width = \bw ex, minimum height=\bh ex, xshift=\j*\bw ex, yshift=-\i*\bh ex]{$0$};}
}
}
{
\FPeval{\i}{0}
\FPeval{\j}{4}
\node (c0) [draw=none, fill=p3clr, minimum width = \bw ex, minimum height=\bh ex, xshift=\j*\bw ex+0.0*\bw ex, yshift=-\i*\bh ex+0.0*\bh ex,scale=1.0]{$\mathbf{1}$};
\FPeval{\i}{1}
\FPeval{\j}{1}
\node (c0) [draw=none, fill=p1clr, minimum width = \bw ex, minimum height=\bh ex, xshift=\j*\bw ex+0.0*\bw ex, yshift=-\i*\bh ex+0.0*\bh ex,scale=1.0]{$\mathbf{3}$};
\FPeval{\i}{2}
\FPeval{\j}{2}
\node (c0) [draw=none, fill=p2clr, minimum width = \bw ex, minimum height=\bh ex, xshift=\j*\bw ex+0.0*\bw ex, yshift=-\i*\bh ex+0.0*\bh ex,scale=1.0]{$\mathbf{2}$};
\FPeval{\i}{4}
\FPeval{\j}{0}
\node (c0) [draw=none, fill=p3clr, minimum width = \bw ex, minimum height=\bh ex, xshift=\j*\bw ex+0.0*\bw ex, yshift=-\i*\bh ex+0.0*\bh ex,scale=1.0	]{$\mathbf{1}$};
}
\FPeval{\ow}{\w+1}
\FPeval{\oh}{\h+1}
\draw [very thick, gridclr] ($(0ex,0ex)+(-0.5*\bw ex, 0.5*\bh ex)$) -- ($(0ex,0ex)+(\ow * \bw ex - 0.5*\bw ex, 0.5*\bh ex)$);
\draw [very thick, gridclr] ($(0ex,0ex)+(-0.5*\bw ex, 0.5*\bh ex- \oh*\bh ex)$) -- ($(0ex,0ex)+(\ow * \bw ex - 0.5*\bw ex, 0.5*\bh ex- \oh*\bh ex)$);
\draw [very thick, gridclr] ($(0ex,0ex)+(-0.5*\bw ex, 0.5*\bh ex)$) -- ($(0ex,0ex)+(-0.5*\bw ex, -0.5*\bh ex - \h * \bh ex)$);
\draw [very thick, gridclr] ($(0ex,0ex)+(-0.5*\bw ex + \ow*\bw ex, 0.5*\bh ex)$) -- ($(0ex,0ex)+(-0.5*\bw ex + \ow*\bw ex, -0.5*\bh ex - \h * \bh ex)$);
\node [thick, dashed, draw=nmsdclr, fill=nmsclr, opacity=0.3, minimum width=3.3*\bw ex,minimum height=3.3*\bh ex, xshift=6.95ex, yshift=-6.95ex]{};
}};
\FPeval{\w}{4}
\FPeval{\h}{4}
\node (im1) [xshift=7.15ex,scale=0.35]
{
\tikz{
\foreach \i in {0,1,...,\h}
{
\foreach \j in {0,1,...,\w}
{
\FPeval{\a}{clip(\i*(\h+1)+\j)}
\ifthenelse{\( \NOT \(\i = 0 \AND \j =4 \) \AND \NOT \(\i = 1 \AND \j = 1 \) \AND \NOT \(\i = 2 \AND \j = 2 \) \AND \NOT \(\i = 4 \AND \j = 0 \) \)}
{\node (c\i\j) [draw=minigriddclr, fill=minigridclr, minimum width = \bw ex, minimum height=\bh ex, xshift=\j*\bw ex, yshift=-\i*\bh ex]{$0$};}
}
}
{
\FPeval{\i}{0}
\FPeval{\j}{4}
\node (c0) [draw=none, fill=p3clr, minimum width = \bw ex, minimum height=\bh ex, xshift=\j*\bw ex+0.0*\bw ex, yshift=-\i*\bh ex+0.0*\bh ex,scale=1.0]{$\mathbf{10}$};
\FPeval{\i}{1}
\FPeval{\j}{1}
\node (c0) [draw=none, fill=p1clr, minimum width = \bw ex, minimum height=\bh ex, xshift=\j*\bw ex+0.0*\bw ex, yshift=-\i*\bh ex+0.0*\bh ex,scale=1.0]{$\mathbf{30}$};
\FPeval{\i}{2}
\FPeval{\j}{2}
\node (c0) [draw=none, fill=p2clr, minimum width = \bw ex, minimum height=\bh ex, xshift=\j*\bw ex+0.0*\bw ex, yshift=-\i*\bh ex+0.0*\bh ex,scale=1.0]{$\mathbf{10}$};
\FPeval{\i}{4}
\FPeval{\j}{0}
\node (c0) [draw=none, fill=p3clr, minimum width = \bw ex, minimum height=\bh ex, xshift=\j*\bw ex+0.0*\bw ex, yshift=-\i*\bh ex+0.0*\bh ex,scale=1.0	]{$\mathbf{10}$};
}
\FPeval{\ow}{\w+1}
\FPeval{\oh}{\h+1}
\draw [very thick, gridclr] ($(0ex,0ex)+(-0.5*\bw ex, 0.5*\bh ex)$) -- ($(0ex,0ex)+(\ow * \bw ex - 0.5*\bw ex, 0.5*\bh ex)$);
\draw [very thick, gridclr] ($(0ex,0ex)+(-0.5*\bw ex, 0.5*\bh ex- \oh*\bh ex)$) -- ($(0ex,0ex)+(\ow * \bw ex - 0.5*\bw ex, 0.5*\bh ex- \oh*\bh ex)$);
\draw [very thick, gridclr] ($(0ex,0ex)+(-0.5*\bw ex, 0.5*\bh ex)$) -- ($(0ex,0ex)+(-0.5*\bw ex, -0.5*\bh ex - \h * \bh ex)$);
\draw [very thick, gridclr] ($(0ex,0ex)+(-0.5*\bw ex + \ow*\bw ex, 0.5*\bh ex)$) -- ($(0ex,0ex)+(-0.5*\bw ex + \ow*\bw ex, -0.5*\bh ex - \h * \bh ex)$);
\node [thick, dashed, draw=nmsdclr, fill=nmsclr, opacity=0.3, minimum width=3.3*\bw ex,minimum height=3.3*\bh ex, xshift=6.95ex, yshift=-6.95ex]{};
}};
}};
%
%
%
%
%
\node (step5) [xshift=94ex]
{
\tikz{
\node (im) [scale=0.35]
{
\tikz{
\foreach \i in {0,1,...,\h}
{
\foreach \j in {0,1,...,\w}
{
\FPeval{\a}{clip(\i*(\h+1)+\j)}
\node (c\i\j) [draw=minigriddclr, fill=minigridclr, minimum width = \bw ex, minimum height=\bh ex, xshift=\j*\bw ex, yshift=-\i*\bh ex]{};
}
}
{
\FPeval{\i}{1}
\FPeval{\j}{1}
\node (c0) [draw=p1dclr, circle,fill=p1clr, minimum width = \bw ex, minimum height=\bh ex, xshift=\j*\bw ex+0.0*\bw ex, yshift=-\i*\bh ex+0.0*\bh ex,scale=0.4]{};
\FPeval{\i}{4}
\FPeval{\j}{0}
\node (c0) [draw=p3dclr, circle,fill=p3clr, minimum width = \bw ex, minimum height=\bh ex, xshift=\j*\bw ex+0.0*\bw ex, yshift=-\i*\bh ex+0.0*\bh ex,scale=0.4]{};
}
\FPeval{\ow}{\w+1}
\FPeval{\oh}{\h+1}
\draw [very thick, gridclr] ($(0ex,0ex)+(-0.5*\bw ex, 0.5*\bh ex)$) -- ($(0ex,0ex)+(\ow * \bw ex - 0.5*\bw ex, 0.5*\bh ex)$);
\draw [very thick, gridclr] ($(0ex,0ex)+(-0.5*\bw ex, 0.5*\bh ex- \oh*\bh ex)$) -- ($(0ex,0ex)+(\ow * \bw ex - 0.5*\bw ex, 0.5*\bh ex- \oh*\bh ex)$);
\draw [very thick, gridclr] ($(0ex,0ex)+(-0.5*\bw ex, 0.5*\bh ex)$) -- ($(0ex,0ex)+(-0.5*\bw ex, -0.5*\bh ex - \h * \bh ex)$);
\draw [very thick, gridclr] ($(0ex,0ex)+(-0.5*\bw ex + \ow*\bw ex, 0.5*\bh ex)$) -- ($(0ex,0ex)+(-0.5*\bw ex + \ow*\bw ex, -0.5*\bh ex - \h * \bh ex)$);
}};
\FPeval{\w}{3}
\FPeval{\h}{3}
\node (im1) [xshift=6.5ex,scale=0.35]
{
\tikz{
\foreach \i in {0,1,...,\h}
{
\foreach \j in {0,1,...,\w}
{
\FPeval{\a}{clip(\i*(\h+1)+\j)}
\node (c\i\j) [draw=minigriddclr, fill=minigridclr, minimum width = \bw ex, minimum height=\bh ex, xshift=\j*\bw ex, yshift=-\i*\bh ex]{};
}
}
{
\FPeval{\i}{1}
\FPeval{\j}{1}
\node (c0) [draw=p1dclr, circle,fill=p1clr, minimum width = \bw ex, minimum height=\bh ex, xshift=\j*\bw ex+0.0*\bw ex, yshift=-\i*\bh ex+0.0*\bh ex,scale=0.4]{};
\FPeval{\i}{0}
\FPeval{\j}{3}
\node (c0) [draw=p3dclr, circle,fill=p3clr, minimum width = \bw ex, minimum height=\bh ex, xshift=\j*\bw ex+0.0*\bw ex, yshift=-\i*\bh ex+0.0*\bh ex,scale=0.4]{};
}
\FPeval{\ow}{\w+1}
\FPeval{\oh}{\h+1}
\draw [very thick, gridclr] ($(0ex,0ex)+(-0.5*\bw ex, 0.5*\bh ex)$) -- ($(0ex,0ex)+(\ow * \bw ex - 0.5*\bw ex, 0.5*\bh ex)$);
\draw [very thick, gridclr] ($(0ex,0ex)+(-0.5*\bw ex, 0.5*\bh ex- \oh*\bh ex)$) -- ($(0ex,0ex)+(\ow * \bw ex - 0.5*\bw ex, 0.5*\bh ex- \oh*\bh ex)$);
\draw [very thick, gridclr] ($(0ex,0ex)+(-0.5*\bw ex, 0.5*\bh ex)$) -- ($(0ex,0ex)+(-0.5*\bw ex, -0.5*\bh ex - \h * \bh ex)$);
\draw [very thick, gridclr] ($(0ex,0ex)+(-0.5*\bw ex + \ow*\bw ex, 0.5*\bh ex)$) -- ($(0ex,0ex)+(-0.5*\bw ex + \ow*\bw ex, -0.5*\bh ex - \h * \bh ex)$);
}};
\FPeval{\w}{2}
\FPeval{\h}{2}
\node (im2) [xshift=11.8ex,scale=0.35]
{
\tikz{
\foreach \i in {0,1,...,\h}
{
\foreach \j in {0,1,...,\w}
{
\FPeval{\a}{clip(\i*(\h+1)+\j)}
\node (c\i\j) [draw=minigriddclr, fill=minigridclr, minimum width = \bw ex, minimum height=\bh ex, xshift=\j*\bw ex, yshift=-\i*\bh ex]{};
}
}
{
\FPeval{\i}{1}
\FPeval{\j}{1}
\node (c0) [draw=p1dclr, circle,fill=p1clr, minimum width = \bw ex, minimum height=\bh ex, xshift=\j*\bw ex+0.0*\bw ex, yshift=-\i*\bh ex+0.0*\bh ex,scale=0.4]{};
}
\FPeval{\ow}{\w+1}
\FPeval{\oh}{\h+1}
\draw [very thick, gridclr] ($(0ex,0ex)+(-0.5*\bw ex, 0.5*\bh ex)$) -- ($(0ex,0ex)+(\ow * \bw ex - 0.5*\bw ex, 0.5*\bh ex)$);
\draw [very thick, gridclr] ($(0ex,0ex)+(-0.5*\bw ex, 0.5*\bh ex- \oh*\bh ex)$) -- ($(0ex,0ex)+(\ow * \bw ex - 0.5*\bw ex, 0.5*\bh ex- \oh*\bh ex)$);
\draw [very thick, gridclr] ($(0ex,0ex)+(-0.5*\bw ex, 0.5*\bh ex)$) -- ($(0ex,0ex)+(-0.5*\bw ex, -0.5*\bh ex - \h * \bh ex)$);
\draw [very thick, gridclr] ($(0ex,0ex)+(-0.5*\bw ex + \ow*\bw ex, 0.5*\bh ex)$) -- ($(0ex,0ex)+(-0.5*\bw ex + \ow*\bw ex, -0.5*\bh ex - \h * \bh ex)$);
}};
}};

\draw [->] (step1.east) -- (step2.west) node [xshift=-2.5ex, yshift=1.0ex,scale=0.5]{Fill $M$};
\draw [->] (step2.east) -- (step3.west) node [xshift=-2.5ex, yshift=1.0ex,scale=0.5]{Fill $K_r,K_l$};
\draw [->] (step3.east) -- (step4.west) node [xshift=-2.5ex, yshift=1.4ex,scale=0.5]{\shortstack{$3\times3$\\ Sparse-NMS}};
\draw [->] (step4.east) -- (step5.west)  node [xshift=-2.75ex, yshift=1.0ex,scale=0.5]{Aggregate};

\draw [->, p1clr] (-6.25ex, 1.25ex) .. controls (4ex,4ex) .. (17.9ex, 1.25ex);
\draw [->, p1clr] (0.82ex, 0.80ex) .. controls (6ex,5ex) .. (25.1ex, 1.25ex);
\draw [->, p1clr] (6.7ex, 0.0ex) .. controls (13ex,4ex) .. (32.2ex, 1.25ex);
\draw [->, p3clr] (3.2ex, 1.95ex) .. controls (13ex,5.5ex) .. (27.5ex, 2.45ex);
\draw [->, p3clr] (-7.5ex, -2.5ex) .. controls (6ex,-4.5ex) .. (16.75ex, -2.5ex);
\draw [->, p2clr] (-5.1ex, -0.1ex) .. controls (8ex,-3.5ex) .. (19.2ex, -0.1ex);
\draw [->, p2clr] (8.0ex, -1.2ex) .. controls (22ex,-2.0ex) .. (33.4ex, -0.1ex);
\node [xshift=-5ex, yshift=-4ex, scale=0.7]{s$_0$};
\node [xshift=1.6ex, yshift=-4ex, scale=0.7]{s$_1$};
\node [xshift=7ex, yshift=-4ex, scale=0.7]{s$_2$};
\draw [<->] (16.15ex, -5ex) -- (36.8ex, -5ex);
\node [fill=white, xshift=26.5ex, yshift=-5ex, scale=0.7]{$M \in \mathbb{R}^{N_s \times H \times W}$};
\node [xshift=51.0ex, yshift=-5ex, scale=0.7]{$K_r, K_l \in \mathbb{R}^{H \times W}$};
%
%
\node [xshift=71.5ex, yshift=-5ex, scale=0.7]{$K_r, K_l \in \mathbb{R}^{H \times W}$};
%
%
\node [xshift=89ex, yshift=-4ex, scale=0.7]{s$_0$};
\node [xshift=95.5ex, yshift=-4ex, scale=0.7]{s$_1$};
\node [xshift=101ex, yshift=-4ex, scale=0.7]{s$_2$};
\draw [<->] (-9.5ex, -5ex) -- (8.5ex, -5ex);
\node [fill=white, xshift=-0.2ex, yshift=-6.5ex, scale=0.7]{Per-Scale Aggregated Feature};
\draw [<->] (85.5ex, -5ex) -- (102.5ex, -5ex);
\node [fill=white, xshift=94.1ex, yshift=-6.5ex, scale=0.7]{Pyramidally Aggregated Feature};
}};

\end{tikzpicture}

\vspace{-1.0ex}
\newcommand{\usept}{\raisebox{0.2ex}{\tikz{\node (c0) [draw=p1dclr, circle,fill=p1clr, minimum width = 1 ex, minimum height=1 ex,scale=0.4]{};}}}
\newcommand{\suppresspt}{\raisebox{0.2ex}{\tikz{\node (c0) [draw=p2dclr, circle,fill=p2clr, minimum width = 1 ex, minimum height=1 ex,scale=0.4]{};}}}
\newcommand{\normalpt}{\raisebox{0.2ex}{\tikz{\node (c0) [draw=p3dclr, circle,fill=p3clr, minimum width = 1 ex, minimum height=1 ex,scale=0.4]{};}}}

\caption{Pyramidal Feature Aggregation (\texttt{PFA}). Steps of a single $5 \times 5$ image at scale factor $1.2$ and three scales. A `\protect\usept' indicates a survived point in the \texttt{PFA} process, a `\protect\suppresspt' indicates a suppressed point, and a `\protect\normalpt' indicates an unaffected point.}
\label{fig:pfa}
\vspace{-2.5ex}
\end{figure*}

%% file: tables/tewa.tex
\begin{table}[!t]
\centering
\caption{Thread-Efficient Warp allocation (\texttt{TEWA}) efficiency (Eq.~\ref{eq:warpefficiency}).}
\label{tab:warpefficiency}
\arrayrulecolor{white!70!black}
\tiny
\setlength{\tabcolsep}{14.8pt}
\vspace{-1ex}
\begin{tabular}{c c c c c}
\hline
\multirow{1}{*}{Cell-size} & \multicolumn{1}{|c}{Warp allocation Scheme} & \multicolumn{1}{|c}{$N_{ta}$}  & \multicolumn{1}{|c}{$N_w$} & \multicolumn{1}{|c}{$\eta_w$} \\ \hline

\multirow{2}{*}{} &  \multicolumn{1}{|l}{\bdota{} Single block per cell}   & \multicolumn{1}{|c}{$9$}  &  \multicolumn{1}{|c}{$1$}   &  \multicolumn{1}{|c}{$28\%$}  \\
\rowcolor{rwclr}
\multirow{-2}{*}{$3 \times 3$}  &  \multicolumn{1}{|l}{\bdotb{} \texttt{TEWA}}   & \multicolumn{1}{|c}{$30$}  &  \multicolumn{1}{|c}{$1$}   &  \multicolumn{1}{|c}{$\mathbf{94\%}$}  \\ \hline
\multirow{2}{*}{} &  \multicolumn{1}{|l}{\bdota{} Single block per cell}   & \multicolumn{1}{|c}{$25$}     &  \multicolumn{1}{|c}{$1$}   &  \multicolumn{1}{|c}{$78\%$} \\
\rowcolor{rwclr}
\multirow{-2}{*}{$5 \times 5$}  &  \multicolumn{1}{|l}{\bdotb{} \texttt{TEWA}}   & \multicolumn{1}{|c}{$30$}  &  \multicolumn{1}{|c}{$1$}   &  \multicolumn{1}{|c}{$\mathbf{94\%}$}  \\ \hline
\multirow{2}{*}{} &  \multicolumn{1}{|l}{\bdota{} Single block per cell}   & \multicolumn{1}{|c}{$49$}     &  \multicolumn{1}{|c}{$2$}  &  \multicolumn{1}{|c}{$76\%$}  \\
\rowcolor{rwclr}
\multirow{-2}{*}{$7 \times 7$}  &  \multicolumn{1}{|l}{\bdotb{} \texttt{TEWA}}   & \multicolumn{1}{|c}{$28$}  &  \multicolumn{1}{|c}{$1$}   &  \multicolumn{1}{|c}{$\mathbf{88\%}$}  \\  
 \hline
\end{tabular}
\vspace{-5.0ex}
\end{table}

%% file: pyca_pfa.tex
\subsubsection{Pyramidal Feature Aggregation (\texttt{PFA})}
\label{sec:pycapfa}
It is the final step of \texttt{PyCA} that processes the features obtained by performing \texttt{FC} at each scale of the image pyramid in a multiscale setting. \texttt{PFA} reduces the overall feature count, which lowers the feature and stereo-matching calculations, thus lowering the overall runtime without affecting the SLAM accuracy.
\par
\texttt{PFA} is motivated by two reasons: \textit{First}, the multiscale setting is crucial in SLAM to obtain robust features which govern tracking, localization, and mapping accuracy. Hence, features that project to a common pixel at the native scale (s$_0$), need to be prioritized because of their uniqueness and consistency across scales. \textit{Second}, the projected features may fall into the vicinity of each other, which confuses the feature-matcher and stereo-matcher, and also increases the computation time of the frontend (feature extraction), middle-end (stereo matching, tracking) and backend (optimization).
\par
\texttt{PFA} is carried out in three steps (Fig.~\ref{fig:pfa}). Firstly, we project all keypoints to s$_0$, i.e. $x_0 = x_n \times \zeta^n $, and copy their responses in a $3$D matrix $M \in \mathbb{R}^{N_s \times H \times W}$ initialized with zeros. Here $x_0$ denotes s$_0$ correspondent of a coordinate $x_n$ at $n^{th}$ scale with scale-factor $\zeta$, and $N_s, H, W$ denote the number of scales, image height and width respectively at the s$_0$. Secondly, we compute two metrics for each keypoint, i.e. the sum of the corner responses across scales if a keypoint is detected at multiple scales ($k_r$), and the total number of levels at which it is detected ($k_l$). These scores are stored as two matrices $K_r, K_l \in \mathbb{R}^{H \times W}$. Finally, we perform non-maximal suppression (NMS) over the $K_r, K_l$ matrices, but only sparsely in a $q \times q$ window ($q=3$ adjustable) around each keypoint, saving a lot of computations. The keypoint is suppressed if its $k_r$ and $k_l$ scores are smaller than any keypoint in the window.  
\par
\texttt{PFA} is crafted such that it runs on GPU while avoiding CPU-GPU memory transfer. On the contrary, if run on CPU, its operation unnecessarily consumes CPU and requires CPU-GPU memory transfer, because the input to \texttt{PFA} resides on GPU. It also keeps the CPU occupied, and reduces the memory-transfer bandwidth due to small-sized data transfer, a point of consideration for Jetson devices. 
%

%% file: system_integration.tex
\input{plots/slam_figure_final}
In this section, we describe our system development contributions i.e. our new Frontend--Middle-end Jetson-SLAM design (Fig.~\ref{fig:slampipeline}), and strategic integration to optimize information flow. These are crucial to achieving resource efficiency despite the frontend achievements because SLAM components now are multi-device residents (CPU and GPU)
\subsubsection{$\mu$-Sec. Efficient FAST Detection}
\label{sec:fastesetgpu}
%
We use two $16$-bit integers $B_b$ and $B_d$ whose each bit denotes one of the $16$ locations on the Bresenhem path (Sec.~\ref{sec:fast}), and is computed via Eq.~\ref{eq:label}. A `$1$' bit corresponds to $L_p =\texttt{bright}$ in $B_b$ and $L_p = \texttt{dark}$ in $B_d$, whereas a `$0$' bit signifies $L_p = \texttt{similar}$. To speed up the process, we construct a lookup table where all the $16$-bit combinations are pre-calculated to be a corner or a non-corner based on the proposed bounded rectification technique (Eq.~\ref{eq:newcorner}). We use the sum of absolute differences ($|I_c-I_p|$) over the circle as the corner response \cite{opencv}.
\subsubsection{Streamlined MultiScale Detection \& Extraction}
\label{sec:fastdetextgpu}
Our frontend is dedicated to high-speed SLAM, however, it can be used as a VO frontend which only performs detection. In contrast, SLAM computes descriptors as well for the map elements, making SLAM slower than VO. The multiscale setting is more compute-intensive, and the scarcity of GPU cores prevents the concurrent execution of multiple jobs. Thus we employ CUDA-streams \cite{cudaguide} for faster execution such that when GPU is released for one scale, its CPU work begins, while at the same time, GPU is allocated to another scale.
\par
For extraction, ORB descriptor \cite{orb} is used due to its speed and uniqueness but existing implementations extract serially for each scale \cite{opencv, arrayfire}. Since we have the multiscale key points ready, we perform the extraction for all scales at once. To do so, we parallelize Gaussian filtering via CUDA streams, which is essential for robustness \cite{orb2}. Then the ORB extraction, which is quite a time-consuming step and prevents memory coalescing, is performed at once for all scales, leading to high-speed multiscale detection and extraction (Fig.~\ref{fig:det_ext}).
%
\subsubsection{Middle-end}
Despite the accuracy, stereo visual SLAM \cite{orb2} poses a high computing burden for Jetson-like devices. In addition, unlike VO, conic projections of the map points in feature tracking are also time-consuming \cite{orb2}. Hence to attain high throughput, we parallelize both of them, which now form the middle-end. However, na\"ively doing so results in inefficiency since stereo matching requires the descriptors and images to be present in the GPU memory. As they are also used by many SLAM components, creating their multiple copies is not desirable for a resource-constraint platform.
\par
Thus, we design an information flow that allows data sharing between the frontend and the middle-end (Fig.~\ref{fig:slampipeline}). It saves memory consumption by preventing duplication, which in turn avoids CPU-GPU data transfer overhead. Achieving this task is programmatically complex, however we tackle it via our synchronized shared memory, as discussed below.
\subsubsection{Synchronized Shared Memory (\texttt{SSM})}
We adopt synchronized memory primitives from \cite{caffe}, and on top of which, we build synchronized shared memory (\texttt{SSM}) that wraps CPU-GPU transfers and memory allocation/de-allocation calls. Since feature count across frames keeps varying, stereo-matching and tracking demand variable memory. In such cases, \texttt{SSM} reduces dynamic memory allocation/de-allocations calls by performing them only when the requested memory exceeds the current size. Also when CPU-GPU memory is accessed, \texttt{SSM} on its own transfers the underlying data to the destination device, reducing the framework's complexity. 
%

%% file: plots/slam_figure_final.tex
\begin{figure}[!t]
\centering

\colorlet{clrouter}{white!95!black}

\colorlet{clrdg}{white!98!cyan}
\colorlet{clrcpu}{white!60!blue}
\colorlet{clrgpu}{white!60!magenta}

\colorlet{dclrouter}{white!60!magenta}

\colorlet{dclr}{white!70!black}
\colorlet{dclrcg}{white!90!black}
\colorlet{attribdclr}{white!70!cyan}
\colorlet{attribclr}{white!99!black}

\colorlet{imdclr}{pink!60!black}
\colorlet{imclr}{white!95!gray}

\colorlet{cdetdclr}{white!20!cyan}
\colorlet{cdetclr}{white!95!gray}

\colorlet{pcadclr}{white!20!magenta}
\colorlet{pcaclr}{white!95!gray}

\colorlet{orbdclr}{yellow!80!black}
\colorlet{orbclr}{white!95!gray}

\colorlet{impdclr}{black!40!green}
\colorlet{impclr}{white!95!gray}

\colorlet{indclr}{red!50!cyan}

\colorlet{compdclr}{black!90!white}
\colorlet{compclr}{white!100!gray}
\colorlet{endclr}{white!100!gray}
\colorlet{algoindclr}{white!100!gray}

\colorlet{smemdclr}{white!10!gray}
\colorlet{smemclr}{white!60!white}

\colorlet{mptclr}{black!20!yellow}

\colorlet{dsdclr}{white!50!cyan}
\colorlet{dsclr}{white!90!gray}

\colorlet{camdclr}{white!50!blue}
\colorlet{camclr}{white!90!blue}


\colorlet{framedclr}{white!40!black}
\colorlet{frameclr}{white!80!gray}

\colorlet{impldclr}{white!85!gray}
\colorlet{implfillclr}{black!5!impldclr}
\colorlet{softdclr}{impldclr}
\colorlet{softfillclr}{implfillclr}
\colorlet{algodclr}{white!0!brown}
\colorlet{algofillclr}{white!60!brown}



\colorlet{clr}{green!40!purple}
\colorlet{cclr}{red!40!blue}

\colorlet{gpudclr}{white!30!gray}
\colorlet{gpuclr}{white!90!gray}

\colorlet{cpudclr}{white!60!cclr}
\colorlet{cpuclr}{white!60!gray}

\colorlet{linkdclr}{red!40!black}
\colorlet{linkclr}{white!30!linkdclr}


\FPeval{\gpuon}{1.45}
\FPeval{\gpuoff}{0.3}

\FPeval{\frameon}{2}
\FPeval{\frameoff}{0}

\FPeval{\compon}{2}
\FPeval{\compoff}{0.0}

\FPeval{\rcorners}{0.2}

\FPeval{\dotradii}{0.1}
\FPeval{\dotdistance}{1.0}
\FPeval{\patternlw}{0.1}

\newcommand{\hatchpattern}{crosshatch}

\colorlet{t}{black!10!white}
\colorlet{outerclr}{white!95!gray}

\begin{tikzpicture}

\node [scale=0.68]{
\resizebox{81ex}{36ex}
{\tikz{
\node (outer) [draw=t, fill=white, rectangle, line width=0.3ex, rounded corners=0.2ex, minimum width=80ex, minimum height=36ex,xshift=4.25ex, yshift=-4.5ex]{};
\node (rect) [draw=algodclr, fill=algofillclr, rectangle, line width=0.1ex, rounded corners=0.2ex, minimum width=1ex, minimum height=1ex,xshift=11ex, yshift=11.7ex, scale=0.6]{Our Algorithmic Conributions};
\node (rect) [draw=impldclr, fill=implfillclr, draw=impldclr, fill=implfillclr, pattern={\hatchpattern[radius=\dotradii ex, angle=45,  distance=\dotdistance ex]}, pattern color=implfillclr,rectangle, line width=0.1ex, rounded corners=0.2ex, minimum width=1ex, minimum height=1ex,xshift=32ex, yshift=11.7ex, scale=0.6]{Our System Development Conributions};
\node (ds)[xshift=0.0ex, yshift=-7ex, scale=0.8] {
\tikz{
\node () [draw=impldclr, fill=implfillclr, pattern={\hatchpattern[radius=\dotradii ex,   angle=45, distance=\dotdistance ex]}, pattern color=implfillclr, rounded corners=0.2ex, minimum width=23.0ex, minimum height=34.9ex, yshift=-6.25ex, scale=1.0]{};
%
%
\node (smem)[yshift=9.3ex, scale=0.9]{\textsc{\shortstack{Synchronized \\ Shared Memory}}};
\node (lframe) [xshift=-5.8ex,yshift=-7.50ex]{
\tikz{
\node (frame) [draw=framedclr, fill=frameclr,rounded corners=0.2ex, minimum width=10.0ex, minimum height=25.9ex]{};
\node (frame) [draw=framedclr, fill=frameclr,rounded corners=0.2ex, dash pattern=on \frameon pt off \frameoff pt,minimum width=1ex, minimum height=1ex, xshift=-1.5ex, yshift=13.8ex, scale=0.6]{Left Frame};
\node (frame) [draw=none, fill=frameclr, minimum width=44.2ex, minimum height=4ex, xshift=-1.5ex, yshift=12.90ex, scale=0.15]{};
}};
\node (rframe) [xshift=5.8ex,yshift=-7.40ex]{
\tikz{
\node (frame) [draw=framedclr, fill=frameclr,rounded corners=0.2ex, minimum width=10.0ex, minimum height=25.9ex]{};
\node (frame) [draw=framedclr, fill=frameclr,rounded corners=0.2ex, dash pattern=on \frameon pt off \frameoff pt,minimum width=1ex, minimum height=1ex, xshift=-1.15ex, yshift=13.9ex, scale=0.6]{Right Frame};
\node (frame) [draw=none, fill=frameclr, minimum width=49.2ex, minimum height=4ex, xshift=-1.15ex, yshift=12.90ex, scale=0.15]{};
}};
\node (sim) [draw=none,xshift=0ex, yshift=2ex]{
\tikz{
\node (image) [draw= imdclr, fill= smemclr, rounded corners=0.3ex,minimum width=17.0ex, minimum height=2.5ex, xshift=0.1ex, yshift=0.1ex]{};
\node (image) [draw= none, xshift=-5ex, scale=0.8]{Image};
\node (image) [draw= none, xshift=5ex, scale=0.8]{Image};
}};
\node (simp) [draw=none,xshift=0ex, yshift=-5.5ex]{
\tikz{
\node (image) [draw= impdclr, fill= smemclr, rounded corners=0.3ex,minimum width=20ex, minimum height=4.0ex, xshift=0.1ex, yshift=0.1ex]{};
\node (image) [draw= none, xshift=-5ex, scale=0.8]{\shortstack{Image \\ Pyramid}};
\node (image) [draw= none, xshift=5ex, scale=0.8]{\shortstack{Image \\ Pyramid}};
}};
\node (skp) [draw=none,xshift=0ex, yshift=-12.0ex]{
\tikz{
\node (image) [draw= pcadclr, fill= smemclr, rounded corners=0.3ex,minimum width=20ex, minimum height=2.5ex, xshift=0.15ex, yshift=0.1ex]{};
\node (image) [draw= none, xshift=-5ex, scale=0.8]{Keypoints};
\node (image) [draw= none, xshift=5ex, scale=0.8]{Keypoints};
}};
\node (sdesc) [draw=none,xshift=0ex, yshift=-19.8ex]{
\tikz{
\node (image) [draw= dsdclr, fill= smemclr, rounded corners=0.3ex,minimum width=20.0ex, minimum height=2.5ex, xshift=0.1ex, yshift=0.1ex]{};
\node (image) [draw= none, xshift=-5ex, scale=0.8]{Descriptors};
\node (image) [draw= none, xshift=5ex, scale=0.8]{Descriptors};
}};
%
%
}};
\node (fe) [xshift=-22.5ex, yshift=-7.0ex, scale=0.8]{
\tikz{
\node () [draw=impldclr, fill=implfillclr, pattern={\hatchpattern[radius=\dotradii ex,  angle=45, distance=\dotdistance ex]}, pattern color=implfillclr, rounded corners=0.2ex, minimum width=30.5ex, minimum height=34.9ex, yshift=-6.25ex, scale=1.0]{};
%
%
\node (smem)[yshift=10ex, scale=0.9]{\textsc{Front-End}};
\node (im) [yshift=6ex]{\tikz{
\node (impb) [draw=indclr, fill=algoindclr,rounded corners=0.3ex, dash pattern=on \compon pt off \compoff pt, minimum width=13ex, minimum height=2.5ex, yshift=0.0ex, scale=1.0]{};
\node (imp) [yshift=0.0ex, scale=0.8]{\shortstack{Image Pyramid}};
}};
\node (corner) [yshift=-2ex]{\tikz{
\node (cdb) [draw=indclr, fill=algoindclr, rounded corners=0.3ex, dash pattern=on \compon pt off \compoff pt,minimum width=23.5ex, minimum height=9.25ex, yshift=-1.3ex, scale=1.0]{};
\node (cd) [yshift=2.5ex, scale=0.70]{\shortstack{\textbf{\texttt{Corner Detection}}}};
\node (ctb) [draw=algodclr,fill=algofillclr, rounded corners=0.3ex, minimum width=20.5ex, minimum height=2.5ex, yshift=-0.25ex, scale=1.0]{};
\node (ct) [yshift=-0.25ex, scale=0.8]{\shortstack{Bounded Rectification}};
\node (crfb) [draw=compdclr,fill=compclr, rounded corners=0.3ex, minimum width=21.0ex, minimum height=2.5ex, yshift=-4.25ex, scale=1.0]{};
\node (crf) [yshift=-4.25ex, scale=0.8]{\shortstack{Corner Response Function}};%
\draw [->] (ctb.south) -- (crfb.north);
}};
\node (pcs) [yshift=-13.4ex]{\tikz{
\node (pcab) [draw=indclr, fill=algoindclr,rounded corners=0.3ex, dash pattern=on \compon pt off \compoff pt, minimum width=29.5ex, minimum height=9.25ex, yshift=-1.3ex, scale=1.0]{};
%
%
\node (pca) [yshift=2.5ex, scale=0.70]{\shortstack{\textbf{\texttt{PyCA}}}};
\node (cab) [draw=algodclr,fill=algofillclr, rounded corners=0.3ex,  minimum width=16.25ex, minimum height=2.5ex, yshift=-0.25ex, scale=1.0]{};
\node (ca) [yshift=-0.25ex, scale=0.8]{\shortstack{Feature Culling (\texttt{FC})}};
\node (pfab) [draw=algodclr,fill=algofillclr, rounded corners=0.3ex,  minimum width=28.5ex, minimum height=2.5ex, yshift=-4.25ex, scale=1.0]{};
\node (pfa) [yshift=-4.25ex, scale=0.8]{\shortstack{Pyramidal Feature Aggregation (\texttt{PFA})}};%
\draw [->] (cab.south) -- (pfab.north);
}};
\node (desc) [yshift=-21.8ex]{\tikz{
\node (orbb) [draw=indclr, fill=algoindclr,rounded corners=0.3ex, dash pattern=on \compon pt off \compoff pt, minimum width=21.5ex, minimum height=2.5ex, yshift=-18.5ex, scale=1.0]{};
\node (orb) [yshift=-18.5ex,scale=0.8]{\shortstack{ORB Extraction}};%
}};
\draw [->] ($(im.south)-(-0ex,-0.8ex)$) -- ($(corner.north)-(-0ex,0.8ex)$);
\draw [->] ($(corner.south)-(-0ex,-0.8ex)$) -- ($(pcs.north)-(-0ex,0.9ex)$);
\draw [->] ($(pcs.south)-(-0ex,-0.8ex)$) -- ($(desc.north)-(-0ex,0.8ex)$);
%
}};
\node (me) [xshift=17.70ex, yshift=-7.0ex, scale=0.8]{
\tikz{
\node () [draw=impldclr, fill=implfillclr, pattern={\hatchpattern[radius=\dotradii ex,  angle=45, distance=\dotdistance ex]}, pattern color=implfillclr, line width=\patternlw ex,rounded corners=0.2ex, minimum width=18.5ex, minimum height=34.9ex, yshift=-6.25ex, scale=1.0]{};
%
%
\node (smem)[yshift=10ex,scale=0.9]{\textsc{Middle-End}};
\node (corner) [yshift=1.1ex]{\tikz{
\node (smb) [draw=indclr, fill=algoindclr,rounded corners=0.2ex, dash pattern=on \compon pt off \compoff pt, minimum width=17.0ex, minimum height=13.5ex, yshift=3ex, scale=1.0]{};
\node (sm) [yshift=8.6ex, scale=0.70]{\shortstack{\texttt{\textbf{Stereo Matching}}}};
\node (cmb) [draw=compdclr, fill=compclr,rounded corners=0.3ex, minimum width=12.5ex, minimum height=4.0ex, yshift=4.9ex, scale=1.0]{};
\node (cm) [yshift=4.9ex, scale=0.8]{\shortstack{Correlation \\ Minimization}};
\node (ddmb) [draw=compdclr, fill=compclr,rounded corners=0.3ex, minimum width=16.0ex, minimum height=4.0ex, yshift=-0.8ex, scale=1.0]{};
\node (ddm) [yshift=-0.8ex, scale=0.8]{\shortstack{Descriptor Distance \\ Minimization}};%
%
%
\draw [->] ($(cmb.south)-(-0ex,-0.0ex)$) -- ($(ddmb.north)-(-0ex,0.0ex)$);
}};
\node (cp) [yshift=-14.5ex]{\tikz{
\node () [draw=indclr, fill=algoindclr,rounded corners=0.3ex, dash pattern=on \compon pt off \compoff pt, minimum width=15.5ex, minimum height=2.5ex, yshift=-18.5ex, scale=1.0]{};
\node () [yshift=-18.5ex,scale=0.8]{\shortstack{Conic Projections}};%
}};
\node (dm) [yshift=-20ex]{\tikz{
\node () [draw=indclr, fill=algoindclr,rounded corners=0.3ex, dash pattern=on \compon pt off \compoff pt, minimum width=15.5ex, minimum height=2.5ex, yshift=-18.5ex, scale=1.0]{};
\node () [yshift=-18.5ex,scale=0.8]{\shortstack{Descriptor Matching}};%
}};
%
\draw [<->, black, rounded corners=0.2ex, line width=0.1ex,] ($(smb.west)-(-0.5ex,5.8ex)$) -| ($(dm.west)-(0.0ex,0.0ex)$) -- ($(dm.west)-(-1.1ex,0.0ex)$);
}};
\node (be) [xshift=35.20ex, yshift=-7.0ex, scale=0.8]{
\tikz{
\node () [draw=black, fill=endclr, line width=\patternlw ex,rounded corners=0.2ex, minimum width=20.5ex, minimum height=34.8ex, yshift=-6.25ex]{};
\node (smem)[yshift=10ex,scale=0.9]{\textsc{Back-End}};
\node (gob) [draw=indclr, fill=algoindclr,rounded corners=0.3ex, dash pattern=on \compon pt off \compoff pt, minimum width=16.5ex, minimum height=2.5ex, yshift=5.5ex, scale=1.0]{};
\node () [yshift=5.5ex, scale=0.8]{\shortstack{Graph Optimization}};
\node (lcb) [draw=indclr, fill=algoindclr,rounded corners=0.3ex, dash pattern=on \compon pt off \compoff pt, minimum width=12.5ex, minimum height=2.5ex, yshift=-1ex, scale=1.0]{};
\node () [yshift=-1ex, scale=0.8]{\shortstack{Loop Closure}};
\node (corner) [yshift=-14.5ex]{\tikz{
\node () [draw=indclr, fill=algoindclr,rounded corners=0.3ex, dash pattern=on \compon pt off \compoff pt, minimum width=18.5ex, minimum height=14.5ex, yshift=-3.5ex, scale=1.0]{};
\node () [yshift=2.8ex, scale=0.70]{\shortstack{\texttt{\textbf{Mapping}}}};
\node (meb) [draw=compdclr, fill=compclr,rounded corners=0.3ex, minimum width=13.5ex, minimum height=2.5ex, yshift=-1ex, scale=1.0]{};
\node () [yshift=-1ex, scale=0.8]{\shortstack{Map Elements}};
\node (mcb) [draw=compdclr, fill=compclr,rounded corners=0.3ex,  minimum width=12.5ex, minimum height=2.5ex, yshift=-5ex, scale=1.0]{};
\node () [yshift=-5ex, scale=0.8]{\shortstack{Map Culling}};%
\node (mptb) [draw=compdclr, fill=compclr,rounded corners=0.3ex, minimum width=17.0ex, minimum height=2.5ex, yshift=-9ex, scale=1.0]{};
\node () [yshift=-9ex, scale=0.8]{\shortstack{Map Point Tracking}};%
\draw [<->] ($(meb.south)-(-0ex,-0.0ex)$) -- ($(mcb.north)-(-0ex,0.0ex)$);
\draw [->] ($(meb.east)-(-0ex,-0.0ex)$) -| ($(mptb.north)-(-7.4ex,0.0ex)$);
}};
\draw [<->] ($(lcb.south)-(-4.5ex,-0.0ex)$) -| ($(corner.north)-(-4.5ex,4.3ex)$);
\draw [<->] ($(gob.south)-(7.0ex,0.1ex)$) -| ($(corner.north)-(7.0ex,0.9ex)$);
}};
\node (cam0) [xshift=-25ex, yshift=12ex, scale=1.2]
{
\tikz{
\node (rect) [draw=camdclr, fill=camclr, rounded corners=0.2ex, rectangle, minimum width=5.5ex, minimum height=1ex, scale=0.5]{Stereo-Rig};
\draw [draw=camdclr,fill=camclr] ($(rect.west)-(0ex,-0.30ex)$) -- ($(rect.north)-(4.0ex, 0.10ex)$) -- ($(rect.south)-(4.0ex,-0.10ex)$) --  ($(rect.west)-(0ex,0.30ex)$) -- ($(rect.west)-(0ex,-0.30ex)$) ;
}};
\node (gpu) [xshift=-5.1ex,yshift=-6.25ex]
{
\tikz{
\node (cpub) [draw=gpudclr, fill=none, rectangle, line width=0.4 ex, rounded corners=0.2ex, minimum width=60.0ex, minimum height=28.0ex,xshift=0ex, yshift=0ex]{};
\node (gpusb) [draw=gpudclr,fill=white!0!gpudclr,  line width=0.07ex,rounded corners=0.2ex, xshift=-28.2ex, yshift=14.6ex,minimum width=5ex, minimum height=2ex,scale=0.8]{};
\node (gpu) [xshift=-28.2ex, yshift=14.6ex,minimum height=1ex,scale=0.8]{GPU};
}};
\node (cpu) [xshift=35.0ex,yshift=-6.25ex]
{
\tikz{
\node (cpub) [draw=cpudclr, fill=none, rectangle, line width=0.4 ex, rounded corners=0.2ex, minimum width=16.5ex, minimum height=28.0ex,xshift=0ex, yshift=0ex]{};
\node (gpusb) [draw=cpudclr,fill=white!0!cpudclr,  line width=0.07ex,rounded corners=0.2ex, xshift=-6.47ex, yshift=14.6ex,minimum width=5ex, minimum height=2ex,scale=0.8]{};
\node (gpu) [xshift=-6.47ex, yshift=14.6ex,minimum height=1ex,scale=0.8]{CPU};
}};
\colorlet{linkdclr}{black!40!black}
\colorlet{linkclr}{white!70!linkdclr}
\FPeval{\bglw}{0.25}
\FPeval{\lw}{0.7*\bglw}
\FPeval{\fgmix}{10}
\FPeval{\bgmix}{0}
\colorlet{bgimdclr}{white!\fgmix!imdclr}
\colorlet{fgimdclr}{black!\bgmix!imdclr}
\colorlet{bgimpdclr}{white!\fgmix!impdclr}
\colorlet{fgimpdclr}{black!\bgmix!impdclr}
\colorlet{bgpcadclr}{white!\fgmix!pcadclr}
\colorlet{fgpcadclr}{black!\bgmix!pcadclr}
\colorlet{bgdsdclr}{white!\fgmix!dsdclr}
\colorlet{fgdsdclr}{black!\bgmix!dsdclr}
\colorlet{bgmptclr}{white!\fgmix!mptclr}
\colorlet{fgmptclr}{black!\bgmix!mptclr}
%
\draw [-, line width=\lw ex, t] ($(outer.west)-(0.0ex,-14.5ex)$) -- ($(outer.east)-(0.0ex,-14.5ex)$) node [fill=outerclr, xshift=-40ex, yshift=-1.3ex,scale=0.8]{\textcolor{black}{\OursAcronym}};
\draw [-, rounded corners=0.2ex,line width=\lw ex, camdclr] ($(cam0.east)-(1.0ex,0.0ex)$) -| ($(ds.west)-(-1.00ex,-7.0ex)$) -- ($(ds.west)-(-3.1ex,-7.0ex)$);
\draw [-, rounded corners=0.2ex,line width=\lw ex, fgimdclr] ($(ds.west)-(-3.1ex,-6.0ex)$) -| ($(fe.east)-(0.2ex,-11.5ex)$) -|  ($(fe.east)-(13.2ex,-10.8ex)$);
 \draw [-, rounded corners=0.2ex,line width=\lw ex,  fgimpdclr] ($(ds.west)-(-1.9ex,-1.5ex)$) -- ($(ds.west)-(0.25ex,-1.5ex)$) |- ($(fe.east)-(7.9ex,-9.8ex)$);
\draw [-, rounded corners=0.2ex,line width=\lw ex, fgimpdclr] ($(fe.east)-(4.5ex,12.0ex)$) -|($(ds.west)-(0.25ex,-1.7ex)$);
\draw [-, rounded corners=0.2ex,line width=\lw ex, fgimpdclr] ($(ds.east)-(1.9ex,-1.5ex)$) -- ($(ds.east)-(0.4ex,-1.5ex)$) |- ($(me.west)-(-3.1ex,-7.9ex)$) ;
\draw [-, rounded corners=0.2ex,line width=\lw ex, fgpcadclr] ($(ds.west)-(-1.9ex,4.1ex)$) -- ($(fe.east)-(1.3ex,4.1ex)$);
\draw [-, rounded corners=0.2ex,line width=\lw ex, fgpcadclr] ($(ds.east)-(1.9ex,4.1ex)$) -- ($(be.west)-(-3.5ex,4.1ex)$);
\draw [-, rounded corners=0.2ex, line width=\lw ex, fgpcadclr] ($(me.west)-(-3.1ex,-6.9ex)$) -| ($(ds.east)-(-0.1ex,4.1ex)$) ;
\draw [-, rounded corners=0.2ex,line width=\lw ex, fgdsdclr] ($(fe.east)-(4.5ex,13.0ex)$) -| ($(ds.west)-(-0.5ex,10.7ex)$) -- ($(ds.west)-(-1.9ex,10.7ex)$);
\draw [-, rounded corners=0.2ex,line width=\lw ex, fgdsdclr] ($(ds.east)-(1.9ex,10.7ex)$) -- ($(ds.east)-(-0.25ex,10.7ex)$) |- ($(be.west)-(-3.5ex,5.0ex)$);
\draw [-, rounded corners=0.2ex, line width=\lw ex, fgdsdclr] ($(me.west)-(-3.45ex,-1.0ex)$) -| ($(me.west)-(-3.45ex,4.9ex)$) ;
%
%
\draw [-, rounded corners=0.2ex, line width=\lw ex, fgmptclr] ($(me.east)-(1.9ex, 6.7ex)$) -|($(me.east)-(0.0ex, 11.1ex)$);
\draw [-, rounded corners=0.2ex, line width=\lw ex, fgmptclr] ($(me.east)-(1.9ex, 11.1ex)$) --  ($(be.west)-(-2.1ex,11.1ex)$);
}
}};

\end{tikzpicture}

\newcommand{\imline}{\raisebox{0.2ex}{\tikz{\node (c0) [draw=none,rectangle,fill=
imdclr, minimum width = 3 ex, minimum height=1 ex,scale=0.4]{};}}}
\newcommand{\impline}{\raisebox{0.2ex}{\tikz{\node (c0) [draw=none,rectangle,fill=impdclr, minimum width = 3 ex, minimum height=1 ex,scale=0.4]{};}}}
\newcommand{\kpline}{\raisebox{0.2ex}{\tikz{\node (c0) [draw=none,rectangle,fill=pcadclr, minimum width = 3 ex, minimum height=1 ex,scale=0.4]{};}}}
\newcommand{\dsline}{\raisebox{0.2ex}{\tikz{\node (c0) [draw=none,rectangle,fill=dsdclr, minimum width = 3 ex, minimum height=1 ex,scale=0.4]{};}}}
\newcommand{\mptline}{\raisebox{0.2ex}{\tikz{\node (c0) [draw=none,rectangle,fill=mptclr, minimum width = 3 ex, minimum height=1 ex,scale=0.4]{};}}}
\newcommand{\arrowline}{\raisebox{0.5ex}{\tikz{\draw[->] (0ex,0ex) -- (2ex,0ex);}}}
\newcommand{\algocontrib}{\raisebox{-1.25ex}{\tikz{\node (c0) [draw=algodclr, fill=algofillclr, rectangle, line width=0.1ex, rounded corners=0.2ex, minimum width=11ex, minimum height=2.25ex,xshift=0ex, yshift=0.1ex, scale=1.0]{}; \node (c0) [xshift=0ex, yshift=0ex, scale=1.0]{algorithmic};}}}
\newcommand{\systemcontrib}{\raisebox{-1.25ex}{\tikz{\node (c0) [draw=impldclr, fill=implfillclr, draw=impldclr, fill=implfillclr, pattern={\hatchpattern[radius=\dotradii ex, angle=45,  distance=\dotdistance ex]}, pattern color=implfillclr,rectangle, line width=0.1ex, rounded corners=0.2ex, minimum width=19ex, minimum height=2.25ex,xshift=0ex, yshift=0.1ex, scale=1.0]{}; \node (c0) [xshift=0ex, yshift=0ex, scale=1.0]{system development};}}}
%
%

\vspace{-0.5ex}
\caption{
%
%
\OursAcronym~design, highlighting our contributions i.e. bounded rectification, \texttt{PyCA}, Synchronized Shared Memory (SSM), and Middle-end. \OursAcronym~utilizes them to achieve high speeds and resource efficiency in multi-scale stereo setting. The lines `\protect\imline', `\protect\impline', `\protect\kpline', `\protect\dsline' depict the consumption of shared SSM objects among various SLAM components, interconnected via the lines `\protect\mptline', `\protect\arrowline'.
}
\label{fig:slampipeline}
\vspace{-3.5ex}
\end{figure}
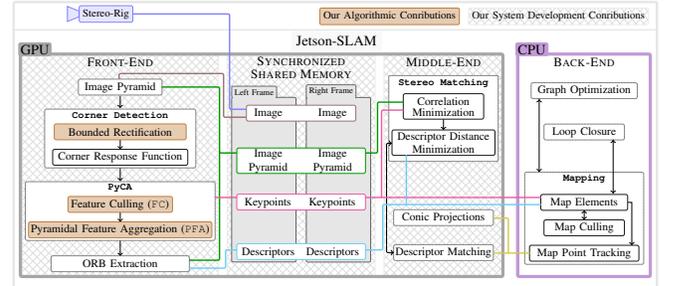

%% file: experiments.tex
\input{tables/gpu_specs}
\input{plots/br_eval}
We evaluate our contributions on a desktop GPU: NVIDIA RTX-$2070$ ($200$W) and an embedded device: Jetson-NX ($10$W). Please see Table~\ref{tab:specs} for the GPU specifications.
\subsection{Bounded Rectification For Corner Detection (Sec.~\ref{sec:fast})}
\label{sec:expcs}
Due to the lack of a labeled corner dataset, we use synthetic data following \cite{cornerbenchmark}. We perform rotation transformation as it is the most challenging case \cite{cornerbenchmark} to analyze Repeatability (F$1$-score) of bounded rectification.
%
%
%
Fig.~\ref{fig:repeatability} shows bounded rectification drastically improves the corner quality while outperforming \cite{opencv,arrayfire}. Fig.~\ref{fig:constrainedthreshres} shows a qualitative analysis. 
%


\vspace{-1.5ex}
\subsection{Pyramidal Culling and Aggregation (\texttt{PyCA}) (Sec.~\ref{sec:pyca})}
\label{sec:exppyca}
We evaluate \texttt{PyCA} comprehensively since it governs the front-end speed. However, due to the lack of aligned baselines, we evaluate its feature culling (\texttt{FC}), and pyramidal feature aggregation (\texttt{PFA}) steps separately.
\subsubsection{Feature Culling (\texttt{FC})}
\input{plots/pyca_eval}
\cite{fastergpu} is loosely comparable with \texttt{PyCA} but in \texttt{FC} mode only. We compare the wall-time (obtained via NVIDIA profiler) for which \texttt{FC} and \cite{fastergpu} hold the computing resources. It was done for different resolutions, cell sizes, and GPUs since these variables govern the computations required and the available computing power. See Fig.~\ref{fig:pyca_fc_eval}.
\par
Notably, \texttt{FC} has a roughly linear timing profile w.r.t the cell size, whereas \cite{fastergpu} grows exponentially. The smaller cells are a major concern in the multiscale setting and core scarcity because they consume more time and cause GPU wastage. However, our \texttt{FC} can handle them easily, as evident that the runtime of \texttt{FC} is drastically lower than \cite{fastergpu} in smaller cells, irrespective of the GPU. This advantage is attributed to our resource-efficient \texttt{MLPT} and \texttt{TEWA} schemes.
\subsubsection{Pyramidal Feature Aggregation \texttt{PFA}}
\label{sec:exppfa}
Due to the lack of matching baselines, we evaluate \texttt{PFA} by studying its runtime on CPU and GPU, and its effect on the number of features by varying resolution and scales, as claimed in Sec.~\ref{sec:pycapfa}.
\par
In the runtime analysis (Fig.~\ref{fig:pfa_cpu_gpu_rtx},~\ref{fig:pfa_cpu_gpu_jetson}), we observe a huge gap in the CPU and GPU modes regardless of the GPU device, thanks to \texttt{PFA}'s parallelizable and sparse-NMS design. While the other analysis (Fig.~\ref{fig:pfa_num_features}) shows that \texttt{PFA} significantly reduces the number of features regardless of the resolution. 
\par
\textit{Conclusively}, the high-speed feature culling (\texttt{FC}) and feature aggregation via \texttt{PFA} drastically improves the \OursAcronym's speed, even in the multiscale stereo mode on low-powered devices having core scarcity. The smaller number of features lowers the \OursAcronym's middle-end and backend runtime by $\sim 2-4$ms even on high-resolution KITTI images, without affecting the SLAM accuracy. This allows timely allocation of the computing resources to other onboard compute-intensive sub-systems, such as deep neural networks. 
\input{plots/fast_det_eval_combined} 
\vspace{-1ex}
\subsection{High-Speed FAST Detection \& Extraction}

\subsubsection{$\mu$-Second Efficient CRF Computations (Sec.~\ref{sec:fastesetgpu})}
Computing CRF-matrix is the foremost step in FAST detection. Hence we compare the GPU kernel performance to compute CRF-matrix, enhanced with bounded rectification against \cite{fastergpu} (Fig.~\ref{fig:crf_eval}). On the RTX GPU, similar runtime is observed due to many cores ($2304$). However, on Jetson-NX with only $384$ cores, ours runs $50\%$ faster regardless of the resolution, thanks to the simplified GPU kernel that avoids warp divergence and guarantees coalesced access to image data. This contrasts to \cite{fastergpu}, which performs additional address extraction steps in the look-up tables, leading to warp divergence and slower speeds in case of GPU core scarcity. 
\subsubsection{High-Speed FAST Detection}
We compare frame rate of our \texttt{PyCA}-based FAST detection with single-scale \cite{arrayfire, opencv} and multi-scale \cite{fastergpu} baselines (See Fig.~\ref{fig:pyca_fastdet_ms_eval},~\ref{fig:pyca_fastdet_ss_eval}). Notably, \texttt{PyCA} is quite faster with predominant gains on Jetson-NX. 
\subsubsection{High-Speed Multiscale Detection-Extraction (Sec.~\ref{sec:fastdetextgpu})}
Since, FAST detection and Extraction are also useful in many applications other than VO or SLAM, such as sparse optical flow or calibration etc., we present the throughput of our FAST detection and ORB extraction modules across resolutions and scales for their use in such applications. The timings include image upload to GPU, image-pyramid, CRF, \texttt{PyCA}, ORB extraction, and keypoint download to CPU.
\par
From Fig.~\ref{fig:det_ext}, 
detection on Jetson-NX at a high resolution of $1242 \times 375$ can run at $1000$FPS for single scale and $250$FPS for eight scales, and the detection-and-extraction can run at $250$ FPS for single scale and $80$FPS for eight scales. For a smaller resolution of $320 \times 240$ (popular in UAV), our method runs at $2000$FPS for single scale detection and at $800$FPS for eight scales, while for detection-and-extraction, it reaches $250$FPS even for eight scales which is huge. It fulfils our goal and claim of using the computing resources for a short duration which is a key requirement in the modern robotic autonomy solutions having many sub-systems \cite{towards}.
\subsection{SLAM Runtime Analysis of \OursAcronym}
We evaluate \OursAcronym{} with two backends: \ORBBE~\cite{orb2}, ICE-BA \cite{iceba}, and three datasets: KITTI \cite{kitti}, EuRoC \cite{euroc}, KAIST-VIO \cite{kaistvio}. KITTI dataset is collected via a self-driving test bed, while EuRoC and KAIST-VIO are collected via a UAV flying indoors. These datasets have several events of saturation \cite{kitti}, severely low lighting and low texture \cite{euroc}, and rapid rotations \cite{kaistvio}, thus sufficient to rigorously analyze a frontend and a SLAM system. Following \cite{orb2}, we report RMSE Absolute Trajectory Error (ATE) and frame rate.
\subsubsection{Effect of \texttt{PyCA} on the Middle-end and the Backend}
Fig~\ref{fig:slamcomponents} shows that \OursAcronym{} with \ORBBE~backend is $85\%$ faster on RTX-$2070$ GPU, whereas $65\%$ faster on Jetson-NX $@752 \times 480$ for eight scales and stereo mode, showing major achievement of this work. Interestingly, the backend still uses CPU and has not been altered to run faster, nonetheless, \texttt{PyCA} improves the backend's speed, justifying that \texttt{PyCA} produces robust features. The overall frame rate is bottlenecked by the backend, especially on Jetson-NX, which opens future possibilities to speed up the backend.
\subsubsection{Throughput Analysis of Jetson-SLAM over Datasets}
We also analyze the SLAM frame rate on high-resolution KITTI and EuRoC images with different backends. See Fig~\ref{fig:slam_kitti},~\ref{fig:slam_euroc}. Notably, \OursAcronym{} achieves an average speed-up of $80\%$ on RTX-$2070$ and $67\%$ on Jetson-NX, even at eight scales and stereo mode, which is huge (see video).
\input{plots/slam_throughput}
\input{tables/kitti}
\subsection{SLAM Metric Performance of \OursAcronym}
\subsubsection{KITTI \cite{kitti} Dataset}
See Table~\ref{tab:kittiate}. \OursAcronym{} performs better in seq. KITTI-$00$. It has a marginally higher error in other sequences that can be traded for speed, and occurs due to the reduced number of features. This is also evident from trajectories in Fig.~\ref{fig:kitti_traj} where \OursAcronym{} remains close to the ground truth. We also test \cite{fastergpu} with \ORBBE{} backend, but observe tracking failures due to its incapability to produce sufficient map points (see video). 
\subsubsection{EuRoC \cite{euroc} Dataset}
We compare \OursAcronym{} with existing VO/VIO/SLAM pipelines in Table~\ref{tab:existingeurociceba}. \OursAcronym{} achieves the lowest error on all sequences with Full-BA backend while remains among Top-$3$ with ICE-BA backend. This indicates \texttt{PyCA} produces reliable features at high speed, turning \OursAcronym{} faster and accurate. Notably, \OursAcronym{} does not fail in any sequence even without an IMU.
\par
We also show the trajectory analysis in Fig.~\ref{fig:euroc_traj}. It indicates a high overlap of \OursAcronym{} with the ground truth, regardless of the backend, validating the small ATE.
\input{tables/euroc}
\input{tables/kaist_vio}
\subsubsection{KAIST-VIO \cite{kaistvio} dataset} 
Table~\ref{tab:kaistvio} shows the analysis on all the $11$ sequences of this dataset. Notably, \OursAcronym{} achieves the lowest error except \texttt{circle\_head}, where it remains among Top-$3$ with only a marginal ATE difference. 
\par
In this dataset, \textit{rapid heading movement} is the most challenging aspect for SLAM systems to handle without using an Inertial-Measurement-Unit (IMU). Nonetheless, \OursAcronym{}, even without using an IMU, outperforms several baselines relying on IMU. The primary reason is the \OursAcronym{}'s capability to process each frame quickly.
\par
For reference, we also show trajectories by \OursAcronym{} in Fig.~\ref{fig:kaist_traj}, which closely overlaps with ground truth. 
\vspace{-2ex}
\subsection{SLAM-Focused Ablation Study of \OursAcronym}
\subsubsection{Bounded Rectification}
It governs the quality of the features reaching the SLAM backend. Therefore, it is crucial to analyze its effect on the SLAM accuracy. See Table~\ref{tab:cs}. Notably, bounded rectification significantly lowers the SLAM error, verifying the robustness of the detected corners. 
\subsubsection{\texttt{PyCA} Cell-Size}
Table~\ref{tab:pycacellsizeslam} shows that smaller cells at the native scale lead to a large number of features because smaller cells become even smaller at lower resolutions in the multiscale setting. It impacts the middle-end and backend performance due to stereo and feature matching ambiguities, as evident from $15 \times 15$ cell having the highest feature count and error. On the contrary, the cell-size $32 \times 32$ results in $8 \times 8$ cells at the $8^{th}$ scale with a scale factor of $1.2$ but have the lowest feature count, ATE, and runtime. This demonstrates \texttt{PyCA}-based frontend yielding fewer but high-quality features.
\input{plots/traj_plots_combined}
%
\input{tables/br_slam}
\input{tables/pyca_cellsize_slam_ablation}
%
\input{tables/num_scales_ablation}
\subsubsection{Effect of Number of Scales}
Table~\ref{tab:pycanumscalesslam} shows that too few scales (i.e. $2$) at large cell-size results in insufficient points which leads to SLAM failure. Smaller cell size helps but increases the number of features and hence the runtime. On the contrary, more scales with larger cell-size results in sufficiently fewer key points, lower ATE, and a high frame-processing rate of \OursAcronym{}. 
%
%
%

\subsection{\OursAcronym~Resource Utilization Analysis}
\OursAcronym{} utilizes roughly $15\%$ CPU, $40\%$ GPU, and $10\%$ RAM on Jetson-NX, which is quite low, owing to \texttt{PyCA}, \texttt{MLPT}, \texttt{TEWA}, and data-sharing. It allows other sub-systems to utilize GPU, such as deep networks, discussed next.
\subsection{\OursAcronym~Co-existing with Deep Neural Networks}
\label{sec:vgg}
Modern autonomy requires deep networks co-existing with other sub-systems. We show that \OursAcronym~marginally affects the runtime of a deep network without sacrificing its own runtime and accuracy, a major achievement of our work.
\par
To verify that, we choose VGG \cite{vgg} deep network, popular in robotics applications, and construct its three variants: VGG-$1.0$, VGG-$0.5$, and VGG-$0.25$. The variant VGG-$1.0$ has five stages with $\{2,2,2,4,4\}$ layers, and $\{32,64,128,256,256\}$ channels, while the others are scaled w.r.t the $1.0$ variant.
\par
Fig.~\ref{fig:slam_on_vgg} depicts the effect of ORB-SLAM$2$ and \OursAcronym{} onto the frame rate of VGG. Despite using GPU, \OursAcronym{} incurs a frame-rate drop similar to the CPU only ORB-SLAM$2$, indicating GPU efficiency of \OursAcronym{}. This experiment was conducted at a resolution of $640 \times 480$ which is quite high, thus for smaller resolutions, \OursAcronym{} will result in negligible frame-rate drop.
\par
Fig.~\ref{fig:vgg_on_slam_b},~\ref{fig:vgg_on_slam_c} shows how VGG affects SLAM performance. Interestingly, ORB-SLAM$2$ faces a drop in frame rate which is already running below $10$FPS and faces higher ATE error due to its failure to process the frames in time. On the contrary, \OursAcronym{} does not face a drop in ATE but has a minimal drop in frame rate that is still well-above real-time ($30$FPS).
\par
This experiment shows the utility of \OursAcronym{} to develop complex UAV autonomy solutions. We use \OursAcronym{} in a UAV $@432\times240$ with VGG-$1.0$, and do not observe FPS drop in VGG or \OursAcronym{}.
\input{plots/deep_network_slam_all_in_one}

%% file: tables/gpu_specs.tex
\begin{table}[!t]
\centering
\caption{GPU devices specifications.}
\label{tab:specs}
\arrayrulecolor{white!70!black}
\tiny
\setlength{\tabcolsep}{20.0pt}
\vspace{-1ex}
\begin{tabular}{c c c}
\hline
\multicolumn{1}{c}{Attribute} & \multicolumn{1}{|c}{RTX-$2070$} & \multicolumn{1}{|c}{Jetson-NX} \\ \cline{1-3}

\multicolumn{1}{c}{GPU grade} & \multicolumn{1}{|c}{Desktop/Laptop} & \multicolumn{1}{|c}{Edge/Embedded} \\ 

\multicolumn{1}{c}{GPU cores} & \multicolumn{1}{|c}{$2304$} & \multicolumn{1}{|c}{$384$} \\ 
\multicolumn{1}{c}{Clock} & \multicolumn{1}{|c}{$1620$ MHz} & \multicolumn{1}{|c}{$1100$ MHz} \\ 
\multicolumn{1}{c}{Memory bandwidth} & \multicolumn{1}{|c}{$448$ GB/s} & \multicolumn{1}{|c}{$59.7$GB/s} \\ 
\multicolumn{1}{c}{Compute performance} & \multicolumn{1}{|c}{$7.4$ TFLOPs} & \multicolumn{1}{|c}{$1$ TFLOPs} \\ 

 \hline
\end{tabular}
\vspace{-3.5ex}

\end{table}

%% file: plots/br_eval.tex
\begin{figure}[t]

\begin{tikzpicture}

\FPeval{\xshifta}{0}
\FPeval{\yshifta}{0-0}

\FPeval{\xshiftb}{27.5}
\FPeval{\yshiftb}{0+0.4}

\FPeval{\scaleb}{0.60}

\FPeval{\pltw}{30}
\FPeval{\plth}{22}

\FPeval{\mrksize}{0.15}
\FPeval{\linew}{0.2}

\FPeval{\netxshft}{5}
\FPeval{\netyshft}{0}

\FPeval{\lxshfta}{10.5}
\FPeval{\lyshfta}{4.2}

\colorlet{desclr}{white!0!magenta}
\colorlet{currclr}{blue!60!green}

\colorlet{trnclr}{white!50!magenta}
\colorlet{dtrnclr}{white!10!magenta}
\colorlet{mtrnclr}{white!10!magenta}
\colorlet{mdtrnclr}{white!10!magenta}

\colorlet{tstclr}{white!50!blue}
\colorlet{dtstclr}{white!10!blue}
\colorlet{mtstclr}{white!50!blue}
\colorlet{mdtstclr}{white!10!blue}

\colorlet{gridclr}{white!90!black}
\colorlet{dlegendclr}{white!80!black}
\colorlet{axisclr}{white!80!black}
\colorlet{axisbgclr}{white!99!black}

\colorlet{dlegendclr}{white!80!black}
\colorlet{legendclr}{white!100!black}

\colorlet{clr1}{white!0!magenta}
\colorlet{clr2}{black!40!green}
\colorlet{clr3}{white!0!cyan}

\FPeval{\bscal}{0.75}
\FPeval{\labelscale}{0.8}
\FPeval{\ticklabelscale}{0.6}
\FPeval{\lscale}{0.5}

\colorlet{bndryclr}{white!80!black}
\node (bn)[draw=bndryclr, rounded corners=0.2ex,minimum width=54.2ex, minimum height=13.0ex, xshift=16.2ex, yshift=0.0ex]{};
\draw [bndryclr, dashed] ($(bn.north)-(5.5ex, 0ex)$) -- ($(bn.south)-(5.5ex, 0ex)$);
\node (n1) [draw=none, scale = 1.0, xshift =0 ex, yshift=0 ex]{
\scalebox{\bscal}{
\tikz
{
\pgfplotsset{width=\pltw ex, height=\plth ex}
\begin{axis}[
   axis background style={fill=axisbgclr},
    title={},
    xlabel={Rotation Angle},
    ylabel={F$1$-Score},
    xmin=-180, xmax=180,
    ymin=0, ymax=1.0,
   xtick={-175, -100, -50, 0,  50,  100,  175},
   xticklabels={-175$^o$, -100$^o$, -50$^o$, 0$^o$, 50$^o$, 100$^o$,  175$^o$},
       ytick={0, 0.2, 0.4, 0.6, 0.8, 1.0},
   yticklabels={0, 0.2, 0.4, 0.6, 0.8, 1.0},
     axis line style={axisclr},
    legend image post style={scale =0.6},
    legend style={at={(\lxshfta ex,\lyshfta ex)},anchor=south, legend columns = 1, draw = {dlegendclr}, fill={legendclr}, nodes={scale=\lscale}},
    ymajorgrids=true, 
    xmajorgrids=true,
    grid style={dashed, gridclr},
    major tick length=1ex,
    tick label style={scale=\ticklabelscale},
    y label style={at={(4 ex, 5 ex)}, scale=\labelscale,},
    x label style={at={(10 ex, 2 ex)}, scale=\labelscale},
    xticklabel style={rotate=0},
    legend cell align={left}
]
\addplot[
    fill=none,draw = clr1,
    very thick,
    line width= \linew ex,
    mark = square*,
    mark size = \mrksize ex,
    ]
    coordinates {
(-175, 0.225814)
(-150, 0.245847)
(-125, 0.229182)
(-100, 0.239917)
(-75, 0.234427)
(-50, 0.225814)
(-25, 0.225402)
(0, 0.233241)
(25, 0.22773)
(50, 0.229341)
(75, 0.224211)
(100, 0.228048)
(125, 0.229073)
(150, 0.228048)
(175, 0.220916)
};
 %
\addplot[
    fill=none,draw = clr2,
    very thick,
    line width= \linew ex,
    mark = square*,
    mark size = \mrksize ex,
    ]
    coordinates {
(-175, 0.211126)
(-150, 0.270805)
(-125, 0.212992)
(-100, 0.266606)
(-75, 0.261016)
(-50, 0.293138)
(-25, 0.260636)
(0, 0.278767)
(25, 0.26539)
(50, 0.258322)
(75, 0.283755)
(100, 0.251621)
(125, 0.27807)
(150, 0.233319)
(175, 0.196183)
};
\addplot[
    fill=none,draw = clr3,
    very thick,
    line width= \linew ex,
    mark = square*,
    mark size = \mrksize ex,
    ]
    coordinates {
(-175, 0.883988)
(-150, 0.854034)
(-125, 0.814404)
(-100, 0.879362)
(-75, 0.821994)
(-50, 0.8566)
(-25, 0.853583)
(0, 0.84262)
(25, 0.823069)
(50, 0.850938)
(75, 0.829727)
(100, 0.8566)
(125, 0.79962)
(150, 0.784716)
(175, 0.873185)
    };
 \legend{FAST-OpenCV \cite{opencv}, FAST-ArrayFire \cite{arrayfire}, FAST-Proposed}
\end{axis}
}
}};
%
%

%
\FPeval{\imw}{12.5}
\FPeval{\imh}{7.5}
\colorlet{bdrwclr}{white!50!magenta}
\colorlet{cbfillclr}{white!65!black}
\colorlet{bfillclr}{white!90!black}
\node (b) [xshift=\xshiftb ex, yshift=\yshiftb ex, scale = \scaleb]
{
\tikz{
\node (n1) [xshift=0ex, yshift=0ex]{\includegraphics[width=\imw ex, height=\imh ex]{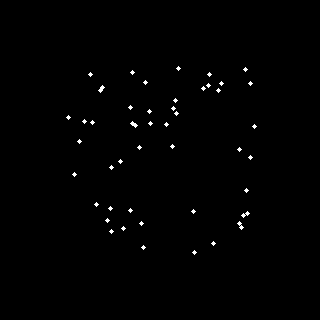}};
\node (n1) [xshift=\imw ex + 0.25ex, yshift=0ex]{\includegraphics[width=\imw ex, height=\imh ex]{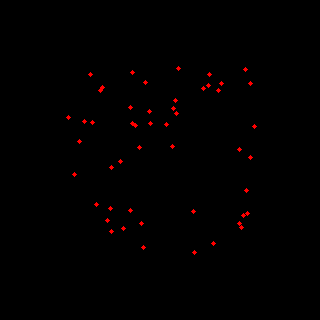}};
\node (n1) [xshift=2*\imw ex + .5ex, yshift=0ex]{\includegraphics[width=\imw ex, height=\imh ex]{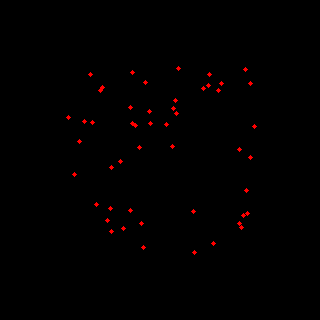}};
\node (n1) [xshift=3*\imw ex + 0.75ex, yshift=0ex]{\includegraphics[width=\imw ex, height=\imh ex]{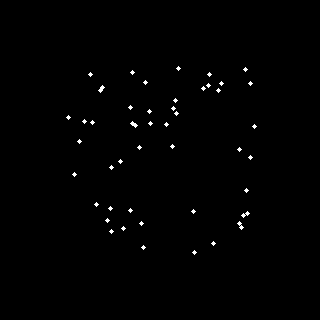}};
\node (n1) [xshift=0ex, yshift=-1 * \imh ex - 0.5ex]{\includegraphics[width=\imw ex, height=\imh ex]{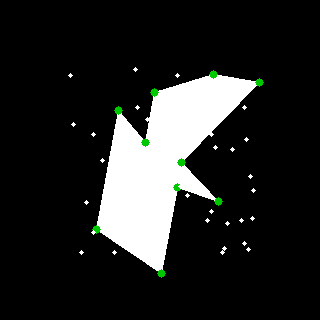}};
\node (n1) [xshift=\imw ex + 0.25ex, yshift=-1 * \imh ex - 0.5ex]{\includegraphics[width=\imw ex, height=\imh ex]{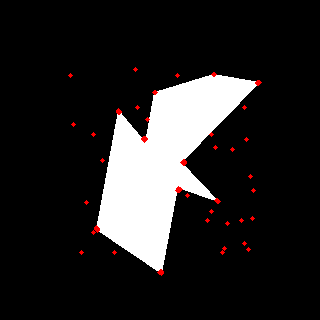}};
\node (n1) [xshift=2*\imw ex + 0.5ex, yshift=-1 * \imh ex - 0.5ex]{\includegraphics[width=\imw ex, height=\imh ex]{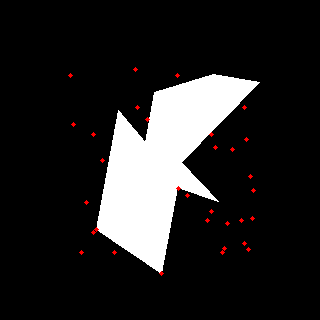}};
\node (n1) [xshift=3*\imw ex + 0.75ex, yshift=-1 * \imh ex - 0.5ex]{\includegraphics[width=\imw ex, height=\imh ex]{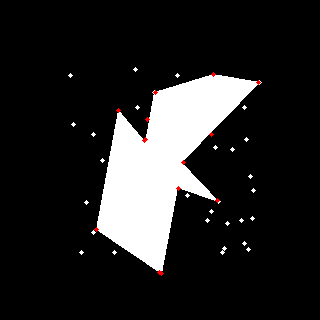}};
\node (a) [draw=none, xshift = -0 ex, yshift=-13.5 ex, scale=1.1]{\footnotesize Input};
\node (b) [draw=none, xshift = 12.9 ex, yshift=-13.5 ex, scale=1.1]{\footnotesize OpenCV \cite{opencv}};
\node (c) [draw=none, xshift = 26.4 ex, yshift=-13.5 ex, scale=1.1]{\footnotesize ArrayFire \cite{arrayfire}};
\node (d) [draw=none, xshift = 39.5 ex, yshift=-13.5 ex, scale=1.1]{\footnotesize Proposed};
}};

\node (a) [draw=none, xshift = -8.5 ex, yshift=-5.7 ex]{\footnotesize (a)};
\node (b) [draw=none, xshift = 13.0 ex, yshift=-5.7 ex]{\footnotesize (b)};

\end{tikzpicture}
\subfloat{\label{fig:repeatability}}
\subfloat{\label{fig:constrainedthreshres}}
\vspace{-0.5ex}
\caption{Evaluation of Bounded Rectification. (a) Repeatability score. (b) Qualitative results. \textit{Top}: only corners where baselines do not face issues. \textit{Bottom}: non-corners and corners where only bounded rectification suppresses non-corners but the baselines \cite{opencv}, \cite{arrayfire} fail. \textcolor{green}{Green} dots are ground truth corners and \textcolor{red}{red} dots are detections.}
\vspace{-3.0ex}
\end{figure}
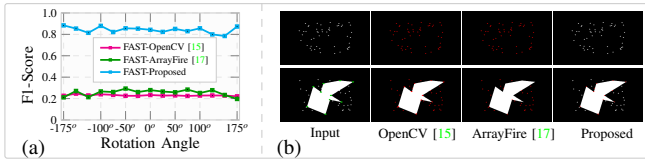

%% file: plots/pyca_eval.tex
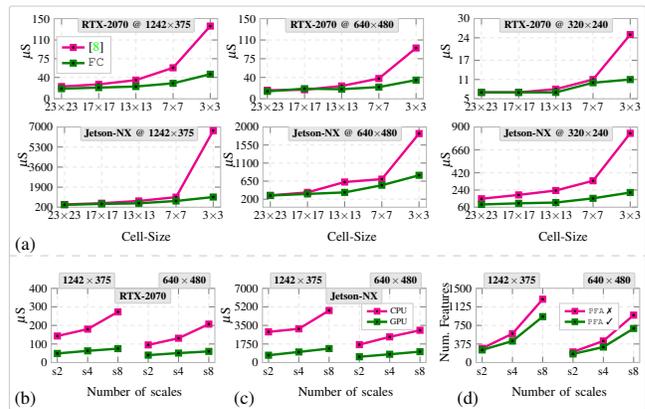
\begin{figure}[!t]
\centering

\begin{tikzpicture}

\FPeval{\xshifta}{0}
\FPeval{\xshiftb}{17.3}
\FPeval{\xshiftc}{35.3}
\FPeval{\xshiftd}{0}
\FPeval{\xshifte}{17.3}
\FPeval{\xshiftf}{35.3}

\FPeval{\yshifta}{0}
\FPeval{\yshiftb}{0}
\FPeval{\yshiftc}{0}
\FPeval{\yshiftd}{0-10.0}
\FPeval{\yshifte}{0-10.0}
\FPeval{\yshiftf}{0-10.0}

\FPeval{\pltw}{37.5}
\FPeval{\plth}{23.5}

\FPeval{\pscal}{0.5}

\FPeval{\ticklabelscale}{0.9}
\FPeval{\labelscale}{1.0}
\FPeval{\xlabelrotate}{0}

\FPeval{\gpuscale}{0.8}
\FPeval{\gpuxshift}{13.8}
\FPeval{\gpuyshift}{12.5}

\FPeval{\barw}{12}
\FPeval{\barsep}{0.8}
\FPeval{\linew}{0.4}
\FPeval{\barcorner}{0.4}

\FPeval{\mrksize}{0.4}
\FPeval{\legendscale}{1.0}
\FPeval{\legendimscale}{1.0}

\FPeval{\legendxshifta}{5.0}
\FPeval{\legendxshiftb}{5.0}
\FPeval{\legendxshiftc}{5.0}
\FPeval{\legendxshiftd}{5.0}
\FPeval{\legendxshifte}{5.0}
\FPeval{\legendxshiftf}{5.0}

\FPeval{\legendyshifta}{10.5}
\FPeval{\legendyshiftb}{10.5}
\FPeval{\legendyshiftc}{10.5}
\FPeval{\legendyshiftd}{10.5}
\FPeval{\legendyshifte}{10.5}
\FPeval{\legendyshiftf}{10.5}

\FPeval{\ylabelxshifta}{3.2}
\FPeval{\ylabelxshiftb}{3.2}
\FPeval{\ylabelxshiftc}{3.2}
\FPeval{\ylabelxshiftd}{2.2}
\FPeval{\ylabelxshifte}{2.2}
\FPeval{\ylabelxshiftf}{3.2}

\FPeval{\ylabelyshifta}{8}
\FPeval{\ylabelyshiftb}{8}
\FPeval{\ylabelyshiftc}{8}
\FPeval{\ylabelyshiftd}{8}
\FPeval{\ylabelyshifte}{8}
\FPeval{\ylabelyshiftf}{8}

\FPeval{\xlabelxshifta}{15}
\FPeval{\xlabelxshiftb}{15}
\FPeval{\xlabelxshiftc}{15}
\FPeval{\xlabelxshiftd}{15}
\FPeval{\xlabelxshifte}{15}
\FPeval{\xlabelxshiftf}{15}

\FPeval{\xlabelyshifta}{1-1.0}
\FPeval{\xlabelyshiftb}{1-1.0}
\FPeval{\xlabelyshiftc}{1-1.0}
\FPeval{\xlabelyshiftd}{1-1.0}
\FPeval{\xlabelyshifte}{1-1.0}
\FPeval{\xlabelyshiftf}{1-1.0}

\colorlet{gridclr}{white!90!black}
\colorlet{dlegendclr}{white!80!black}
\colorlet{axisclr}{white!70!black}
\colorlet{axisbgclr}{white!100!black}

\colorlet{dlegendclr}{white!80!black}
\colorlet{legendclr}{white!100!black}

\colorlet{clr1}{white!0!magenta}
\colorlet{clr2}{black!50!green}
\colorlet{clr3}{white!0!cyan}

\colorlet{clr1fill}{black!30!clr1}
\colorlet{clr2fill}{black!30!clr2}
\colorlet{clr3fill}{black!30!clr3}

\colorlet{gpudclr}{white!80!black}
\colorlet{gpuclr}{white!90!black}

\colorlet{deltaclr}{white!0!black}
\colorlet{gputxtclr}{white!0!black}
\colorlet{bndryclr}{white!80!black}
\node (bn)[draw=bndryclr, rounded corners=0.2ex,minimum width=53.5ex, minimum height=34.0ex, xshift=17.2ex, yshift=-12.2ex]{};
\node (a)[draw=none, xshift=-8.2ex, yshift=-15.3ex,scale=0.7]{(a)};
\node (b)[draw=none, xshift=-8.2ex, yshift=-28.3ex,scale=0.7]{(b)};
\node (c)[draw=none, xshift= 10.2ex, yshift=-28.3ex,scale=0.7]{(c)};
\node (d)[draw=none, xshift=29.0ex, yshift=-28.3ex,scale=0.7]{(d)};
\draw [bndryclr, dashed] ($(bn.west)-(0ex, 4.0ex)$) -- ($(bn.east)-(0, 4.0ex)$);
\node (na) [draw=none, xshift =\xshifta ex, yshift= \yshifta ex]{
\scalebox{\pscal}
{\tikz{
\pgfplotsset{width=\pltw ex, height=\plth ex}
\begin{axis}[
   axis background style={fill=axisbgclr},
    title={},
    xlabel={},
    ylabel={$\mu$S},
    xmin=-0.2, xmax=4.2,
    ymin=0, ymax=150,
    xtick={0, 1, 2, 3, 4},
    ytick={0, 40, 75, 110, 150},
   xticklabels={$23$$\times$$23$, $17$$\times$$17$, $13$$\times$$13$, $7$$\times$$7$, $3$$\times$$3$},
   yticklabels={$0$, $40$, $75$, $110$, $150$},
     axis line style={axisclr},
    legend image post style={scale =\legendimscale},
    legend style={at={(\legendxshifta ex,\legendyshifta ex)},anchor=north, legend columns = 1, draw = {dlegendclr}, fill={legendclr}, nodes={scale=\legendscale}},
    ymajorgrids=true, 
    xmajorgrids=true,
    grid style={dashed, gridclr},
    major tick length=1ex,
    y label style={at={(\ylabelxshifta ex, \ylabelyshifta ex)}, scale=\labelscale},
    x label style={at={(\xlabelxshifta ex, \xlabelyshifta ex)}, scale=\labelscale},
    xticklabel style={rotate=\xlabelrotate, scale= \ticklabelscale},
    yticklabel style={xshift=0.0ex, scale= \ticklabelscale},
]
\addplot[
    fill=none,draw = clr1,
    very thick,
    line width= \linew ex,
    mark = square*,
    mark size = \mrksize ex,
    ]
    coordinates {
(0, 23)
(1, 27)
(2, 35)
(3, 58)
(4, 136) 
    };
\addplot[
    fill=none,draw = clr2,
    very thick,
    line width= \linew ex,
    mark = square*,
    mark size = \mrksize ex,
    ]
    coordinates {
(0, 19)
(1, 21)
(2, 23)
(3, 29)
(4, 46)
};
 %
 \legend{\cite{fastergpu}, \texttt{FC}}
 \node [draw=gpudclr,fill=gpuclr,rounded corners=0.2ex, minimum width=23.7ex, xshift=\gpuxshift ex, yshift=\gpuyshift ex, scale=\gpuscale]{\textcolor{gputxtclr}{\textbf{RTX-}$\mathbf{2070~@~1242}$$\mathbf{\times}$$\mathbf{375}$}};
\end{axis}
}
}};
\node (nb) [draw=none, xshift =\xshiftb ex, yshift= \yshiftb ex]{
\scalebox{\pscal}
{\tikz{
\pgfplotsset{width=\pltw ex, height=\plth ex}
\begin{axis}[
   axis background style={fill=axisbgclr},
    title={},
    xlabel={},
    ylabel={$\mu$S},
    xmin=-0.2, xmax=4.2,
    ymin=0, ymax=150,
    xtick={0, 1, 2, 3, 4},
    ytick={0, 40, 75, 110, 150},
   xticklabels={$23$$\times$$23$, $17$$\times$$17$, $13$$\times$$13$, $7$$\times$$7$, $3$$\times$$3$},
      yticklabels={$0$, $40$, $75$, $110$, $150$},
     axis line style={axisclr},
    legend image post style={scale =\legendimscale},
    legend style={at={(\legendxshiftb ex,\legendyshiftb ex)},anchor=north, legend columns = 1, draw = {dlegendclr}, fill={legendclr}, nodes={scale=\legendscale}},
    ymajorgrids=true, 
    xmajorgrids=true,
    grid style={dashed, gridclr},
    major tick length=1ex,
    y label style={at={(\ylabelxshiftb ex, \ylabelyshiftb ex)}, scale=\labelscale},
    x label style={at={(\xlabelxshiftb ex, \xlabelyshiftb ex)}, scale=\labelscale},
    xticklabel style={rotate=\xlabelrotate, scale= \ticklabelscale},
    yticklabel style={xshift=0.0ex, scale= \ticklabelscale},
]
\addplot[
    fill=none,draw = clr1,
    very thick,
    line width= \linew ex,
    mark = square*,
    mark size = \mrksize ex,
    ]
    coordinates {
(0, 16)
(1, 17)
(2, 24)
(3, 38)
(4, 95)
    };
\addplot[
    fill=none,draw = clr2,
    very thick,
    line width= \linew ex,
    mark = square*,
    mark size = \mrksize ex,
    ]
    coordinates {
(0, 14)
(1, 19)
(2, 18)
(3, 22)
(4, 35)
};
 %
 \node [draw=gpudclr,fill=gpuclr,rounded corners=0.2ex, minimum width=23.7ex, xshift=\gpuxshift ex, yshift=\gpuyshift ex, scale=\gpuscale]{\textcolor{gputxtclr}{\textbf{RTX-}$\mathbf{2070~@~640}$$\mathbf{\times}$$\mathbf{480}$}};
\end{axis}
}
}};
\node (nc) [draw=none, xshift =\xshiftc ex, yshift= \yshiftc ex]{
\scalebox{\pscal}
{\tikz{
\pgfplotsset{width=\pltw ex, height=\plth ex}
\begin{axis}[
   axis background style={fill=axisbgclr},
    title={},
    xlabel={},
    ylabel={$\mu$S},
    xmin=-0.2, xmax=4.2,
    ymin=5, ymax=30,
    xtick={0, 1, 2, 3, 4},
    ytick={5, 11, 17, 24, 30},
   xticklabels={$23$$\times$$23$, $17$$\times$$17$, $13$$\times$$13$, $7$$\times$$7$, $3$$\times$$3$},
    yticklabels={$5$, $11$, $17$, $24$, $30$},
     axis line style={axisclr},
    legend image post style={scale =\legendimscale},
    legend style={at={(\legendxshiftc ex,\legendyshiftc ex)},anchor=north, legend columns = 1, draw = {dlegendclr}, fill={legendclr}, nodes={scale=\legendscale}},
    ymajorgrids=true, 
    xmajorgrids=true,
    grid style={dashed, gridclr},
    major tick length=1ex,
    y label style={at={(\ylabelxshiftc ex, \ylabelyshiftc ex)}, scale=\labelscale},
    x label style={at={(\xlabelxshiftc ex, \xlabelyshiftc ex)}, scale=\labelscale},
    xticklabel style={rotate=\xlabelrotate, scale= \ticklabelscale},
    yticklabel style={xshift=0.0ex, scale= \ticklabelscale},
]
\addplot[
    fill=none,draw = clr1,
    very thick,
    line width= \linew ex,
    mark = square*,
    mark size = \mrksize ex,
    ]
    coordinates {
(0, 7)
(1, 7)
(2, 8)
(3, 11)
(4, 25)
    };
\addplot[
    fill=none,draw = clr2,
    very thick,
    line width= \linew ex,
    mark = square*,
    mark size = \mrksize ex,
    ]
    coordinates {
(0, 7)
(1, 7)
(2, 7)
(3, 10)
(4, 11)
};
 %
 \node [draw=gpudclr,fill=gpuclr,rounded corners=0.2ex, minimum width=23.7ex, xshift=\gpuxshift ex, yshift=\gpuyshift ex, scale=\gpuscale]{\textcolor{gputxtclr}{\textbf{RTX-}$\mathbf{2070~@~320}$$\mathbf{\times}$$\mathbf{240}$}};
\end{axis}
}
}};
\node (nd) [draw=none, xshift =\xshiftd ex, yshift= \yshiftd ex]{
\scalebox{\pscal}
{\tikz{
\pgfplotsset{width=\pltw ex, height=\plth ex}
\begin{axis}[
   axis background style={fill=axisbgclr},
    title={},
    xlabel={Cell-Size},
    ylabel={$\mu$S},
    xmin=-0.2, xmax=4.2,
    ymin=200, ymax=7000,
    xtick={0, 1, 2, 3, 4},
    ytick={200, 1900, 3600, 5300 ,7000},
   xticklabels={$23$$\times$$23$, $17$$\times$$17$, $13$$\times$$13$, $7$$\times$$7$, $3$$\times$$3$},
    yticklabels={$200$, $1900$, $3600$, $5300$, $7000$},
     axis line style={axisclr},
    legend image post style={scale =\legendimscale},
    legend style={at={(\legendxshiftd ex,\legendyshiftd ex)},anchor=north, legend columns = 1, draw = {dlegendclr}, fill={legendclr}, nodes={scale=\legendscale}},
    ymajorgrids=true, 
    xmajorgrids=true,
    grid style={dashed, gridclr},
    major tick length=1ex,
    y label style={at={(\ylabelxshiftd ex, \ylabelyshiftd ex)}, scale=\labelscale},
    x label style={at={(\xlabelxshiftd ex, \xlabelyshiftd ex)}, scale=\labelscale},
    xticklabel style={rotate=\xlabelrotate, scale= \ticklabelscale},
    yticklabel style={xshift=0.0ex, scale= \ticklabelscale},
]
\addplot[
    fill=none,draw = clr1,
    very thick,
    line width= \linew ex,
    mark = square*,
    mark size = \mrksize ex,
    ]
    coordinates {
(0, 431)
(1, 540)
(2, 730)
(3, 1052)
(4, 6687)
    };
\addplot[
    fill=none,draw = clr2,
    very thick,
    line width= \linew ex,
    mark = square*,
    mark size = \mrksize ex,
    ]
    coordinates {
(0, 409)
(1, 480)
(2, 532)
(3, 729)
(4, 1062)
};
 %
 \node [draw=gpudclr,fill=gpuclr,rounded corners=0.2ex, minimum width=23.7ex, xshift=\gpuxshift ex, yshift=\gpuyshift ex, scale=\gpuscale]{\textcolor{gputxtclr}{\textbf{Jetson-NX}$\mathbf{~@~1242}$$\mathbf{\times}$$\mathbf{375}$}};
\end{axis}
}
}};
\node (ne) [draw=none, xshift =\xshifte ex, yshift= \yshifte ex]{
\scalebox{\pscal}
{\tikz{
\pgfplotsset{width=\pltw ex, height=\plth ex}
\begin{axis}[
   axis background style={fill=axisbgclr},
    title={},
    xlabel={Cell-Size},
    ylabel={$\mu$S},
    xmin=-0.2, xmax=4.2,
    ymin=0, ymax=2000,
    xtick={0, 1, 2, 3, 4},
    ytick={200, 650, 1100, 1550, 2000},
    xticklabels={$23$$\times$$23$, $17$$\times$$17$, $13$$\times$$13$, $7$$\times$$7$, $3$$\times$$3$},
    yticklabels={$200$, $650$, $1100$, $1550$, $2000$},
     axis line style={axisclr},
    legend image post style={scale =\legendimscale},
    legend style={at={(\legendxshifte ex,\legendyshifte ex)},anchor=north, legend columns = 1, draw = {dlegendclr}, fill={legendclr}, nodes={scale=\legendscale}},
    ymajorgrids=true, 
    xmajorgrids=true,
    grid style={dashed, gridclr},
    major tick length=1ex,
    y label style={at={(\ylabelxshifte ex, \ylabelyshifte ex)}, scale=\labelscale},
    x label style={at={(\xlabelxshifte ex, \xlabelyshifte ex)}, scale=\labelscale},
    xticklabel style={rotate=\xlabelrotate, scale= \ticklabelscale},
    yticklabel style={xshift=0.0ex, scale= \ticklabelscale},
]
\addplot[
    fill=none,draw = clr1,
    very thick,
    line width= \linew ex,
    mark = square*,
    mark size = \mrksize ex,
    ]
    coordinates {
(0, 295)
(1, 369)
(2, 628)
(3, 700)
(4, 1835)
    };
\addplot[
    fill=none,draw = clr2,
    very thick,
    line width= \linew ex,
    mark = square*,
    mark size = \mrksize ex,
    ]
    coordinates {
(0, 292)
(1, 331)
(2, 369)
(3, 548)
(4, 796)
};
 %
 \node [draw=gpudclr,fill=gpuclr,rounded corners=0.2ex, minimum width=23.7ex, xshift=\gpuxshift ex, yshift=\gpuyshift ex, scale=\gpuscale]{\textcolor{gputxtclr}{\textbf{Jetson-NX}$\mathbf{~@~640}$$\mathbf{\times}$$\mathbf{480}$}};
\end{axis}
}
}};
\node (nf) [draw=none, xshift =\xshiftf ex, yshift= \yshiftf ex]{
\scalebox{\pscal}
{\tikz{
\pgfplotsset{width=\pltw ex, height=\plth ex}
\begin{axis}[
   axis background style={fill=axisbgclr},
    title={},
    xlabel={Cell-Size},
    ylabel={$\mu$S},
    xmin=-0.2, xmax=4.2,
    ymin=60, ymax=900,
    xtick={0, 1, 2, 3, 4},
    ytick={60, 240, 420, 660, 900},
   xticklabels={$23$$\times$$23$, $17$$\times$$17$, $13$$\times$$13$, $7$$\times$$7$, $3$$\times$$3$},
    yticklabels={$60$, $240$, $420$, $660$, $900$},
     axis line style={axisclr},
    legend image post style={scale =\legendimscale},
    legend style={at={(\legendxshiftf ex,\legendyshiftf ex)},anchor=north, legend columns = 1, draw = {dlegendclr}, fill={legendclr}, nodes={scale=\legendscale}},
    ymajorgrids=true, 
    xmajorgrids=true,
    grid style={dashed, gridclr},
    major tick length=1ex,
    y label style={at={(\ylabelxshiftf ex, \ylabelyshiftf ex)}, scale=\labelscale},
    x label style={at={(\xlabelxshiftf ex, \xlabelyshiftf ex)}, scale=\labelscale},
    xticklabel style={xshift=0ex, rotate=\xlabelrotate, scale= \ticklabelscale},
    yticklabel style={xshift=0.0ex, scale= \ticklabelscale},
]
\addplot[
    fill=none,draw = clr1,
    very thick,
    line width= \linew ex,
    mark = square*,
    mark size = \mrksize ex,
    ]
    coordinates {
(0, 149)
(1, 189)
(2, 235)
(3, 337)
(4, 834)
    };
\addplot[
    fill=none,draw = clr2,
    very thick,
    line width= \linew ex,
    mark = square*,
    mark size = \mrksize ex,
    ]
    coordinates {
(0, 89)
(1, 101)
(2, 109)
(3, 153)
(4, 214)
};
 %
 \node [draw=gpudclr,fill=gpuclr,rounded corners=0.2ex, minimum width=23.7ex, xshift=\gpuxshift ex, yshift=\gpuyshift ex, scale=\gpuscale]{\textcolor{gputxtclr}{\textbf{Jetson-NX}$\mathbf{~@~320}$$\mathbf{\times}$$\mathbf{240}$}};
\end{axis}
}
}};
%
%
%
%
%
%
%
\FPeval{\xshifta}{0-0.5}
\FPeval{\xshiftb}{17.0}
\FPeval{\xshiftc}{35.2}
\FPeval{\xshiftd}{0}
\FPeval{\xshifte}{18.5}
\FPeval{\xshiftf}{37}

\FPeval{\yshifta}{0-23}
\FPeval{\yshiftb}{0-23}
\FPeval{\yshiftc}{0-23}
\FPeval{\yshiftd}{0-10.0}
\FPeval{\yshifte}{0-10.0}
\FPeval{\yshiftf}{0-10.0}

\FPeval{\pltw}{37.5}
\FPeval{\plth}{22.5}

\FPeval{\bscale}{0.5}

\FPeval{\ticklabelscale}{0.9}
\FPeval{\labelscale}{1.0}
\FPeval{\xlabelrotate}{0}


\FPeval{\gpuscale}{0.8}
\FPeval{\gpuxshift}{15.4}
\FPeval{\gpuyshift}{11.2}
\FPeval{\resaxshift}{6.0}
\FPeval{\resbxshift}{22.4}
\FPeval{\rescxshift}{38.6}
\FPeval{\resyshift}{13.8}

\FPeval{\barw}{12}
\FPeval{\barsep}{0.8}
\FPeval{\linew}{0.4}
\FPeval{\barcorner}{0.4}

\FPeval{\mrksize}{0.4}
\FPeval{\legendscale}{0.65}
\FPeval{\legendimscale}{0.9}

\FPeval{\legendxshifta}{38.0}
\FPeval{\legendxshiftb}{21.0}
\FPeval{\legendxshiftc}{19.0}
\FPeval{\legendxshiftd}{5.0}
\FPeval{\legendxshifte}{5.0}
\FPeval{\legendxshiftf}{5.0}

\FPeval{\legendyshifta}{9.0}
\FPeval{\legendyshiftb}{10.0}
\FPeval{\legendyshiftc}{10.0}
\FPeval{\legendyshiftd}{10.5}
\FPeval{\legendyshifte}{10.5}
\FPeval{\legendyshiftf}{10.5}

\FPeval{\ylabelxshifta}{3-0.20}
\FPeval{\ylabelxshiftb}{2-0.20}
\FPeval{\ylabelxshiftc}{2.5}
\FPeval{\ylabelxshiftd}{2-0.20}
\FPeval{\ylabelxshifte}{2-0.20}
\FPeval{\ylabelxshiftf}{3-0.20}

\FPeval{\ylabelyshifta}{8}
\FPeval{\ylabelyshiftb}{8}
\FPeval{\ylabelyshiftc}{6.8}
\FPeval{\ylabelyshiftd}{8}
\FPeval{\ylabelyshifte}{8}
\FPeval{\ylabelyshiftf}{8}

\FPeval{\xlabelxshifta}{14}
\FPeval{\xlabelxshiftb}{14}
\FPeval{\xlabelxshiftc}{14}
\FPeval{\xlabelxshiftd}{15}
\FPeval{\xlabelxshifte}{15}
\FPeval{\xlabelxshiftf}{15}

\FPeval{\xlabelyshifta}{1-1.0}
\FPeval{\xlabelyshiftb}{1-1.0}
\FPeval{\xlabelyshiftc}{1-1.0}
\FPeval{\xlabelyshiftd}{1-1.0}
\FPeval{\xlabelyshifte}{1-1.0}
\FPeval{\xlabelyshiftf}{1-1.0}

\colorlet{gridclr}{white!90!black}
\colorlet{dlegendclr}{white!80!black}
\colorlet{axisclr}{white!70!black}
\colorlet{axisbgclr}{white!100!black}

\colorlet{dlegendclr}{white!80!black}
\colorlet{legendclr}{white!100!black}

\colorlet{clr1}{white!0!magenta}
\colorlet{clr2}{black!50!green}
\colorlet{clr3}{white!0!cyan}

\colorlet{clr1fill}{black!30!clr1}
\colorlet{clr2fill}{black!30!clr2}
\colorlet{clr3fill}{black!30!clr3}

\colorlet{gpudclr}{white!80!black}
\colorlet{gpuclr}{white!90!black}

\colorlet{deltaclr}{white!0!black}
\colorlet{gputxtclr}{white!0!black}
\node (na) [draw=none, xshift =\xshifta ex, yshift= \yshifta ex]{
\scalebox{\bscale}
{\tikz{
\pgfplotsset{width=\pltw ex, height=\plth ex}
\begin{axis}[
   axis background style={fill=axisbgclr},
    title={},
    xlabel={Number of scales},
    ylabel={$\mu$S},
    xmin=-0.2, xmax=5.2,
    ymin=0, ymax=400,
    xtick={0, 1, 2, 3, 4, 5},
    ytick={0, 100, 200, 300, 400},
   xticklabels={s$2$, s$4$, s$8$, s$2$, s$4$, s$8$},
   yticklabels={$0$, $100$, $200$, $300$, $400$},
     axis line style={axisclr},
    legend image post style={scale =\legendimscale},
    legend style={draw=gpuclr, at={(\legendxshifta ex,\legendyshifta ex)},anchor=north, legend columns = 1, draw = {dlegendclr}, fill={legendclr}, nodes={scale=\legendscale}},
    ymajorgrids=true, 
    xmajorgrids=true,
    grid style={dashed, gridclr},
    major tick length=1ex,
    y label style={at={(\ylabelxshifta ex, \ylabelyshifta ex)}, scale=\labelscale},
    x label style={at={(\xlabelxshifta ex, \xlabelyshifta ex)}, scale=\labelscale},
    xticklabel style={rotate=\xlabelrotate, scale=\ticklabelscale},
    yticklabel style={xshift=0.0ex, scale=\ticklabelscale},
]
\addplot[
    fill=none,draw = clr1,
    very thick,
    line width= \linew ex,
    mark = square*,
    mark size = \mrksize ex,
    ]
    coordinates {
(0, 143)
(1, 180)
(2, 273)
    };
\addplot[
    fill=none,draw = clr2,
    very thick,
    line width= \linew ex,
    mark = square*,
    mark size = \mrksize ex,
    ]
    coordinates {
(0, 49)
(1, 63)
(2, 75)
};
 %

 %
\addplot[
    fill=none,draw = clr1,
    very thick,
    line width= \linew ex,
    mark = square*,
    mark size = \mrksize ex,
    ]
    coordinates {
(3, 95)
(4, 131)
(5, 207)
    };
\addplot[
    fill=none,draw = clr2,
    very thick,
    line width= \linew ex,
    mark = square*,
    mark size = \mrksize ex,
    ]
    coordinates {
(3, 40)
(4, 51)
(5, 60)
};
%
\end{axis}
\node [draw=gpudclr,fill=gpuclr,rounded corners=0.2ex, minimum width=11.7ex, xshift=\gpuxshift ex, yshift=\gpuyshift ex, scale=\gpuscale]{\textcolor{gputxtclr}{\textbf{RTX-}$\mathbf{2070}$}};
\node [draw=gpudclr,fill=gpuclr,rounded corners=0.2ex, minimum width=10.7ex, xshift=\resaxshift ex, yshift=\resyshift ex, scale=\gpuscale]{\textcolor{gputxtclr}{$\mathbf{1242 \times 375}$}};
\node [draw=gpudclr,fill=gpuclr,rounded corners=0.2ex, minimum width=10.7ex, xshift=\resbxshift ex, yshift=\resyshift ex, scale=\gpuscale]{\textcolor{gputxtclr}{$\mathbf{640 \times 480}$}};
}
}};
\node (nb) [draw=none, xshift =\xshiftb ex, yshift= \yshiftb ex]{
\scalebox{\bscale}
{\tikz{
\pgfplotsset{width=\pltw ex, height=\plth ex}
\begin{axis}[
   axis background style={fill=axisbgclr},
    title={},
    xlabel={Number of scales},
    ylabel={$\mu$S},
    xmin=-0.2, xmax=5.2,
    ymin=0, ymax=7000,
    xtick={0, 1, 2, 3, 4, 5},
    ytick={0, 1750, 3500, 5250 ,7000},
   xticklabels={s$2$, s$4$, s$8$, s$2$, s$4$, s$8$},
    yticklabels={$0$, $1750$, $3500$, $5200$, $7000$},
     axis line style={axisclr},
    legend image post style={scale =\legendimscale},
    legend style={at={(\legendxshiftb ex,\legendyshiftb ex)},anchor=north, legend columns = 1, draw = {dlegendclr}, fill={legendclr}, nodes={scale=\legendscale}},
    ymajorgrids=true, 
    xmajorgrids=true,
    grid style={dashed, gridclr},
    major tick length=1ex,
    y label style={at={(\ylabelxshiftb ex, \ylabelyshiftb ex)}, scale=\labelscale},
    x label style={at={(\xlabelxshiftb ex, \xlabelyshiftb ex)}, scale=\labelscale},
    xticklabel style={rotate=\xlabelrotate, scale=\ticklabelscale},
    yticklabel style={xshift=0.0ex, scale=\ticklabelscale},
]
\addplot[
    fill=none,draw = clr1,
    very thick,
    line width= \linew ex,
    mark = square*,
    mark size = \mrksize ex,
    ]
    coordinates {
(0, 2892)
(1, 3178)
(2, 4906)
    };
\addplot[
    fill=none,draw = clr2,
    very thick,
    line width= \linew ex,
    mark = square*,
    mark size = \mrksize ex,
    ]
    coordinates {
(0, 691)
(1, 999)
(2, 1317)
};
%
\addplot[
    fill=none,draw = clr1,
    very thick,
    line width= \linew ex,
    mark = square*,
    mark size = \mrksize ex,
    ]
    coordinates {
(3, 1689)
(4, 2419)
(5, 3042)
    };
\addplot[
    fill=none,draw = clr2,
    very thick,
    line width= \linew ex,
    mark = square*,
    mark size = \mrksize ex,
    ]
    coordinates {
(3, 546)
(4, 780)
(5, 1027)
};
 \legend{CPU, GPU}
 \end{axis}
\node [draw=gpudclr,fill=gpuclr,rounded corners=0.2ex, minimum width=11.7ex, xshift=\gpuxshift ex, yshift=\gpuyshift ex, scale=\gpuscale]{\textcolor{gputxtclr}{\textbf{Jetson-NX}}};
\node [draw=gpudclr,fill=gpuclr,rounded corners=0.2ex, minimum width=10.7ex, xshift=\resaxshift ex, yshift=\resyshift ex, scale=\gpuscale]{\textcolor{gputxtclr}{$\mathbf{1242 \times 375}$}};
\node [draw=gpudclr,fill=gpuclr,rounded corners=0.2ex, minimum width=10.7ex, xshift=\resbxshift ex, yshift=\resyshift ex, scale=\gpuscale]{\textcolor{gputxtclr}{$\mathbf{640 \times 480}$}};
}
}};
\node (nc) [draw=none, xshift =\xshiftc ex, yshift= \yshiftc ex]{
\scalebox{\bscale}
{\tikz{
\pgfplotsset{width=\pltw ex, height=\plth ex}
\begin{axis}[
   axis background style={fill=axisbgclr},
    title={},
    xlabel={Number of scales},
    ylabel={Num. Features},
    xmin=-0.2, xmax=5.2,
    ymin=0, ymax=1500,
    xtick={0, 1, 2, 3, 4, 5},
    ytick={0, 375, 750, 1125, 1500},
   xticklabels={s$2$, s$4$, s$8$, s$2$, s$4$, s$8$},
   yticklabels={$0$, $375$, $750$, $1125$, $1500$},
     axis line style={axisclr},
    legend image post style={scale =\legendimscale},
    legend style={at={(\legendxshiftc ex,\legendyshiftc ex)},anchor=north, legend columns = 1, draw = {dlegendclr}, fill={legendclr}, nodes={scale=\legendscale}},
    ymajorgrids=true, 
    xmajorgrids=true,
    grid style={dashed, gridclr},
    major tick length=1ex,
    y label style={at={(\ylabelxshiftc ex, \ylabelyshiftc ex)}, scale=\labelscale},
    x label style={at={(\xlabelxshiftc ex, \xlabelyshiftc ex)}, scale=\labelscale},
    xticklabel style={rotate=\xlabelrotate, scale= \ticklabelscale},
    yticklabel style={xshift=0.0ex, scale= \ticklabelscale},
]
\addplot[
    fill=none,draw = clr1,
    very thick,
    line width= \linew ex,
    mark = square*,
    mark size = \mrksize ex,
    ]
    coordinates {
(0, 287)
(1, 583)
(2, 1279)
    };
\addplot[
    fill=none,draw = clr2,
    very thick,
    line width= \linew ex,
    mark = square*,
    mark size = \mrksize ex,
    ]
    coordinates {
(0, 255)
(1, 434)
(2, 927)
};
 %
%
\addplot[
    fill=none,draw = clr1,
    very thick,
    line width= \linew ex,
    mark = square*,
    mark size = \mrksize ex,
    ]
    coordinates {
(3, 215)
(4, 436)
(5, 958)
    };
\addplot[
    fill=none,draw = clr2,
    very thick,
    line width= \linew ex,
    mark = square*,
    mark size = \mrksize ex,
    ]
    coordinates {
(3, 174)
(4, 313)
(5, 688)
};
  \legend{\texttt{PFA}~\xmark, \texttt{PFA}~\cmark}
\end{axis}
\node [draw=gpudclr,fill=gpuclr,rounded corners=0.2ex, minimum width=10.7ex, xshift=\resaxshift ex, yshift=\resyshift ex, scale=\gpuscale]{\textcolor{gputxtclr}{$\mathbf{1242 \times 375}$}};
\node [draw=gpudclr,fill=gpuclr,rounded corners=0.2ex, minimum width=10.7ex, xshift=\resbxshift ex, yshift=\resyshift ex, scale=\gpuscale]{\textcolor{gputxtclr}{$\mathbf{640 \times 480}$}};
}
}};
\end{tikzpicture}
\subfloat{\label{fig:pyca_fc_eval}}
\subfloat{\label{fig:pfa_cpu_gpu_rtx}}
\subfloat{\label{fig:pfa_cpu_gpu_jetson}}
\subfloat{\label{fig:pfa_num_features}}
\vspace{-0.5ex}
\caption{Evaluation of \texttt{PyCA}'s components. (a) Runtime of Feature Culling (\texttt{FC}) \textit{vs} \cite{fastergpu} for different cell-size, resolution and GPUs since smaller cells are a concern in multiscale SLAM (Sec.~\ref{sec:tewa}). Our \texttt{FC} handles them easily, as notable via its linear profile compared to exponentially growing \cite{fastergpu} at smaller cells. (b)-(c) \texttt{PFA}'s runtime on CPU and GPU and (d) \texttt{PFA}'s effect on features count.}
\vspace{-3.0ex}
\end{figure}

%% file: plots/fast_det_eval_combined.tex
\begin{figure}[!t]
\centering

\begin{tikzpicture}

\FPeval{\xshifta}{0-2}
\FPeval{\xshiftb}{24.5-2}

\FPeval{\yshifta}{0}
\FPeval{\yshiftb}{0}

\FPeval{\pltw}{38}
\FPeval{\plth}{23.5}

\FPeval{\bscale}{0.5}

\FPeval{\ticklabelscale}{0.8}
\FPeval{\labelscale}{1.0}
\FPeval{\xlabelrotate}{0}

\FPeval{\gpuscale}{0.8}
\FPeval{\gpuxshift}{13.8}
\FPeval{\gpuyshift}{12.5}

\FPeval{\barw}{7}
\FPeval{\barsep}{0.8}
\FPeval{\linew}{1.0}
\FPeval{\barcorner}{0.2}

\FPeval{\mrksize}{0.25}
\FPeval{\legendscale}{1.0}
\FPeval{\legendimscale}{1.0}

\FPeval{\legendxshifta}{6.0}
\FPeval{\legendxshiftb}{38.0}

\FPeval{\legendyshifta}{10.0}
\FPeval{\legendyshiftb}{10.0}

\FPeval{\ylabelxshifta}{3-0.20}
\FPeval{\ylabelxshiftb}{3-0.20}

\FPeval{\ylabelyshifta}{8}
\FPeval{\ylabelyshiftb}{8}

\FPeval{\xlabelxshifta}{15}
\FPeval{\xlabelxshiftb}{15}

\FPeval{\xlabelyshifta}{1-1.0}
\FPeval{\xlabelyshiftb}{1-1.0}

\colorlet{gridclr}{white!90!black}
\colorlet{dlegendclr}{white!80!black}
\colorlet{axisclr}{white!70!black}
\colorlet{axisbgclr}{white!100!black}

\colorlet{dlegendclr}{white!80!black}
\colorlet{legendclr}{white!100!black}

\colorlet{clr1}{white!0!magenta}
\colorlet{clr2}{black!50!green}
\colorlet{clr3}{white!0!cyan}

\colorlet{clr1fill}{black!30!clr1}
\colorlet{clr2fill}{black!30!clr2}
\colorlet{clr3fill}{black!30!clr3}

\colorlet{gpudclr}{white!80!black}
\colorlet{gpuclr}{white!90!black}

\colorlet{deltaclr}{white!0!black}
\colorlet{gputxtclr}{white!0!black}
\colorlet{bndryclr}{white!80!black}

\node (bn)[draw=bndryclr, rounded corners=0.2ex,minimum width=54ex, minimum height=48ex, xshift=13.0ex, yshift=-18.2ex]{};
\node (a)[draw=none, xshift=-12.8ex, yshift=-4.5ex,scale=0.7]{(a)};
\node (b)[draw=none, xshift=-12.8ex, yshift=-16.3ex,scale=0.7]{(b)};
\node (c)[draw=none, xshift=-12.8ex, yshift=-29.0ex,scale=0.7]{(c)};
\node (d)[draw=none, xshift=-12.8ex, yshift=-41.3ex,scale=0.7]{(d)};
\draw [bndryclr, dashed] ($(bn.west)-(0ex, -12.6ex)$) -- ($(bn.east)-(0, -12.6ex)$);
\draw [bndryclr, dashed] ($(bn.west)-(0ex, -0.8ex)$) -- ($(bn.east)-(0, -0.8ex)$);
\draw [bndryclr, dashed] ($(bn.west)-(0ex, 12.0ex)$) -- ($(bn.east)-(0, 12.0ex)$);
\node (na) [draw=none, xshift =\xshifta ex, yshift= \yshifta ex]{
\scalebox{\bscale}
{\tikz{
\pgfplotsset{width=\pltw ex, height=\plth ex}
\begin{axis}[
   axis background style={fill=axisbgclr},
    title={},
    xlabel={Resolution},
    ylabel={$\mu$S},
    xmin=-0.2, xmax=2.2,
    ymin=0, ymax=30,
    xtick={0, 1, 2},
    ytick={0, 8, 15, 22, 30},
   xticklabels={ $320$$\times$$240$, $640$$\times$$480$, $1242$$\times$$375$},
   yticklabels={ $0$, $8$, $15$, $22$, $30$},
     axis line style={axisclr},
    legend image post style={scale =\legendimscale},
    legend style={at={(\legendxshifta ex,\legendyshifta ex)},anchor=north, legend columns = 1, draw = {dlegendclr}, fill={legendclr}, nodes={scale=\legendscale}},
    ymajorgrids=true, 
    xmajorgrids=true,
    grid style={dashed, gridclr},
    major tick length=-1ex,
    ytick align=outside,
    y label style={at={(\ylabelxshifta ex, \ylabelyshifta ex)},scale=\labelscale},
    x label style={at={(\xlabelxshifta ex, \xlabelyshifta ex)},scale=\labelscale},
    xticklabel style={yshift=-1ex,rotate=\xlabelrotate,scale=\ticklabelscale},
    yticklabel style={xshift=-1.0ex,scale=\ticklabelscale},
    scaled y ticks=false,
    ybar=\barsep ex,
    bar width = \barw pt,
    legend cell align={left}
]
\addplot[
    fill=clr1fill,draw = clr1,
    very thick,
    line width= \linew pt,
    rounded corners=\barcorner ex
    ]
    coordinates {
(0, 6)
(1, 15)
(2, 20)
    };
\addplot[
    fill=clr2fill,draw = clr2,
    very thick,
    line width= \linew pt,
    rounded corners=\barcorner ex
    ]
    coordinates {
(0, 6)
(1, 15)
(2, 20)
};
 %
 \node [draw=gpudclr,fill=gpuclr,rounded corners=0.2ex, minimum width=23.7ex, xshift=\gpuxshift ex, yshift=\gpuyshift ex, scale=\gpuscale]{\textcolor{gputxtclr}{\textbf{RTX-}$\mathbf{2070~@}$\textbf{~Scales~}$\mathbf{=1}$}};
\end{axis}
}
}};
\node (nb) [draw=none, xshift =\xshiftb ex, yshift= \yshiftb ex]{
\scalebox{\bscale}
{\tikz{
\pgfplotsset{width=\pltw ex, height=\plth ex}
\begin{axis}[
   axis background style={fill=axisbgclr},
    title={},
    xlabel={Resolution},
    ylabel={$\mu$S},
    xmin=-0.2, xmax=2.2,
    ymin=0, ymax=1000,
    xtick={0, 1, 2, 3, 4},
    ytick={0, 250, 500, 750, 1000},
   xticklabels={$320$$\times$$240$, $640$$\times$$480$, $1242$$\times$$375$},
         yticklabels={$0$, $250$, $500$, $750$, $1000$},
     axis line style={axisclr},
    legend image post style={scale =\legendimscale},
    legend style={at={(\legendxshiftb ex,\legendyshiftb ex)},anchor=north, legend columns = 1, draw = {dlegendclr}, fill={legendclr}, nodes={scale=\legendscale}},
    ymajorgrids=true, 
    xmajorgrids=true,
    grid style={dashed, gridclr},
    major tick length=-1ex,
    ytick align=outside,
    y label style={at={(\ylabelxshiftb ex, \ylabelyshiftb ex)},scale=\labelscale},
    x label style={at={(\xlabelxshiftb ex, \xlabelyshiftb ex)},scale=\labelscale},
    xticklabel style={yshift=-1ex,rotate=\xlabelrotate,scale=\ticklabelscale},
    yticklabel style={xshift=-1.0ex,scale=\ticklabelscale},
    scaled y ticks=false,
    ybar=\barsep ex,
    bar width = \barw pt,
    legend cell align={left}
]
\addplot[
    fill=clr1fill,draw = clr1,
    very thick,
    line width= \linew pt,
    rounded corners=\barcorner ex
    ]
    coordinates {
(0, 217)
(1, 597)
(2, 728)
    };
\addplot[
    fill=clr2fill,draw = clr2,
    very thick,
    line width= \linew pt,
    rounded corners=\barcorner ex
    ]
    coordinates {
(0, 109)
(1, 288)
(2, 362)
};
 \legend{\cite{fastergpu}, Proposed}
 \node [draw=gpudclr,fill=gpuclr,rounded corners=0.2ex, minimum width=23.7ex, xshift=\gpuxshift ex, yshift=\gpuyshift ex, scale=\gpuscale]{\textcolor{gputxtclr}{\textbf{Jetson-NX}$\mathbf{~@}$\textbf{~Scales~}$\mathbf{=1}$}};
\end{axis}
}
}};
%
%
%
%
%
\FPeval{\xshifta}{0-3}
\FPeval{\xshiftb}{27.0-3}

\FPeval{\yshifta}{0-11.7}
\FPeval{\yshiftb}{0-11.7}

\FPeval{\pltw}{38}
\FPeval{\plth}{23.5}

\FPeval{\bscale}{0.5}

\FPeval{\ticklabelscale}{0.8}
\FPeval{\labelscale}{1.0}
\FPeval{\xlabelrotate}{0}

\FPeval{\gpuscale}{0.8}
\FPeval{\gpuxshift}{13.8}
\FPeval{\gpuyshift}{12.5}

\FPeval{\barw}{12}
\FPeval{\barsep}{0.8}
\FPeval{\linew}{0.4}
\FPeval{\barcorner}{0.4}

\FPeval{\mrksize}{0.25}
\FPeval{\legendscale}{1.0}
\FPeval{\legendimscale}{1.0}

\FPeval{\legendxshifta}{6.0}
\FPeval{\legendxshiftb}{45.0}

\FPeval{\legendyshifta}{10.0}
\FPeval{\legendyshiftb}{12.0}

\FPeval{\ylabelxshifta}{1-0.20}
\FPeval{\ylabelxshiftb}{1-0.20}

\FPeval{\ylabelyshifta}{8}
\FPeval{\ylabelyshiftb}{8}

\FPeval{\xlabelxshifta}{15}
\FPeval{\xlabelxshiftb}{15}

\FPeval{\xlabelyshifta}{1-1.0}
\FPeval{\xlabelyshiftb}{1-1.0}

\colorlet{gridclr}{white!90!black}
\colorlet{dlegendclr}{white!80!black}
\colorlet{axisclr}{white!70!black}
\colorlet{axisbgclr}{white!100!black}

\colorlet{dlegendclr}{white!80!black}
\colorlet{legendclr}{white!100!black}

\colorlet{clr1}{white!0!magenta}
\colorlet{clr2}{black!50!green}
\colorlet{clr3}{white!0!cyan}

\colorlet{clr1fill}{black!30!clr1}
\colorlet{clr2fill}{black!30!clr2}
\colorlet{clr3fill}{black!30!clr3}

\colorlet{gpudclr}{white!80!black}
\colorlet{gpuclr}{white!90!black}

\colorlet{deltaclr}{white!0!black}
\colorlet{gputxtclr}{white!0!black}
\node (na) [draw=none, xshift =\xshifta ex, yshift= \yshifta ex]{
\scalebox{\bscale}
{\tikz{
\pgfplotsset{width=\pltw ex, height=\plth ex}
\begin{axis}[
   axis background style={fill=axisbgclr},
    title={},
    xlabel={Resolution},
    ylabel={FPS (Hz)},
    xmin=-0.2, xmax=2.2,
    ymin=0, ymax=60000,
    xtick={0, 1, 2},
    ytick={0, 15000, 30000, 45000, 60000},
   xticklabels={$1242$$\times$$375$, $640$$\times$$480$, $320$$\times$$240$},
   yticklabels={$0$, $15000$, $30000$, $45000$, $60000$},
     axis line style={axisclr},
    legend image post style={scale =\legendimscale},
    legend style={at={(\legendxshifta ex,\legendyshifta ex)},anchor=north, legend columns = 1, draw = {dlegendclr}, fill={legendclr}, nodes={scale=\legendscale}},
    ymajorgrids=true, 
    xmajorgrids=true,
    grid style={dashed, gridclr},
    major tick length=1ex,
    y label style={at={(\ylabelxshifta ex, \ylabelyshifta ex)}, scale=\labelscale},
    x label style={at={(\xlabelxshifta ex, \xlabelyshifta ex)}, scale=\labelscale},
    xticklabel style={rotate=\xlabelrotate},
    yticklabel style={xshift=0.5ex},
    xticklabel style={rotate=\xlabelrotate, scale=\ticklabelscale},
    yticklabel style={xshift=-0.5ex, scale=\ticklabelscale},
    scaled y ticks=false,
    legend cell align={left}
]
\addplot[
    fill=none,draw = clr1,
    very thick,
    line width= \linew ex,
    mark = square*,
    mark size = \mrksize ex,
    ]
    coordinates {
(0, 3194)
 (1, 3378)
 (2, 7246)
};
\addplot[
    fill=none,draw = clr2,
    very thick,
    line width= \linew ex,
    mark = square*,
    mark size = \mrksize ex,
    ]
    coordinates {
 (0, 10204)
 (1, 11627)
 (2, 13513)
    };
\addplot[
    fill=none,draw = clr3,
    very thick,
    line width= \linew ex,
    mark = square*,
    mark size = \mrksize ex,
    ]
    coordinates {
(0, 23255)
 (1, 28571)
 (2, 50000)
};
 %
 %
 \node [draw=gpudclr,fill=gpuclr,rounded corners=0.2ex, minimum width=23.7ex, xshift=\gpuxshift ex, yshift=\gpuyshift ex, scale=\gpuscale]{\textcolor{gputxtclr}{\textbf{RTX-}$\mathbf{2070}$\textbf{,~Scales~}$\mathbf{=1}$}};
\end{axis}
}
}};
\node (nb) [draw=none, xshift =\xshiftb ex, yshift= \yshiftb ex]{
\scalebox{\bscale}
{\tikz{
\pgfplotsset{width=\pltw ex, height=\plth ex}
\begin{axis}[
   axis background style={fill=axisbgclr},
    title={},
    xlabel={Resolution},
    ylabel={},
    xmin=-0.2, xmax=2.2,
    ymin=0, ymax=4000,
    xtick={0, 1, 2, 3, 4},
    ytick={0, 1000, 2000, 3000 ,4000},
   xticklabels={$1242$$\times$$375$, $640$$\times$$480$, $320$$\times$$240$},
    yticklabels={$0$, $1000$, $2000$, $3000$, $4000$},
     axis line style={axisclr},
    legend image post style={scale =\legendimscale},
    legend style={at={(\legendxshiftb ex,\legendyshiftb ex)},anchor=north, legend columns = 1, draw = {dlegendclr}, fill={legendclr}, nodes={scale=\legendscale}},
    ymajorgrids=true, 
    xmajorgrids=true,
    grid style={dashed, gridclr},
    major tick length=1ex,
    y label style={at={(\ylabelxshiftb ex, \ylabelyshiftb ex)}, scale=\labelscale},
    x label style={at={(\xlabelxshiftb ex, \xlabelyshiftb ex)}, scale=\labelscale},
    xticklabel style={rotate=\xlabelrotate, scale=\ticklabelscale},
    yticklabel style={xshift=-0.5ex, scale=\ticklabelscale},
    scaled y ticks=false,
    legend cell align={left}
]
%
\addplot[
    fill=none,draw = clr1,
    very thick,
    line width= \linew ex,
    mark = square*,
    mark size = \mrksize ex,
    ]
    coordinates {
(0, 394)
 (1, 390)
 (2, 623)
};
\addplot[
    fill=none,draw = clr2,
    very thick,
    line width= \linew ex,
    mark = square*,
    mark size = \mrksize ex,
    ]
    coordinates {
 (0, 615)
 (1, 834)
 (2, 1136)
    };
\addplot[
    fill=none,draw = clr3,
    very thick,
    line width= \linew ex,
    mark = square*,
    mark size = \mrksize ex,
    ]
    coordinates {
(0, 1272)
 (1, 1626)
 (2, 2994)
};
 \legend{OpenCV-GPU \cite{opencv}, ArrayFire-GPU \cite{arrayfire}, Proposed-GPU}
 \node [draw=gpudclr,fill=gpuclr,rounded corners=0.2ex, minimum width=23.7ex, xshift=\gpuxshift ex, yshift=\gpuyshift ex, scale=\gpuscale]{\textcolor{gputxtclr}{\textbf{Jetson-NX}\textbf{,~Scales~}$\mathbf{=1}$}};
\end{axis}
}
}};
%
%
%
%
%
\FPeval{\xshifta}{0-0.2}
\FPeval{\xshiftb}{26.8-0.2}
\FPeval{\xshiftc}{37}
\FPeval{\xshiftd}{0}
\FPeval{\xshifte}{18.5}
\FPeval{\xshiftf}{37}

\FPeval{\yshifta}{0-24}
\FPeval{\yshiftb}{0-24}
\FPeval{\yshiftc}{0-24}
\FPeval{\yshiftd}{0-26.0}
\FPeval{\yshifte}{0-26.0}
\FPeval{\yshiftf}{0-26.0}

\FPeval{\pltw}{54}
\FPeval{\plth}{22.5}

\FPeval{\bscale}{0.5}

\FPeval{\ticklabelscale}{0.8}
\FPeval{\labelscale}{1.0}
\FPeval{\xlabelrotate}{0}


\FPeval{\gpuscale}{0.8}
\FPeval{\gpuxshift}{22.4}
\FPeval{\gpuyshift}{11.2}
\FPeval{\resaxshift}{6.0}
\FPeval{\resbxshift}{22.4}
\FPeval{\rescxshift}{38.6}
\FPeval{\resyshift}{13.8}

\FPeval{\barw}{12}
\FPeval{\barsep}{0.8}
\FPeval{\linew}{0.4}
\FPeval{\barcorner}{0.4}

\FPeval{\mrksize}{0.4}
\FPeval{\legendscale}{0.65}
\FPeval{\legendimscale}{1.0}

\FPeval{\legendxshifta}{38.0}
\FPeval{\legendxshiftb}{6.0}
\FPeval{\legendxshiftc}{5.0}
\FPeval{\legendxshiftd}{5.0}
\FPeval{\legendxshifte}{5.0}
\FPeval{\legendxshiftf}{5.0}

\FPeval{\legendyshifta}{9.0}
\FPeval{\legendyshiftb}{11.5}
\FPeval{\legendyshiftc}{10.5}
\FPeval{\legendyshiftd}{10.5}
\FPeval{\legendyshifte}{10.5}
\FPeval{\legendyshiftf}{10.5}

\FPeval{\ylabelxshifta}{1-0.20}
\FPeval{\ylabelxshiftb}{2-0.20}
\FPeval{\ylabelxshiftc}{3-0.20}
\FPeval{\ylabelxshiftd}{2-0.20}
\FPeval{\ylabelxshifte}{2-0.20}
\FPeval{\ylabelxshiftf}{3-0.20}

\FPeval{\ylabelyshifta}{8}
\FPeval{\ylabelyshiftb}{8}
\FPeval{\ylabelyshiftc}{8}
\FPeval{\ylabelyshiftd}{8}
\FPeval{\ylabelyshifte}{8}
\FPeval{\ylabelyshiftf}{8}

\FPeval{\xlabelxshifta}{22}
\FPeval{\xlabelxshiftb}{22}
\FPeval{\xlabelxshiftc}{15}
\FPeval{\xlabelxshiftd}{15}
\FPeval{\xlabelxshifte}{15}
\FPeval{\xlabelxshiftf}{15}

\FPeval{\xlabelyshifta}{1-1.0}
\FPeval{\xlabelyshiftb}{1-1.0}
\FPeval{\xlabelyshiftc}{1-1.0}
\FPeval{\xlabelyshiftd}{1-1.0}
\FPeval{\xlabelyshifte}{1-1.0}
\FPeval{\xlabelyshiftf}{1-1.0}

\colorlet{gridclr}{white!90!black}
\colorlet{dlegendclr}{white!80!black}
\colorlet{axisclr}{white!70!black}
\colorlet{axisbgclr}{white!100!black}

\colorlet{dlegendclr}{white!80!black}
\colorlet{legendclr}{white!100!black}

\colorlet{clr1}{white!0!magenta}
\colorlet{clr2}{black!50!green}
\colorlet{clr3}{white!0!cyan}

\colorlet{clr1fill}{black!30!clr1}
\colorlet{clr2fill}{black!30!clr2}
\colorlet{clr3fill}{black!30!clr3}

\colorlet{gpudclr}{white!80!black}
\colorlet{gpuclr}{white!90!black}

\colorlet{deltaclr}{white!0!black}
\colorlet{gputxtclr}{white!0!black}
\node (na) [draw=none, xshift =\xshifta ex, yshift= \yshifta ex]{
\scalebox{\bscale}
{\tikz{
\pgfplotsset{width=\pltw ex, height=\plth ex}
\begin{axis}[
   axis background style={fill=axisbgclr},
    title={},
    xlabel={Number of scales},
    ylabel={FPS (Hz)},
    xmin=-0.2, xmax=8.2,
    ymin=0, ymax=48000,
    xtick={0, 1, 2, 3, 4, 5, 6, 7, 8},
    xticklabels={s$1$, s$2$, s$4$, s$1$, s$2$, s$4$, s$1$, s$2$, s$4$},
    ytick={0, 12000, 24000, 36000, 48000},
    yticklabels={$0$, $12000$, $24000$, $36000$, $48000$},
     axis line style={axisclr},
    legend image post style={scale =\legendimscale},
    legend style={at={(\legendxshifta ex,\legendyshifta ex)},anchor=north, legend columns = 3, draw = {dlegendclr}, fill={legendclr}, nodes={scale=\legendscale}},
    ymajorgrids=true, 
    xmajorgrids=true,
    grid style={dashed, gridclr},
    major tick length=1ex,
    y label style={at={(\ylabelxshifta ex, \ylabelyshifta ex)}, scale=\labelscale},
    x label style={at={(\xlabelxshifta ex, \xlabelyshifta ex)}, scale=\labelscale},
    xticklabel style={rotate=\xlabelrotate, scale=\ticklabelscale},
    yticklabel style={xshift=-0.5ex, scale=\ticklabelscale},
    scaled y ticks=false,
    legend cell align={left}
]
\addplot[
    fill=none,draw = clr1,
    very thick,
    line width= \linew ex,
    mark = square*,
    mark size = \mrksize ex,
    ]
    coordinates {
(0, 19607)
(1, 17241)
(2, 13157)
};
\addplot[
    fill=none,draw = clr2,
    very thick,
    line width= \linew ex,
    mark = square*,
    mark size = \mrksize ex,
    ]
    coordinates {
(0, 21739)
(1, 20000)
(2, 18181)
    };
\addplot[
    fill=none,draw = clr3,
    very thick,
    line width= \linew ex,
    mark = square*,
    mark size = \mrksize ex,
    ]
    coordinates {
(1, 14705)
(2, 14285)
};
\addplot[
    fill=none,draw = clr1,
    very thick,
    line width= \linew ex,
    mark = square*,
    mark size = \mrksize ex,
    ]
    coordinates {
(3, 21739)
(4, 20408)
(5, 15151)
};
%
%
\addplot[
    fill=none,draw = clr2,
    very thick,
    line width= \linew ex,
    mark = square*,
    mark size = \mrksize ex,
    ]
    coordinates {
(3, 25641)
(4, 19230)
(5, 22222)
};
\addplot[
    fill=none,draw = clr3,
    very thick,
    line width= \linew ex,
    mark = square*,
    mark size = \mrksize ex,
    ]
    coordinates {
(4, 13157)
(5, 11764)
};
\addplot[
    fill=none,draw = clr1,
    very thick,
    line width= \linew ex,
    mark = square*,
    mark size = \mrksize ex,
    ]
    coordinates {
(6, 29411)
(7, 23809)
(8, 14492)
};
%
%
\addplot[
    fill=none,draw = clr2,
    very thick,
    line width= \linew ex,
    mark = square*,
    mark size = \mrksize ex,
    ]
    coordinates {
(6, 35714)
(7, 38461)
(8, 35714)
};
%
\addplot[
    fill=none,draw = clr3,
    very thick,
    line width= \linew ex,
    mark = square*,
    mark size = \mrksize ex,
    ]
    coordinates {
(7, 19607)
(8, 20833)
};
%
%
\end{axis}
\node [draw=gpudclr,fill=gpuclr,rounded corners=0.2ex, minimum width=11.7ex, xshift=\gpuxshift ex, yshift=\gpuyshift ex, scale=\gpuscale]{\textcolor{gputxtclr}{\textbf{RTX-}$\mathbf{2070}$}};
\node [draw=gpudclr,fill=gpuclr,rounded corners=0.2ex, minimum width=10.7ex, xshift=\resaxshift ex, yshift=\resyshift ex, scale=\gpuscale]{\textcolor{gputxtclr}{$\mathbf{1242 \times 375}$}};
\node [draw=gpudclr,fill=gpuclr,rounded corners=0.2ex, minimum width=10.7ex, xshift=\resbxshift ex, yshift=\resyshift ex, scale=\gpuscale]{\textcolor{gputxtclr}{$\mathbf{640 \times 480}$}};
\node [draw=gpudclr,fill=gpuclr,rounded corners=0.2ex, minimum width=10.7ex, xshift=\rescxshift ex, yshift=\resyshift ex, scale=\gpuscale]{\textcolor{gputxtclr}{$\mathbf{320 \times 240}$}};
}
}};
\node (nb) [draw=none, xshift =\xshiftb ex, yshift= \yshiftb ex]{
\scalebox{\bscale}
{\tikz{
\pgfplotsset{width=\pltw ex, height=\plth ex}
\begin{axis}[
   axis background style={fill=axisbgclr},
    title={},
    xlabel={Number of scales},
    ylabel={FPS (Hz)},
    xmin=-0.2, xmax=8.2,
    ymin=0, ymax=4000,
    xtick={0, 1, 2, 3, 4, 5, 6, 7, 8},
    xticklabels={s$1$, s$2$, s$4$, s$1$, s$2$, s$4$, s$1$, s$2$, s$4$},
    ytick={0, 1000, 2000, 3000, 4000},
    yticklabels={$0$, $1000$, $2000$, $3000$, $4000$},
     axis line style={axisclr},
    legend image post style={scale =\legendimscale},
    legend style={at={(\legendxshiftb ex,\legendyshiftb ex)},anchor=north, legend columns = 1, draw = {dlegendclr}, fill={legendclr}, nodes={scale=\legendscale}},
    ymajorgrids=true, 
    xmajorgrids=true,
    grid style={dashed, gridclr},
    major tick length=1ex,
    y label style={at={(\ylabelxshiftb ex, \ylabelyshiftb ex)}, scale=\labelscale},
    x label style={at={(\xlabelxshiftb ex, \xlabelyshiftb ex)}, scale=\labelscale},
    xticklabel style={rotate=\xlabelrotate, scale=\ticklabelscale},
    yticklabel style={xshift=-0.5ex, scale=\ticklabelscale},
    scaled y ticks=false,
]
\addplot[
    fill=none,draw = clr1,
    very thick,
    line width= \linew ex,
    mark = square*,
    mark size = \mrksize ex,
    ]
    coordinates {
(0, 1187)
(1, 888)
(2, 733)
};
%
%
\addplot[
    fill=none,draw = clr2,
    very thick,
    line width= \linew ex,
    mark = square*,
    mark size = \mrksize ex,
    ]
    coordinates {
(0, 1322)
(1, 1072)
(2, 903)
};
%
\addplot[
    fill=none,draw = clr3,
    very thick,
    line width= \linew ex,
    mark = square*,
    mark size = \mrksize ex,
    ]
    coordinates {
(1, 758)
(2, 676)
};
%
\addplot[
    fill=none,draw = clr1,
    very thick,
    line width= \linew ex,
    mark = square*,
    mark size = \mrksize ex,
    ]
    coordinates {
(3, 1492)
(4, 1042)
(5, 931)
};
%
%
\addplot[
    fill=none,draw = clr2,
    very thick,
    line width= \linew ex,
    mark = square*,
    mark size = \mrksize ex,
    ]
    coordinates {
(3, 1709)
(4, 1424)
(5, 1298)
};
%
\addplot[
    fill=none,draw = clr3,
    very thick,
    line width= \linew ex,
    mark = square*,
    mark size = \mrksize ex,
    ]
    coordinates {
(4, 1013)
(5, 874)
};
%
\addplot[
    fill=none,draw = clr1,
    very thick,
    line width= \linew ex,
    mark = square*,
    mark size = \mrksize ex,
    ]
    coordinates {
(6, 3215)
(7, 2610)
(8, 2564)
};
%
%
\addplot[
    fill=none,draw = clr2,
    very thick,
    line width= \linew ex,
    mark = square*,
    mark size = \mrksize ex,
    ]
    coordinates {
(6, 3401)
(7, 3134)
(8, 2915)
};
%
\addplot[
    fill=none,draw = clr3,
    very thick,
    line width= \linew ex,
    mark = square*,
    mark size = \mrksize ex,
    ]
    coordinates {
(7, 1984)
(8, 1798)
};
%
\legend{\cite{fastergpu}, \texttt{FC}, \texttt{PyCA}}
\end{axis}
\node [draw=gpudclr,fill=gpuclr,rounded corners=0.2ex, minimum width=11.7ex, xshift=\gpuxshift ex, yshift=\gpuyshift ex, scale=\gpuscale]{\textcolor{gputxtclr}{\textbf{Jetson-NX}}};
\node [draw=gpudclr,fill=gpuclr,rounded corners=0.2ex, minimum width=10.7ex, xshift=\resaxshift ex, yshift=\resyshift ex, scale=\gpuscale]{\textcolor{gputxtclr}{$\mathbf{1242 \times 375}$}};
\node [draw=gpudclr,fill=gpuclr,rounded corners=0.2ex, minimum width=10.7ex, xshift=\resbxshift ex, yshift=\resyshift ex, scale=\gpuscale]{\textcolor{gputxtclr}{$\mathbf{640 \times 480}$}};
\node [draw=gpudclr,fill=gpuclr,rounded corners=0.2ex, minimum width=10.7ex, xshift=\rescxshift ex, yshift=\resyshift ex, scale=\gpuscale]{\textcolor{gputxtclr}{$\mathbf{320 \times 240}$}};
}
}};
%
%
%
%
%
%
%
\FPeval{\xshifta}{0-0.2}
\FPeval{\xshiftb}{26.8-0.2}
\FPeval{\xshiftc}{37}
\FPeval{\xshiftd}{0}
\FPeval{\xshifte}{18.5}
\FPeval{\xshiftf}{37}

\FPeval{\yshifta}{0-36.5}
\FPeval{\yshiftb}{0-36.5}
\FPeval{\yshiftc}{0-36.5}
\FPeval{\yshiftd}{0-50.0}
\FPeval{\yshifte}{0-50.0}
\FPeval{\yshiftf}{0-50.0}

\FPeval{\pltw}{54}
\FPeval{\plth}{22.5}

\FPeval{\bscale}{0.5}

\FPeval{\ticklabelscale}{0.8}
\FPeval{\labelscale}{1.0}
\FPeval{\xlabelrotate}{0}

\FPeval{\gpuscale}{0.8}
\FPeval{\gpuxshift}{22.4}
\FPeval{\gpuyshift}{13.8}
\FPeval{\resaxshift}{6.0}
\FPeval{\resbxshift}{22.4}
\FPeval{\rescxshift}{38.6}
\FPeval{\resyshift}{11.2}

\FPeval{\barw}{6.5}
\FPeval{\barsep}{0.0-1.4}
\FPeval{\linew}{0.2}
\FPeval{\barcorner}{0.25}

\FPeval{\mrksize}{0.4}
\FPeval{\legendscale}{0.6}
\FPeval{\legendimscale}{1.0}

\FPeval{\legendxshifta}{8.0}
\FPeval{\legendxshiftb}{8.0}
\FPeval{\legendxshiftc}{23.0}
\FPeval{\legendxshiftd}{23.0}
\FPeval{\legendxshifte}{23.0}
\FPeval{\legendxshiftf}{23.0}

\FPeval{\legendyshifta}{9.0}
\FPeval{\legendyshiftb}{9.0}
\FPeval{\legendyshiftc}{10.5}
\FPeval{\legendyshiftd}{10.5}
\FPeval{\legendyshifte}{10.5}
\FPeval{\legendyshiftf}{10.5}

\FPeval{\ylabelxshifta}{2-0.20}
\FPeval{\ylabelxshiftb}{2-0.20}
\FPeval{\ylabelxshiftc}{3-0.20}
\FPeval{\ylabelxshiftd}{2-0.20}
\FPeval{\ylabelxshifte}{2-0.20}
\FPeval{\ylabelxshiftf}{3-0.20}

\FPeval{\ylabelyshifta}{7}
\FPeval{\ylabelyshiftb}{7}
\FPeval{\ylabelyshiftc}{8}
\FPeval{\ylabelyshiftd}{8}
\FPeval{\ylabelyshifte}{8}
\FPeval{\ylabelyshiftf}{8}

\FPeval{\xlabelxshifta}{22}
\FPeval{\xlabelxshiftb}{22}
\FPeval{\xlabelxshiftc}{15}
\FPeval{\xlabelxshiftd}{15}
\FPeval{\xlabelxshifte}{15}
\FPeval{\xlabelxshiftf}{15}

\FPeval{\xlabelyshifta}{1-1.0}
\FPeval{\xlabelyshiftb}{1-1.0}
\FPeval{\xlabelyshiftc}{1-1.0}
\FPeval{\xlabelyshiftd}{1-1.0}
\FPeval{\xlabelyshifte}{1-1.0}
\FPeval{\xlabelyshiftf}{1-1.0}

\colorlet{gridclr}{white!90!black}
\colorlet{dlegendclr}{white!80!black}
\colorlet{axisclr}{white!70!black}
\colorlet{axisbgclr}{white!100!black}

\colorlet{dlegendclr}{white!80!black}
\colorlet{legendclr}{white!100!black}

\colorlet{clr1}{white!0!magenta}
\colorlet{clr2}{black!50!green}
\colorlet{clr3}{white!0!cyan}

\colorlet{clr1fill}{black!30!clr1}
\colorlet{clr2fill}{black!30!clr2}
\colorlet{clr3fill}{black!30!clr3}

\colorlet{gpudclr}{white!80!black}
\colorlet{gpuclr}{white!90!black}

\colorlet{deltaclr}{white!0!black}
\colorlet{gputxtclr}{white!0!black}
\node (na) [draw=none, xshift =\xshifta ex, yshift= \yshifta ex]{
\scalebox{\bscale}
{\tikz{
\pgfplotsset{width=\pltw ex, height=\plth ex}
\begin{axis}[
   axis background style={fill=axisbgclr},
    title={},
    xlabel={Number of scales},
    ylabel={FPS (Hz)},
    xmin=-0.75, xmax=11.5,
    ymin=0, ymax=20000,
    xtick={0, 0.75, 1.5, 2.25, 4.5, 5.25, 6., 6.75,  8.5, 9.25, 10, 10.75},
    xticklabels={s$1$, s$2$, s$4$, s$8$, s$1$, s$2$, s$4$, s$8$, s$1$, s$2$, s$4$, s$8$},
    ytick={0, 5000, 10000, 15000, 20000},
    yticklabels={$0$, $5000$, $10000$, $15000$, $20000$},
     axis line style={axisclr},
    legend image post style={scale =\legendimscale},
    legend style={draw=gpuclr, at={(\legendxshifta ex,\legendyshifta ex)},anchor=north, legend columns = 2, draw = {dlegendclr}, fill={legendclr}, nodes={scale=\legendscale}},
    ymajorgrids=true, 
    xmajorgrids=true,
    grid style={dashed, gridclr},
    major tick length=1ex,
    y label style={at={(\ylabelxshifta ex, \ylabelyshifta ex)}, scale=\labelscale},
    x label style={at={(\xlabelxshifta ex, \xlabelyshifta ex)}, scale=\labelscale},
    xticklabel style={yshift=0ex,rotate=\xlabelrotate,scale=\ticklabelscale},
    yticklabel style={xshift=0ex,scale=\ticklabelscale},
        scaled y ticks=false,
  ybar=\barsep ex,
   bar width = \barw pt,
 legend cell align={left}
]
\addplot[
    fill=clr1fill,draw = clr1,
    very thick,
    line width= \linew ex,
    rounded corners=\barcorner ex
    ]
    coordinates {
(0, 7518)
(0.75, 4975)
(1.5, 3906)
(2.25, 3278)
    };
\addplot[
    fill=clr2fill,draw = clr2,
    very thick,
    line width= \linew ex,
    rounded corners=\barcorner ex
    ]
    coordinates {
(0, 3030)
(0.75, 2252)
(1.5, 1766)
(2.25, 1101)
};
%
%
\addplot[
    fill=clr1fill,draw = clr1,
    very thick,
    line width= \linew ex,
    rounded corners=\barcorner ex
    ]
    coordinates {
(4.5, 9345)
(5.25, 5962)
(6.0, 4926)
(6.75, 4132)
    };
\addplot[
    fill=clr2fill,draw = clr2,
    very thick,
    line width= \linew ex,
    rounded corners=\barcorner ex
    ]
    coordinates {
(4.5, 3257)
(5.25, 2801)
(6.0, 2272)
(6.75, 1763)
};
 %
\addplot[
    fill=clr1fill,draw = clr1,
    very thick,
    line width= \linew ex,
    rounded corners=\barcorner ex
    ]
    coordinates {
(8.5, 17543)
(9.25, 10000)
(10, 9345)
(10.75, 8130)
    };
\addplot[
    fill=clr2fill,draw = clr2,
    very thick,
    line width= \linew ex,
     rounded corners=\barcorner ex
   ]
    coordinates {
(8.5, 6211)
(9.25, 4901)
(10, 4219)
(10.75, 3184)
};
  %
%
 \legend{F$_D$, F$_{DE}$}
 \node [draw=gpudclr,fill=gpuclr,rounded corners=0.2ex, minimum width=23.7ex, xshift=\gpuxshift ex, yshift=\gpuyshift ex, scale=\gpuscale]{\textcolor{gputxtclr}{\textbf{RTX-}$\mathbf{2070~@~1242}$$\mathbf{\times}$$\mathbf{375}$}};
\end{axis}
\node [draw=gpudclr,fill=gpuclr,rounded corners=0.2ex, minimum width=11.7ex, xshift=\gpuxshift ex, yshift=\gpuyshift ex, scale=\gpuscale]{\textcolor{gputxtclr}{\textbf{RTX-}$\mathbf{2070}$}};
\node [draw=gpudclr,fill=gpuclr,rounded corners=0.2ex, minimum width=10.7ex, xshift=\resaxshift ex, yshift=\resyshift ex, scale=\gpuscale]{\textcolor{gputxtclr}{$\mathbf{1242 \times 375}$}};
\node [draw=gpudclr,fill=gpuclr,rounded corners=0.2ex, minimum width=10.7ex, xshift=\resbxshift ex, yshift=\resyshift ex, scale=\gpuscale]{\textcolor{gputxtclr}{$\mathbf{640 \times 480}$}};
\node [draw=gpudclr,fill=gpuclr,rounded corners=0.2ex, minimum width=10.7ex, xshift=\rescxshift ex, yshift=\resyshift ex, scale=\gpuscale]{\textcolor{gputxtclr}{$\mathbf{320 \times 240}$}};
}
}};
\node (nb) [draw=none, xshift =\xshiftb ex, yshift= \yshiftb ex]{
\scalebox{\bscale}
{\tikz{
\pgfplotsset{width=\pltw ex, height=\plth ex}
\begin{axis}[
   axis background style={fill=axisbgclr},
    title={},
    xlabel={Number of scales},
    ylabel={FPS (Hz)},
    xmin=-0.75, xmax=11.5,
    ymin=0, ymax=2200,
    xtick={0, 0.75, 1.5, 2.25, 4.5, 5.25, 6., 6.75,  8.5, 9.25, 10, 10.75}, 
    xticklabels={s$1$, s$2$, s$4$, s$8$, s$1$, s$2$, s$4$, s$8$, s$1$, s$2$, s$4$, s$8$},
    ytick={0, 550, 1100, 1650, 2200},
    yticklabels={$0$, $550$, $1100$, $1650$, $2200$},
     axis line style={axisclr},
    legend image post style={scale =\legendimscale},
    legend style={at={(\legendxshiftb ex,\legendyshiftb ex)},anchor=north, legend columns = 2, draw = {dlegendclr}, fill={legendclr}, nodes={scale=\legendscale}},
    ymajorgrids=true, 
    xmajorgrids=true,
    grid style={dashed, gridclr},
    major tick length=1ex,
    y label style={at={(\ylabelxshiftb ex, \ylabelyshiftb ex)}, scale=\labelscale},
    x label style={at={(\xlabelxshiftb ex, \xlabelyshiftb ex)}, scale=\labelscale},
    xticklabel style={yshift=0ex,rotate=\xlabelrotate,scale=\ticklabelscale},
    yticklabel style={xshift=0ex,scale=\ticklabelscale},
        scaled y ticks=false,
   ybar=\barsep ex,
   bar width = \barw pt,
 legend cell align={left}
]
\addplot[
    fill=clr1fill,draw = clr1,
    very thick,
    line width= \linew ex,
    rounded corners=\barcorner ex
    ]
    coordinates {
(0, 826)
(0.75, 444)
(1.5, 311)
(2.25, 245)
    };
\addplot[
    fill=clr2fill,draw = clr2,
    very thick,
    line width= \linew ex,
    rounded corners=\barcorner ex
    ]
    coordinates {
(0, 243)
(0.75, 150)
(1.5, 104)
(2.25, 82)
};
%
%
\addplot[
    fill=clr1fill,draw = clr1,
    very thick,
    line width= \linew ex,
    rounded corners=\barcorner ex
    ]
    coordinates {
(4.5, 1079)
(5.25, 650)
(6, 398)
(6.75, 323)
    };
\addplot[
    fill=clr2fill,draw = clr2,
    very thick,
    line width= \linew ex,
    rounded corners=\barcorner ex
    ]
    coordinates {
(4.5, 324)
(5.25, 201)
(6, 143)
(6.75, 112)
};
%
%
\addplot[
    fill=clr1fill,draw = clr1,
    very thick,
    line width= \linew ex,
    rounded corners=\barcorner ex
    ]
    coordinates {
(8.5, 1893)
(9.25, 1108)
(10, 892)
(10.75, 729)
    };
\addplot[
    fill=clr2fill,draw = clr2,
    very thick,
    line width= \linew ex,
     rounded corners=\barcorner ex
   ]
    coordinates {
(8.5, 643)
(9.25, 451)
(10, 344)
(10.75, 274)
};
%
 \legend{F$_D$, F$_{DE}$}
\end{axis}
\node [draw=gpudclr,fill=gpuclr,rounded corners=0.2ex, minimum width=11.7ex, xshift=\gpuxshift ex, yshift=\gpuyshift ex, scale=\gpuscale]{\textcolor{gputxtclr}{\textbf{Jetson-NX}}};
\node [draw=gpudclr,fill=gpuclr,rounded corners=0.2ex, minimum width=10.7ex, xshift=\resaxshift ex, yshift=\resyshift ex, scale=\gpuscale]{\textcolor{gputxtclr}{$\mathbf{1242 \times 375}$}};
\node [draw=gpudclr,fill=gpuclr,rounded corners=0.2ex, minimum width=10.7ex, xshift=\resbxshift ex, yshift=\resyshift ex, scale=\gpuscale]{\textcolor{gputxtclr}{$\mathbf{640 \times 480}$}};
\node [draw=gpudclr,fill=gpuclr,rounded corners=0.2ex, minimum width=10.7ex, xshift=\rescxshift ex, yshift=\resyshift ex, scale=\gpuscale]{\textcolor{gputxtclr}{$\mathbf{320 \times 240}$}};
}
}};
\end{tikzpicture}
\subfloat{\label{fig:crf_eval}}
\subfloat{\label{fig:pyca_fastdet_ss_eval}}
\subfloat{\label{fig:pyca_fastdet_ms_eval}}
\subfloat{\label{fig:det_ext}}
\vspace{-0.5ex}
\caption{Evaluation of FAST Detection. (a) Performance of our GPU kernel for computing Corner Response Function (CRF) \textit{vs} \cite{fastergpu}. (b) The proposed FAST detection \textit{vs} open-source single-scale baselines \cite{opencv, arrayfire}. (c) multi-scale baseline \cite{fastergpu} comparable with \texttt{FC} only since \cite{fastergpu} does not perform feature aggregation. However, \texttt{PyCA} i.e. \texttt{FC}+\texttt{PFA} is shown for reference. (d) Throughput of our \texttt{PyCA} based frontend across resolutions, scales and GPUs. `F$_D$': Detection frame rate, `F$_{DE}$': Detection-Extraction frame rate.}
%
%
%
%
\vspace{-3.5ex}
\end{figure}
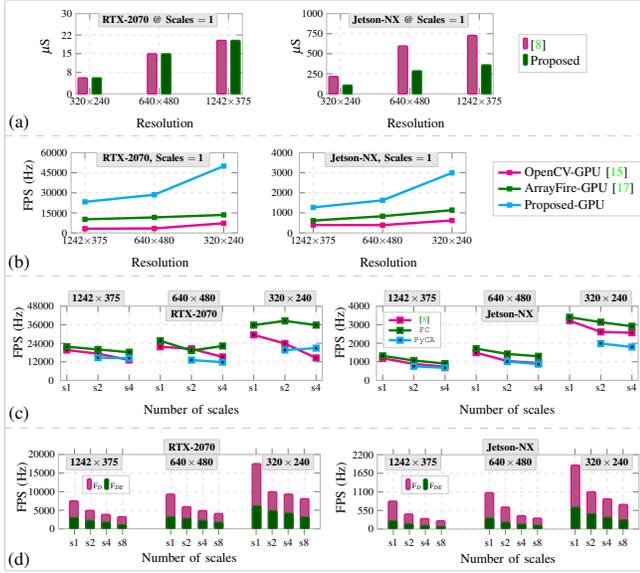

%% file: plots/slam_throughput.tex
\begin{figure}[!t]
\centering

\begin{tikzpicture}

\FPeval{\xshifta}{0-0}
\FPeval{\xshiftb}{27.0-0}

\FPeval{\yshifta}{0}
\FPeval{\yshiftb}{0}

\FPeval{\pltw}{53}
\FPeval{\plth}{23.5}

\FPeval{\bscale}{0.5}

\FPeval{\ticklabelscale}{0.8}
\FPeval{\labelscale}{1.0}
\FPeval{\xlabelrotate}{0}

\FPeval{\gpuscale}{0.8}
\FPeval{\gpuxshift}{22.0}
\FPeval{\gpuyshift}{12.1}

\FPeval{\barw}{7}
\FPeval{\barsep}{0.8}
\FPeval{\linew}{1.0}
\FPeval{\barcorner}{0.2}

\FPeval{\mrksize}{0.25}
\FPeval{\legendscale}{0.8}
\FPeval{\legendimscale}{0.8}

\FPeval{\legendxshifta}{11.0}
\FPeval{\legendxshiftb}{11.0}

\FPeval{\legendyshifta}{10.0}
\FPeval{\legendyshiftb}{10.0}

\FPeval{\ylabelxshifta}{1.5-0.20}
\FPeval{\ylabelxshiftb}{2.5-0.20}

\FPeval{\ylabelyshifta}{8}
\FPeval{\ylabelyshiftb}{8}

\FPeval{\xlabelxshifta}{20}
\FPeval{\xlabelxshiftb}{20}

\FPeval{\xlabelyshifta}{1-1.0}
\FPeval{\xlabelyshiftb}{1-1.0}

\colorlet{gridclr}{white!90!black}
\colorlet{dlegendclr}{white!80!black}
\colorlet{axisclr}{white!70!black}
\colorlet{axisbgclr}{white!100!black}

\colorlet{dlegendclr}{white!80!black}
\colorlet{legendclr}{white!100!black}

\colorlet{clr1}{white!0!magenta}
\colorlet{clr2}{black!50!green}
\colorlet{clr3}{white!0!cyan}

\colorlet{clr1fill}{black!30!clr1}
\colorlet{clr2fill}{black!30!clr2}
\colorlet{clr3fill}{black!30!clr3}

\colorlet{gpudclr}{white!80!black}
\colorlet{gpuclr}{white!90!black}

\colorlet{deltaclr}{white!0!black}
\colorlet{gputxtclr}{white!0!black}
\colorlet{bndryclr}{white!80!black}

\node (bn)[draw=bndryclr, rounded corners=0.2ex,minimum width=54ex, minimum height=23.0ex, xshift=13.0ex, yshift=-5.8ex]{};
\node (a)[draw=none, xshift=14.2ex, yshift=-4.9ex,scale=0.7]{(a)};
\node (b)[draw=none, xshift=-3.8ex, yshift=-16.3ex,scale=0.7]{(b)};
\node (c)[draw=none, xshift= 23.2ex, yshift=-16.3ex,scale=0.7]{(c)};
\draw [bndryclr, dashed] ($(bn.west)-(0ex, 0.0ex)$) -- ($(bn.east)-(0, 0.0ex)$);
\node (na) [draw=none, xshift =\xshifta ex, yshift= \yshifta ex]{
\scalebox{\bscale}
{\tikz{
\pgfplotsset{width=\pltw ex, height=\plth ex}
\begin{axis}[
   axis background style={fill=axisbgclr},
    title={},
    xlabel={SLAM Component},
    ylabel={FPS (Hz)},
    xmin=-0.2, xmax=3.2,
    ymin=0, ymax=4000,
    xtick={0, 1, 2, 3},
    ytick={0, 1000, 2000, 3000, 4000},
    xticklabels={T$_{DE}$, Stereo Match, Track, Back-end},
   yticklabels={$0$, $1000$, $2000$, $3000$, $4000$},
     axis line style={axisclr},
    legend image post style={scale =\legendimscale},
    legend style={at={(\legendxshifta ex,\legendyshifta ex)},anchor=north, legend columns = 1, draw = {dlegendclr}, fill={legendclr}, nodes={scale=\legendscale}},
    ymajorgrids=true, 
    xmajorgrids=true,
    grid style={dashed, gridclr},
    major tick length=1ex,
    y label style={at={(\ylabelxshifta ex, \ylabelyshifta ex)}, scale=\labelscale},
    x label style={at={(\xlabelxshifta ex, \xlabelyshifta ex)}, scale=\labelscale},
    xticklabel style={yshift=0ex,rotate=\xlabelrotate,scale=\ticklabelscale},
    yticklabel style={xshift=-0.5ex,scale=\ticklabelscale},
    ybar=\barsep ex,
    bar width = \barw pt,
	legend cell align={left}
]
\addplot[
    fill=clr1fill,draw = clr1,
    very thick,
    line width= \linew pt,
    rounded corners=\barcorner ex
    ]
    coordinates {
(0, 45)
(1, 76)
(2, 1000)
(3, 666)
    };
\addplot[
    fill=clr2fill,draw = clr2,
    very thick,
    line width= \linew pt,
    rounded corners=\barcorner ex
    ]
    coordinates {
(0, 833)
(1, 571)
(2, 2857)
(3, 1428)
};
 %
 \legend{ORB-SLAM$2$ \cite{orb2}, \OursAcronym{}+Full-BA}
 \node [draw=gpudclr,fill=gpuclr,rounded corners=0.2ex, minimum width=23.7ex, xshift=\gpuxshift ex, yshift=\gpuyshift ex, scale=\gpuscale]{\textcolor{gputxtclr}{\textbf{RTX-}$\mathbf{2070~@}$~$\mathbf{752}$$\mathbf{\times}$$\mathbf{480}$\textbf{,~Scales~}$\mathbf{=8}$}};
\end{axis}
}
}};
\node (nb) [draw=none, xshift =\xshiftb ex, yshift= \yshiftb ex]{
\scalebox{\bscale}
{\tikz{
\pgfplotsset{width=\pltw ex, height=\plth ex}
\begin{axis}[
   axis background style={fill=axisbgclr},
    title={},
    xlabel={SLAM Component},
    ylabel={FPS (Hz)},
    xmin=-0.2, xmax=3.2,
    ymin=0, ymax=600,
    xtick={0, 1, 2, 3, 4},
    ytick={0, 150, 300, 450, 600},
    xticklabels={T$_{DE}$, Stereo Match, Track, Back-end},
         yticklabels={$0$, $150$, $300$, $450$, $600$},
     axis line style={axisclr},
    legend image post style={scale =\legendimscale},
    legend style={at={(\legendxshiftb ex,\legendyshiftb ex)},anchor=north, legend columns = 1, draw = {dlegendclr}, fill={legendclr}, nodes={scale=\legendscale}},
    ymajorgrids=true, 
    xmajorgrids=true,
    grid style={dashed, gridclr},
    major tick length=1ex,
    y label style={at={(\ylabelxshiftb ex, \ylabelyshiftb ex)}, scale=\labelscale},
    x label style={at={(\xlabelxshiftb ex, \xlabelyshiftb ex)}, scale=\labelscale},
    xticklabel style={yshift=0ex,rotate=\xlabelrotate,scale=\ticklabelscale},
    yticklabel style={xshift=-0.5ex,scale=\ticklabelscale},
    ybar=\barsep ex,
    bar width = \barw pt,
	legend cell align={left}
]
\addplot[
    fill=clr1fill,draw = clr1,
    very thick,
    line width= \linew pt,
    rounded corners=\barcorner ex
    ]
    coordinates {
(0, 24)
(1, 50)
(2, 250)
(3, 142)
    };
\addplot[
    fill=clr2fill,draw = clr2,
    very thick,
    line width= \linew pt,
    rounded corners=\barcorner ex
    ]
    coordinates {
(0, 100)
(1, 142)
(2, 400)
(3, 200)
};
  %
 \legend{ORB-SLAM$2$ \cite{orb2}, \OursAcronym{}+Full-BA}
 \node [draw=gpudclr,fill=gpuclr,rounded corners=0.2ex, minimum width=23.7ex, xshift=\gpuxshift ex, yshift=\gpuyshift ex, scale=\gpuscale]{\textcolor{gputxtclr}{\textbf{Jetson-NX}$\mathbf{~@}$~$\mathbf{752}$$\mathbf{\times}$$\mathbf{480}$\textbf{,~Scales~}$\mathbf{=8}$}};
\end{axis}
}
}};
%
%
%
%
%
%
\FPeval{\xshifta}{0-4.8}
\FPeval{\xshiftb}{27.0-4.8}
\FPeval{\xshiftc}{36.0-4.8}

\FPeval{\yshifta}{0-11}
\FPeval{\yshiftb}{0-11.4}
\FPeval{\yshiftc}{0-11}

\FPeval{\pltwb}{70}
\FPeval{\pltw}{39}
\FPeval{\plth}{23.5}

\FPeval{\bscale}{0.5}

\FPeval{\ticklabelscale}{0.8}
\FPeval{\labelscale}{1.0}
\FPeval{\xlabelrotate}{0}

\FPeval{\gpuscale}{0.78}
\FPeval{\gpuxshifta}{14.8}
\FPeval{\gpuxshift}{28.8}
\FPeval{\gpuyshift}{12.3}

\FPeval{\barw}{7}
\FPeval{\barsep}{0.8}
\FPeval{\linew}{1.0}
\FPeval{\barcorner}{0.2}

\FPeval{\mrksize}{0.4}
\FPeval{\legendscale}{1.0}
\FPeval{\legendimscale}{0.6}

\FPeval{\legendxshifta}{5.8}
\FPeval{\legendxshiftb}{5.8}
\FPeval{\legendxshiftc}{7.8}

\FPeval{\legendyshifta}{10.5}
\FPeval{\legendyshiftb}{10.5}
\FPeval{\legendyshiftc}{11.0}

\FPeval{\ylabelxshifta}{3.5-0.20}
\FPeval{\ylabelxshiftb}{4.1-0.20}
\FPeval{\ylabelxshiftc}{4.1-0.20}

\FPeval{\ylabelyshifta}{7}
\FPeval{\ylabelyshiftb}{7}
\FPeval{\ylabelyshiftc}{7}

\FPeval{\xlabelxshifta}{15}
\FPeval{\xlabelxshiftb}{15}
\FPeval{\xlabelxshiftc}{15}

\FPeval{\xlabelyshifta}{1-1.0}
\FPeval{\xlabelyshiftb}{1-1.0}
\FPeval{\xlabelyshiftc}{1-1.0}

\colorlet{gridclr}{white!90!black}
\colorlet{dlegendclr}{white!80!black}
\colorlet{axisclr}{white!70!black}
\colorlet{axisbgclr}{white!100!black}

\colorlet{dlegendclr}{white!80!black}
\colorlet{legendclr}{white!100!black}

\colorlet{clr1}{white!0!magenta}
\colorlet{clr2}{black!50!green}
\colorlet{clr3}{white!0!cyan}

\colorlet{clr1fill}{black!30!clr1}
\colorlet{clr2fill}{black!30!clr2}
\colorlet{clr3fill}{black!30!clr3}

\colorlet{gpudclr}{white!80!black}
\colorlet{gpuclr}{white!90!black}

\colorlet{deltaclr}{white!0!black}
\colorlet{gputxtclr}{white!0!black}
\node (na) [draw=none, xshift =\xshifta ex, yshift= \yshifta ex]{
\scalebox{\bscale}
{\tikz{
\pgfplotsset{width=\pltw ex, height=\plth ex}
\begin{axis}[
   axis background style={fill=axisbgclr},
    title={},
    xlabel={},
    ylabel={FPS (Hz)},
    xmin=-1.2, xmax=3.2,
    ymin=0, ymax=200,
    xtick={-0.2, 2.5},
    xticklabels={ ORB-SLAM$2$\cite{orb2}, \OursAcronym},
   ytick={0, 50, 100, 150, 200},
   yticklabels={$0$, $50$, $100$, $150$, $200$},
     axis line style={axisclr},
    legend image post style={scale =\legendimscale},
    legend style={at={(\legendxshifta ex,\legendyshifta ex)},anchor=north, legend columns = 1, draw = {dlegendclr}, fill={legendclr}, nodes={scale=\legendscale}},
    ymajorgrids=true, 
    xmajorgrids=true,
    grid style={dashed, gridclr},
    major tick length=-1ex,
    ytick align=outside,
    y label style={at={(\ylabelxshifta ex, \ylabelyshifta ex)},scale=\labelscale},
    x label style={at={(\xlabelxshifta ex, \xlabelyshifta ex)},scale=\labelscale},
    xticklabel style={yshift=-1ex,rotate=\xlabelrotate,scale=\ticklabelscale},
    yticklabel style={xshift=-0.5ex,scale=\ticklabelscale},
    scaled y ticks=false,
    ybar=\barsep ex,
    bar width = \barw pt,
    legend cell align={left}
]
\addplot[
    fill=clr1fill,draw = clr1,
    very thick,
    line width= \linew pt,
    rounded corners=\barcorner ex
    ]
    coordinates {
(0, 19)
    };
\addplot[
    fill=clr2fill,draw = clr2,
    very thick,
    line width= \linew pt,
    rounded corners=\barcorner ex
    ]
    coordinates {
(0,8)
};

\addplot[
    fill=clr1fill,draw = clr1,
    very thick,
    line width= \linew pt,
    rounded corners=\barcorner ex
    ]
    coordinates {
(2, 100)
};
\addplot[
    fill=clr2fill,draw = clr2,
    very thick,
    line width= \linew pt,
    rounded corners=\barcorner ex
    ]
    coordinates {
(2, 22)
};
 %
%
%
  %
\legend{\footnotesize RTX-2070, \footnotesize Jetson-NX}
  \node [draw=gpudclr,fill=gpuclr,rounded corners=0.2ex, minimum width=23.7ex, xshift=\gpuxshifta ex, yshift=\gpuyshift ex, scale=\gpuscale]{\textcolor{gputxtclr}{$\mathbf{@1242 {\times} 375}$\textbf{, Scales}$\mathbf{=8}$}};
\end{axis}
}
}};
%
%
%
\node (nb) [draw=none, xshift =\xshiftb ex, yshift= \yshiftb ex]{
\scalebox{\bscale}
{\tikz{
\pgfplotsset{width=\pltwb ex, height=\plth ex}
\begin{axis}[
   axis background style={fill=axisbgclr},
    title={},
    xlabel={},
    ylabel={FPS (Hz)},
    xmin=-2.2, xmax=6.2,
    ymin=0, ymax=200,
     xtick={-1.2, 0.5,  2, 3.5, 5.2},
     xticklabels={ ORB-SLAM$2$\cite{orb2}, ICE-BA\cite{iceba}, \cite{fastergpu}+ICE-BA, \shortstack{\OursAcronym \\ +Full-BA}, \shortstack{\OursAcronym \\ + ICE-BA}},
      ytick={0, 50, 100, 150, 200},
   yticklabels={$0$, $50$, $100$, $150$, $200$},
     axis line style={axisclr},
    legend image post style={scale =\legendimscale},
    legend style={at={(\legendxshiftb ex,\legendyshiftb ex)},anchor=north, legend columns = 1, draw = {axisclr}, fill={none}, nodes={scale=\legendscale}},
    ymajorgrids=true, 
    xmajorgrids=true,
    grid style={dashed, gridclr},
    major tick length=-1ex,
    ytick align=outside,
    y label style={at={(\ylabelxshiftb ex, \ylabelyshiftb ex)},scale=\labelscale},
    x label style={at={(\xlabelxshiftb ex, \xlabelyshiftb ex)},scale=\labelscale},
    xticklabel style={yshift=-1ex,rotate=\xlabelrotate,scale=\ticklabelscale},
    yticklabel style={xshift=-0.5ex,scale=\ticklabelscale},
    scaled y ticks=false,
    ybar=\barsep ex,
    bar width = \barw pt,
    legend cell align={left}
]
\addplot[
    fill=clr1fill,draw = clr1,
    very thick,
    line width= \linew pt,
    rounded corners=\barcorner ex
    ]
    coordinates {
(0, 22)
    };
\addplot[
    fill=clr2fill,draw = clr2,
    very thick,
    line width= \linew pt,
    rounded corners=\barcorner ex
    ]
    coordinates {
(0, 11)
    };
\addplot[
    fill=clr1fill,draw = clr1,
    very thick,
    line width= \linew pt,
    rounded corners=\barcorner ex
    ]
    coordinates {
(1, 40)
    };
 \addplot[
    fill=clr2fill,draw = clr2,
    very thick,
    line width= \linew pt,
    rounded corners=\barcorner ex
    ]
    coordinates {
(1, 16)
    };
\addplot[
    fill=clr1fill,draw = clr1,
    very thick,
    line width= \linew pt,
    rounded corners=\barcorner ex
    ]
    coordinates {
(2, 55)
};
 %
\addplot[
    fill=clr2fill,draw = clr2,
    very thick,
    line width= \linew pt,
    rounded corners=\barcorner ex
    ]
    coordinates {
(2, 25)
};
 %
 %
\addplot[
    fill=clr1fill,draw = clr1,
    very thick,
    line width= \linew pt,
    rounded corners=\barcorner ex
    ]
    coordinates {
(3, 158)
};
 %
\addplot[
    fill=clr2fill,draw = clr2,
    very thick,
    line width= \linew pt,
    rounded corners=\barcorner ex
    ]
    coordinates {
(3, 31)
};
\addplot[
    fill=clr1fill,draw = clr1,
    very thick,
    line width= \linew pt,
    rounded corners=\barcorner ex
    ]
    coordinates {
(4, 83)
};
 %

\addplot[
    very thick,
    fill=clr2fill,draw = clr2,
    line width= \linew pt,
    rounded corners=\barcorner ex
    ]
    coordinates {
(4, 45)
};
 %
  %
\legend{\footnotesize RTX-2070, \footnotesize Jetson-NX}
  \node [draw=gpudclr,fill=gpuclr,rounded corners=0.2ex, minimum width=23.7ex, xshift=\gpuxshift ex, yshift=\gpuyshift ex, scale=\gpuscale]{\textcolor{gputxtclr}{$\mathbf{@752 {\times} 480}$\textbf{, Scales}$\mathbf{=8}$}};
 \end{axis}
}
}};

\end{tikzpicture}
\subfloat{\label{fig:slamcomponents}}
\subfloat{\label{fig:slam_kitti}}
\subfloat{\label{fig:slam_euroc}}
\vspace{-2ex}
\caption{SLAM throughput analysis. (a) Frame-rates of SLAM components on EuRoC \cite{euroc} resolution. Notice how the frontend efficiency also transfers to the backend. (b) SLAM frame rate at KITTI resolution \cite{kitti}, and (c) SLAM frame rate EuRoC resolution.}
\vspace{-0.5ex}
\end{figure}
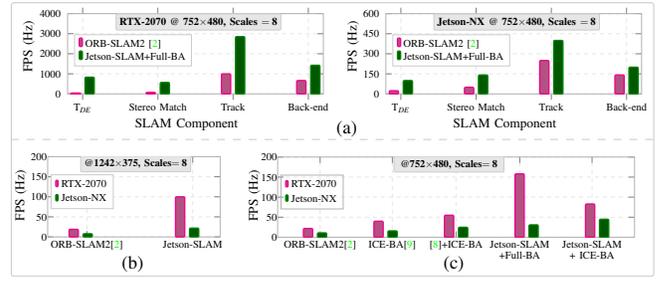

%% file: tables/kitti.tex
\begin{table}[!t]
\centering
\caption{SLAM Performance on KITTI \cite{kitti} . `\xmark': Tracking Failure.}
\label{tab:kittiate}
\arrayrulecolor{white!70!black}
\tiny
\setlength{\tabcolsep}{16.0pt}

\vspace{-1ex}
\begin{tabular}{l c c c}
\hline
\multicolumn{1}{l|}{\multirow{2}{*}{Approach}} &  \multicolumn{3}{c}{RMSE ATE (m)} \\ \cline{2-4}

\multicolumn{1}{l|}{} & \multicolumn{1}{c}{ KITTI-$00$} & \multicolumn{1}{|c}{KITTI-$01$} & \multicolumn{1}{|c}{KITTI-$02$} \\ \hline
\multicolumn{1}{l|}{\bdota{} ORB-SLAM~\cite{orb2}}  & \multicolumn{1}{c}{$0.70$m}   & \multicolumn{1}{c}{$1.39$m}   & \multicolumn{1}{c}{$0.76$m} \\
\multicolumn{1}{l|}{\bdota{} \cite{fastergpu}+Full-BA \cite{orb2}} & \multicolumn{1}{c}{\xmark}   & \multicolumn{1}{c}{\xmark}   & \multicolumn{1}{c}{\xmark} \\
\rowcolor{rwclr}
\multicolumn{1}{l|}{\bdotb{} Jetson-SLAM+Full-BA \cite{orb2}} & \multicolumn{1}{c}{$0.66$m}   & \multicolumn{1}{c}{$1.96$m}   & \multicolumn{1}{c}{$0.96$m} \\
\hline
\end{tabular}
\vspace{-4.0ex}
\end{table}

%% file: tables/euroc.tex
\begin{table}[!t]
\centering
\caption{SLAM Performance of VO/VIO/SLAM pipelines on EuRoC \cite{euroc}. boldface denotes Top-$3$ scores. `\xmark': Tracking Failure.}
\label{tab:existingeurociceba}
\arrayrulecolor{white!70!black}
\tiny
\setlength{\tabcolsep}{16.0pt}

\vspace{-1ex}
\begin{tabular}{l c c c}
\hline
\multicolumn{1}{l|}{\multirow{2}{*}{Approach}} &  \multicolumn{3}{c}{RMSE ATE (m)} \\ \cline{2-4}

\multicolumn{1}{l|}{} & \multicolumn{1}{c}{ MH$01$} & \multicolumn{1}{|c}{MH$02$} & \multicolumn{1}{|c}{MH$03$} \\ \hline
\multicolumn{1}{l|}{\bdota{} VINS-Stereo \cite{vinsfusion}}  & \multicolumn{1}{c}{$0.54$m}   & \multicolumn{1}{c}{$0.46$m}   & \multicolumn{1}{c}{$0.33$m} \\
\multicolumn{1}{l|}{\bdota{} MSCKF-Stereo \cite{msckf}} & \multicolumn{1}{c}{$0.42$m}   & \multicolumn{1}{c}{$0.45$m}   & \multicolumn{1}{c}{$0.23$m} \\
\multicolumn{1}{l|}{\bdota{} VINS-Stereo + IMU \cite{vinsfusion}}  & \multicolumn{1}{c}{$0.24$m}   & \multicolumn{1}{c}{$0.18$m}   & \multicolumn{1}{c}{$0.23$m} \\
\multicolumn{1}{l|}{\bdota{} ICE-BA \cite{iceba}} &   \multicolumn{1}{c}{$\mathbf{0.05}$m}   & \multicolumn{1}{c}{$\mathbf{0.04}$m}   & \multicolumn{1}{c}{$0.10$m} \\
\multicolumn{1}{l|}{\bdota{} \cite{fastergpu}$+$ICE-BA \cite{iceba}}  & \multicolumn{1}{c}{$0.16$m}   & \multicolumn{1}{c}{$\mathbf{0.06}$m}   & \multicolumn{1}{c}{$0.16$m} \\
\multicolumn{1}{l|}{\bdota{} \cite{fastergpu}$+$Full-BA \cite{iceba}}  & \multicolumn{1}{c}{\xmark}   & \multicolumn{1}{c}{\xmark}   & \multicolumn{1}{c}{\xmark} \\
\multicolumn{1}{l|}{\bdota{} SVO-GTSAM \cite{svogtsam}}  & \multicolumn{1}{c}{$\mathbf{0.05}$m}   & \multicolumn{1}{c}{$\mathbf{0.03}$m}   & \multicolumn{1}{c}{$\mathbf{0.12}$m} \\
\multicolumn{1}{l|}{\bdota{} KIMERA-VIO-FULL \cite{kimera}} & \multicolumn{1}{c}{$\mathbf{0.04}$m}   & \multicolumn{1}{c}{$0.07$m}   & \multicolumn{1}{c}{$\mathbf{0.12}$m} \\
\multicolumn{1}{l|}{\bdota{} KIMERA-RPGO \cite{kimera}}  & \multicolumn{1}{c}{$0.08$m}   & \multicolumn{1}{c}{$0.09$m}   & \multicolumn{1}{c}{$\mathbf{0.11}$m} \\
\rowcolor{rwclr}
\multicolumn{1}{l|}{\bdotb{} \OursAcronym + IMU + ICE-BA \cite{iceba}} &  \multicolumn{1}{c}{$\mathbf{0.07}$m}   & \multicolumn{1}{c}{$\mathbf{0.04}$m}   & \multicolumn{1}{c}{$\mathbf{0.07}$m} \\
\rowcolor{rwclr}
\multicolumn{1}{l|}{\bdotb{} \OursAcronym + \ORBBE~\cite{orb2}} &  \multicolumn{1}{c}{$\mathbf{0.04}$m}   & \multicolumn{1}{c}{$\mathbf{0.03}$m}   & \multicolumn{1}{c}{$\mathbf{0.03}$m} \\

\hline
\end{tabular}
\vspace{-0.5ex}
\end{table}

%% file: tables/kaist_vio.tex
\begin{table}[!t]
\centering
\caption{SLAM Performance on KAIST-VIO \cite{kaistvio} on Jetson-NX. Baselines results are borrowed from \cite{kaistvio}.}
\label{tab:kaistvio}
\arrayrulecolor{white!70!black}
\scriptsize
\tiny
\setlength{\tabcolsep}{0.6pt}

\vspace{-1ex}
\begin{tabular}{l c c c c c c c c c c c}
\hline 
\multicolumn{1}{c}{\multirow{3}{*}{Approach}} &  \multicolumn{11}{|c}{KAIST-VIO Sequence} \\ \cline{2-12}

 & \multicolumn{3}{|c}{\texttt{circle}}  & \multicolumn{3}{|c}{\texttt{infinite}} & \multicolumn{3}{|c}{\texttt{square}} & \multicolumn{2}{|c}{\texttt{rotation}} \\  \cline{2-12}
 
 & \multicolumn{1}{|c}{\texttt{normal}} & \multicolumn{1}{c}{\texttt{fast}} & \multicolumn{1}{c}{\texttt{head}}  & \multicolumn{1}{|c}{\texttt{normal}} & \multicolumn{1}{c}{\texttt{fast}} & \multicolumn{1}{c}{\texttt{head}}  & \multicolumn{1}{|c}{\texttt{normal}} & \multicolumn{1}{c}{\texttt{fast}} & \multicolumn{1}{c}{\texttt{head}}  & \multicolumn{1}{|c}{\texttt{normal}} & \multicolumn{1}{c}{\texttt{head}} \\  \cline{1-12}
\bdota{}
VINS-Fusion \cite{vinsfusion}  & \multicolumn{1}{|c}{$0.06$m}   & $0.12$m   & $\mathbf{0.08}$m   & \multicolumn{1}{|c}{$0.05$m}   & $0.09$m   & $0.12$m   & \multicolumn{1}{|c}{$0.17$m}   & $0.07$m   & $0.19$m   & \multicolumn{1}{|c}{$0.11$m}   & $0.28$m \\
\bdota{}
MSCKF-Stereo \cite{msckf}  & \multicolumn{1}{|c}{$0.12$m}   & $0.19$m   & $0.21$m   & \multicolumn{1}{|c}{$0.32$m}   & $0.17$m   & $0.60$m   & \multicolumn{1}{|c}{$0.10$m}   & $0.30$m   & $0.30$m   & \multicolumn{1}{|c}{$0.10$m}   & $0.29$m \\
\bdota{}
KIMERA-VIO \cite{kimera}  & \multicolumn{1}{|c}{$0.12$m}   & $0.07$m   & $0.28$m  & \multicolumn{1}{|c}{$0.05$m}   & $0.14$m   & $1.08$m   & \multicolumn{1}{|c}{$0.17$m}   & $0.19$m   & $1.57$m    & \multicolumn{1}{|c}{$0.17$m }   & $0.74$m  \\
\bdota{}
VINS-Fusion + IMU \cite{vinsfusion}  & \multicolumn{1}{|c}{$0.11$m}   & $0.10$m   & $0.13$m   & \multicolumn{1}{|c}{$0.08$m}   & $0.08$m   & $0.12$m   & \multicolumn{1}{|c}{$0.21$m}   & $0.13$m   & $0.20$m   & \multicolumn{1}{|c}{$0.16$m}   & $0.10$m \\
\bdota{}
VINS-Fusion + GPU \cite{vinsfusion}  & \multicolumn{1}{|c}{$0.09$m}   & $0.13$m   & $0.11$m   & \multicolumn{1}{|c}{$0.09$m}   & $0.05$m   & $0.14$m   & \multicolumn{1}{|c}{$0.12$m}   & $0.11$m   & $0.15$m   & \multicolumn{1}{|c}{$0.12$m}   & $0.11$m \\
\bdota{}
ORB-SLAM$2$ \cite{orb2}  & \multicolumn{1}{|c}{$0.09$m}   & $0.11$m   & $0.13$m   & \multicolumn{1}{|c}{$0.08$m}   & $0.10$m   & $0.12$m   & \multicolumn{1}{|c}{$0.09$m}   & $0.09$m   & $0.16$m   & \multicolumn{1}{|c}{$0.17$m}   & $0.21$m \\

\rowcolor{rwclr}
\bdotb{}
\OursAcronym{}  & \multicolumn{1}{|c}{$\mathbf{0.014}$m}   & $\mathbf{0.017}$m   & $0.12$m   & \multicolumn{1}{|c}{$\mathbf{0.017}$m}   & $\mathbf{0.016}$m   & $\mathbf{0.09}$m   & \multicolumn{1}{|c}{$\mathbf{0.016}$m}   & $\mathbf{0.017}$m   & $\mathbf{0.04}$m   & \multicolumn{1}{|c}{$\mathbf{0.07}$m}   & $\mathbf{0.09}$m 

\\

\hline
\end{tabular}
\vspace{-4.5ex}
\end{table}

%% file: plots/traj_plots_combined.tex
\begin{figure}[!t]
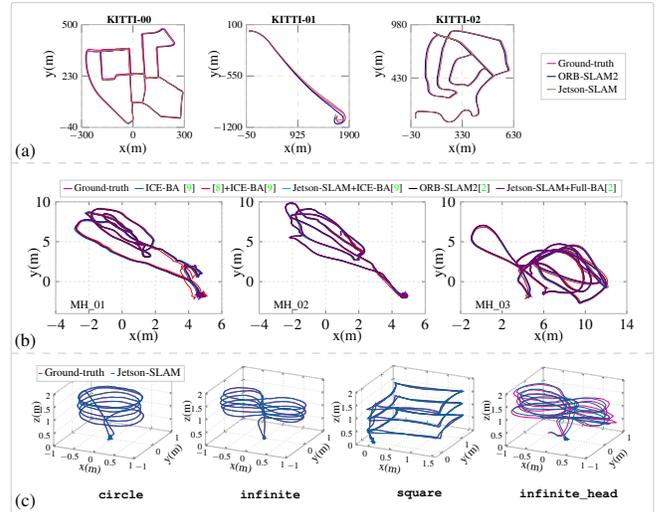

    \centering
\begin{tikzpicture}

\colorlet{frontendclr}{white!60!gray}
\colorlet{frontenddclr}{white!10!gray}

\colorlet{cpuclr}{white!60!magenta}
\colorlet{cpudclr}{white!20!magenta}

\colorlet{gpuclr}{white!80!blue}
\colorlet{gpudclr}{white!20!blue}

\colorlet{boundaryclr}{white!95!black}

\FPeval{\nodescale}{0.95}

\FPeval{\xshifta}{0-21}
\FPeval{\xshiftb}{0.0}
\FPeval{\xshiftc}{0+27}

\FPeval{\yshifta}{0-0.0}
\FPeval{\yshiftb}{0-0}
\FPeval{\yshiftc}{0-0.0}

\FPeval{\xshftd}{75}
\FPeval{\yshftd}{0-0.09}

\FPeval{\pltw}{35}
\FPeval{\plth}{25}

\FPeval{\scal}{0.70}

\FPeval{\mrksize}{0.25}

\FPeval{\netxshft}{5}
\FPeval{\netyshft}{0}

\FPeval{\lxshfta}{17.8}
\FPeval{\lyshfta}{7.75}

\FPeval{\lxshftb}{21.8}
\FPeval{\lyshftb}{3.5}

\FPeval{\lxshftc}{18.5}
\FPeval{\lyshftc}{0-5.5}

\FPeval{\lxshftd}{0-00.5}
\FPeval{\lyshftd}{0.5}

\FPeval{\lscale}{0.60}

\FPeval{\ticklabelscale}{0.6}
\FPeval{\labelscale}{0.8}

\colorlet{gtclr}{white!0!magenta}
\colorlet{predclr}{blue!10!brown}
\colorlet{orbclr}{blue!50!black}

\colorlet{desclr}{white!0!magenta}
\colorlet{currclr}{blue!60!green}

\colorlet{trnclr}{white!50!magenta}
\colorlet{dtrnclr}{white!10!magenta}
\colorlet{mtrnclr}{white!10!magenta}
\colorlet{mdtrnclr}{white!10!magenta}

\colorlet{tstclr}{white!50!blue}
\colorlet{dtstclr}{white!10!blue}
\colorlet{mtstclr}{white!50!blue}
\colorlet{mdtstclr}{white!10!blue}

\colorlet{gridclr}{white!90!black}
\colorlet{dlegendclr}{white!80!black}
\colorlet{axisclr}{white!70!black}
\colorlet{axisbgclr}{white!100!black}

\colorlet{dlegendclr}{white!80!black}
\colorlet{legendclr}{white!100!black}

\colorlet{comnetclr}{white!0!magenta}
\colorlet{repvggclr}{white!0!blue}
\colorlet{resnetclr}{black!50!green}
\colorlet{resnextclr}{white!0!cyan}
\colorlet{effnetclr}{white!0!brown}

%
%
\colorlet{comnetfillclr}{black!30!comnetclr}
\colorlet{resnetfillclr}{black!30!resnetclr}
%
%

\colorlet{gpudclr}{white!80!black}
\colorlet{gpuclr}{white!90!black}


\colorlet{deltaclr}{white!0!black}

\colorlet{gputxtclr}{white!0!black}

\FPeval{\xlabelrotate}{0}
\FPeval{\gpuscale}{0.6}
\FPeval{\gpuxshift}{11.5}
\FPeval{\gpuyshift}{9.0}
\FPeval{\jetspongpuyshift}{14.02}

\FPeval{\barw}{6}
\FPeval{\linew}{0.5}

\FPeval{eurocyshift}{0-15}
\FPeval{kaistyshift}{0-29}

\colorlet{bndryclr}{white!80!black}

%

\node (bn)[draw=bndryclr, rounded corners=0.2ex,minimum width=54ex, minimum height=43ex, xshift=0ex, yshift=-14.5ex]{};
\node (a)[draw=none, xshift=-25.8ex, yshift=-5.5ex,scale=0.7]{(a)};
\node (b)[draw=none, xshift=-25.8ex, yshift=-21.5ex,scale=0.7]{(b)};
\node (c)[draw=none, xshift=-25.8ex, yshift=-35.0ex,scale=0.7]{(c)};
\draw [bndryclr, dashed] ($(bn.west)-(0ex, -8.0ex)$) -- ($(bn.east)-(0, -8.0ex)$);
\draw [bndryclr, dashed] ($(bn.west)-(0ex, 8.0ex)$) -- ($(bn.east)-(0, 8.0ex)$);

\node [xshift=0.0ex,yshift=0ex,scale=\nodescale]{
\tikz{
%
%
\FPeval{\pltw}{23}
\FPeval{\plth}{23}
\node (n1) [draw=none, scale = \scal, xshift =\xshifta ex, yshift=\yshifta ex]{
\tikz{
\pgfplotsset{width=\pltw ex, height=\plth ex}
\begin{axis}[
   axis background style={fill=axisbgclr},
    title={},
    xlabel={x(m)},
    ylabel={y(m)},
    xmin=-300, xmax=300,
    ymin=-40, ymax=500,
    xtick={-300,  0,  300},
    ytick={-40, 230,  500},
   xticklabels={$-300$, $0$, $300$},
    yticklabels={$-40$, $230$,  $500$},
     axis line style={axisclr},
    legend image post style={scale =0.3},
    legend style={at={(\lxshftb ex,\lyshftb ex)},anchor=south, legend columns = 3, draw = {dlegendclr}, fill={legendclr}, nodes={scale=\lscale}},
    ymajorgrids=true, 
    xmajorgrids=true,
    grid style={dashed, gridclr},
    major tick length=1ex,
    y label style={at={(3.75 ex, 8.0 ex)}, scale=\labelscale},
    x label style={at={(6.5 ex, 2.0 ex)}, scale=\labelscale},
    xticklabel style={scale= \ticklabelscale},
    yticklabel style={scale= \ticklabelscale},
    legend cell align={left}
]
\input{plots/traj_coords_kitti_00}
%
%
 %
 %
\end{axis}
\node [xshift=6ex,yshift=13.7ex, scale=0.6]{\textbf{KITTI-}$\mathbf{00}$};
}};
\node (n2) [draw=none, scale = \scal, xshift =\xshiftb ex, yshift=\yshiftb ex]{
\tikz{
\pgfplotsset{width=\pltw ex, height=\plth ex}
\begin{axis}[
   axis background style={fill=axisbgclr},
    title={},
    xlabel={x(m)},
    ylabel={y(m)},
    xmin=-50, xmax=1900,
    ymin=-1200, ymax=100,
    xtick={-50, 925, 1900},
    ytick={-1200, -550, 100},
   xticklabels={$-50$, $925$, $1900$},
    yticklabels={$-1200$,$-550$, $100$},
     axis line style={axisclr},
    legend image post style={scale =0.3},
    legend style={at={(\lxshftb ex,\lyshftb ex)},anchor=south, legend columns = 1, draw = {dlegendclr}, fill={legendclr}, nodes={scale=\lscale}},
    ymajorgrids=true, 
    xmajorgrids=true,
    grid style={dashed, gridclr},
    major tick length=1ex,
    y label style={at={(3.75 ex, 8.0 ex)}, scale=\labelscale},
    x label style={at={(6.5 ex, 2.0 ex)}, scale=\labelscale},
    xticklabel style={scale= \ticklabelscale},
    yticklabel style={scale= \ticklabelscale},
    legend cell align={left}
]
\input{plots/traj_coords_kitti_01}
%
%
 %
  %
\end{axis}
\node [xshift=6ex,yshift=13.7ex, scale=0.6]{\textbf{KITTI-}$\mathbf{01}$};
}};
\node (n3) [draw=none, scale = \scal, xshift =\xshiftc ex, yshift=\yshiftc ex]{
\tikz{
\pgfplotsset{width=\pltw ex, height=\plth ex}
\begin{axis}[
   axis background style={fill=axisbgclr},
    title={},
    xlabel={x(m)},
    ylabel={y(m)},
    xmin=-30, xmax=630,
    ymin=-80, ymax=980,
    xtick={-30, 300, 630},
    ytick={-100, 430,  980},
   xticklabels={$-30$, $330$, $630$},
    yticklabels={$-80$,$430$, $980$},
     axis line style={axisclr},
    legend image post style={scale =0.3},
    legend style={at={(\lxshftb ex,\lyshftb ex)},anchor=south, legend columns = 1, draw = {dlegendclr}, fill={legendclr}, nodes={scale=\lscale}},
    ymajorgrids=true, 
    xmajorgrids=true,
    grid style={dashed, gridclr},
    major tick length=1ex,
    y label style={at={(3.75 ex, 8.0 ex)}, scale=\labelscale},
    x label style={at={(6.5 ex, 2.0 ex)}, scale=\labelscale},
    xticklabel style={scale= \ticklabelscale},
    yticklabel style={scale= \ticklabelscale},
    legend cell align={left}
]
\input{plots/traj_coords_kitti_02}
%
%
\legend{Ground-truth, ORB-SLAM$2$, \OursAcronym}
\end{axis}
\node [xshift=6ex,yshift=13.7ex, scale=0.6]{\textbf{KITTI-}$\mathbf{02}$};
}};
}};
%
%
%
%
%
%
%
\node [xshift=0.0ex,yshift=\eurocyshift ex,scale=\nodescale]{
\tikz{
%
%
\FPeval{\xshft}{18.0}
\FPeval{\yshft}{0-7.0}
\FPeval{\pltw}{59}
\FPeval{\plth}{43}
\FPeval{\xshifta}{0*\xshft}
\FPeval{\xshiftb}{1*\xshft}
\FPeval{\xshiftc}{2*\xshft}
\FPeval{\xshiftd}{19}
\FPeval{\yshifta}{0*\yshft}
\FPeval{\yshiftb}{0*\yshft}
\FPeval{\yshiftc}{0*\yshft}
\FPeval{\yshiftd}{5.5}
\FPeval{\bscal}{0.3}
\FPeval{\bscald}{0.41}
\FPeval{\ticklabelscale}{2.0}
\FPeval{\labelscale}{2.0}
\FPeval{\xlabelrotate}{0}
\FPeval{\gpuscale}{1.3}
\FPeval{\gpuxshift}{25.8}
\FPeval{\gpuyshift}{29.8}
\FPeval{\barw}{12}
\FPeval{\barsep}{0.8}
\FPeval{\linew}{0.1}
\FPeval{\barcorner}{0.4}
\FPeval{\mrksize}{0.4}
\FPeval{\legendscale}{1.0}
\FPeval{\legendimscale}{0.5}
\FPeval{\legendcolumna}{1}
\FPeval{\legendcolumnb}{1}
\FPeval{\legendcolumnc}{2}
\FPeval{\legendcolumnd}{0-1}
\FPeval{\legendxshifta}{15.0}
\FPeval{\legendxshiftb}{15.0}
\FPeval{\legendxshiftc}{24.0}
\FPeval{\legendyshifta}{7.0}
\FPeval{\legendyshiftb}{10.0}
\FPeval{\legendyshiftc}{25.0}
\FPeval{\ylabelxshifta}{1.5}
\FPeval{\ylabelxshiftb}{1.5}
\FPeval{\ylabelxshiftc}{1.5}
\FPeval{\ylabelyshifta}{15}
\FPeval{\ylabelyshiftb}{15}
\FPeval{\ylabelyshiftc}{15}
\FPeval{\xlabelxshifta}{26}
\FPeval{\xlabelxshiftb}{26}
\FPeval{\xlabelxshiftc}{26}
\FPeval{\xlabelyshifta}{0-0.0}
\FPeval{\xlabelyshiftb}{0-0.0}
\FPeval{\xlabelyshiftc}{0-0.0}
\colorlet{gridclr}{white!90!black}
\colorlet{dlegendclr}{white!80!black}
\colorlet{axisclr}{white!70!black}
\colorlet{axisbgclr}{white!100!black}
\colorlet{dlegendclr}{white!80!black}
\colorlet{legendclr}{white!100!black}
\colorlet{clr1}{white!0!magenta}
\colorlet{clr2}{blue!60!green}
\colorlet{clr3}{white!0!red}
\colorlet{clr4}{white!0!cyan}
\colorlet{clr5}{white!0!black}
\colorlet{clr6}{black!0!violet}
\colorlet{clr1fill}{black!30!clr1}
\colorlet{clr2fill}{black!30!clr2}
\colorlet{clr3fill}{black!30!clr3}
\colorlet{gpudclr}{white!80!black}
\colorlet{gpuclr}{white!90!black}
\colorlet{deltaclr}{white!0!black}
\colorlet{gputxtclr}{white!0!black}
\node (na) [draw=none, xshift =\xshifta ex, yshift= \yshifta ex]{
\scalebox{\bscal}
{\tikz{
\pgfplotsset{width=\pltw ex, height=\plth ex}
\begin{axis}[
   axis background style={fill=axisbgclr},
    title={},
    xlabel={x(m)},
    ylabel={y(m)},
    xmin=-4, xmax=6,
    ymin=-4, ymax=10,
    xtick={-4, -2, 0, 2, 4,  6},
    ytick={0, 5, 10},
     axis line style={axisclr},
    legend image post style={scale =\legendimscale},
    legend style={at={(\legendxshifta ex,\legendyshifta ex)},anchor=south, legend columns = \legendcolumna, draw = {dlegendclr}, fill={legendclr}, nodes={scale=\legendscale}},
    ymajorgrids=true, 
    xmajorgrids=true,
    grid style={dashed, gridclr},
    major tick length=1ex,
    y label style={at={(\ylabelxshifta ex, \ylabelyshifta ex)},scale=\labelscale},
    x label style={at={(\xlabelxshifta ex, \xlabelyshifta ex)},scale=\labelscale},
    yticklabel style={scale=\ticklabelscale},
    xticklabel style={scale=\ticklabelscale},
    legend cell align={left}
]
\input{plots/euroc_traj_coords/mh01_gt}
\input{plots/euroc_traj_coords/mh01_ice_ba}
\input{plots/euroc_traj_coords/mh01_ice_ba_fastergpu}
\input{plots/euroc_traj_coords/mh01_ice_ba_jetson_slam}
\input{plots/euroc_traj_coords/mh01_orbslam2}
\input{plots/euroc_traj_coords/mh01_full_ba_jetson_slam}
%
 %
\end{axis}
%
\node [xshift=9.5ex,yshift=2.5ex,scale=1.5]{MH\_$01$};
}
}};
\node (nb) [draw=none, xshift =\xshiftb ex, yshift= \yshiftb ex]{
\scalebox{\bscal}
{\tikz{
\pgfplotsset{width=\pltw ex, height=\plth ex}
\begin{axis}[
   axis background style={fill=axisbgclr},
    title={},
    xlabel={x(m)},
    ylabel={y(m)},
    xmin=-4, xmax=6,
    ymin=-4, ymax=10,
    xtick={-4, -2, 0, 2, 4,  6},
    ytick={0, 5, 10},
     axis line style={axisclr},
    legend image post style={scale =\legendimscale},
    legend style={at={(\legendxshiftb ex,\legendyshiftb ex)},anchor=south, legend columns = \legendcolumnb, draw = {dlegendclr}, fill={legendclr}, nodes={scale=\legendscale}},
    ymajorgrids=true, 
    xmajorgrids=true,
    grid style={dashed, gridclr},
    major tick length=1ex,
    y label style={at={(\ylabelxshiftb ex, \ylabelyshiftb ex)},scale=\labelscale},
    x label style={at={(\xlabelxshiftb ex, \xlabelyshiftb ex)},scale=\labelscale},
    yticklabel style={scale=\ticklabelscale},
    xticklabel style={scale=\ticklabelscale},
    legend cell align={left}
]
\input{plots/euroc_traj_coords/mh02_gt}
\input{plots/euroc_traj_coords/mh02_ice_ba}
\input{plots/euroc_traj_coords/mh02_ice_ba_fastergpu}
\input{plots/euroc_traj_coords/mh02_ice_ba_jetson_slam}
\input{plots/euroc_traj_coords/mh02_orbslam2}
\input{plots/euroc_traj_coords/mh02_full_ba_jetson_slam}
%
%
 %
\end{axis}
%
\node [xshift=9.5ex,yshift=2.5ex,scale=1.5]{MH\_$02$};
}
}};

\node (nc) [draw=none, xshift =\xshiftc ex, yshift= \yshiftc ex]{
\scalebox{\bscal}
{\tikz{
\pgfplotsset{width=\pltw ex, height=\plth ex}
\begin{axis}[
   axis background style={fill=axisbgclr},
    title={},
    xlabel={x(m)},
    ylabel={y(m)},
    xmin=-2, xmax=14,
    ymin=-4, ymax=10,
    xtick={-2, 2, 6, 10, 14},
    ytick={0, 5, 10},
     axis line style={axisclr},
    legend image post style={scale =\legendimscale},
    legend style={at={(\legendxshiftc ex,\legendyshiftc ex)},anchor=south, legend columns = \legendcolumnc, draw = {dlegendclr}, fill={legendclr}, nodes={scale=\legendscale}},
    ymajorgrids=true, 
    xmajorgrids=true,
    grid style={dashed, gridclr},
    major tick length=1ex,
    y label style={at={(\ylabelxshiftc ex, \ylabelyshiftc ex)},scale=\labelscale},
    x label style={at={(\xlabelxshiftc ex, \xlabelyshiftc ex)},scale=\labelscale},
    yticklabel style={scale=\ticklabelscale},
    xticklabel style={scale=\ticklabelscale},
    legend cell align={left}
]
\input{plots/euroc_traj_coords/mh03_gt}
\input{plots/euroc_traj_coords/mh03_ice_ba}
\input{plots/euroc_traj_coords/mh03_ice_ba_fastergpu}
\input{plots/euroc_traj_coords/mh03_ice_ba_jetson_slam}
\input{plots/euroc_traj_coords/mh03_orbslam2}
\input{plots/euroc_traj_coords/mh03_full_ba_jetson_slam}
%
 %
\end{axis}
%
\node [xshift=9.5ex,yshift=2.5ex,scale=1.5]{MH\_$03$};
}
}};

\node (nc) [draw=none, xshift =\xshiftd ex, yshift= \yshiftd ex]{
\scalebox{\bscald}
{\tikz{
\pgfplotsset{width=20 ex, height=20 ex}
\begin{axis}[
   hide axis,
	ticks=none,
	xmin=0,
	xmax=0.001,
	ymin=0,
	ymax=0.001,
	legend image post style={scale =\legendimscale},
    legend style={at={(0 ex,12 ex)},anchor=north, legend columns = \legendcolumnd, draw = {dlegendclr}, fill={legendclr}, nodes={scale=\legendscale}},
    legend cell align={left}
]
\addplot[
    fill=none,draw = clr1,
    very thick,
    line width= \linew ex,
    ]
    coordinates {
(0,0)
(0,0.0001)
};
\addplot[
    fill=none,draw = clr2,
    very thick,
    line width= \linew ex,
    ]
    coordinates {
(0,0)
(0,0.0001)
};
\addplot[
    fill=none,draw = clr3,
    very thick,
    line width= \linew ex,
    ]
    coordinates {
(0,0)
(0,0.0001)
};
\addplot[
    fill=none,draw = clr4,
    very thick,
    line width= \linew ex,
    ]
    coordinates {
(0,0)
(0,0.0001)
};
\addplot[
    fill=none,draw = clr5,
    very thick,
    line width= \linew ex,
    ]
    coordinates {
(0,0)
(0,0.0001)
};
\addplot[
    fill=none,draw = clr6,
    very thick,
    line width= \linew ex,
    ]
    coordinates {
(0,0)
(0,0.0001)
};
\legend{Ground-truth~~, ICE-BA \cite{iceba}~~,\cite{fastergpu}+ICE-BA\cite{iceba}~~, \OursAcronym{}+ICE-BA\cite{iceba}~~, ORB-SLAM$2$\cite{orb2}~~, \OursAcronym{}+Full-BA\cite{orb2}}
\end{axis}
%
%
}
}};
}};
%
%
%
%
%
%
\node [xshift=0.0ex,yshift=\kaistyshift ex,scale=\nodescale]{
\tikz{
%
%
\FPeval{\xshft}{13.2}
\FPeval{\yshft}{0-13.5}
\FPeval{\pltw}{68}
\FPeval{\plth}{55}
\FPeval{\xshifta}{0*\xshft}
\FPeval{\xshiftb}{1*\xshft}
\FPeval{\xshiftc}{2*\xshft}
\FPeval{\xshiftd}{3*\xshft}
\FPeval{\xshifte}{4*\xshft}
\FPeval{\yshifta}{0*\yshft}
\FPeval{\yshiftb}{0*\yshft}
\FPeval{\yshiftc}{0*\yshft}
\FPeval{\yshiftd}{0*\yshft}
\FPeval{\yshifte}{0*\yshft}
\FPeval{\pscal}{0.40}
\FPeval{\spltw}{\pltw*\pscal}
\FPeval{\splth}{\plth*\pscal}
\FPeval{\ticklabelscale}{2.0}
\FPeval{\labelscale}{2.5}
\FPeval{\xlabelrotate}{0}
\FPeval{\gpuscale}{1.3}
\FPeval{\gpuxshift}{25.8}
\FPeval{\gpuyshift}{29.8}
\FPeval{\barw}{12}
\FPeval{\barsep}{0.8}
\FPeval{\linew}{0.25}
\FPeval{\barcorner}{0.4}
\FPeval{\mrksize}{0.4}
\FPeval{\legendscale}{2.5}
\FPeval{\legendimscale}{0.5}
\FPeval{\legendxshifta}{28.0}
\FPeval{\legendxshiftb}{28.0}
\FPeval{\legendxshiftc}{28.0}
\FPeval{\legendxshiftd}{32.0}
\FPeval{\legendxshifte}{28.0}
\FPeval{\legendyshifta}{32.0}
\FPeval{\legendyshiftb}{32.0}
\FPeval{\legendyshiftc}{32.0}
\FPeval{\legendyshiftd}{32.0}
\FPeval{\legendyshifte}{32.0}
\FPeval{\ylabelxshifta}{56.0}
\FPeval{\ylabelxshiftb}{54.5}
\FPeval{\ylabelxshiftc}{54.5}
\FPeval{\ylabelxshiftd}{57.5}
\FPeval{\ylabelxshifte}{54.5}
\FPeval{\ylabelyshifta}{0-0.3}
\FPeval{\ylabelyshiftb}{0-0.3}
\FPeval{\ylabelyshiftc}{0-0.3}
\FPeval{\ylabelyshiftd}{0-3.3}
\FPeval{\ylabelyshifte}{0-0.3}
\FPeval{\xlabelxshifta}{20}
\FPeval{\xlabelxshiftb}{20}
\FPeval{\xlabelxshiftc}{20}
\FPeval{\xlabelxshiftd}{20}
\FPeval{\xlabelxshifte}{20}
\FPeval{\xlabelyshifta}{0-6.8}
\FPeval{\xlabelyshiftb}{0-6.8}
\FPeval{\xlabelyshiftc}{0-6.8}
\FPeval{\xlabelyshiftd}{0-6.8}
\FPeval{\xlabelyshifte}{0-6.8}
\FPeval{\zlabelxshifta}{0-4}
\FPeval{\zlabelxshiftb}{0-4}
\FPeval{\zlabelxshiftc}{0-4}
\FPeval{\zlabelxshiftd}{0-0}
\FPeval{\zlabelxshifte}{0-4}
\FPeval{\zlabelyshifta}{18}
\FPeval{\zlabelyshiftb}{18}
\FPeval{\zlabelyshiftc}{18}
\FPeval{\zlabelyshiftd}{18}
\FPeval{\zlabelyshifte}{18}
\FPeval{\xlabelrotate}{0-10}
\FPeval{\ylabelrotate}{0+45}
\colorlet{gridclr}{white!90!black}
\colorlet{dlegendclr}{white!80!black}
\colorlet{axisclr}{white!70!black}
\colorlet{axisbgclr}{white!100!black}
\colorlet{dlegendclr}{white!80!black}
\colorlet{legendclr}{white!100!black}
\colorlet{clr1}{white!0!magenta}
\colorlet{clr2}{blue!60!green}
\colorlet{clr3}{white!0!cyan}
\colorlet{clr1fill}{black!30!clr1}
\colorlet{clr2fill}{black!30!clr2}
\colorlet{clr3fill}{black!30!clr3}
\colorlet{gpudclr}{white!100!black}
\colorlet{gpuclr}{white!100!black}
\colorlet{deltaclr}{white!0!black}
\colorlet{gputxtclr}{white!0!black}
\FPeval{\gpuscale}{0.5}
\FPeval{\gpuxshift}{1}
\FPeval{\gpuyshift}{0+0}
\FPeval{\bscal}{0.18}
\node (na) [draw=none, xshift =\xshifta ex, yshift= \yshifta ex]{
\scalebox{\bscal}
{\tikz{
\pgfplotsset{width=\pltw ex, height=\plth ex}
\begin{axis}[
   axis background style={fill=axisbgclr},
    title={},
    xlabel={x(m)},
    ylabel={y(m)},
    zlabel={z(m)},
    xmin=-1.0, xmax=1.0,
    ymin=-1.0, ymax=1.0,
    zmin=0, zmax=2,
    xtick={-1, -0.5, 0, 0.5,  1},
    ytick={-1, 0,   1},
     axis line style={axisclr},
    legend image post style={scale =\legendimscale},
    legend style={at={(\legendxshifta ex,\legendyshifta ex)},anchor=south, legend columns = 2, draw = {dlegendclr}, fill={legendclr}, nodes={scale=\legendscale}},
    ymajorgrids=true, 
    xmajorgrids=true,
    zmajorgrids=true, 
    grid style={dashed, gridclr},
    major tick length=1ex,
    x label style={at={(\xlabelxshifta ex, \xlabelyshifta ex)},scale=\labelscale, rotate=\xlabelrotate},
    y label style={at={(\ylabelxshifta ex, \ylabelyshifta ex)},scale=\labelscale, rotate=\ylabelrotate},
    z label style={at={(\zlabelxshifta ex, \zlabelyshifta ex)},scale=\labelscale},
    xticklabel style={scale=\ticklabelscale},
    yticklabel style={scale=\ticklabelscale},
    zticklabel style={scale=\ticklabelscale},
]
\input{plots/kaist_traj_coords/circle_gt}
\input{plots/kaist_traj_coords/circle_jetson_slam}
\legend{Ground-truth~~, \OursAcronym}
\end{axis}
}
}};
%
%
%
%
%
%
%
%
\node (nb) [draw=none, xshift =\xshiftb ex, yshift= \yshiftb ex]{
\scalebox{\bscal}
{\tikz{
\pgfplotsset{width=\pltw ex, height=\plth ex}
\begin{axis}[
   axis background style={fill=axisbgclr},
    title={},
    xlabel={x(m)},
    ylabel={y(m)},
    zlabel={z(m)},
    xmin=-1.0, xmax=1.0,
    ymin=-1.0, ymax=1.0,
    zmin=0, zmax=2,
     axis line style={axisclr},
    legend image post style={scale =\legendimscale},
    legend style={at={(\legendxshiftb ex,\legendyshiftb ex)},anchor=south, legend columns = 2, draw = {dlegendclr}, fill={legendclr}, nodes={scale=\legendscale}},
    ymajorgrids=true, 
    xmajorgrids=true,
    zmajorgrids=true,
    grid style={dashed, gridclr},
    major tick length=1ex,
    x label style={at={(\xlabelxshiftb ex, \xlabelyshiftb ex)},scale=\labelscale, rotate=\xlabelrotate},
    y label style={at={(\ylabelxshiftb ex, \ylabelyshiftb ex)},scale=\labelscale, rotate=\ylabelrotate},
    z label style={at={(\zlabelxshiftb ex, \zlabelyshiftb ex)},scale=\labelscale},
    xticklabel style={scale=\ticklabelscale},
    yticklabel style={scale=\ticklabelscale},
    zticklabel style={scale=\ticklabelscale},
]
\input{plots/kaist_traj_coords/infinite_gt}
\input{plots/kaist_traj_coords/infinite_jetson_slam}
 %
%
 %
\end{axis}
}
}};
%
%
%
%
%
%
%
%
%
\node (nc) [draw=none, xshift =\xshiftc ex, yshift= \yshiftc ex]{
\scalebox{\bscal}
{\tikz{
\pgfplotsset{width=\pltw ex, height=\plth ex}
\begin{axis}[
   axis background style={fill=axisbgclr},
    title={},
    xlabel={x(m)},
    ylabel={y(m)},
    zlabel={z(m)},
    xmin=-0.3, xmax=1.6,
    ymin=-0.3, ymax=1.6,
    zmin=0, zmax=2,
     axis line style={axisclr},
    legend image post style={scale =\legendimscale},
    legend style={at={(\legendxshiftc ex,\legendyshiftc ex)},anchor=south, legend columns = 2, draw = {dlegendclr}, fill={legendclr}, nodes={scale=\legendscale}},
    ymajorgrids=true, 
    xmajorgrids=true,
    zmajorgrids=true,
    grid style={dashed, gridclr},
    major tick length=1ex,
    x label style={at={(\xlabelxshiftc ex, \xlabelyshiftc ex)},scale=\labelscale, rotate=\xlabelrotate},
    y label style={at={(\ylabelxshiftc ex, \ylabelyshiftc ex)},scale=\labelscale, rotate=\ylabelrotate},
    z label style={at={(\zlabelxshiftc ex, \zlabelyshiftc ex)},scale=\labelscale},
    xticklabel style={scale=\ticklabelscale},
    yticklabel style={scale=\ticklabelscale},
    zticklabel style={scale=\ticklabelscale},
]
\input{plots/kaist_traj_coords/infinite_head_gt}
\input{plots/kaist_traj_coords/infinite_head_jetson_slam}
%
 %
 %
\end{axis}
}
}};
%
%
%
%
%
%
%
\node (nd) [draw=none, xshift =\xshiftd ex, yshift= \yshiftd ex]{
\scalebox{\bscal}
{\tikz{
\pgfplotsset{width=\pltw ex, height=\plth ex}
\begin{axis}[
   axis background style={fill=axisbgclr},
    title={},
    xlabel={x(m)},
    ylabel={y(m)},
    zlabel={z(m)},
    xmin=-1.0, xmax=1.0,
    ymin=-1.0, ymax=1.0,
    zmin=0, zmax=2,
     axis line style={axisclr},
    legend image post style={scale =\legendimscale},
    legend style={at={(\legendxshiftd ex,\legendyshiftd ex)},anchor=south, legend columns = 2, draw = {dlegendclr}, fill={legendclr}, nodes={scale=\legendscale}},
    ymajorgrids=true, 
    xmajorgrids=true,
    zmajorgrids=true,
    grid style={dashed, gridclr},
    major tick length=1ex,
    x label style={at={(\xlabelxshiftd ex, \xlabelyshiftd ex)},scale=\labelscale, rotate=\xlabelrotate},
    y label style={at={(\ylabelxshiftd ex, \ylabelyshiftd ex)},scale=\labelscale, rotate=\ylabelrotate},
    z label style={at={(\zlabelxshiftd ex, \zlabelyshiftd ex)},scale=\labelscale},
    xticklabel style={scale=\ticklabelscale},
    yticklabel style={scale=\ticklabelscale},
    zticklabel style={scale=\ticklabelscale},
]
 %
 %
\input{plots/kaist_traj_coords/square_gt}
\input{plots/kaist_traj_coords/square_jetson_slam}
\end{axis}
}
}};
%
%
%
%
%
%
%
 %
\node [below of=na, draw=none,fill=none,rounded corners=0.2ex, minimum width=23.7ex,minimum height=3.5ex, xshift=\gpuxshift ex, yshift=\gpuyshift ex, scale=\gpuscale]{\textcolor{gputxtclr}{\textbf{\texttt{circle}}}};
\node [below of=nb, draw=none,fill=none,rounded corners=0.2ex, minimum width=23.7ex,minimum height=3.5ex, xshift=\gpuxshift ex, yshift=\gpuyshift ex, scale=\gpuscale]{\textcolor{gputxtclr}{\textbf{\texttt{infinite}}}};
\node [below of=nc, draw=none,fill=none,rounded corners=0.2ex, minimum width=23.7ex,minimum height=3.5ex, xshift=\gpuxshift ex, yshift=\gpuyshift ex, scale=\gpuscale]{\textcolor{gputxtclr}{\textbf{\texttt{square}}}};
\node [below of=nd, draw=none,fill=none,rounded corners=0.2ex, minimum width=23.7ex,minimum height=3.5ex, xshift=\gpuxshift ex, yshift=\gpuyshift ex, scale=\gpuscale]{\textcolor{gputxtclr}{\textbf{\texttt{infinite\_head}}}};
%
 %
%
%
}};

\end{tikzpicture}
\subfloat{\label{fig:kitti_traj}}
\subfloat{\label{fig:euroc_traj}}
\subfloat{\label{fig:kaist_traj}}
\vspace{-0.0ex}
\caption{\textbf{Zoom in}. Trajectory output on different datasets. (a) KITTI \cite{kitti}, (b) EuRoC \cite{euroc}, and (c)  KAIST-VIO \cite{kaistvio}.}
\label{fig:all_traj_plots}
\vspace{-1.0ex}
\end{figure}

%% file: tables/br_slam.tex
%
%
%
%
%
\begin{table}[!t]
\centering
\caption{Effect of bounded rectification on SLAM performance.}
\label{tab:cs}
\arrayrulecolor{white!70!black}
\tiny
\setlength{\tabcolsep}{11.5pt}

\vspace{-1ex}
\begin{tabular}{c c c c c}
\hline
\multicolumn{1}{c|}{\multirow{3}{*}{Approach}} & \multicolumn{1}{c|}{\multirow{3}{*}{Dataset}} & \multicolumn{1}{c}{\multirow{3}{*}{Sequence}} & \multicolumn{2}{|c}{RMSE ATE (m)} \\ \cline{4-5}

\multicolumn{1}{c|}{} & \multicolumn{1}{c|}{} &  & \multicolumn{2}{|c}{Bounded Rectification} \\ \cline{4-5}
\multicolumn{1}{c|}{} & \multicolumn{1}{c|}{} & & \multicolumn{1}{|c}{\xmark} & \multicolumn{1}{|c}{\cmark} \\ \hline
\multicolumn{1}{c|}{\multirow{3}{*}{\makecell{\OursAcronym \\$+$ICE-BA \cite{iceba}}}} & \multicolumn{1}{c|}{\multirow{3}{*}{EuRoC \cite{euroc}}} & \multicolumn{1}{c}{MH$01$} & \multicolumn{1}{|c}{$0.11$m}   & $0.07$m \\
 \multicolumn{1}{c|}{} & \multicolumn{1}{c|}{} & \multicolumn{1}{c}{MH$02$} & \multicolumn{1}{|c}{$0.07$m}   & $0.04$m \\
 \multicolumn{1}{c|}{} & \multicolumn{1}{c|}{} & \multicolumn{1}{c}{MH$03$} & \multicolumn{1}{|c}{$0.16$m}   & $0.07$m \\ \hline
 \multicolumn{1}{c|}{\makecell{\OursAcronym \\$+$FULL-BA \cite{orb2}}} & \multicolumn{1}{c|}{KAIST-VIO \cite{kaistvio}} & \multicolumn{1}{c}{\texttt{infinite\_fast}} & \multicolumn{1}{|c}{$0.10$m}  & $0.09$m \\
\hline

\end{tabular}
\vspace{-1ex}

\end{table}

%% file: tables/pyca_cellsize_slam_ablation.tex
\begin{table}[!t]
\centering
\caption{Effect of \texttt{PyCA} cell-size on SLAM for challenging \texttt{infinite\_head} sequence of KAIST-VIO. GPU: RTX-$2070$.}
\label{tab:pycacellsizeslam}
\arrayrulecolor{white!70!black}
\tiny
\setlength{\tabcolsep}{15.5pt}

\vspace{-1ex}
\begin{tabular}{c c c c}
\hline
\multicolumn{1}{c|}{\multirow{1}{*}{Cell Size}} & \multicolumn{1}{c}{Average Feature Count} &  \multicolumn{1}{|c}{RMSE ATE (m)} &\multicolumn{1}{|c}{FPS (Hz)} \\ \cline{1-4}
\multicolumn{1}{c|}{$15 \times 15$}  & $1800$ & $0.20$m & $52$  \\
\multicolumn{1}{c|}{$20 \times 20$}  & $1385$ & $0.13$m & $62$  \\
\multicolumn{1}{c|}{$25 \times 25$}  & $925$ & $0.10$m & $90$  \\
\rowcolor{rwclr}
\multicolumn{1}{c|}{$32 \times 32$}  & $\mathbf{600}$ & $\mathbf{0.09}$m & $200$  \\
\hline
\end{tabular}
\vspace{-0ex}
\end{table}
%

%% file: tables/num_scales_ablation.tex
%
%

   %
\begin{table}[!t]
\centering
\caption{Effect of number of scales on SLAM for seq. \texttt{infinite\_head} \cite{kaistvio}. GPU: RTX-$2070$. `\xmark': Tracking Failure.}
\label{tab:pycanumscalesslam}
\arrayrulecolor{white!70!black}
\tiny
\setlength{\tabcolsep}{8.8pt}

\vspace{-1ex}
\begin{tabular}{c c c c c}
\hline
\multicolumn{1}{c|}{\multirow{1}{*}{Number of scales}} & \multicolumn{1}{c}{\multirow{1}{*}{Cell Size}}  & \multicolumn{1}{|c}{Average Feature Count} &  \multicolumn{1}{|c}{RMSE ATE (m)} &\multicolumn{1}{|c}{FPS (Hz)} \\ \cline{1-5}
\multicolumn{1}{c|}{$2$} & $32 \times 32$ & \xmark & \xmark & \xmark  \\
\multicolumn{1}{c|}{$2$} & $10 \times 10$ & $1200$ & $0.09$m & $66$  \\
\multicolumn{1}{c|}{$4$} & $32 \times 32$ & \xmark & \xmark & \xmark  \\
\multicolumn{1}{c|}{$4$} & $20 \times 20$ & $800$ & $0.13$m & $125$  \\
\multicolumn{1}{c|}{$6$} & $32 \times 32$ & $700$ & $0.11$m & $166$  \\
\rowcolor{rwclr}
\multicolumn{1}{c|}{$8$} & $32 \times 32$ & $\mathbf{600}$ & $\mathbf{0.09}$m & $\mathbf{200}$  \\
\hline
\end{tabular}
\vspace{-4ex}
\end{table}

%% file: plots/deep_network_slam_all_in_one.tex
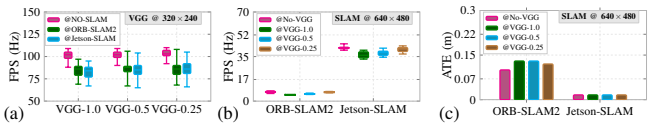
\begin{figure}[!t]
\centering

\begin{tikzpicture}

\FPeval{\xshifta}{0}
\FPeval{\xshiftb}{17.8}
\FPeval{\xshiftc}{36.5}

\FPeval{\yshifta}{0}
\FPeval{\yshiftb}{0}
\FPeval{\yshiftc}{0}

\FPeval{\pltw}{66}
\FPeval{\plth}{40}

\FPeval{\bscal}{0.25}

\FPeval{\ticklabelscale}{1.8}
\FPeval{\labelscale}{2.0}
\FPeval{\xlabelrotate}{0}

\FPeval{\gpuscale}{1.3}
\FPeval{\gpuxshift}{40.0}
\FPeval{\gpuyshift}{27.9}

\FPeval{\barw}{15}
\FPeval{\barsep}{1.5}
\FPeval{\linew}{0.3}
\FPeval{\barcorner}{0.4}

\FPeval{\linew}{2.0}
\FPeval{\boxw}{0.2}
\FPeval{\medianlinew}{0.5}

\FPeval{\mrksize}{0.4}
\FPeval{\legendscale}{1.2}
\FPeval{\legendimscale}{1.0}

\FPeval{\legendxshifta}{12.9}
\FPeval{\legendxshiftb}{11.9}
\FPeval{\legendxshiftc}{11.9}

\FPeval{\legendyshifta}{29.5}
\FPeval{\legendyshiftb}{29.5}
\FPeval{\legendyshiftc}{29.5}

\FPeval{\ylabelxshifta}{0-2.0}
\FPeval{\ylabelxshiftb}{0-0.20}
\FPeval{\ylabelxshiftc}{0-3.00}

\FPeval{\ylabelyshifta}{15}
\FPeval{\ylabelyshiftb}{15}
\FPeval{\ylabelyshiftc}{15}

\FPeval{\xlabelxshifta}{26}
\FPeval{\xlabelxshiftb}{26}
\FPeval{\xlabelxshiftc}{26}

\FPeval{\xlabelyshifta}{0-1.0}
\FPeval{\xlabelyshiftb}{0-1.0}
\FPeval{\xlabelyshiftc}{0-1.0}

\colorlet{medianclr}{white!0!blue}

\colorlet{gridclr}{white!90!black}
\colorlet{dlegendclr}{white!80!black}
\colorlet{axisclr}{white!70!black}
\colorlet{axisbgclr}{white!100!black}

\colorlet{dlegendclr}{white!80!black}
\colorlet{legendclr}{white!100!black}

\colorlet{clr1}{white!0!magenta}
\colorlet{clr2}{black!50!green}
\colorlet{clr3}{white!0!cyan}
\colorlet{clr4}{white!0!brown}

\colorlet{clr1fill}{black!30!clr1}
\colorlet{clr2fill}{black!30!clr2}
\colorlet{clr3fill}{black!30!clr3}
\colorlet{clr4fill}{black!30!clr4}

\colorlet{gpudclr}{white!80!black}
\colorlet{gpuclr}{white!90!black}

\colorlet{deltaclr}{white!0!black}
\colorlet{gputxtclr}{white!0!black}
\node (na) [draw=none, xshift =\xshifta ex, yshift= \yshifta ex]{
\scalebox{\bscal}
{\tikz{
\pgfplotsset{width=\pltw ex, height=\plth ex}
\begin{axis}[
   axis background style={fill=axisbgclr},
    title={},
    xlabel={},
    ylabel={FPS (Hz)},
    xmin=-0.2, xmax=4.2,
    ymin=50, ymax=150,
    xtick={0.7, 2.0, 3.3},
    ytick={50, 75, 100, 125, 150},
   xticklabels={VGG-$1.0$, VGG-$0.5$, VGG-$0.25$},
    yticklabels={$50$, $75$, $100$, $125$, $150$},
     axis line style={axisclr},
    ymajorgrids=true, 
    xmajorgrids=true,
    grid style={dashed, gridclr},
    major tick length=1ex,
    y label style={at={(\ylabelxshifta ex, \ylabelyshifta ex)}, scale=\labelscale},
    x label style={at={(\xlabelxshifta ex, \xlabelyshifta ex)}, scale=\labelscale},
    xticklabel style={rotate=\xlabelrotate, scale= \ticklabelscale},
    yticklabel style={xshift=0.0ex, scale= \ticklabelscale},
    legend image post style={scale =\legendimscale},
    legend style={at={(\legendxshifta ex,\legendyshifta ex)},anchor=north, legend columns = 1, draw = {dlegendclr}, fill={legendclr}, nodes={scale=\legendscale}},
 legend cell align=left
]
\addplot[
        area legend,
		boxplot/draw direction = y,
		boxplot/box extend=\boxw,
		boxplot/draw position=0.43,
		boxplot prepared={
			lower whisker=88,
			lower quartile=98,
			median=102,
			upper quartile=105,
			upper whisker=109,
		},
	    fill=clr1fill,draw = clr1,
    		very thick,
	    line width= \linew pt,
      rounded corners=\barcorner ex
    ]
coordinates {};
\addplot[
        area legend,
		boxplot/draw direction = y,
		boxplot/box extend=\boxw,
		boxplot/draw position=0.7,
		boxplot prepared={
			lower whisker=69,
			lower quartile=79,
			median=86,
			upper quartile=89,
			upper whisker=97,
		},
	    fill=clr2fill,draw = clr2,
    		very thick,
	    line width= \linew pt,
      rounded corners=\barcorner ex
    ]
coordinates {};
\addplot[
        area legend,
		area legend,
		boxplot/draw direction = y,
		boxplot/box extend=\boxw,
		boxplot/draw position=0.97,
		boxplot prepared={
			lower whisker=67,
			lower quartile=77,
			median=84,
			upper quartile=88,
			upper whisker=95	,
		},
	    fill=clr3fill,draw = clr3,
    		very thick,
	    line width= \linew pt,
      rounded corners=\barcorner ex
    ]
coordinates {};
\addplot[
        area legend,
		boxplot/draw direction = y,
		boxplot/box extend=\boxw,
		boxplot/draw position=1.73,
		boxplot prepared={
			lower whisker=90,
			lower quartile=99,
			median=102,
			upper quartile=105,
			upper whisker=109,
		},
	    fill=clr1fill,draw = clr1,
    		very thick,
	    line width= \linew pt,
      rounded corners=\barcorner ex
    ]
coordinates {};
\addplot[
        area legend,
		boxplot/draw direction = y,
		boxplot/box extend=\boxw,
		boxplot/draw position=2.0,
		boxplot prepared={
			lower whisker=67,
			lower quartile=83,
			median=88,
			upper quartile=89,
			upper whisker=106,
		},
	    fill=clr2fill,draw = clr2,
    		very thick,
	    line width= \linew pt,
      rounded corners=\barcorner ex
    ]
coordinates {};
\addplot[
        area legend,
		boxplot/draw direction = y,
		boxplot/box extend=\boxw,
		boxplot/draw position=2.27,
		boxplot prepared={
			lower whisker=65,
			lower quartile=80,
			median=85,
			upper quartile=90,
			upper whisker=104,
		},
	    fill=clr3fill,draw = clr3,
    		very thick,
	    line width= \linew pt,
      rounded corners=\barcorner ex
    ]
coordinates {};
\addplot[
		area legend,
		boxplot/draw direction = y,
		boxplot/box extend=\boxw,
		boxplot/draw position=3.03,
		boxplot prepared={
			lower whisker=92,
			lower quartile=101,
			median=103,
			upper quartile=107,
			upper whisker=110,
		},
	    fill=clr1fill,draw = clr1,
    		very thick,
	    line width= \linew pt,
      rounded corners=\barcorner ex
    ]
coordinates {};
\addplot[
		area legend,
		boxplot/draw direction = y,
		boxplot/box extend=\boxw,
		boxplot/draw position=3.3,
		boxplot prepared={
			lower whisker=68,
			lower quartile=80,
			median=86,
			upper quartile=90,
			upper whisker=108,
		},
	    fill=clr2fill,draw = clr2,
    		very thick,
	    line width= \linew pt,
      rounded corners=\barcorner ex
    ]
coordinates {};
\addplot[
        area legend,
		boxplot/draw direction = y,
		boxplot/box extend=\boxw,
		boxplot/draw position=3.57,
		boxplot prepared={
			lower whisker=66,
			lower quartile=81,
			median=86,
			upper quartile=92,
			upper whisker=105,
		},
	    fill=clr3fill,draw = clr3,
    		very thick,
	    line width= \linew pt,
      rounded corners=\barcorner ex
    ]
coordinates {};
%
%
\legend{~@NO-SLAM, ~@ORB-SLAM$2$, ~@\OursAcronym}
\node [draw=gpudclr,fill=gpuclr,rounded corners=0.2ex, minimum width=19.7ex, xshift=\gpuxshift ex, yshift=\gpuyshift ex, scale=\gpuscale]{\textcolor{gputxtclr}{\textbf{\shortstack{VGG$\mathbf{~@~320\times240}$}}}};
\end{axis}
}
}};
\node (nb) [draw=none, xshift =\xshiftb ex, yshift= \yshiftb ex]{
\scalebox{\bscal}
{\tikz{
\pgfplotsset{width=\pltw ex, height=\plth ex}
\begin{axis}[
   axis background style={fill=axisbgclr},
    title={},
    xlabel={},
    ylabel={FPS (Hz)},
    xmin=-0.2, xmax=3.2,
    ymin=0, ymax=70,
    xtick={0.75, 2.25},
    ytick={0, 17, 35, 52, 70},
   xticklabels={ORB-SLAM$2$, Jetson-SLAM},
    yticklabels={$0$, $17$, $35$, $52$, $70$},
     axis line style={axisclr},
    legend image post style={scale =\legendimscale},
    legend style={at={(\legendxshiftb ex,\legendyshiftb ex)},anchor=north, legend columns = 1, draw = {dlegendclr}, fill={legendclr}, nodes={scale=\legendscale}},
    ymajorgrids=true, 
    xmajorgrids=true,
    grid style={dashed, gridclr},
    major tick length=1ex,
    y label style={at={(\ylabelxshiftb ex, \ylabelyshiftb ex)}, scale=\labelscale},
    x label style={at={(\xlabelxshiftb ex, \xlabelyshiftb ex)}, scale=\labelscale},
    xticklabel style={rotate=\xlabelrotate, scale= \ticklabelscale},
    yticklabel style={xshift=0.0ex, scale= \ticklabelscale},
    legend cell align=left
]
\addplot[
		area legend,
		boxplot/draw direction = y,
		boxplot/box extend=\boxw,
		boxplot/draw position=0.15,
		boxplot prepared={
			lower whisker=6,
			lower quartile=6.6,
			median=7,
			upper quartile=7.8,
			upper whisker=8,
		},
	    fill=clr1fill,draw = clr1,
    		very thick,
	    line width= \linew pt,
      rounded corners=\barcorner ex
    ]
coordinates {};
\addplot[
		area legend,
		boxplot/draw direction = y,
		boxplot/box extend=\boxw,
		boxplot/draw position=0.55,
		boxplot prepared={
			lower whisker=4.76,
			lower quartile=4.8,
			median=5.05,
			upper quartile=5.1,
			upper whisker=5.26,
		},
	    fill=clr2fill,draw = clr2,
    		very thick,
	    line width= \linew pt,
      rounded corners=\barcorner ex
    ]
coordinates {};
\addplot[
		area legend,
		boxplot/draw direction = y,
		boxplot/box extend=\boxw,
		boxplot/draw position=0.95,
		boxplot prepared={
			lower whisker=5.26,
			lower quartile=5.3,
			median=5.5,
			upper quartile=6.0,
			upper whisker=6.25,
		},
	    fill=clr3fill,draw = clr3,
    		very thick,
	    line width= \linew pt,
      rounded corners=\barcorner ex
    ]
coordinates {};
\addplot[
		area legend,
		boxplot/draw direction = y,
		boxplot/box extend=\boxw,
		boxplot/draw position=1.35,
		boxplot prepared={
			lower whisker=6.45,
			lower quartile=6.8,	
			median=7,
			upper quartile=7.2,
			upper whisker=7.6,
		},
	    fill=clr4fill,draw = clr4,
    		very thick,
	    line width= \linew pt,
      rounded corners=\barcorner ex
    ]
coordinates {};
\addplot[
		area legend,
		boxplot/draw direction = y,
		boxplot/box extend=\boxw,
		boxplot/draw position=1.65,
		boxplot prepared={
			lower whisker=40,
			lower quartile=41.25,
			median=42,
			upper quartile=42.3,
			upper whisker=45,
		},
	    fill=clr1fill,draw = clr1,
    		very thick,
	    line width= \linew pt,
      rounded corners=\barcorner ex
    ]
coordinates {};
\addplot[
		area legend,
		boxplot/draw direction = y,
		boxplot/box extend=\boxw,
		boxplot/draw position=2.05,
		boxplot prepared={
			lower whisker=33,
			lower quartile=34,
			median=36,
			upper quartile=38.7,
			upper whisker=40,
		},
	    fill=clr2fill,draw = clr2,
    		very thick,
	    line width= \linew pt,
      rounded corners=\barcorner ex
    ]
coordinates {};
\addplot[
		area legend,
		boxplot/draw direction = y,
		boxplot/box extend=\boxw,
		boxplot/draw position=2.45,
		boxplot prepared={
			lower whisker=34.48,
			lower quartile=36,
			median=38,
			upper quartile=39.2,
			upper whisker=41.6,
		},
	    fill=clr3fill,draw = clr3,
    		very thick,
	    line width= \linew pt,
      rounded corners=\barcorner ex
    ]
coordinates {};
\addplot[
		area legend,
		boxplot/draw direction = y,
		boxplot/box extend=\boxw,
		boxplot/draw position=2.85,
		boxplot prepared={
			lower whisker=37,
			lower quartile=39,
			median=41,
			upper quartile=42.1,
			upper whisker=43.4,
		},
	    fill=clr4fill,draw = clr4,
    		very thick,
	    line width= \linew pt,
      rounded corners=\barcorner ex
    ]
coordinates {};
%
 %
\node [draw=gpudclr,fill=gpuclr,rounded corners=0.2ex, minimum width=19.7ex, xshift=\gpuxshift ex, yshift=\gpuyshift ex, scale=\gpuscale]{\textcolor{gputxtclr}{\textbf{\shortstack{SLAM$\mathbf{~@~640\times480}$}}}};
\legend{~@No-VGG, ~@VGG-$1.0$, ~@VGG-$0.5$, ~@VGG-$0.25$}
\end{axis}
}
}};
\node (nc) [draw=none, xshift =\xshiftc ex, yshift= \yshiftc ex]{
\scalebox{\bscal}
{\tikz{

\pgfplotsset{width=\pltw ex, height=\plth ex}
\begin{axis}[
   axis background style={fill=axisbgclr},
    title={},
    xlabel={},
    ylabel={ATE (m)},
    xmin=-0.2, xmax=3.2,
    ymin=0, ymax=0.3,
    xtick={0.75, 2.25},
    ytick={0, 0.07, 0.15, 0.22, 0.3},
   xticklabels={ORB-SLAM$2$, Jetson-SLAM},
    yticklabels={$0$, $0.07$, $0.15$, $0.22$, $0.3$},
     axis line style={axisclr},
    legend image post style={scale =\legendimscale},
    legend style={at={(\legendxshiftc ex,\legendyshiftc ex)},anchor=north, legend columns = 1, draw = {dlegendclr}, fill={legendclr}, nodes={scale=\legendscale}},
    ymajorgrids=true, 
    xmajorgrids=true,
    grid style={dashed, gridclr},
    major tick length=1ex,
    y label style={at={(\ylabelxshiftc ex, \ylabelyshiftc ex)}, scale=\labelscale},
    x label style={at={(\xlabelxshiftc ex, \xlabelyshiftc ex)}, scale=\labelscale},
    xticklabel style={rotate=\xlabelrotate, scale= \ticklabelscale},
    yticklabel style={xshift=0.0ex, scale= \ticklabelscale},
    ybar=\barsep ex,
    bar width = \barw pt,
    legend cell align=left
    ]
\addplot[
        area legend,
	    fill=clr1fill,draw = clr1,
    		very thick,
	    line width= \linew pt,
      rounded corners=\barcorner ex
    ]
coordinates {
(0.75, 0.10)
(2.25, 0.016)
};
\addplot[
        area legend,
	    fill=clr2fill,draw = clr2,
    		very thick,
	    line width= \linew pt,
      rounded corners=\barcorner ex
    ]
coordinates {
(0.75, 0.13)
(2.25, 0.016)
};
\addplot[
        area legend,
	    fill=clr3fill,draw = clr3,
    		very thick,
	    line width= \linew pt,
      rounded corners=\barcorner ex
    ]
coordinates {
(0.75, 0.13)
(2.25, 0.016)
};
\addplot[
        area legend,
	    fill=clr4fill,draw = clr4,
    		very thick,
	    line width= \linew pt,
      rounded corners=\barcorner ex
    ]
coordinates {
(0.75, 0.12)
(2.25, 0.016)
};
%
 \node [draw=gpudclr,fill=gpuclr,rounded corners=0.2ex, minimum width=19.7ex, xshift=\gpuxshift ex, yshift=\gpuyshift ex, scale=\gpuscale]{\textcolor{gputxtclr}{\textbf{\shortstack{SLAM$\mathbf{~@~640\times480}$}}}};
\legend{~@No-VGG, ~@VGG-$1.0$, ~@VGG-$0.5$, ~@VGG-$0.25$}
\end{axis}
}
}};
\node (a) [xshift=-8ex, yshift=-4.5ex, scale=0.7]{(a)};
\node (b) [xshift=10.0ex, yshift=-4.5ex, scale=0.7]{(b)};
\node (c) [xshift=28.6ex, yshift=-4.5ex, scale=0.7]{(c)};
\end{tikzpicture}
\subfloat{\label{fig:slam_on_vgg}}
\subfloat{\label{fig:vgg_on_slam_b}}
\subfloat{\label{fig:vgg_on_slam_c}}
%
\vspace{-3.5ex}
\caption{Effect of SLAM on the FPS of a deep network VGG~\cite{vgg} on Jetson-NX, stereo-mode, eight scales $@640 \times 480$, \texttt{infinite\_fast} sequence \cite{kaistvio}, and (b)-(c) Effect of VGG onto the FPS and accuracy of ORB-SLAM$2$ \cite{orb2} \textit{vs} \OursAcronym.}
\vspace{-2.5ex}
\end{figure}

%% file: conclusion.tex
We present a resource-efficient and accurate GPU-accelerated \OursAcronym{} for low-powered computing devices. We proposed \emph{Bounded Rectification} to prevent non-corners from being classified as corners in the FAST corner detection process, and \emph{Pyramidal Culling and Aggregation} \texttt{PyCA} which yields high-quality features at very high speeds in multiscale and stereo setting. \texttt{PyCA} is based on our Feature Culling (\texttt{FC}), Pyramidal Feature Aggregation (\texttt{PFA}), Multi-Location Per-Thread (\texttt{MLPT}) culling, and Thread Efficient Warp Allocation (\texttt{TEWA}) techniques. We also design \emph{Middle-end} in visual SLAM and develop a \OursAcronym{} library that utilizes synchronized shared memory to achieve resource efficiency. \OursAcronym{} exhibits a very high frame rate, suitable for modern autonomous robotic systems having several sub-systems. \OursAcronym{} outperforms many prominent SLAM pipelines by a large margin even in multi-scale and stereo settings on Jetson devices. 